\documentclass{article} 
\usepackage{iclr2026/iclr2026_conference,times}

\usepackage{amsmath,amsfonts,bm}

\def\eqref#1{equation~\ref{#1}}

\def\1{\bm{1}}

\DeclareMathAlphabet{\mathsfit}{\encodingdefault}{\sfdefault}{m}{sl}
\SetMathAlphabet{\mathsfit}{bold}{\encodingdefault}{\sfdefault}{bx}{n}

\usepackage{hyperref}
\usepackage{url}
\usepackage{graphicx}
\usepackage{array} 
\usepackage{booktabs}       
\usepackage{amsfonts}       
\usepackage{nicefrac}       
\usepackage{microtype}      
\usepackage{xcolor}
\usepackage{enumitem}
\usepackage{natbib}
\usepackage{setspace}
\usepackage{multirow} 
\usepackage{subcaption}
\usepackage{siunitx}
\usepackage{chngcntr}
\usepackage{float}
\title{Representational Alignment Across Model Layers and Brain Regions with Multi-Level \\ Optimal Transport}

\author{Shaan Shah\\
{\rm University of California San Diego}\\
\And
Meenakshi Khosla\\
{\rm University of California San Diego}\\
}

\iclrfinalcopy 
\begin{document}
\maketitle
\begin{abstract}
Standard representational similarity methods align each layer of a network to its best match in another independently, producing asymmetric results, lacking a global alignment score, and struggling with networks of different depths. These limitations arise from ignoring global activation structure and restricting mappings to rigid one-to-one layer correspondences.
We propose Multi-Level Optimal Transport (MOT), a unified framework that jointly infers soft, globally consistent layer-to-layer couplings and neuron-level transport plans. MOT allows source neurons to distribute mass across multiple target layers while minimizing total transport cost under marginal constraints. This yields both a single alignment score for the entire network comparison and a soft transport plan that naturally handles depth mismatches through mass distribution.
We evaluate MOT on vision models, large language models, and human visual cortex recordings. Across all domains, MOT matches or surpasses standard pairwise matching in alignment quality. Moreover, it reveals smooth, fine-grained hierarchical correspondences: early layers map to early layers, deeper layers maintain relative positions, and depth mismatches are resolved by distributing representations across multiple layers. These structured patterns emerge naturally from global optimization without being imposed, yet are absent in greedy layer-wise methods. MOT thus enables richer, more interpretable comparisons between representations, particularly when networks differ in architecture or depth. We further extend our method to a three-level MOT framework, providing a proof-of-concept alignment of two networks across their training trajectories and demonstrating that MOT uncovers checkpoint-wise correspondences missed by greedy layer-wise matching.
\end{abstract}
\section{Introduction}

Understanding high-dimensional neural activity is a shared challenge in neuroscience and artificial intelligence (AI). In neuroscience, comparing neural responses across individuals reveals which computations are universally shared versus idiosyncratic. In AI, comparing representations across models reveals how architectural choices, training objectives, and learning dynamics shape learned features, and helps identify principles of universality i.e. representational properties that emerge consistently across diverse network architectures and objectives. Comparing models to brains extends this logic further: while we cannot rerun biological evolution, we can simulate ``evolution in silico'' by training artificial networks with different constraints, inputs, and objectives. When such models converge on brain-like representations, they offer mechanistic hypotheses for why the brain may have adopted its computational strategies which is a deeply important theoretical question. 
These comparisons have revealed striking similarities between biological and artificial networks~(\cite{yamins2014performance,eickenberg2017seeing,gucclu2015deep,cichy2016comparison,KriegeskorteRSA, schrimpf2018brain, schrimpf2020integrative, storrs2021diverse, kell2018task}), common computational motifs across diverse architectures and objectives~\cite{huh2024platonic, kornblith2019similarity, bansal2021revisiting, dravid2023rosetta}, and other universal representational dimensions~\cite{chen2025universal, hosseini2024universality}.

The standard approach to representational comparison is layer-wise matching: each source layer is paired with the single best-matched target layer under some similarity measure (e.g., Representational Similarity Analysis~\citep{kriegeskorte2008representational}, Centered Kernel Alignment~\citep{kornblith2019similarity}, Procrustes distance~\citep{williams2021generalized} or linear predictivity). Despite its widespread use, this approach has fundamental limitations. It enforces rigid one-to-one correspondences that fail when networks differ in depth or when a source layer corresponds to features distributed across multiple target layers. It produces asymmetric layer mappings depending on the direction of comparison and yields no unified score for global network alignment. Most importantly, by optimizing each match independently, it ignores the global activation structure and risks overfitting to noise.

We propose Multi-Level Optimal Transport (MOT), a framework for globally consistent representational alignment. Optimal transport–based methods, such as Soft-Matching distance~\citep{khosla2024soft, khosla2024privileged}, have recently emerged as powerful metrics for comparing neural representations. Unlike metrics such as RSA, CKA, or linear predictivity—which are rotation-invariant and thus unable to capture similarities in neuron-level tuning—OT-based methods are rotation-sensitive. They explicitly match neurons based on their tuning profiles and, by relaxing hard permutation constraints into fractional couplings, can also handle layers of unequal size. This enables richer, more flexible neuron-level alignments than either rotation-invariant similarity metrics or strict permutation-based approaches. However, Soft Matching also remains limited to pairwise layer comparisons like other methods and does not capture global structure across networks. MOT fills this gap by operating across multiple levels: it simultaneously infers soft neuron-to-neuron couplings within layers and a soft, globally consistent layer-to-layer coupling across the two levels. Rather than forcing each source layer to match exactly one target layer, MOT allows source layers to distribute their representational ``mass'' across multiple target layers while minimizing the total transport cost under marginal constraints; that is, each source layer must distribute exactly 100$\%$ of its mass across target layers (no information is lost), and the total mass each target layer receives from all source layers must sum to a balanced allocation (no target is over- or under-utilized). These conservation laws ensure a balanced alignment where every layer contributes meaningfully to the global correspondence, preventing any layer from being arbitrarily overweighted or ignored in the global matching. The result is a single network-level alignment score and a soft transport plan that naturally handles depth mismatches. 


We evaluate MOT on three diverse domains: comparisons between foundation models in vision (Vision Transformers like DINOv2 and ViT-MAE), large language models of varying scales (LLaMA, Qwen), and fMRI recordings from human visual cortex across different participants. Our key contributions are:

\begin{itemize}
\item \textbf{A principled global alignment framework (theoretical):} MOT jointly optimizes all layer correspondences to produce symmetric, globally consistent assignments with a single alignment score. In contrast, greedy pairwise approaches are asymmetric, can overweight certain layers while completely ignoring others, and may spuriously treat wide layers as similar to every layer they are compared against, since their high dimensionality allows them to fit or partially overlap with many different representational subspaces, obscuring more meaningful correspondences. 

\item \textbf{Natural handling of depth mismatches (theoretical):} By allowing soft, many-to-many mappings between layers, MOT aligns networks of different depths without forcing inappropriate one-to-one correspondences.  

\item \textbf{Rotation-invariant extension (theoretical):} We propose an extension of MOT that incorporates additional orthogonal transformations, making the framework rotation-invariant. This ensures that correspondences can be recovered even when shared representational features are embedded in rotated subspaces, and yields consistently high-quality alignments.

\item \textbf{Improved alignment scores (empirical):} Across domains (vision models, large language models, and brain data), MOT matches or surpasses standard pairwise methods, yielding higher alignment.

\item \textbf{Emergent hierarchical structure (empirical):} Without imposing ordering constraints, MOT recovers known hierarchical organization in visual cortex data across subjects and reveals clean layer-to-layer correspondences in model–model comparisons where early layers map to early layers and deeper layers preserve their relative ordering. By contrast, greedy pairwise methods fail to reveal hierarchical structure, often leaving many layers unmatched while a single or few layers dominate the mappings.

\item \textbf{Featural distribution across depth (empirical):} MOT reveals how deeper networks spread computations across multiple layers that shallower networks compress into fewer stages. Concretely, a single layer in a shallower network often distributes its mass across several neighboring layers in a deeper network, an effect that greedy pairwise methods completely miss.

\end{itemize}

\section{Methods}

\subsection{Problem Setup and Existing Approaches}

Comparing the internal representations of neural networks often proceeds by a \emph{pairwise} layer search under some similarity or distance measure. In general, let $T$ denote the number of stimuli (e.g., images, text sequences) used to probe both models. We consider two neural networks:

\begin{itemize}
    \item The first network has $L$ layers. Layer $\ell$ contains $n_\ell$ units, and its activations across the $T$ stimuli are represented by
    \[
    X_\ell \in \mathbb{R}^{T \times n_\ell}, \quad \ell = 1, \ldots, L.
    \]
    \item The second network has $M$ layers. Layer $m$ contains $n_m$ units, with activations
    \[
    Y_m \in \mathbb{R}^{T \times n_m}, \quad m = 1, \ldots, M.
    \]
\end{itemize}
Each row of $X_\ell$ or $Y_m$ corresponds to the response of all units in that layer to a single stimulus, while each column corresponds to the activity of a single unit across stimuli. For any chosen alignment metric $S(\cdot, \cdot)$ (e.g., linear predictivity, Procrustes distance, RSA, CKA, or Soft Matching), one typically does:
\[
m^*(\ell) = \arg\max_m S(X_\ell, Y_m)
\]
and then reports the layer-wise score $S(X_\ell, Y_{m^*(\ell)})$. This enforces a hard one-to-one mapping from each source layer $\ell$ to a single target layer $m^*(\ell)$. However, when $L \neq M$, or when the set of features represented in one layer of network~A is distributed across multiple layers of network~B, such a rigid per-layer pairing is not well-suited: a source layer may genuinely correspond to a mixture of multiple target layers. Moreover, by optimizing each layer independently, this approach ignores the \emph{global} structure of all activations and can overfit to noise in any single layer's responses.

\begin{figure}[H]
\centering
\includegraphics[scale=0.15]{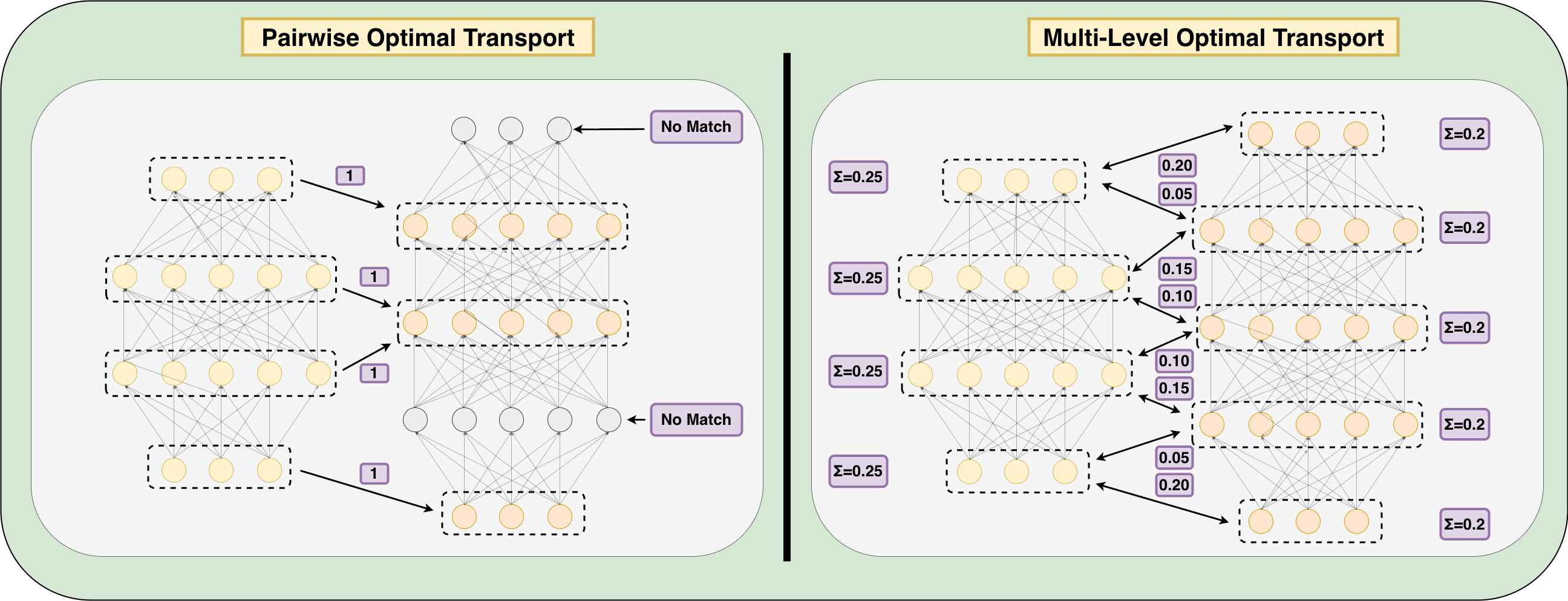}
\caption{\textit{Left}: \textbf{Pairwise OT}. Layers are matched independently, so multiple target layers can be mapped to the same source while other sources remain unused, yielding asymmetric, unbalanced mappings. \textit{Right}: \textbf{Multi-Level OT}. MOT infers a globally consistent transport plan where each source layer distributes all its mass and each target layer receives exactly one unit, ensuring balanced, symmetric alignments and handling depth mismatches.}

\label{fig:mainfigure}
\end{figure}

\subsection{Multi-Level Optimal Transport}

Our framework operates at two levels: an inner level that aligns neurons within each layer pair, and an outer level that determines how layers should be coupled globally.

\subsubsection{Inner Level: Neuron-to-Neuron Transport}

For each candidate layer pair $(\ell, m)$, MOT computes a neuron-level alignment cost by applying the soft-matching OT (Appendix \ref{sec:OTExplanation} for more details) formulation to the tuning functions in $X_\ell$ and $Y_m$. We first construct a pairwise dissimilarity matrix
\[
C_{\ell m}^{\text{inner}}[i,j] = c\!\left(X_\ell[:,i],\, Y_m[:,j]\right),
\]
where $c(\cdot,\cdot)$ is the chosen tuning-based dissimilarity (here, correlation distance). The inner OT objective then computes
\[
C_{\ell m}
= \min_{Q_{\ell m} \in \mathcal{T}(n_\ell,n'_m)}
   \langle C_{\ell m}^{\text{inner}},\, Q_{\ell m} \rangle,
\]
yielding a soft coupling $Q_{\ell m}$ that specifies how strongly each neuron in layer $\ell$ corresponds to neurons in layer $m$.

Because $Q_{\ell m}$ lies in the transportation polytope, its row and column marginals ensure that each source neuron distributes its representational mass uniformly and each target neuron receives its appropriate share. When the two layers have the same width ($n_\ell = n'_m$), the optimal solution reduces to a permutation (by the Birkhoff–von Neumann theorem~\citep{vonNeumann, de2013combinatorics}); when they differ, soft assignments naturally emerge, allowing neurons in one layer to map to mixtures of neurons in the other. The resulting inner OT cost $C_{\ell m}$ serves as the layer-to-layer dissimilarity used by the outer level of MOT.

\subsubsection{Outer Level: Layer-to-Layer Transport}

The inner costs $C_{\ell m}$ form a layer-to-layer cost matrix $C \in \mathbb{R}^{L \times M}$. We solve a second optimal transport problem to find the global layer coupling:
\[
P = \arg\min_{P \in \mathcal{T}(L, M)} \langle C, P \rangle
\]
where $\mathcal{T}(L, M)$ is the transportation polytope for layers:
\[
\mathcal{T}(L, M) = \left\{ P \in \mathbb{R}^{L \times M} : \sum_\ell P_{\ell m} = \tfrac{1}{M}, \; \sum_m P_{\ell m} = \tfrac{1}{L}, \; P_{\ell m} \geq 0 \right\}.
\]

The element $P_{\ell m}$ represents the fraction of layer $\ell$'s representation that is explained by layer $m$. These marginal constraints ensure mass conservation: each source layer distributes its unit mass across target layers (after normalization by $L$), and the total mass received by all target layers is balanced.

When the two networks have the same number of layers ($L = M$), we can show that the optimal solution $P$ is a (scaled) permutation matrix: the objective $\langle C, P \rangle$ is linear in $P$, and the constraints define the transportation polytope whose vertices are permutation matrices (scaled by $1/L$) by the Birkhoff–von Neumann theorem. Since linear programs achieve their optima at vertices, the solution finds a one-to-one layer matching when $L = M$. However, when $L \neq M$, the soft coupling allows source layers to distribute their mass across multiple target layers, naturally handling depth mismatches.

\subsubsection{Reconstruction and Evaluation}
\label{subsection:evalmethods}

Given the layer coupling $P$ and neuron couplings $\{Q_{\ell m}\}$, we reconstruct layer $\ell$ as:
\[
\hat{X}_\ell = L \sum_{m=1}^{M} P_{\ell m} Y_m Q_{\ell m}^{\top}.
\]
The factor $L$ appears because $P_{\ell m}$ sums to $1/L$ across $m$. We evaluate alignment quality using the mean correlation between original and reconstructed neurons on held-out data:
\[
\text{Score}_\ell = \frac{1}{n_\ell} \sum_{i=1}^{n_\ell} \rho(X_\ell[:, i], \hat{X}_\ell[:, i]).
\]
The global MOT score is the average across all layers:
\[
\text{MOT} = \frac{1}{L} \sum_\ell \text{Score}_\ell.
\]



\subsection{Rotation-Invariant Extension}
\label{subsection: rotationalmethods}
Most existing metrics of representational similarity, such as RSA, CKA, and Procrustes distance, are designed to be rotation-invariant. The rationale is that when comparing population codes, we often care less about the tuning of individual neurons and more about the geometry of the representational space: distances, angles, and relative positions between stimulus responses. Two networks can encode essentially the same geometry while using different coordinate bases (for example, rotated versions of one another). Requiring strict unit-to-unit correspondence in such cases would artificially inflate dissimilarity, even though the representational geometry and information content are preserved. A rotation-invariant extension of MOT (\textbf{MOT + R} henceforth) therefore provides a way to capture this geometric equivalence while still enforcing a globally consistent layer- and neuron-level coupling. We introduce rotation matrices $R_{\ell m} \in O(n_\ell)$ for each layer pair and minimize the reconstruction error:
\[
C_{\ell m} = \min_{Q_{\ell m}, R_{\ell m}} \|X_\ell R_{\ell m} - Y_m Q_{\ell m}^{\top}\|_F^2.
\]

We optimize via alternating minimization:
\begin{itemize}
    \item \textbf{Fix $R$, update $Q$:} Solve optimal transport using correlation distance on rotated features $X_\ell R_{\ell m}$.
    \item \textbf{Fix $Q$, update $R$:} Solve orthogonal Procrustes via SVD of $X_\ell^{\top}(Y_m Q_{\ell m}^{\top})$.
    \item \textbf{Update $P$:} Refresh the outer coupling using the Frobenius reconstruction costs.
\end{itemize}

Predictions incorporate the learned rotations:
\[
\hat{X}_\ell = L \sum_{m=1}^{M} P_{\ell m} Y_m Q_{\ell m}^{\top} R_{\ell m}^{\top}.
\]

\subsection{Baseline Comparisons}

We compare MOT against several baselines to isolate the contribution of each component:

\paragraph{Random Layer Assignment (Perm-P):} We randomly permute the rows of $P$, breaking the optimized layer correspondences while preserving the neuron-level optimal transport within each layer pair. This control is expected to perform reasonably well because it still finds optimal neuron matches for each (now randomly assigned) layer pairing—it only disrupts which layers are matched, not how well neurons align within those matches.

\paragraph{MOT Top-1 Layer Transport Plan (Single-Best OT):} Each source layer $\ell$ maps only to its highest-weight target layer 
\[
m^*(\ell) = \arg\max_m P_{\ell m},
\]
converting the soft layer coupling to a hard one-to-one assignment while keeping the soft neuron-level transport.

\paragraph{Independent Pairwise OT (Pairwise Best OT):} The standard greedy approach where each source layer is matched to the single target layer with minimum inner OT cost, computed independently on training data. Formally,
\[
m^*(\ell) = \arg\min_m C_{\ell m},
\]
and layer $\ell$ is reconstructed using only $Y_{m^*(\ell)}$. This baseline ignores global structure and can result in multiple source layers matching to the same target layer while leaving others unmatched.

\paragraph{Rotation-aware variants:} We evaluate rotation-invariant versions of the pairwise OT baseline (henceforth, `Pairwise Best + R'), where predictions use $Y_m Q_{\ell m}^{\top} R_{\ell m}^{\top}$ with orthogonal transformations optimized via Procrustes alignment.

\section{Results}
We evaluate representational similarity under the MOT metric across four distinct alignment setups: (i) large language models of different families and scales, (ii) fMRI responses from the visual cortex of four human subjects, (iii) pretrained transformer-based vision models spanning different families and scales, (iv) cross-domain comparisons between human visual cortex and vision transformers (Appendix Section \ref{sec:FMRIModelsExtraResults}). Across all settings, MOT matches or surpasses baseline reconstruction (prediction) scores. Importantly, the transport plans inferred by MOT naturally reveal systematic layer-wise correspondences across models and cortex, despite not being explicitly optimized for such structure; specifically, earlier layers or regions tend to align with earlier counterparts, while deeper layers or regions map to progressively higher levels; patterns that greedy pairwise methods fail to capture. Furthermore, we demonstrate a proof-of-concept extension of MOT to a Three-level MOT framework by aligning two networks across their training trajectories.

\subsection{Representational Similarity Between Large Language Models}
 
\textbf{Experimental Setup.} We extract layer-wise representations by averaging token activations across 2,552 prompts from the STSB dataset ~\citep{huggingface:dataset:stsb_multi_mt, enevoldsen2025mmtebmassivemultilingualtext, muennighoff2022mteb}. For each model, this yields a sequence of representation matrices $X$ and $Y$, corresponding to successive layers. We evaluate models of varying sizes from the LLaMA-3.2 ~\citep{grattafiori2024llama3herdmodels} and Qwen-2.5 ~\citep{qwen2025qwen25technicalreport} families. Representations are compared using MOT and baseline metrics, and the resulting transport plans are analyzed. To quantify alignment quality, we reconstruct representations on a held-out validation split (20$\%$) using the learned transport maps and report the correlation with ground-truth activations, as detailed in the Methods Section\ref{subsection:evalmethods}.

\begin{table}[H]
\centering
\small
\begin{tabular}{@{}ll S S S S@{}}
\toprule
\multicolumn{1}{c}{Model 1} & \multicolumn{1}{c}{Model 2} &
\multicolumn{1}{c}{MOT Metric} &
\multicolumn{1}{c}{Random (Perm-P)} &
\multicolumn{1}{c}{Single-Best OT} &
\multicolumn{1}{c}{Pairwise Best OT} \\
\midrule
Llama-3.2 1B   & Llama-3.2 3B   & \textbf{0.558} & 0.510 & 0.502 & 0.505 \\
Qwen-2.5 0.5B  & Qwen-2.5 3B    & \textbf{0.510} & 0.494 & 0.467 & 0.477 \\
\midrule
\addlinespace[2pt]
Qwen-2.5 0.5B  & Llama-3.2 1B   & \textbf{0.522} & 0.500 & 0.502 & 0.511 \\
Qwen-2.5 0.5B  & Llama-3.2 3B   & \textbf{0.531} & 0.513 & 0.498 & 0.524 \\
Llama-3.2 1B   & Qwen-2.5 3B    & \textbf{0.432} & 0.411 & 0.345 & 0.380 \\
Llama-3.2 3B   & Qwen-2.5 3B    & \textbf{0.383} & 0.374 & 0.338 & 0.346 \\
\bottomrule
\end{tabular}
\caption{\textbf{LLM alignment performance}. Comparison of MOT against baseline metrics, evaluated by reconstruction correlation on held-out data.}
\label{tab:llms}
\end{table}

\begin{figure}[H]
\centering
\includegraphics[width=0.495\columnwidth]{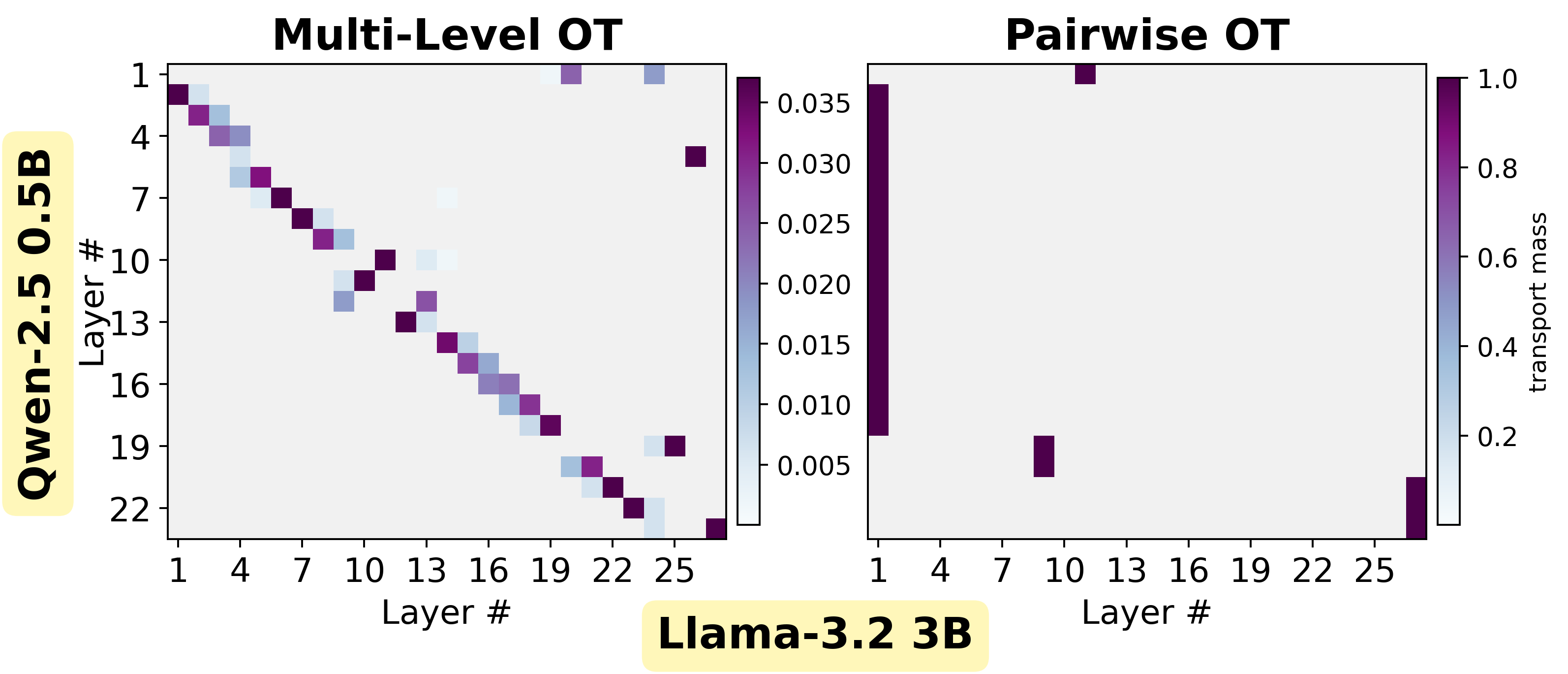}\hfill
\includegraphics[width=0.495\columnwidth]{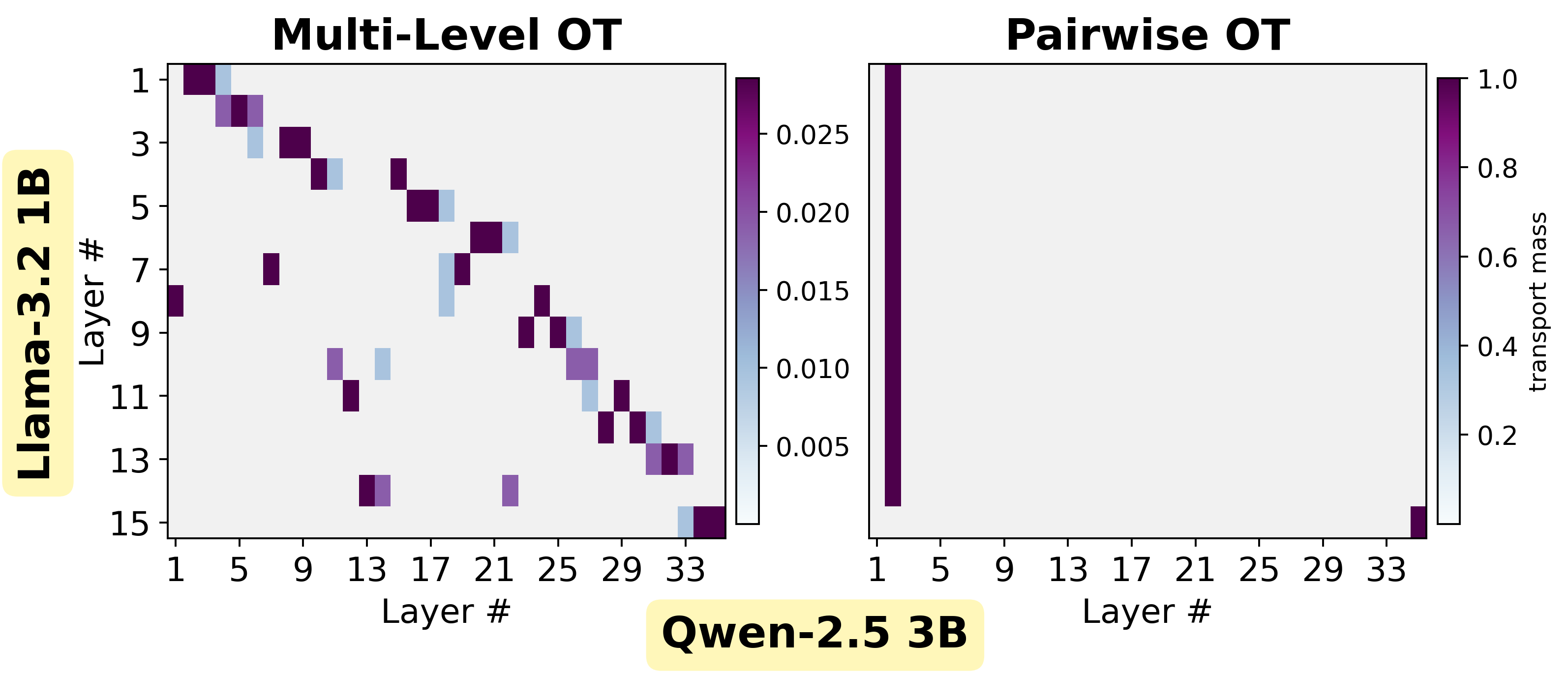}
\caption{\textbf{Transport plans for LLM alignment.} 
Multi-Level OT (left) versus pairwise OT (right) for two cross-model comparisons: 
(a) Qwen-2.5 0.5B $\leftrightarrow$ LLaMA-3.2 3B and 
(b) LLaMA-3.2 1B $\leftrightarrow$ Qwen-2.5 3B. 
MOT reveals smooth, diagonal correspondences across layers, while pairwise OT produces noisier and less structured mappings.}
\label{fig:llms}
\end{figure}

\textbf{Results.} MOT consistently achieves higher reconstruction accuracy than baseline methods. As shown in Table~\ref{tab:llms}, reconstruction scores on the held-out validation set are significantly higher under MOT, indicating that the alignment plans it learns are more robust and generalizable than those obtained from pairwise baselines. This improvement stems from MOT’s (i) enforcement of global consistency across all layers and (ii) ability to distribute representational mass across multiple target layers, allowing it to recover alignments that remain hidden to pairwise methods when features from a single layer are distributed across several layers in another model. Beyond quantitative gains, MOT uncovers clear hierarchical correspondences between models. As illustrated in Figures~\ref{fig:llms} and~\ref{fig:llms_all}, the transport plans produced by MOT exhibit strong diagonal structure: early layers in one model align with early layers in the other, while deeper layers align with deeper layers. Such structured correspondences are absent in pairwise OT, which often yields noisy mappings. Furthermore, when comparing shallower with deeper models, MOT reveals that single layers in the shallower model distribute their mass across multiple consecutive layers in the deeper model. This soft many-to-many mapping reflects how additional depth refines and spreads computations, suggesting that representational stages compressed into one layer in a shallower model are decomposed across multiple processing steps in a deeper model—an organizational principle that greedy baselines fail to uncover. 

To rule out the possibility that this effect stems from the multi-level formulation and not from OT, we also perform pairwise linear predictivity and RSA analyses. As shown in Figures~\ref{fig:linearpredictivity_transport} and ~\ref{fig:rsa_transport}, these methods also fail to reveal hierarchial correspondence.

\subsection{Representational Similarity Between Visual Cortex across Subjects}

\textbf{Experimental Setup.} We analyze fMRI responses from the Natural Scenes Dataset (NSD; ~\citep{allen2022massive}), which contains recordings from 8 participants who each viewed up to 10,000 natural images. Of these, 4 subjects viewed the full set of 10,000 images three times, with 1,000 images shared across all participants. We focus on these 4 subjects and restrict our analysis to their responses to the shared 1,000 images, ensuring that alignment is evaluated on common stimuli.

We target visual cortex regions (V1–V4, Lateral, Dorsal, Ventral), treating each region as a ``layer'' and individual voxels as ``neurons''. The visual cortex provides a strong testbed for multi-level alignment because it is one of the best-characterized cortical systems: early areas (V1, V2) are known to encode low-level features such as orientation and contrast, while higher areas (V3, V4) encode progressively more complex shapes and object features~\citep{hubel1968receptive, pasupathy2001shape, desimone1987visual, desimone1984stimulus}. This hierarchical progression is well-established across individuals, so correspondences between homologous regions are strongly expected.
MOT is applied to align responses across subjects by generating transport maps between regions. To evaluate alignment quality, we reconstruct held-out responses (20$\%$ validation split) and report correlations with ground-truth activity. We repeat this evaluation across 5 random train-validation splits to ensure robustness.
\begin{table}[H]
\centering
\small
\begin{tabular}{@{}ll S S S S@{}}
\toprule
\multicolumn{1}{c}{Model 1} & \multicolumn{1}{c}{Model 2} & \multicolumn{1}{c}{MOT Metric} & \multicolumn{1}{c}{Random (Perm-P)} & \multicolumn{1}{c}{Single-Best OT} & \multicolumn{1}{c}{Pairwise Best OT} \\
\midrule
Subject A & Subject B & 0.244 & 0.135 & 0.244 & \textbf{0.245} \\
Subject A & Subject C & 0.199 & 0.110 & 0.199 & \textbf{0.202} \\
Subject A & Subject D & 0.198 & 0.109 & 0.198 & \textbf{0.201} \\
Subject B & Subject C & \textbf{0.212} & 0.126 & \textbf{0.212} & \textbf{0.212} \\
Subject C & Subject D & 0.197 & 0.112 & 0.197 & \textbf{0.199} \\
Subject B & Subject D & 0.201 & 0.121 & 0.201 & \textbf{0.204} \\

\bottomrule
\end{tabular}
\caption{\textbf{Visual cortex alignment performance.} Comparison of MOT against baseline methods, evaluated by reconstruction correlation on held-out fMRI responses.}
\label{tab:vision}
\end{table}

\begin{figure}[H]
\centering
\includegraphics[width=0.495\columnwidth]{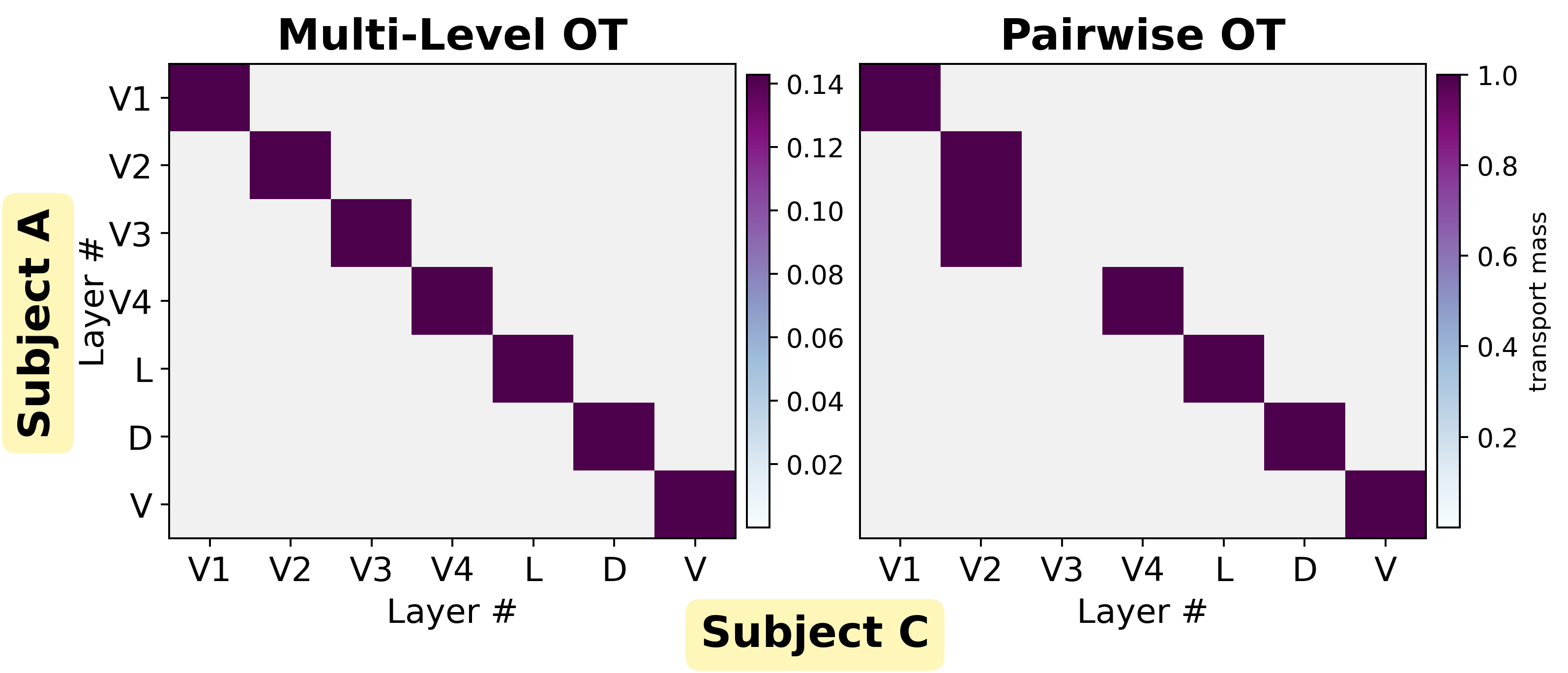}\hfill
\includegraphics[width=0.495\columnwidth]{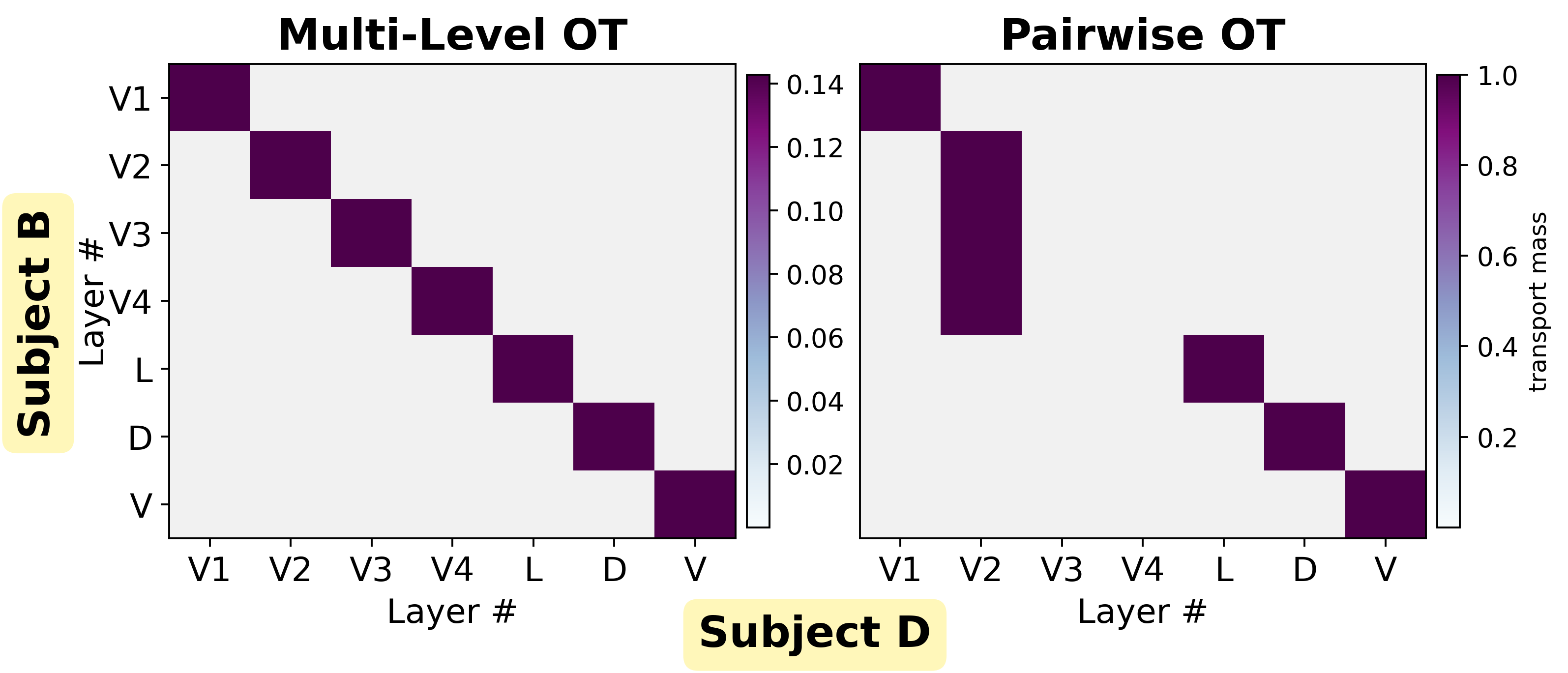}
\caption{\textbf{Transport plans for cross-subject brain alignment.} 
Multi-Level OT (left) versus pairwise OT (right) for two randomly selected subject pairs: 
(a) Subject A $\leftrightarrow$ Subject C and 
(b) Subject B $\leftrightarrow$ Subject D. 
MOT recovers structured region-to-region correspondences that are absent in pairwise OT. 
Other subject pairs show similar trends (see Appendix~\ref{fig:fmris_all}).}
\label{fig:fmris}
\end{figure}

\textbf{Results.} As shown in Tables~\ref{tab:vision} and~\ref{tab:vision_mean_std}, MOT achieves reconstruction scores comparable to pairwise OT, with only a marginal decrease in correlation. More importantly, Figures~\ref{fig:fmris} and~\ref{fig:fmris_all} show that the transport maps inferred by MOT recover the expected cross-subject correspondences: cortical regions in one subject consistently align to the same regions in another. In contrast, pairwise OT does not produce such structured mappings in any subject pair. This indicates that MOT’s global alignment scheme captures region-to-region correspondences that pairwise methods miss. We further assess pairwise linear predictivity and RSA, with the results as shown in Figures~\ref{fig:linearpredictivity_transport} and ~\ref{fig:rsa_transport}. The layers selected as best-matching by these approaches fail to align corresponding regions as well, reinforcing that the multi-level alignment mechanism in MOT is crucial for obtaining robust and generalizable correspondences. Overall, these results demonstrate that MOT yields transport plans that are both interpretable and biologically meaningful, while maintaining reconstruction performance on par with baseline methods.

\subsection{Representation Similarity between Vision Models}
\textbf{Experimental Setup.} We extract layer-wise representations from Vision Transformers by averaging patch activations for each input image. We use 20,000 images randomly sampled from the ImageNet validation set ~\citep{5206848, ILSVRC15}, sampled to ensure a uniform coverage across all classes in Imagenet. For each model, this yields a sequence of representation matrices that serve as inputs to the MOT framework. We evaluate two families of pretrained vision transformers—DINOv2 and ViT-MAE across multiple model scales. Alignment quality is quantified by reconstructing representations on a held-out validation split (20$\%$) using the learned transport plans, with reconstruction–ground truth correlation as the metric.

Prior work has shown that the residual stream in Transformers lacks privileged axes and is invariant up to rotations of its basis \citep{khosla2024privileged}. Since MOT, like other OT-based methods, is rotation-sensitive, we additionally evaluate a rotation-augmented variant (MOT+R) and its baselines (see Methods Section \ref{subsection: rotationalmethods}). We restrict this analysis to Vision Transformers, where rotational invariances are especially relevant and where the computational cost of MOT+R remains tractable. For large language models and fMRI data, the added optimization over rotations was computationally prohibitive, so we report only the rotation-sensitive results in those domains.
In MOT+R, the learned rotation matrices are incorporated into both transport optimization and evaluation, ensuring that geometric equivalences induced by rotations are properly captured.

\begin{table}[H]
\centering
\small
\begin{tabular}{@{}ll S S S S@{}}
\toprule
\multicolumn{1}{c}{Model 1} & \multicolumn{1}{c}{Model 2} & \multicolumn{1}{c}{MOT Metric} & \multicolumn{1}{c}{Pairwise Best OT} & \multicolumn{1}{c}{MOT + R} & \multicolumn{1}{c}{Pairwise Best + R} \\
\midrule
DINOv2 Small & ViT-MAE Base  & 0.289 & 0.301 & \textbf{0.600} & 0.526 \\
DINOv2 Small & DINOv2 Large  & 0.353 & 0.340 & \textbf{0.778} & 0.394 \\
DINOv2 Small & DINOv2 Giant  & 0.466 & 0.433 & \textbf{0.790} & 0.418 \\
DINOv2 Small & ViT-MAE Large & 0.381 & 0.354 & \textbf{0.633} & 0.509 \\
DINOv2 Small & ViT-MAE Huge  & 0.411 & 0.386 & \textbf{0.657} & 0.508 \\
\midrule
ViT-MAE Base & DINOv2 Large  & 0.577 & 0.624 & \textbf{0.732} & 0.283 \\
ViT-MAE Base & DINOv2 Giant  & 0.202 & 0.180 & \textbf{0.580} & 0.293 \\
ViT-MAE Base & ViT-MAE Large & 0.588 & 0.598 & \textbf{0.850} & 0.596 \\
ViT-MAE Base & ViT-MAE Huge  & 0.149 & 0.417 & \textbf{0.788} & 0.571 \\
ViT-MAE Huge & DINOv2 Giant  & 0.317 & 0.352 & \textbf{0.614} & 0.359 \\
\bottomrule
\end{tabular}
\caption{\textbf{Vision model alignment performance} Reconstruction accuracy under MOT and pairwise OT, reported both in the standard (rotation-sensitive) and rotation-augmented (MOT+R) settings.}
\label{tab:subset-hot-pairwise-rotation}
\end{table}

\begin{figure}[H]
\centering
\includegraphics[width=0.48\textwidth]{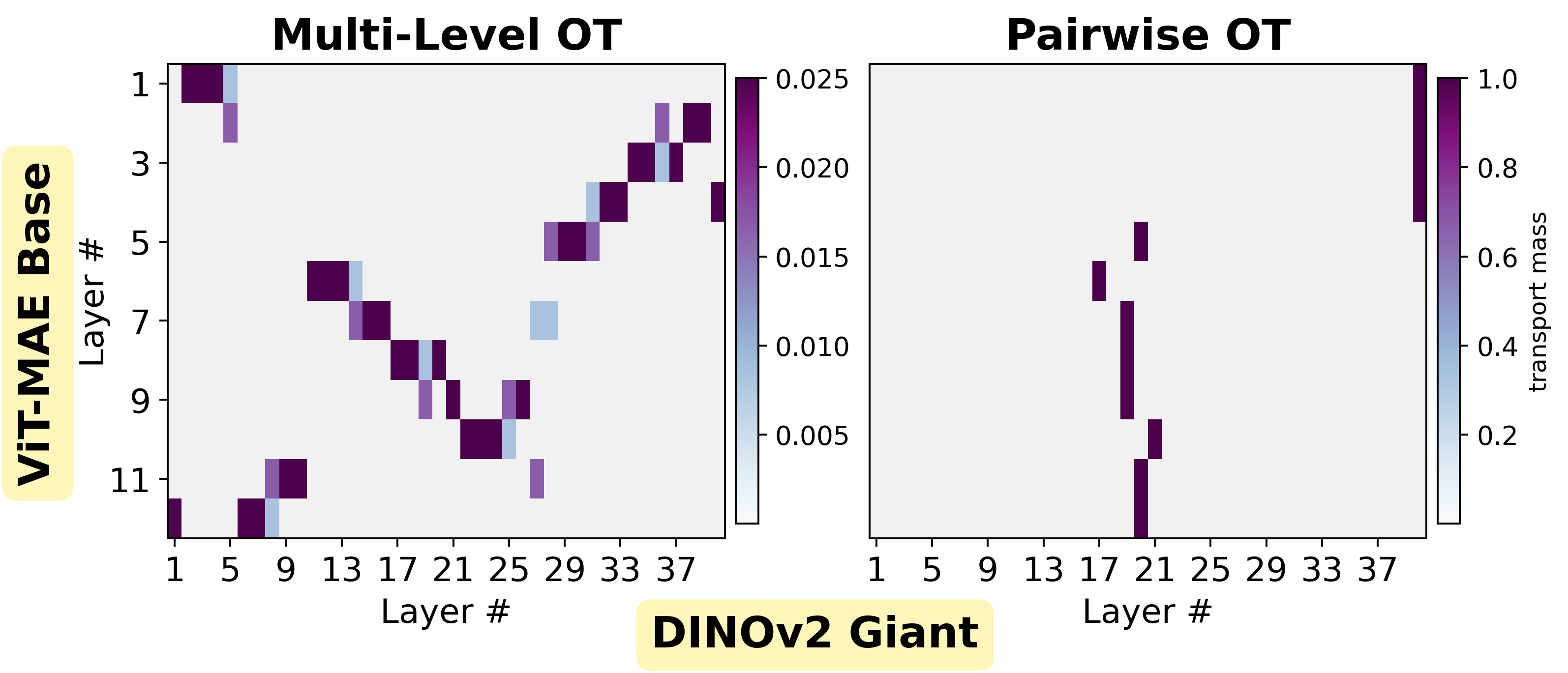}\hfill
\includegraphics[width=0.48\textwidth]{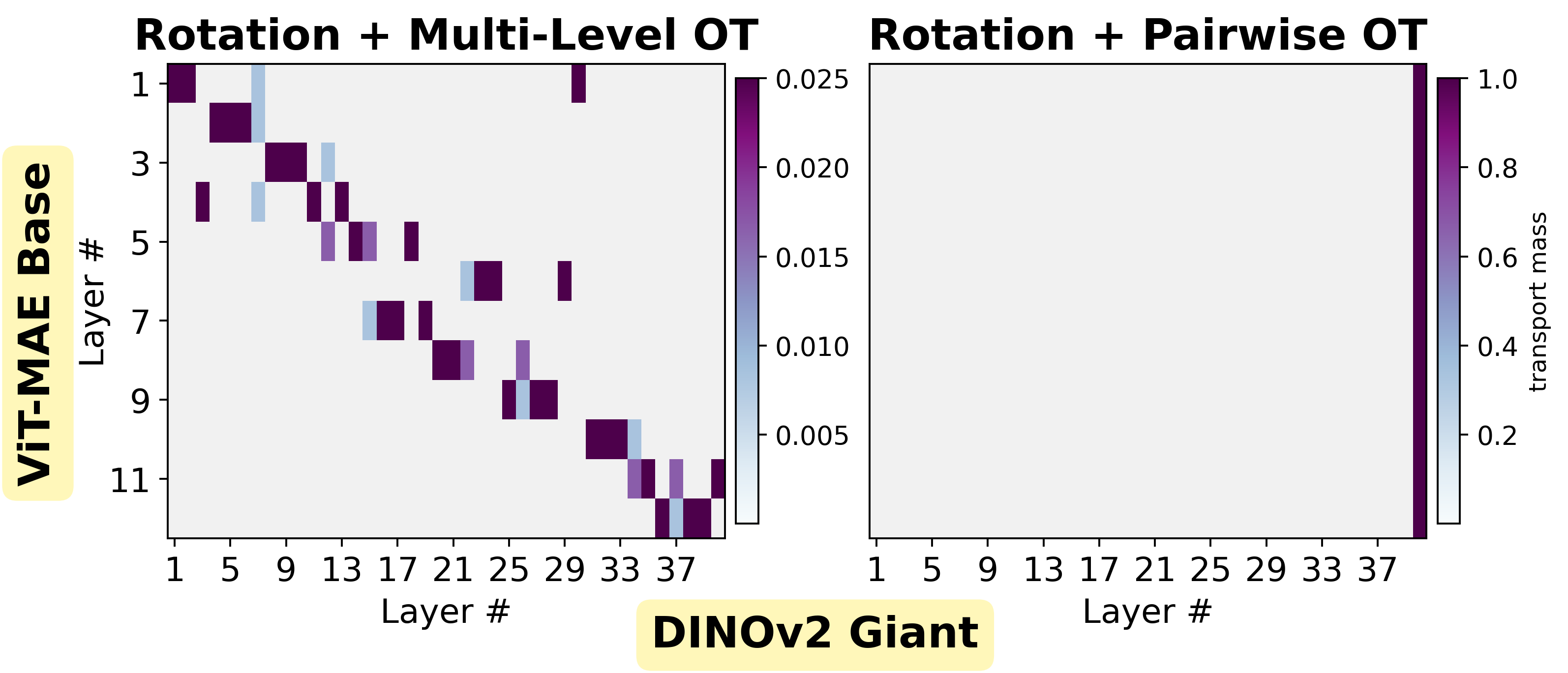}
\caption{\textbf{Transport plans for vision model alignment.} 
ViT-MAE Base $\leftrightarrow$ DINOv2 Giant 
(a) without rotation (MOT) and 
(b) with rotation augmentation (MOT+R). 
MOT+R captures geometric equivalences induced by rotations, yielding clearer correspondences than the rotation-sensitive variant.}
\label{fig:vit-mae-base-dinov2-giant-resultsvisionmodels}
\end{figure}

\textbf{Results.} Table~\ref{tab:subset-hot-pairwise-rotation} shows that, on its own, vanilla MOT does not consistently outperform pairwise OT in reconstruction accuracy. The transport plans inferred by MOT (Figure ~\ref{fig:vit-mae-base-dinov2-giant-resultsvisionmodels} and Figures ~\ref{fig:dinov2-small-vit-mae-base} - ~\ref{fig:vit-mae-huge-dinov2-giant}) only partially reveal layer-wise correspondences: in some cases, clear diagonal structure emerges, but in others the mappings are noisier. By contrast, the rotation-augmented variant (MOT+R) yields substantially higher reconstruction scores surpassing both vanilla MOT and pairwise vanilla and rotational baselines (Tables ~\ref{tab:subset-hot-pairwise-rotation}, ~\ref{tab:hot-and-baselines} and ~\ref{tab:rotation-and-baselines}). Importantly, the transport plans produced by MOT+R consistently exhibit strong hierarchical correspondences, recovering clean layer-to-layer alignment even in settings where vanilla MOT fails to do so. These findings indicate that incorporating rotation into the OT framework not only improves quantitative alignment quality but also produces more generalizable and interpretable mappings, particularly in domains like Vision Transformers where representations are known to be rotation-invariant.

\subsection{Three-Level MOT for Aligning Training Trajectories}

\textbf{Formulation of Three-Level MOT}. We extend the two-level MOT formulation described above to an additional level over checkpoints. Let the first model have checkpoints $c = 1,\dots,C_A$ and the second $d = 1,\dots,C_B$, with checkpoint-$c$ activations $X^{(c)}_\ell$ and checkpoint-$d$ activations $Y^{(d)}_m$.

For each checkpoint pair $(c,d)$, we treat $X^{(c)}_\ell$ and $Y^{(d)}_m$ as two networks and run the same two-level MOT procedure as above, yielding neuron- and layer-level couplings together with a single scalar cost
\[
C^{\text{chkpt}}_{cd}
=
\mathrm{MOT}\big(X^{(c)}_\ell ; Y^{(d)}_m\big),
\]
where $\mathrm{MOT}(\cdot,\cdot)$ denotes the two-level objective (inner neuron OT followed by outer layer OT). This defines a checkpoint-level cost matrix $C^{\text{chkpt}} \in \mathbb{R}^{C_A \times C_B}$. To obtain a globally consistent alignment between the two training trajectories, we then solve a third OT problem over checkpoints, reusing the same uniform-marginal formulation as before:
\[
R
\;=\;
\arg\min_{R \in \mathcal{T}(C_A, C_B)}
\;\langle C^{\text{chkpt}}, R \rangle,
\]
and $R_{cd}$ can be interpreted as a soft assignment between checkpoint $c$ in the first model and checkpoint $d$ in the second. For comparison, we also construct a simple pairwise-best checkpoint map that ignores the global OT constraint and matches each checkpoint independently to its lowest-cost partner.

\begin{figure}[H]
\centering
\includegraphics[width=0.95\columnwidth]{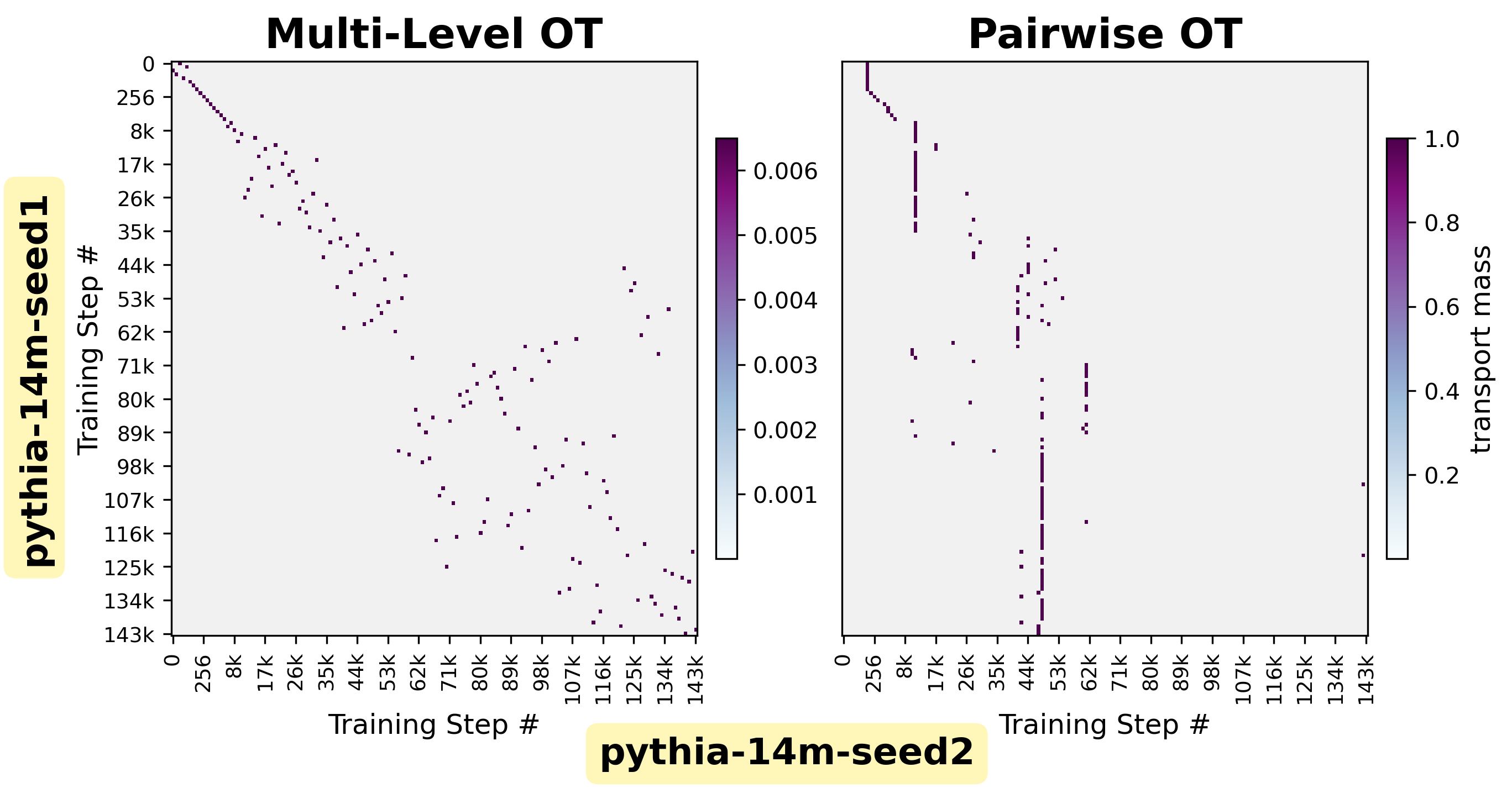}
\caption{\textbf{Transport plans for training-trajectory alignment.} 
Three-level MOT (left) and pairwise OT (right) for Pythia-14M-seed1 $\leftrightarrow$ Pythia-14M-seed2 across training checkpoints. 
MOT yields an, approximately diagonal checkpoint correspondence, whereas pairwise OT shows no clear trajectory-level alignment.}
\label{fig:trainingloops}
\end{figure}

\textbf{Experimental Setup and Results}. We perform three-level OT (and the corresponding pairwise baseline) as a proof of concept for 154 checkpoints of Pythia-14M models with seed 1 and seed 2 (\cite{biderman2023pythia}), where the 154 checkpoints correspond to steps 0 (initialization), 1, 2, 4, 8, 16, 32, 64, 128, 256, 512, 1000, and then every 1,000 subsequent steps. The layer-wise representations for each checkpoint are collected across 2,552 prompts from the STSB dataset ~\citep{huggingface:dataset:stsb_multi_mt, enevoldsen2025mmtebmassivemultilingualtext, muennighoff2022mteb}. We plot the checkpoint-level transport maps generated by three-level MOT and by the pairwise-best alignment in Figure~\ref{fig:trainingloops}. MOT reveals a coherent checkpoint-wise correspondence between the two training trajectories, whereas pairwise matching fails to do so. This demonstrates a proof-of-concept method for aligning entire training trajectories of two models. In principle, the MOT framework can be generalized to any number of levels, depending on the type of alignment setting being studied.

\section{Discussion}  
Multi-Level OT provides a principled framework for aligning two networks with arbitrary depths. By softly coupling all layer objectives, MOT (i) allows a neuron in one layer of a source network to align with a soft combination of units distributed across multiple layers of a target network, (ii) enforces global consistency that mitigates overfitting to noise in any single layer, and (iii) produces a single network-level—rather than layer-level—alignment score. In our experiments, MOT achieves higher alignment scores than greedy pairwise matches and yields intuitive transport plans that reveal hierarchical correspondences both in artificial networks and in the human brain.  

\paragraph{Limitations.} Despite these strengths, several limitations remain. First, MOT is computationally demanding: solving the inner OT scales as $O(n^3 \log n)$ for $n$ neurons in a pair of layers, and the overall procedure is $O(L^2 n^3 \log n)$ for networks with $L$ layers. This makes scaling to very wide or deep models challenging without further algorithmic improvements. Second, our evaluation was limited to a subset of models and brain datasets; follow-up analyses on more models and diverse neural data will be critical to assess generality. Finally, our focus here is on developing a principled method for quantifying representational alignment, rather than explaining why different systems converge on similar representations. The latter is a separate and substantially harder theoretical question. Recent ideas, such as the contravariance principle (which frames convergence as a consequence of shared task or ecological constraints)~\citep{cao2024explanatory2} and the Platonic representation hypothesis (which proposes that training on rich naturalistic data drives different systems toward similar underlying structure in the world)~\citep{huh2024platonic}, offer intriguing conceptual directions but remain challenging to evaluate empirically. By providing a rigorous way to characterize alignment, MOT offers the measurement foundation needed for such explanatory hypotheses to be tested in future work.


\paragraph{Future directions.} Several extensions are natural. One is to study training-run level alignment in more depth, examining how model architecture, training data order, learning rate, and other optimization choices shape the resulting alignment plan. Another is to incorporate priors on the transport plan (e.g. smoothness) to guide alignment toward more interpretable solutions. 
Ultimately, as both biological and artificial networks grow in scale and complexity, methods like MOT that respect global structure while revealing network correspondences will be essential for understanding the universal principles governing intelligent systems.

\newpage
\bibliographystyle{plainnat}
\bibliography{references}

\begin{thebibliography}{51}
\providecommand{\natexlab}[1]{#1}
\providecommand{\url}[1]{\texttt{#1}}
\expandafter\ifx\csname urlstyle\endcsname\relax
  \providecommand{\doi}[1]{doi: #1}\else
  \providecommand{\doi}{doi: \begingroup \urlstyle{rm}\Url}\fi

\bibitem[Abdelhack and Kamitani(2018)]{abdelhack2018sharpening}
Mohamed Abdelhack and Yukiyasu Kamitani.
\newblock Sharpening of hierarchical visual feature representations of blurred images.
\newblock \emph{eneuro}, 5\penalty0 (3), 2018.

\bibitem[Albrecht and Hamilton(1982)]{albrecht1982striate}
Duane~G Albrecht and David~B Hamilton.
\newblock Striate cortex of monkey and cat: contrast response function.
\newblock \emph{Journal of neurophysiology}, 48\penalty0 (1):\penalty0 217--237, 1982.

\bibitem[Allen et~al.(2022)Allen, St-Yves, Wu, Breedlove, Prince, Dowdle, Nau, Caron, Pestilli, Charest, et~al.]{allen2022massive}
Emily~J Allen, Ghislain St-Yves, Yihan Wu, Jesse~L Breedlove, Jacob~S Prince, Logan~T Dowdle, Matthias Nau, Brad Caron, Franco Pestilli, Ian Charest, et~al.
\newblock A massive 7t fmri dataset to bridge cognitive neuroscience and artificial intelligence.
\newblock \emph{Nature neuroscience}, 25\penalty0 (1):\penalty0 116--126, 2022.

\bibitem[Brincat and Connor(2004)]{brincat2004underlying}
Scott~L Brincat and Charles~E Connor.
\newblock Underlying principles of visual shape selectivity in posterior inferotemporal cortex.
\newblock \emph{Nature neuroscience}, 7\penalty0 (8):\penalty0 880--886, 2004.

\bibitem[Cichy et~al.(2016)Cichy, Khosla, Pantazis, Torralba, and Oliva]{cichy2016comparison}
Radoslaw~Martin Cichy, Aditya Khosla, Dimitrios Pantazis, Antonio Torralba, and Aude Oliva.
\newblock Comparison of deep neural networks to spatio-temporal cortical dynamics of human visual object recognition reveals hierarchical correspondence.
\newblock \emph{Scientific reports}, 6\penalty0 (1):\penalty0 27755, 2016.

\bibitem[Conwell et~al.(2022)Conwell, Prince, Kay, Alvarez, and Konkle]{conwell2022can}
Colin Conwell, Jacob~S Prince, Kendrick~N Kay, George~A Alvarez, and Talia Konkle.
\newblock What can 1.8 billion regressions tell us about the pressures shaping high-level visual representation in brains and machines?
\newblock \emph{BioRxiv}, pages 2022--03, 2022.

\bibitem[Dapello et~al.(2022)Dapello, Kar, Schrimpf, Geary, Ferguson, Cox, and DiCarlo]{dapello2022aligning}
Joel Dapello, Kohitij Kar, Martin Schrimpf, Robert Geary, Michael Ferguson, David~D Cox, and James~J DiCarlo.
\newblock Aligning model and macaque inferior temporal cortex representations improves model-to-human behavioral alignment and adversarial robustness.
\newblock \emph{bioRxiv}, pages 2022--07, 2022.

\bibitem[Desimone et~al.(1984)Desimone, Albright, Gross, and Bruce]{desimone1984stimulus}
Robert Desimone, Thomas~D Albright, Charles~G Gross, and Charles Bruce.
\newblock Stimulus-selective properties of inferior temporal neurons in the macaque.
\newblock \emph{Journal of Neuroscience}, 4\penalty0 (8):\penalty0 2051--2062, 1984.

\bibitem[Doerig et~al.(2024)Doerig, Kietzmann, Allen, Wu, Naselaris, Kay, and Charest]{doerig2024Visual}
Adrien Doerig, Tim~C Kietzmann, Emily Allen, Yihan Wu, Thomas Naselaris, Kendrick Kay, and Ian Charest.
\newblock Visual representations in the human brain are aligned with large language models.
\newblock \emph{arXiv preprint arXiv:2209.11737}, 2024.

\bibitem[Eickenberg et~al.(2017)Eickenberg, Gramfort, Varoquaux, and Thirion]{eickenberg2017seeing}
Michael Eickenberg, Alexandre Gramfort, Ga{\"e}l Varoquaux, and Bertrand Thirion.
\newblock Seeing it all: Convolutional network layers map the function of the human visual system.
\newblock \emph{NeuroImage}, 152:\penalty0 184--194, 2017.

\bibitem[Federer et~al.(2020)Federer, Xu, Fyshe, and Zylberberg]{federer2020improved}
Callie Federer, Haoyan Xu, Alona Fyshe, and Joel Zylberberg.
\newblock Improved object recognition using neural networks trained to mimic the brain’s statistical properties.
\newblock \emph{Neural Networks}, 131:\penalty0 103--114, 2020.

\bibitem[Gallant et~al.(1993)Gallant, Braun, and Van~Essen]{gallant1993selectivity}
Jack~L Gallant, Jochen Braun, and David~C Van~Essen.
\newblock Selectivity for polar, hyperbolic, and cartesian gratings in macaque visual cortex.
\newblock \emph{Science}, 259\penalty0 (5091):\penalty0 100--103, 1993.

\bibitem[G{\"u}{\c{c}}l{\"u} and Van~Gerven(2015)]{gucclu2015deep}
Umut G{\"u}{\c{c}}l{\"u} and Marcel~AJ Van~Gerven.
\newblock Deep neural networks reveal a gradient in the complexity of neural representations across the ventral stream.
\newblock \emph{Journal of Neuroscience}, 35\penalty0 (27):\penalty0 10005--10014, 2015.

\bibitem[Horikawa and Kamitani(2017)]{horikawa2017generic}
Tomoyasu Horikawa and Yukiyasu Kamitani.
\newblock Generic decoding of seen and imagined objects using hierarchical visual features.
\newblock \emph{Nature communications}, 8\penalty0 (1):\penalty0 15037, 2017.

\bibitem[Hubel and Wiesel(1962)]{hubel1962receptive}
David~H Hubel and Torsten~N Wiesel.
\newblock Receptive fields, binocular interaction and functional architecture in the cat's visual cortex.
\newblock \emph{The Journal of physiology}, 160\penalty0 (1):\penalty0 106, 1962.

\bibitem[Hubel and Wiesel(1968)]{hubel1968receptive}
David~H Hubel and Torsten~N Wiesel.
\newblock Receptive fields and functional architecture of monkey striate cortex.
\newblock \emph{The Journal of physiology}, 195\penalty0 (1):\penalty0 215--243, 1968.

\bibitem[Jaderberg et~al.(2015)Jaderberg, Simonyan, Zisserman, et~al.]{jaderberg2015spatial}
Max Jaderberg, Karen Simonyan, Andrew Zisserman, et~al.
\newblock Spatial transformer networks.
\newblock \emph{Advances in neural information processing systems}, 28, 2015.

\bibitem[Khaligh-Razavi and Kriegeskorte(2014)]{khaligh2014deep}
Seyed-Mahdi Khaligh-Razavi and Nikolaus Kriegeskorte.
\newblock Deep supervised, but not unsupervised, models may explain it cortical representation.
\newblock \emph{PLoS computational biology}, 10\penalty0 (11):\penalty0 e1003915, 2014.

\bibitem[Khosla and Wehbe(2022)]{khosla2022high}
Meenakshi Khosla and Leila Wehbe.
\newblock High-level visual areas act like domain-general filters with strong selectivity and functional specialization.
\newblock \emph{bioRxiv}, pages 2022--03, 2022.

\bibitem[Khosla et~al.(2022)Khosla, Jamison, Kuceyeski, and Sabuncu]{khosla2022characterizing}
Meenakshi Khosla, Keith Jamison, Amy Kuceyeski, and Mert Sabuncu.
\newblock Characterizing the ventral visual stream with response-optimized neural encoding models.
\newblock \emph{Advances in Neural Information Processing Systems}, 35:\penalty0 9389--9402, 2022.

\bibitem[Klindt et~al.(2017)Klindt, Ecker, Euler, and Bethge]{klindt2017neural}
DA~Klindt, AS~Ecker, T~Euler, and M~Bethge.
\newblock Neural system identification for large 579 populations separating “what” and “where.”.
\newblock \emph{Advances in Neural Information Processing 580 Systems}, 2017.

\bibitem[Kobatake and Tanaka(1994)]{kobatake1994neuronal}
Eucaly Kobatake and Keiji Tanaka.
\newblock Neuronal selectivities to complex object features in the ventral visual pathway of the macaque cerebral cortex.
\newblock \emph{Journal of neurophysiology}, 71\penalty0 (3):\penalty0 856--867, 1994.

\bibitem[Kobatake et~al.(1998)Kobatake, Wang, and Tanaka]{kobatake1998effects}
Eucaly Kobatake, Gang Wang, and Keiji Tanaka.
\newblock Effects of shape-discrimination training on the selectivity of inferotemporal cells in adult monkeys.
\newblock \emph{Journal of neurophysiology}, 80\penalty0 (1):\penalty0 324--330, 1998.

\bibitem[Kriegeskorte et~al.(2008)Kriegeskorte, Mur, Ruff, Kiani, Bodurka, Esteky, Tanaka, and Bandettini]{kriegeskorte2008matching}
Nikolaus Kriegeskorte, Marieke Mur, Douglas~A Ruff, Roozbeh Kiani, Jerzy Bodurka, Hossein Esteky, Keiji Tanaka, and Peter~A Bandettini.
\newblock Matching categorical object representations in inferior temporal cortex of man and monkey.
\newblock \emph{Neuron}, 60\penalty0 (6):\penalty0 1126--1141, 2008.

\bibitem[Lin et~al.(2014)Lin, Maire, Belongie, Hays, Perona, Ramanan, Doll{\'a}r, and Zitnick]{lin2014microsoft}
Tsung-Yi Lin, Michael Maire, Serge Belongie, James Hays, Pietro Perona, Deva Ramanan, Piotr Doll{\'a}r, and C~Lawrence Zitnick.
\newblock Microsoft coco: Common objects in context.
\newblock In \emph{Computer Vision--ECCV 2014: 13th European Conference, Zurich, Switzerland, September 6-12, 2014, Proceedings, Part V 13}, pages 740--755. Springer, 2014.

\bibitem[Livingstone and Hubel(1984)]{livingstone1984anatomy}
Margaret~S Livingstone and David~H Hubel.
\newblock Anatomy and physiology of a color system in the primate visual cortex.
\newblock \emph{Journal of Neuroscience}, 4\penalty0 (1):\penalty0 309--356, 1984.

\bibitem[Lurz et~al.(2020)Lurz, Bashiri, Willeke, Jagadish, Wang, Walker, Cadena, Muhammad, Cobos, Tolias, et~al.]{lurz2020generalization}
Konstantin-Klemens Lurz, Mohammad Bashiri, Konstantin Willeke, Akshay~K Jagadish, Eric Wang, Edgar~Y Walker, Santiago~A Cadena, Taliah Muhammad, Erick Cobos, Andreas~S Tolias, et~al.
\newblock Generalization in data-driven models of primary visual cortex.
\newblock \emph{BioRxiv}, pages 2020--10, 2020.

\bibitem[Miyashita(1988)]{miyashita1988neuronal}
Yasushi Miyashita.
\newblock Neuronal correlate of visual associative long-term memory in the primate temporal cortex.
\newblock \emph{Nature}, 335\penalty0 (6193):\penalty0 817--820, 1988.

\bibitem[Moran and Desimone(1985)]{moran1985selective}
Jeffrey Moran and Robert Desimone.
\newblock Selective attention gates visual processing in the extrastriate cortex.
\newblock \emph{Science}, 229\penalty0 (4715):\penalty0 782--784, 1985.

\bibitem[Pasupathy and Connor(1999)]{pasupathy1999responses}
Anitha Pasupathy and Charles~E Connor.
\newblock Responses to contour features in macaque area v4.
\newblock \emph{Journal of neurophysiology}, 82\penalty0 (5):\penalty0 2490--2502, 1999.

\bibitem[Pasupathy and Connor(2001)]{pasupathy2001shape}
Anitha Pasupathy and Charles~E Connor.
\newblock Shape representation in area v4: position-specific tuning for boundary conformation.
\newblock \emph{Journal of neurophysiology}, 2001.

\bibitem[Pasupathy and Connor(2002)]{pasupathy2002population}
Anitha Pasupathy and Charles~E Connor.
\newblock Population coding of shape in area v4.
\newblock \emph{Nature neuroscience}, 5\penalty0 (12):\penalty0 1332--1338, 2002.

\bibitem[Radford et~al.(2021)Radford, Kim, Hallacy, Ramesh, Goh, Agarwal, Sastry, Askell, Mishkin, Clark, et~al.]{radford2021learning}
Alec Radford, Jong~Wook Kim, Chris Hallacy, Aditya Ramesh, Gabriel Goh, Sandhini Agarwal, Girish Sastry, Amanda Askell, Pamela Mishkin, Jack Clark, et~al.
\newblock Learning transferable visual models from natural language supervision.
\newblock In \emph{International conference on machine learning}, pages 8748--8763. PMLR, 2021.

\bibitem[Rust and DiCarlo(2010)]{rust2010selectivity}
Nicole~C Rust and James~J DiCarlo.
\newblock Selectivity and tolerance (“invariance”) both increase as visual information propagates from cortical area v4 to it.
\newblock \emph{Journal of Neuroscience}, 30\penalty0 (39):\penalty0 12978--12995, 2010.

\bibitem[Safarani et~al.(2021)Safarani, Nix, Willeke, Cadena, Restivo, Denfield, Tolias, and Sinz]{safarani2021towards}
Shahd Safarani, Arne Nix, Konstantin Willeke, Santiago Cadena, Kelli Restivo, George Denfield, Andreas Tolias, and Fabian Sinz.
\newblock Towards robust vision by multi-task learning on monkey visual cortex.
\newblock \emph{Advances in Neural Information Processing Systems}, 34:\penalty0 739--751, 2021.

\bibitem[Schrimpf et~al.(2020)Schrimpf, Kubilius, Lee, Murty, Ajemian, and DiCarlo]{schrimpf2020integrative}
Martin Schrimpf, Jonas Kubilius, Michael~J Lee, N~Apurva~Ratan Murty, Robert Ajemian, and James~J DiCarlo.
\newblock Integrative benchmarking to advance neurally mechanistic models of human intelligence.
\newblock \emph{Neuron}, 108\penalty0 (3):\penalty0 413--423, 2020.

\bibitem[Schwartz et~al.(2019)Schwartz, Toneva, and Wehbe]{schwartz2019inducing}
Dan Schwartz, Mariya Toneva, and Leila Wehbe.
\newblock Inducing brain-relevant bias in natural language processing models.
\newblock \emph{Advances in neural information processing systems}, 32, 2019.

\bibitem[Seeliger et~al.(2021)Seeliger, Ambrogioni, G{\"u}{\c{c}}l{\"u}t{\"u}rk, van~den Bulk, G{\"u}{\c{c}}l{\"u}, and van Gerven]{seeliger2021end}
Katja Seeliger, Luca Ambrogioni, Ya{\u{g}}mur G{\"u}{\c{c}}l{\"u}t{\"u}rk, Leonieke~M van~den Bulk, Umut G{\"u}{\c{c}}l{\"u}, and Marcel~AJ van Gerven.
\newblock End-to-end neural system identification with neural information flow.
\newblock \emph{PLOS Computational Biology}, 17\penalty0 (2):\penalty0 e1008558, 2021.

\bibitem[Song et~al.(2020)Song, Tan, Qin, Lu, and Liu]{song2020mpnet}
Kaitao Song, Xu~Tan, Tao Qin, Jianfeng Lu, and Tie-Yan Liu.
\newblock Mpnet: Masked and permuted pre-training for language understanding.
\newblock \emph{Advances in neural information processing systems}, 33:\penalty0 16857--16867, 2020.

\bibitem[St-Yves et~al.(2023)St-Yves, Allen, Wu, Kay, and Naselaris]{st2023brain}
Ghislain St-Yves, Emily~J Allen, Yihan Wu, Kendrick Kay, and Thomas Naselaris.
\newblock Brain-optimized deep neural network models of human visual areas learn non-hierarchical representations.
\newblock \emph{Nature communications}, 14\penalty0 (1):\penalty0 3329, 2023.

\bibitem[Storrs et~al.(2021)Storrs, Kietzmann, Walther, Mehrer, and Kriegeskorte]{storrs2021diverse}
Katherine~R Storrs, Tim~C Kietzmann, Alexander Walther, Johannes Mehrer, and Nikolaus Kriegeskorte.
\newblock Diverse deep neural networks all predict human inferior temporal cortex well, after training and fitting.
\newblock \emph{Journal of cognitive neuroscience}, 33\penalty0 (10):\penalty0 2044--2064, 2021.

\bibitem[Tanaka et~al.(1991)Tanaka, Saito, Fukada, and Moriya]{tanaka1991coding}
Keiji Tanaka, Hide-aki Saito, Yoshiro Fukada, and Madoka Moriya.
\newblock Coding visual images of objects in the inferotemporal cortex of the macaque monkey.
\newblock \emph{Journal of neurophysiology}, 66\penalty0 (1):\penalty0 170--189, 1991.

\bibitem[Tang et~al.(2024)Tang, Du, Vo, Lal, and Huth]{tang2024brain}
Jerry Tang, Meng Du, Vy~Vo, Vasudev Lal, and Alexander Huth.
\newblock Brain encoding models based on multimodal transformers can transfer across language and vision.
\newblock \emph{Advances in Neural Information Processing Systems}, 36, 2024.

\bibitem[Tsunoda et~al.(2001)Tsunoda, Yamane, Nishizaki, and Tanifuji]{tsunoda2001complex}
Kazushige Tsunoda, Yukako Yamane, Makoto Nishizaki, and Manabu Tanifuji.
\newblock Complex objects are represented in macaque inferotemporal cortex by the combination of feature columns.
\newblock \emph{Nature neuroscience}, 4\penalty0 (8):\penalty0 832--838, 2001.

\bibitem[Wang et~al.(2022)Wang, Kay, Naselaris, Tarr, and Wehbe]{wang2022incorporating}
Aria~Y Wang, Kendrick Kay, Thomas Naselaris, Michael~J Tarr, and Leila Wehbe.
\newblock Incorporating natural language into vision models improves prediction and understanding of higher visual cortex.
\newblock \emph{BioRxiv}, pages 2022--09, 2022.

\bibitem[Weiler and Cesa(2019)]{weiler2019general}
Maurice Weiler and Gabriele Cesa.
\newblock General e (2)-equivariant steerable cnns.
\newblock \emph{Advances in neural information processing systems}, 32, 2019.

\bibitem[Wen et~al.(2018)Wen, Shi, Zhang, Lu, Cao, and Liu]{wen2018neural}
Haiguang Wen, Junxing Shi, Yizhen Zhang, Kun-Han Lu, Jiayue Cao, and Zhongming Liu.
\newblock Neural encoding and decoding with deep learning for dynamic natural vision.
\newblock \emph{Cerebral cortex}, 28\penalty0 (12):\penalty0 4136--4160, 2018.

\bibitem[Yamins et~al.(2014)Yamins, Hong, Cadieu, Solomon, Seibert, and DiCarlo]{yamins2014performance}
Daniel~LK Yamins, Ha~Hong, Charles~F Cadieu, Ethan~A Solomon, Darren Seibert, and James~J DiCarlo.
\newblock Performance-optimized hierarchical models predict neural responses in higher visual cortex.
\newblock \emph{Proceedings of the national academy of sciences}, 111\penalty0 (23):\penalty0 8619--8624, 2014.

\bibitem[Yang et~al.(2023)Yang, Gao, L{\'\i}u, Xiao, Zhao, and Qin]{yang2023unimo}
Hao Yang, Can Gao, Hao L{\'\i}u, Xinyan Xiao, Yanyan Zhao, and Bing Qin.
\newblock Unimo-3: Multi-granularity interaction for vision-language representation learning.
\newblock \emph{arXiv preprint arXiv:2305.13697}, 2023.

\bibitem[Yue et~al.(2020)Yue, Robert, and Ungerleider]{yue2020curvature}
Xiaomin Yue, Sophia Robert, and Leslie~G Ungerleider.
\newblock Curvature processing in human visual cortical areas.
\newblock \emph{NeuroImage}, 222:\penalty0 117295, 2020.

\bibitem[Zeki(1973)]{zeki1973colour}
Semir~M Zeki.
\newblock Colour coding in rhesus monkey prestriate cortex.
\newblock \emph{Brain research}, 53\penalty0 (2):\penalty0 422--427, 1973.

\end{thebibliography}


\begin{thebibliography}{35}
\providecommand{\natexlab}[1]{#1}
\providecommand{\url}[1]{\texttt{#1}}
\expandafter\ifx\csname urlstyle\endcsname\relax
  \providecommand{\doi}[1]{doi: #1}\else
  \providecommand{\doi}{doi: \begingroup \urlstyle{rm}\Url}\fi

\bibitem[Allen et~al.(2022)Allen, St-Yves, Wu, Breedlove, Prince, Dowdle, Nau, Caron, Pestilli, Charest, et~al.]{allen2022massive}
Emily~J Allen, Ghislain St-Yves, Yihan Wu, Jesse~L Breedlove, Jacob~S Prince, Logan~T Dowdle, Matthias Nau, Brad Caron, Franco Pestilli, Ian Charest, et~al.
\newblock A massive 7t fmri dataset to bridge cognitive neuroscience and artificial intelligence.
\newblock \emph{Nature neuroscience}, 25\penalty0 (1):\penalty0 116--126, 2022.

\bibitem[Bansal et~al.(2021)Bansal, Nakkiran, and Barak]{bansal2021revisiting}
Yamini Bansal, Preetum Nakkiran, and Boaz Barak.
\newblock Revisiting model stitching to compare neural representations.
\newblock \emph{Advances in neural information processing systems}, 34:\penalty0 225--236, 2021.

\bibitem[Biderman et~al.(2023)Biderman, Schoelkopf, Anthony, Bradley, O’Brien, Hallahan, Khan, Purohit, Prashanth, Raff, et~al.]{biderman2023pythia}
Stella Biderman, Hailey Schoelkopf, Quentin~Gregory Anthony, Herbie Bradley, Kyle O’Brien, Eric Hallahan, Mohammad~Aflah Khan, Shivanshu Purohit, USVSN~Sai Prashanth, Edward Raff, et~al.
\newblock Pythia: A suite for analyzing large language models across training and scaling.
\newblock In \emph{International Conference on Machine Learning}, pages 2397--2430. PMLR, 2023.

\bibitem[Cao and Yamins(2024)]{cao2024explanatory2}
Rosa Cao and Daniel Yamins.
\newblock Explanatory models in neuroscience, part 2: Functional intelligibility and the contravariance principle.
\newblock \emph{Cognitive Systems Research}, 85:\penalty0 101200, 2024.

\bibitem[Chen and Bonner(2025)]{chen2025universal}
Zirui Chen and Michael~F Bonner.
\newblock Universal dimensions of visual representation.
\newblock \emph{Science Advances}, 11\penalty0 (27):\penalty0 eadw7697, 2025.

\bibitem[Cichy et~al.(2016)Cichy, Khosla, Pantazis, Torralba, and Oliva]{cichy2016comparison}
Radoslaw~Martin Cichy, Aditya Khosla, Dimitrios Pantazis, Antonio Torralba, and Aude Oliva.
\newblock Comparison of deep neural networks to spatio-temporal cortical dynamics of human visual object recognition reveals hierarchical correspondence.
\newblock \emph{Scientific reports}, 6\penalty0 (1):\penalty0 27755, 2016.

\bibitem[De~Loera and Kim(2013)]{de2013combinatorics}
Jes{\'u}s~A De~Loera and Edward~D Kim.
\newblock Combinatorics and geometry of transportation polytopes: An update.
\newblock \emph{Discrete geometry and algebraic combinatorics}, 625:\penalty0 37--76, 2013.

\bibitem[Deng et~al.(2009)Deng, Dong, Socher, Li, Li, and Fei-Fei]{5206848}
Jia Deng, Wei Dong, Richard Socher, Li-Jia Li, Kai Li, and Li~Fei-Fei.
\newblock Imagenet: A large-scale hierarchical image database.
\newblock In \emph{2009 IEEE Conference on Computer Vision and Pattern Recognition}, pages 248--255, 2009.
\newblock \doi{10.1109/CVPR.2009.5206848}.

\bibitem[Desimone and Schein(1987)]{desimone1987visual}
Robert Desimone and Stanley~J Schein.
\newblock Visual properties of neurons in area v4 of the macaque: sensitivity to stimulus form.
\newblock \emph{Journal of neurophysiology}, 57\penalty0 (3):\penalty0 835--868, 1987.

\bibitem[Desimone et~al.(1984)Desimone, Albright, Gross, and Bruce]{desimone1984stimulus}
Robert Desimone, Thomas~D Albright, Charles~G Gross, and Charles Bruce.
\newblock Stimulus-selective properties of inferior temporal neurons in the macaque.
\newblock \emph{Journal of Neuroscience}, 4\penalty0 (8):\penalty0 2051--2062, 1984.

\bibitem[Dravid et~al.(2023)Dravid, Gandelsman, Efros, and Shocher]{dravid2023rosetta}
Amil Dravid, Yossi Gandelsman, Alexei~A Efros, and Assaf Shocher.
\newblock Rosetta neurons: Mining the common units in a model zoo.
\newblock In \emph{Proceedings of the IEEE/CVF International Conference on Computer Vision}, pages 1934--1943, 2023.

\bibitem[Eickenberg et~al.(2017)Eickenberg, Gramfort, Varoquaux, and Thirion]{eickenberg2017seeing}
Michael Eickenberg, Alexandre Gramfort, Ga{\"e}l Varoquaux, and Bertrand Thirion.
\newblock Seeing it all: Convolutional network layers map the function of the human visual system.
\newblock \emph{NeuroImage}, 152:\penalty0 184--194, 2017.

\bibitem[Enevoldsen et~al.(2025)Enevoldsen, Chung, Kerboua, Kardos, Mathur, Stap, Gala, Siblini, Krzemiński, Winata, Sturua, Utpala, Ciancone, Schaeffer, Sequeira, Misra, Dhakal, Rystrøm, Solomatin, Ömer Çağatan, Kundu, Bernstorff, Xiao, Sukhlecha, Pahwa, Poświata, GV, Ashraf, Auras, Plüster, Harries, Magne, Mohr, Hendriksen, Zhu, Gisserot-Boukhlef, Aarsen, Kostkan, Wojtasik, Lee, Šuppa, Zhang, Rocca, Hamdy, Michail, Yang, Faysse, Vatolin, Thakur, Dey, Vasani, Chitale, Tedeschi, Tai, Snegirev, Günther, Xia, Shi, Lù, Clive, Krishnakumar, Maksimova, Wehrli, Tikhonova, Panchal, Abramov, Ostendorff, Liu, Clematide, Miranda, Fenogenova, Song, Safi, Li, Borghini, Cassano, Su, Lin, Yen, Hansen, Hooker, Xiao, Adlakha, Weller, Reddy, and Muennighoff]{enevoldsen2025mmtebmassivemultilingualtext}
Kenneth Enevoldsen, Isaac Chung, Imene Kerboua, Márton Kardos, Ashwin Mathur, David Stap, Jay Gala, Wissam Siblini, Dominik Krzemiński, Genta~Indra Winata, Saba Sturua, Saiteja Utpala, Mathieu Ciancone, Marion Schaeffer, Gabriel Sequeira, Diganta Misra, Shreeya Dhakal, Jonathan Rystrøm, Roman Solomatin, Ömer Çağatan, Akash Kundu, Martin Bernstorff, Shitao Xiao, Akshita Sukhlecha, Bhavish Pahwa, Rafał Poświata, Kranthi~Kiran GV, Shawon Ashraf, Daniel Auras, Björn Plüster, Jan~Philipp Harries, Loïc Magne, Isabelle Mohr, Mariya Hendriksen, Dawei Zhu, Hippolyte Gisserot-Boukhlef, Tom Aarsen, Jan Kostkan, Konrad Wojtasik, Taemin Lee, Marek Šuppa, Crystina Zhang, Roberta Rocca, Mohammed Hamdy, Andrianos Michail, John Yang, Manuel Faysse, Aleksei Vatolin, Nandan Thakur, Manan Dey, Dipam Vasani, Pranjal Chitale, Simone Tedeschi, Nguyen Tai, Artem Snegirev, Michael Günther, Mengzhou Xia, Weijia Shi, Xing~Han Lù, Jordan Clive, Gayatri Krishnakumar, Anna Maksimova, Silvan Wehrli, Maria Tikhonova, Henil
  Panchal, Aleksandr Abramov, Malte Ostendorff, Zheng Liu, Simon Clematide, Lester~James Miranda, Alena Fenogenova, Guangyu Song, Ruqiya~Bin Safi, Wen-Ding Li, Alessia Borghini, Federico Cassano, Hongjin Su, Jimmy Lin, Howard Yen, Lasse Hansen, Sara Hooker, Chenghao Xiao, Vaibhav Adlakha, Orion Weller, Siva Reddy, and Niklas Muennighoff.
\newblock Mmteb: Massive multilingual text embedding benchmark.
\newblock \emph{arXiv preprint arXiv:2502.13595}, 2025.
\newblock \doi{10.48550/arXiv.2502.13595}.
\newblock URL \url{https://arxiv.org/abs/2502.13595}.

\bibitem[Grattafiori et~al.(2024)]{grattafiori2024llama3herdmodels}
Aaron Grattafiori et~al.
\newblock The llama 3 herd of models, 2024.
\newblock URL \url{https://arxiv.org/abs/2407.21783}.

\bibitem[G{\"u}{\c{c}}l{\"u} and Van~Gerven(2015)]{gucclu2015deep}
Umut G{\"u}{\c{c}}l{\"u} and Marcel~AJ Van~Gerven.
\newblock Deep neural networks reveal a gradient in the complexity of neural representations across the ventral stream.
\newblock \emph{Journal of Neuroscience}, 35\penalty0 (27):\penalty0 10005--10014, 2015.

\bibitem[Hosseini et~al.(2024)Hosseini, Casto, Zaslavsky, Conwell, Richardson, and Fedorenko]{hosseini2024universality}
Eghbal Hosseini, Colton Casto, Noga Zaslavsky, Colin Conwell, Mark Richardson, and Evelina Fedorenko.
\newblock Universality of representation in biological and artificial neural networks.
\newblock \emph{bioRxiv}, 2024.

\bibitem[Hubel and Wiesel(1968)]{hubel1968receptive}
David~H Hubel and Torsten~N Wiesel.
\newblock Receptive fields and functional architecture of monkey striate cortex.
\newblock \emph{The Journal of physiology}, 195\penalty0 (1):\penalty0 215--243, 1968.

\bibitem[Huh et~al.(2024)Huh, Cheung, Wang, and Isola]{huh2024platonic}
Minyoung Huh, Brian Cheung, Tongzhou Wang, and Phillip Isola.
\newblock The platonic representation hypothesis.
\newblock \emph{arXiv preprint arXiv:2405.07987}, 2024.

\bibitem[Kell et~al.(2018)Kell, Yamins, Shook, Norman-Haignere, and McDermott]{kell2018task}
Alexander~JE Kell, Daniel~LK Yamins, Erica~N Shook, Sam~V Norman-Haignere, and Josh~H McDermott.
\newblock A task-optimized neural network replicates human auditory behavior, predicts brain responses, and reveals a cortical processing hierarchy.
\newblock \emph{Neuron}, 98\penalty0 (3):\penalty0 630--644, 2018.

\bibitem[Khaligh-Razavi and Kriegeskorte(2014)]{KriegeskorteRSA}
Seyed-Mehdi Khaligh-Razavi and Nikolaus Kriegeskorte.
\newblock Deep supervised, but not unsupervised, models may explain it cortical representation.
\newblock \emph{PLOS Computational Biology}, 2014.
\newblock URL \url{https://doi.org/10.1371/journal.pcbi.1003915}.

\bibitem[Khosla and Williams(2024)]{khosla2024soft}
Meenakshi Khosla and Alex~H Williams.
\newblock Soft matching distance: A metric on neural representations that captures single-neuron tuning.
\newblock In \emph{Proceedings of UniReps: the First Workshop on Unifying Representations in Neural Models}, pages 326--341. PMLR, 2024.

\bibitem[Khosla et~al.(2024)Khosla, Williams, McDermott, and Kanwisher]{khosla2024privileged}
Meenakshi Khosla, Alex~H Williams, Josh McDermott, and Nancy Kanwisher.
\newblock Privileged representational axes in biological and artificial neural networks.
\newblock \emph{bioRxiv}, pages 2024--06, 2024.

\bibitem[Kornblith et~al.(2019)Kornblith, Norouzi, Lee, and Hinton]{kornblith2019similarity}
Simon Kornblith, Mohammad Norouzi, Honglak Lee, and Geoffrey Hinton.
\newblock Similarity of neural network representations revisited.
\newblock In \emph{International conference on machine learning}, pages 3519--3529. PMLR, 2019.

\bibitem[Kriegeskorte et~al.(2008)Kriegeskorte, Mur, and Bandettini]{kriegeskorte2008representational}
Nikolaus Kriegeskorte, Marieke Mur, and Peter~A Bandettini.
\newblock Representational similarity analysis-connecting the branches of systems neuroscience.
\newblock \emph{Frontiers in systems neuroscience}, 2:\penalty0 249, 2008.

\bibitem[May(2021)]{huggingface:dataset:stsb_multi_mt}
Philip May.
\newblock Machine translated multilingual sts benchmark dataset.
\newblock 2021.
\newblock URL \url{https://github.com/PhilipMay/stsb-multi-mt}.

\bibitem[Muennighoff et~al.(2022)Muennighoff, Tazi, Magne, and Reimers]{muennighoff2022mteb}
Niklas Muennighoff, Nouamane Tazi, Lo{\"\i}c Magne, and Nils Reimers.
\newblock Mteb: Massive text embedding benchmark.
\newblock \emph{arXiv preprint arXiv:2210.07316}, 2022.
\newblock \doi{10.48550/ARXIV.2210.07316}.
\newblock URL \url{https://arxiv.org/abs/2210.07316}.

\bibitem[Pasupathy and Connor(2001)]{pasupathy2001shape}
Anitha Pasupathy and Charles~E Connor.
\newblock Shape representation in area v4: position-specific tuning for boundary conformation.
\newblock \emph{Journal of neurophysiology}, 86\penalty0 (5):\penalty0 2505--2519, 2001.

\bibitem[Qwen et~al.(2025)Qwen, :, Yang, Yang, Zhang, Hui, Zheng, Yu, Li, Liu, Huang, Wei, Lin, Yang, Tu, Zhang, Yang, Yang, Zhou, Lin, Dang, Lu, Bao, Yang, Yu, Li, Xue, Zhang, Zhu, Men, Lin, Li, Tang, Xia, Ren, Ren, Fan, Su, Zhang, Wan, Liu, Cui, Zhang, and Qiu]{qwen2025qwen25technicalreport}
Qwen, :, An~Yang, Baosong Yang, Beichen Zhang, Binyuan Hui, Bo~Zheng, Bowen Yu, Chengyuan Li, Dayiheng Liu, Fei Huang, Haoran Wei, Huan Lin, Jian Yang, Jianhong Tu, Jianwei Zhang, Jianxin Yang, Jiaxi Yang, Jingren Zhou, Junyang Lin, Kai Dang, Keming Lu, Keqin Bao, Kexin Yang, Le~Yu, Mei Li, Mingfeng Xue, Pei Zhang, Qin Zhu, Rui Men, Runji Lin, Tianhao Li, Tianyi Tang, Tingyu Xia, Xingzhang Ren, Xuancheng Ren, Yang Fan, Yang Su, Yichang Zhang, Yu~Wan, Yuqiong Liu, Zeyu Cui, Zhenru Zhang, and Zihan Qiu.
\newblock Qwen2.5 technical report, 2025.
\newblock URL \url{https://arxiv.org/abs/2412.15115}.

\bibitem[Russakovsky et~al.(2015)Russakovsky, Deng, Su, Krause, Satheesh, Ma, Huang, Karpathy, Khosla, Bernstein, Berg, and Fei-Fei]{ILSVRC15}
Olga Russakovsky, Jia Deng, Hao Su, Jonathan Krause, Sanjeev Satheesh, Sean Ma, Zhiheng Huang, Andrej Karpathy, Aditya Khosla, Michael Bernstein, Alexander~C. Berg, and Li~Fei-Fei.
\newblock {ImageNet Large Scale Visual Recognition Challenge}.
\newblock \emph{International Journal of Computer Vision (IJCV)}, 115\penalty0 (3):\penalty0 211--252, 2015.
\newblock \doi{10.1007/s11263-015-0816-y}.

\bibitem[Schrimpf et~al.(2018)Schrimpf, Kubilius, Hong, Majaj, Rajalingham, Issa, Kar, Bashivan, Prescott-Roy, Geiger, et~al.]{schrimpf2018brain}
Martin Schrimpf, Jonas Kubilius, Ha~Hong, Najib~J Majaj, Rishi Rajalingham, Elias~B Issa, Kohitij Kar, Pouya Bashivan, Jonathan Prescott-Roy, Franziska Geiger, et~al.
\newblock Brain-score: Which artificial neural network for object recognition is most brain-like?
\newblock \emph{BioRxiv}, page 407007, 2018.

\bibitem[Schrimpf et~al.(2020)Schrimpf, Kubilius, Lee, Murty, Ajemian, and DiCarlo]{schrimpf2020integrative}
Martin Schrimpf, Jonas Kubilius, Michael~J Lee, N~Apurva~Ratan Murty, Robert Ajemian, and James~J DiCarlo.
\newblock Integrative benchmarking to advance neurally mechanistic models of human intelligence.
\newblock \emph{Neuron}, 2020.
\newblock URL \url{https://www.cell.com/neuron/fulltext/S0896-6273(20)30605-X}.

\bibitem[Storrs et~al.(2021)Storrs, Kietzmann, Walther, Mehrer, and Kriegeskorte]{storrs2021diverse}
Katherine~R Storrs, Tim~C Kietzmann, Alexander Walther, Johannes Mehrer, and Nikolaus Kriegeskorte.
\newblock Diverse deep neural networks all predict human inferior temporal cortex well, after training and fitting.
\newblock \emph{Journal of cognitive neuroscience}, 33\penalty0 (10):\penalty0 2044--2064, 2021.

\bibitem[von Neumann(1953)]{vonNeumann}
John von Neumann.
\newblock \emph{A Certain Zero-sum Two-person Game Equivalent to the Optimal Assignment Problem}, pages 5--12.
\newblock Princeton University Press, Princeton, 1953.

\bibitem[Williams et~al.(2021)Williams, Kunz, Kornblith, and Linderman]{williams2021generalized}
Alex~H Williams, Erin Kunz, Simon Kornblith, and Scott Linderman.
\newblock Generalized shape metrics on neural representations.
\newblock \emph{Advances in Neural Information Processing Systems}, 34:\penalty0 4738--4750, 2021.

\bibitem[Yamins et~al.(2014)Yamins, Hong, Cadieu, Solomon, Seibert, and DiCarlo]{yamins2014performance}
Daniel~LK Yamins, Ha~Hong, Charles~F Cadieu, Ethan~A Solomon, Darren Seibert, and James~J DiCarlo.
\newblock Performance-optimized hierarchical models predict neural responses in higher visual cortex.
\newblock \emph{Proceedings of the national academy of sciences}, 111\penalty0 (23):\penalty0 8619--8624, 2014.

\end{thebibliography}
\clearpage
\appendix

\counterwithin{figure}{section}
\counterwithin{table}{section}
\counterwithin{equation}{section}

\section*{Appendix}

\section{Optimal Transport for Representational Alignment}
\label{sec:OTExplanation}
Optimal Transport (OT) provides a principled way to compare two neural populations by framing each representation as a probability distribution over its tuning functions and measuring the minimal “effort’’ required to transform one representational distribution into another. Concretely, if $X \in \mathbb{R}^{T \times N_x}$ and $Y \in \mathbb{R}^{T \times N_y}$ denote two sets of tuning curves across $T$ stimuli, we represent each population as a uniform mixture of Dirac masses over its neurons:
\[
\mu_X = \frac{1}{N_x}\sum_{i=1}^{N_x} \delta_{x_i}, \qquad 
\mu_Y = \frac{1}{N_y}\sum_{j=1}^{N_y} \delta_{y_j}.
\]
The cost function $C_{ij} = c(x_i, y_j)$ quantifies the dissimilarity between the tuning functions of neuron $i$ in $X$ and neuron $j$ in $Y$; depending on the application, $c(\cdot,\cdot)$ may be squared Euclidean distance, correlation distance, or any appropriate tuning-based dissimilarity. The 2-Wasserstein or the soft-matching~(\cite{khosla2024soft}) distance between $\mu_X$ and $\mu_Y$ is
\[
d_T(X,Y)
= \min_{P \in \mathcal{T}(N_x,N_y)} 
\sum_{ij} P_{ij}\, C_{ij},
\]
where $P$ ranges over the \emph{transportation polytope} 
$\mathcal{T}(N_x, N_y)
= \Big\{
P \in \mathbb{R}^{N_x \times N_y} :
P_{ij} \ge 0,\;
\sum_i P_{ij} = \tfrac{1}{N_y},\;
\sum_j P_{ij} = \tfrac{1}{N_x}
\Big\}.
$
Any $P \in \mathcal{T}(N_x,N_y)$ is a \emph{doubly stochastic coupling}: it assigns a non-negative “mass’’ $P_{ij}$ to transporting representational weight from neuron $i$ in $X$ to neuron $j$ in $Y$, with uniform row and column marginals. Unlike rotation-invariant metrics such as RSA or CKA, OT preserves single-neuron tuning structure and yields explicit, interpretable neuron-to-neuron correspondences.

MOT builds directly on this OT foundation by elevating the coupling to the level of layers, simultaneously inferring soft layer-to-layer couplings and per-neuron transport plans to produce a globally balanced alignment across networks of arbitrary depths (Figure~\ref{fig:mainfigure}).

\newpage
\section{Comparing LLM-LLM Representations}
\begin{figure}[H]
\centering
\includegraphics[width=0.48\columnwidth]{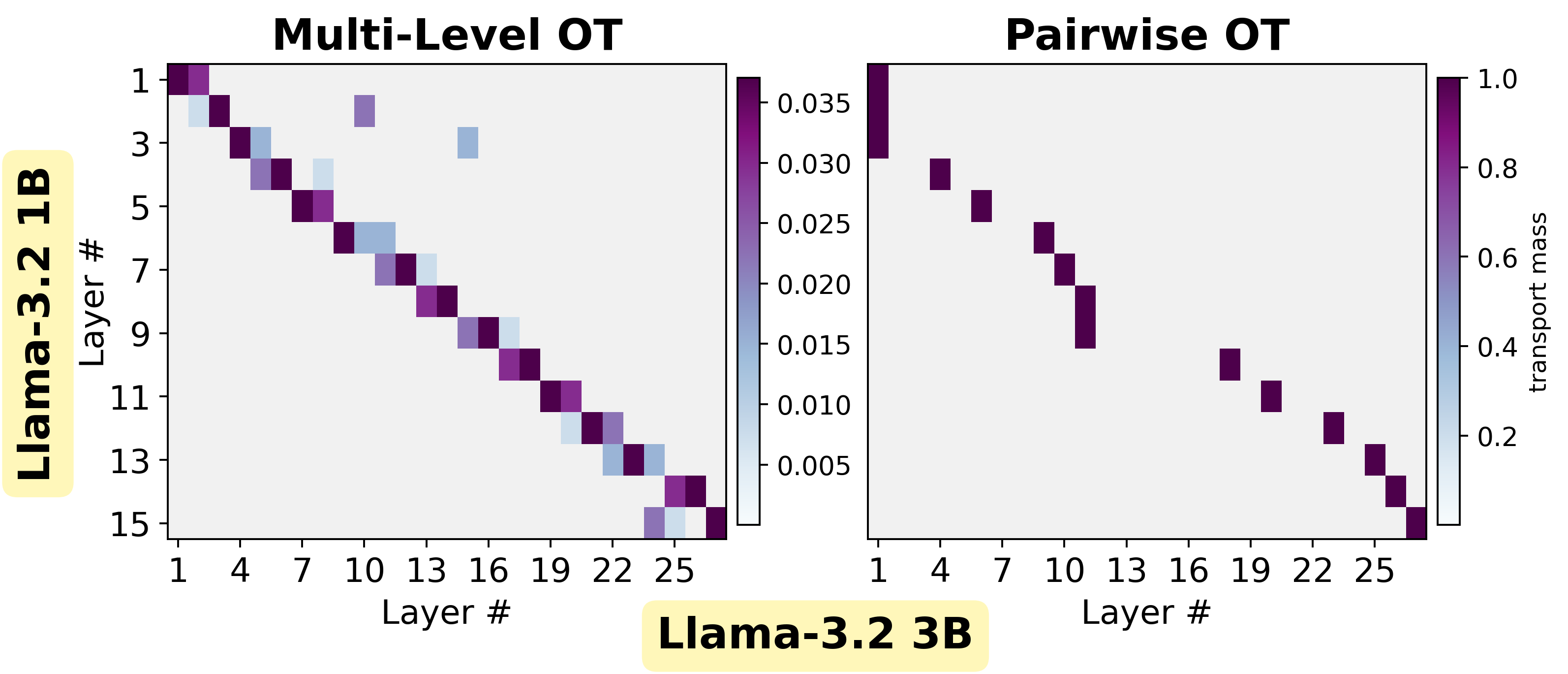}\hfill
\includegraphics[width=0.48\columnwidth]{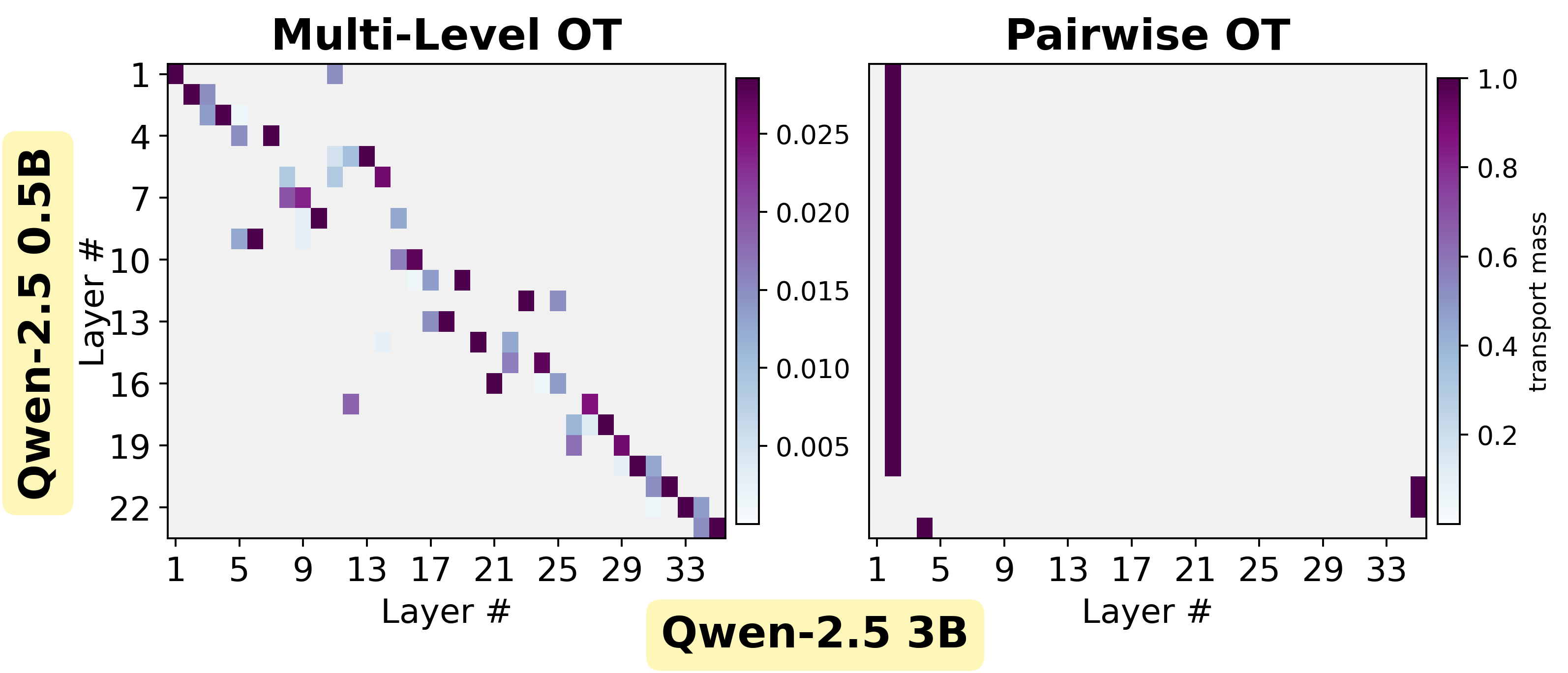}
\medskip
\includegraphics[width=0.48\columnwidth]{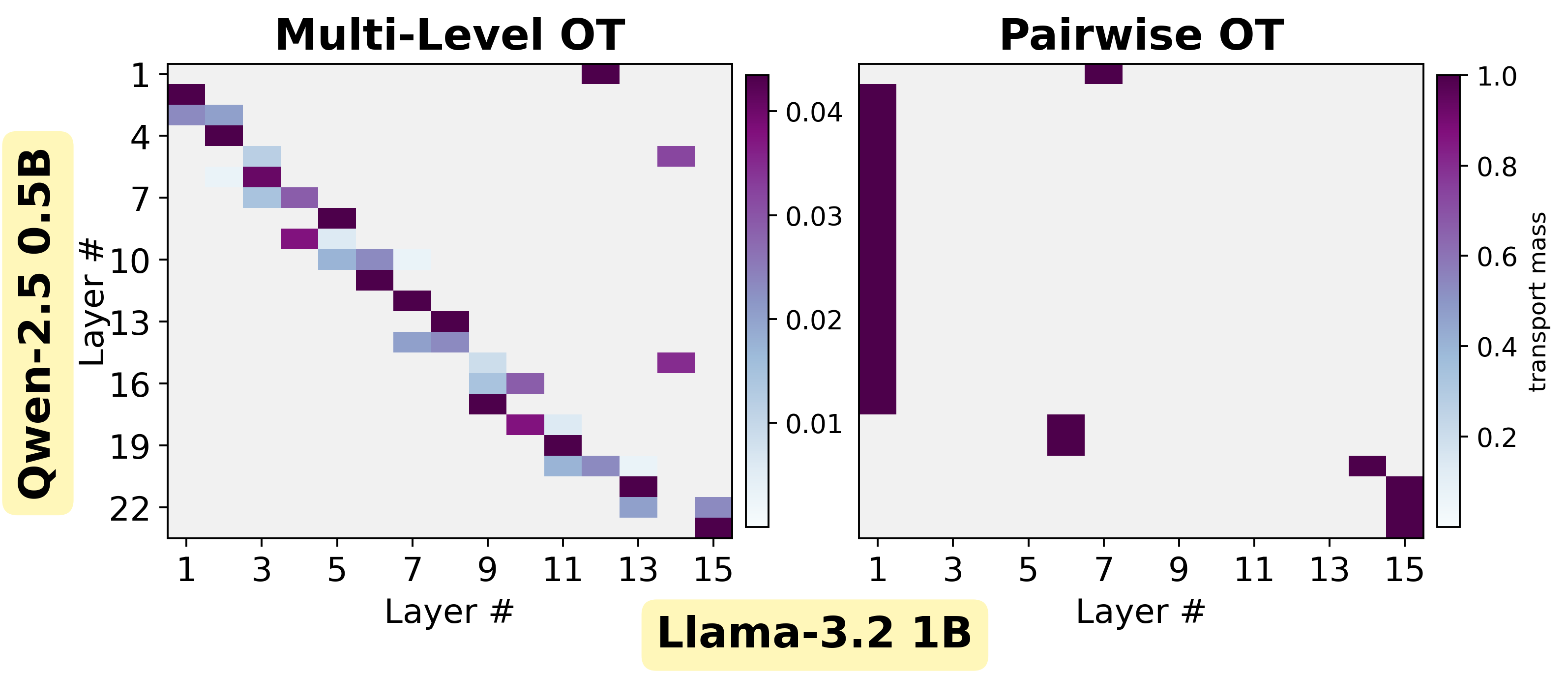}\hfill
\includegraphics[width=0.48\columnwidth]{Figures/LLMs/Qwen_0.5B-Llama_3B.png}
\medskip
\includegraphics[width=0.48\columnwidth]{Figures/LLMs/Llama_1B-Qwen_3B.png}\hfill
\includegraphics[width=0.48\columnwidth]{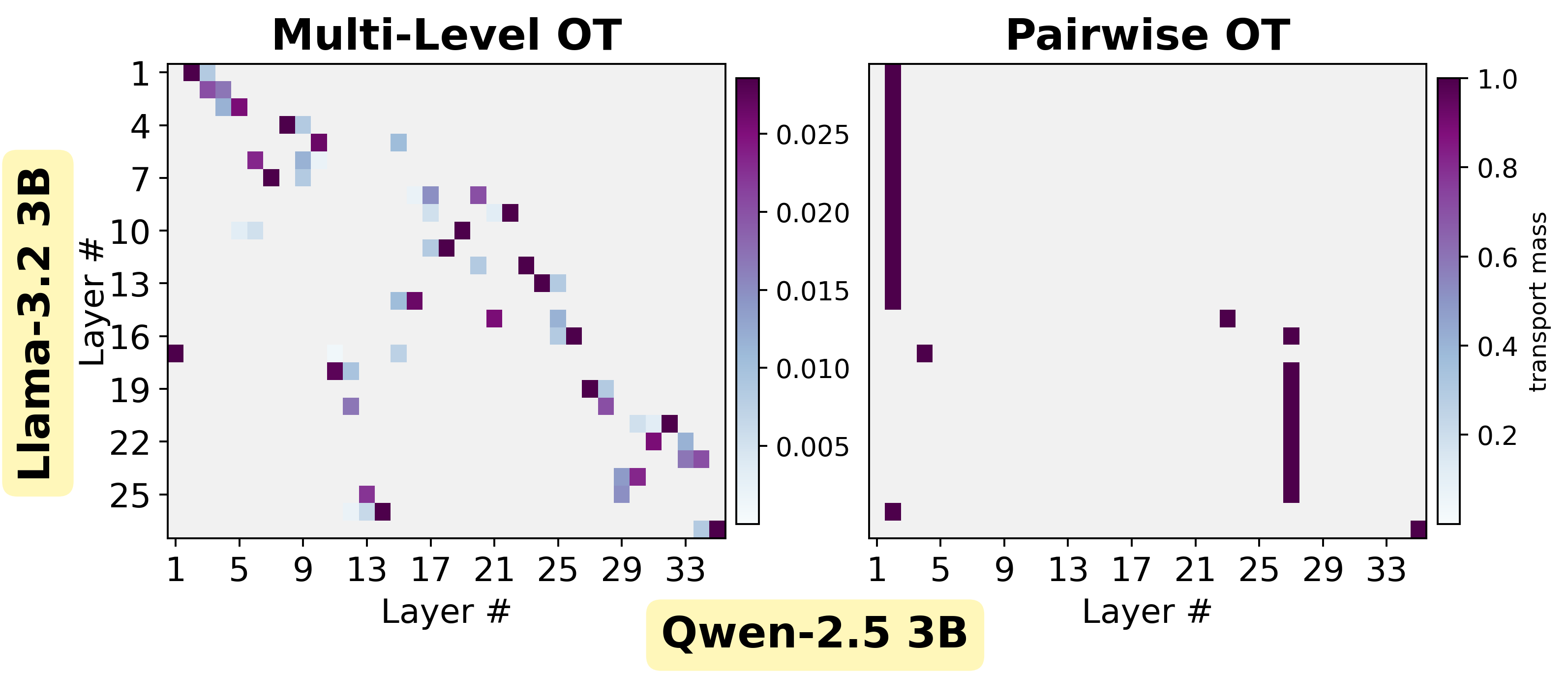}
\caption{\textbf{Transport plans across LLM families and scales.} 
Multi-Level OT (MOT) mappings are shown for six cross-model comparisons: 
(a) LLaMA-3.2 1B $\leftrightarrow$ LLaMA-3.2 3B, 
(b) Qwen-2.5 0.5B $\leftrightarrow$ Qwen-2.5 3B, 
(c) Qwen-2.5 0.5B $\leftrightarrow$ LLaMA-3.2 1B, 
(d) Qwen-2.5 0.5B $\leftrightarrow$ LLaMA-3.2 3B, 
(e) LLaMA-3.2 1B $\leftrightarrow$ Qwen-2.5 3B, and 
(f) LLaMA-3.2 3B $\leftrightarrow$ Qwen-2.5 3B. 
MOT uncovers structured, near-diagonal correspondences that persist across both intra-family (a,b) and cross-family (c-f) alignments, illustrating its robustness compared to pairwise OT.}
\label{fig:llms_all}
\end{figure}

\newpage
\section{Representation Similarity between Vision Cortex Response}
\begin{table}[H]
\centering
\small
\begin{tabular}{@{}ll c c c c@{}}
\toprule
\multicolumn{1}{c}{Model 1} & \multicolumn{1}{c}{Model 2} & \multicolumn{1}{c}{MOT Metric} & \multicolumn{1}{c}{Random (Perm-P)} & \multicolumn{1}{c}{Single-Best OT} & \multicolumn{1}{c}{Pairwise Best OT} \\
\midrule
Subject A & Subject B & 0.244 $\pm$ 0.005 & 0.135 $\pm$ 0.015 & 0.244 $\pm$ 0.005 & \textbf{0.245 $\pm$ 0.004} \\
Subject A & Subject C & 0.199 $\pm$ 0.003 & 0.110 $\pm$ 0.013 & 0.199 $\pm$ 0.003 & \textbf{0.202 $\pm$ 0.003} \\
Subject A & Subject D & 0.198 $\pm$ 0.008 & 0.109 $\pm$ 0.016 & 0.198 $\pm$ 0.008 & \textbf{0.201 $\pm$ 0.007} \\
Subject B & Subject C & \textbf{0.212 $\pm$ 0.007} & 0.126 $\pm$ 0.014 & \textbf{0.212 $\pm$ 0.007} & \textbf{0.212 $\pm$ 0.007} \\
Subject C & Subject D & 0.197 $\pm$ 0.007 & 0.112 $\pm$ 0.016 & 0.197 $\pm$ 0.007 & \textbf{0.199 $\pm$ 0.006} \\

Subject B & Subject D & 0.201 $\pm$ 0.008 & 0.121 $\pm$ 0.014 & 0.201 $\pm$ 0.008 & \textbf{0.204 $\pm$ 0.007} \\

\bottomrule
\end{tabular}
\caption{\textbf{Visual cortex alignment performance.} Comparison of MOT against baseline methods, evaluated by reconstruction correlation on held-out fMRI responses: mean $\pm$ standard deviation across seeds.}
\label{tab:vision_mean_std}
\end{table}

\subsection{Transport plans for OT based mapping pairs}

\begin{figure}[H]
\centering
\includegraphics[width=0.48\columnwidth]{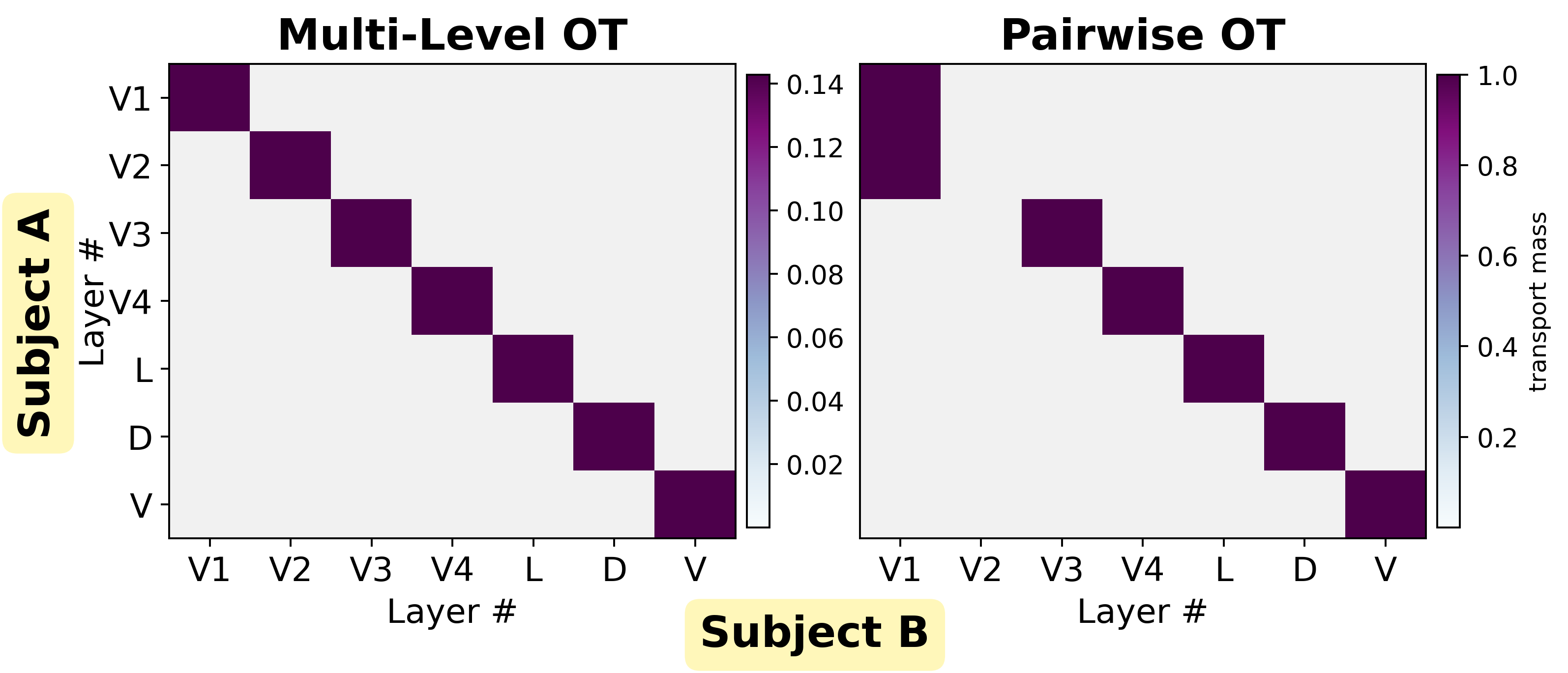}\hfill
\includegraphics[width=0.48\columnwidth]{Figures/FMRIs/Subject_A-Subject_C.png}
\medskip
\includegraphics[width=0.48\columnwidth]{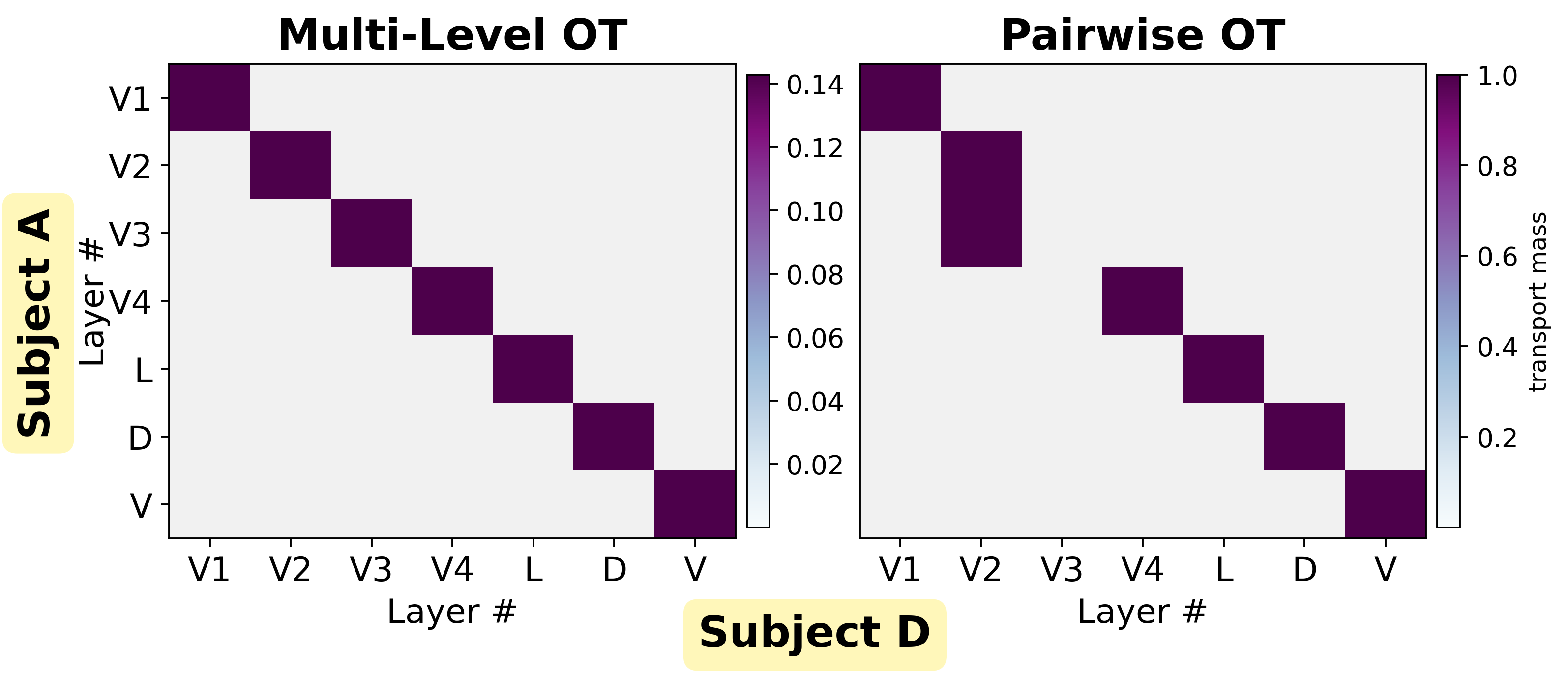}\hfill
\includegraphics[width=0.48\columnwidth]{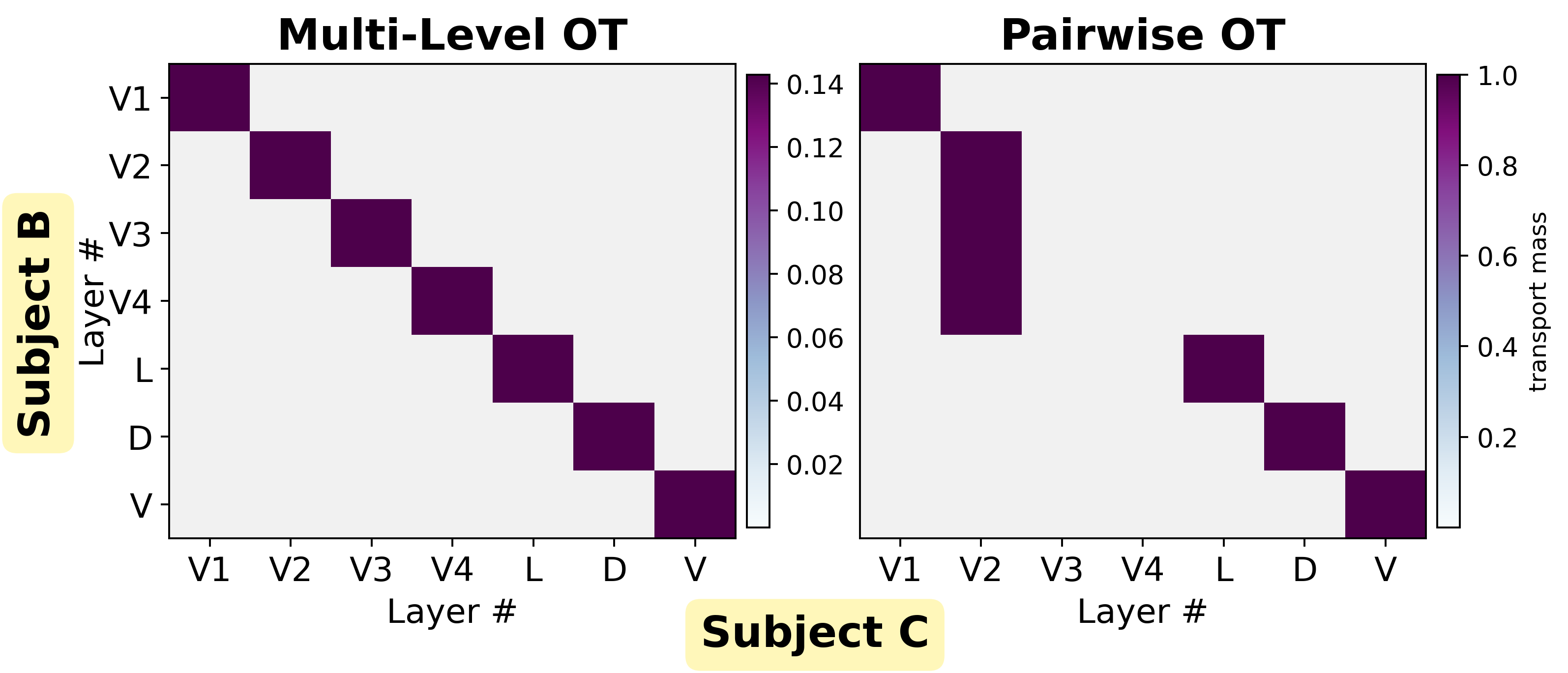}
\medskip
\includegraphics[width=0.48\columnwidth]{Figures/FMRIs/Subject_B-Subject_D.png}\hfill
\includegraphics[width=0.48\columnwidth]{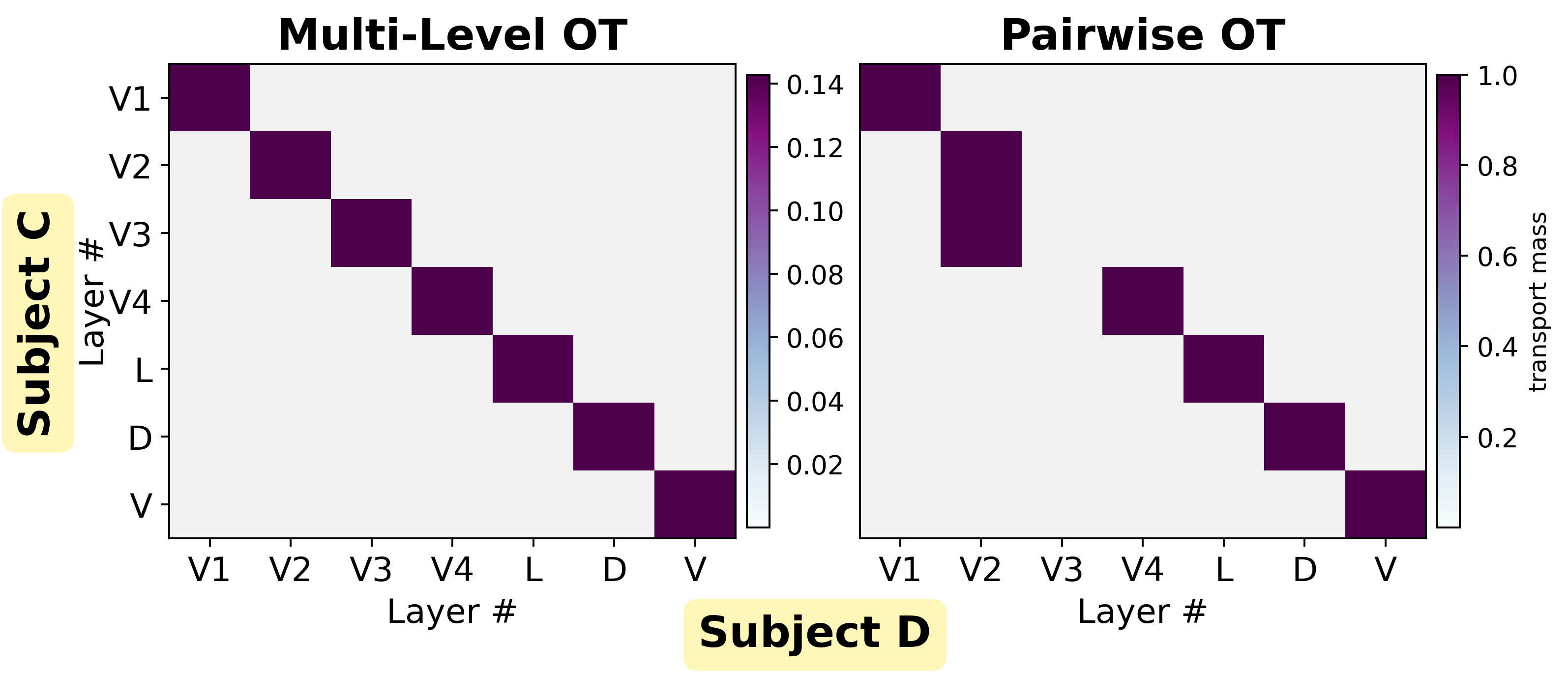}
\caption{\textbf{Transport plans for OT-based mapping pairs.} 
Pairwise OT mappings across all subject pairs: 
(a) Subject A $\leftrightarrow$ Subject B, 
(b) Subject A $\leftrightarrow$ Subject C, 
(c) Subject A $\leftrightarrow$ Subject D, 
(d) Subject B $\leftrightarrow$ Subject C, 
(e) Subject B $\leftrightarrow$ Subject D, and 
(f) Subject C $\leftrightarrow$ Subject D. 
MOT recovers structured region-to-region correspondences that are absent in pairwise OT. }
\label{fig:fmris_all}
\end{figure}

\newpage
\section{Representation Similarity between Vision Models}
\begin{table}[H]
\centering
\small
\begin{tabular}{@{}ll S S S S@{}}
\toprule
\multicolumn{1}{c}{Model 1} & \multicolumn{1}{c}{Model 2} & \multicolumn{1}{c}{MOT Metric} & \multicolumn{1}{c}{Random (Perm-P)} & \multicolumn{1}{c}{Single-Best OT} & \multicolumn{1}{c}{Pairwise Best OT} \\
\midrule
DINOv2 Small & ViT-MAE Base  & 0.289 & 0.259 & 0.289 & \textbf{0.301} \\
DINOv2 Small & DINOv2 Large  & \textbf{0.353} & 0.298 & 0.312 & 0.340 \\
DINOv2 Small & DINOv2 Giant  & \textbf{0.466} & 0.385 & 0.413 & 0.433 \\
DINOv2 Small & ViT-MAE Large & \textbf{0.381} & 0.348 & 0.350 & 0.354 \\
DINOv2 Small & ViT-MAE Huge  & \textbf{0.411} & 0.372 & 0.359 & 0.386 \\
\midrule
ViT-MAE Base & DINOv2 Large  & 0.577 & 0.394 & 0.531 & \textbf{0.624} \\
ViT-MAE Base & DINOv2 Giant  & \textbf{0.202} & 0.197 & 0.148 & 0.180 \\
ViT-MAE Base & ViT-MAE Large & 0.588 & 0.528 & 0.539 & \textbf{0.598} \\
ViT-MAE Base & ViT-MAE Huge  & 0.149 & 0.116 & 0.181 & \textbf{0.417} \\
ViT-MAE Huge & DINOv2 Giant  & 0.317 & 0.261 & 0.293 & \textbf{0.352} \\
\bottomrule
\end{tabular}
\caption{MOT Metric with its baseline comparisons.}
\label{tab:hot-and-baselines}
\end{table}

\begin{table}[H]
\centering
\small
\begin{tabular}{@{}ll S S S@{}}
\toprule
\multicolumn{1}{c}{Model 1} & \multicolumn{1}{c}{Model 2} & \multicolumn{1}{c}{MOT + R} & \multicolumn{1}{c}{Single-Best + R} & \multicolumn{1}{c}{Pairwise Best + R} \\
\midrule
DINOv2 Small & ViT-MAE Base  & \textbf{0.600} & \textbf{0.600} & 0.526 \\
DINOv2 Small & DINOv2 Large  & \textbf{0.778} & 0.708 & 0.394 \\
DINOv2 Small & DINOv2 Giant  & \textbf{0.790} & 0.746 & 0.418 \\
DINOv2 Small & ViT-MAE Large & \textbf{0.633} & 0.599 & 0.509 \\
DINOv2 Small & ViT-MAE Huge  & \textbf{0.657} & 0.614 & 0.508 \\
\midrule
ViT-MAE Base & DINOv2 Large  & \textbf{0.732} & 0.712 & 0.283 \\
ViT-MAE Base & DINOv2 Giant  & \textbf{0.580} & 0.605 & 0.293 \\
ViT-MAE Base & ViT-MAE Large & \textbf{0.850} & 0.848 & 0.596 \\
ViT-MAE Base & ViT-MAE Huge  & \textbf{0.788} & 0.760 & 0.571 \\
ViT-MAE Huge & DINOv2 Giant  & \textbf{0.614} & 0.582 & 0.359 \\
\bottomrule
\end{tabular}
\caption{Rotation (MOT + R) with its baseline comparisons.}
\label{tab:rotation-and-baselines}
\end{table}

\begin{figure}[H]
\centering
\includegraphics[width=0.48\textwidth]{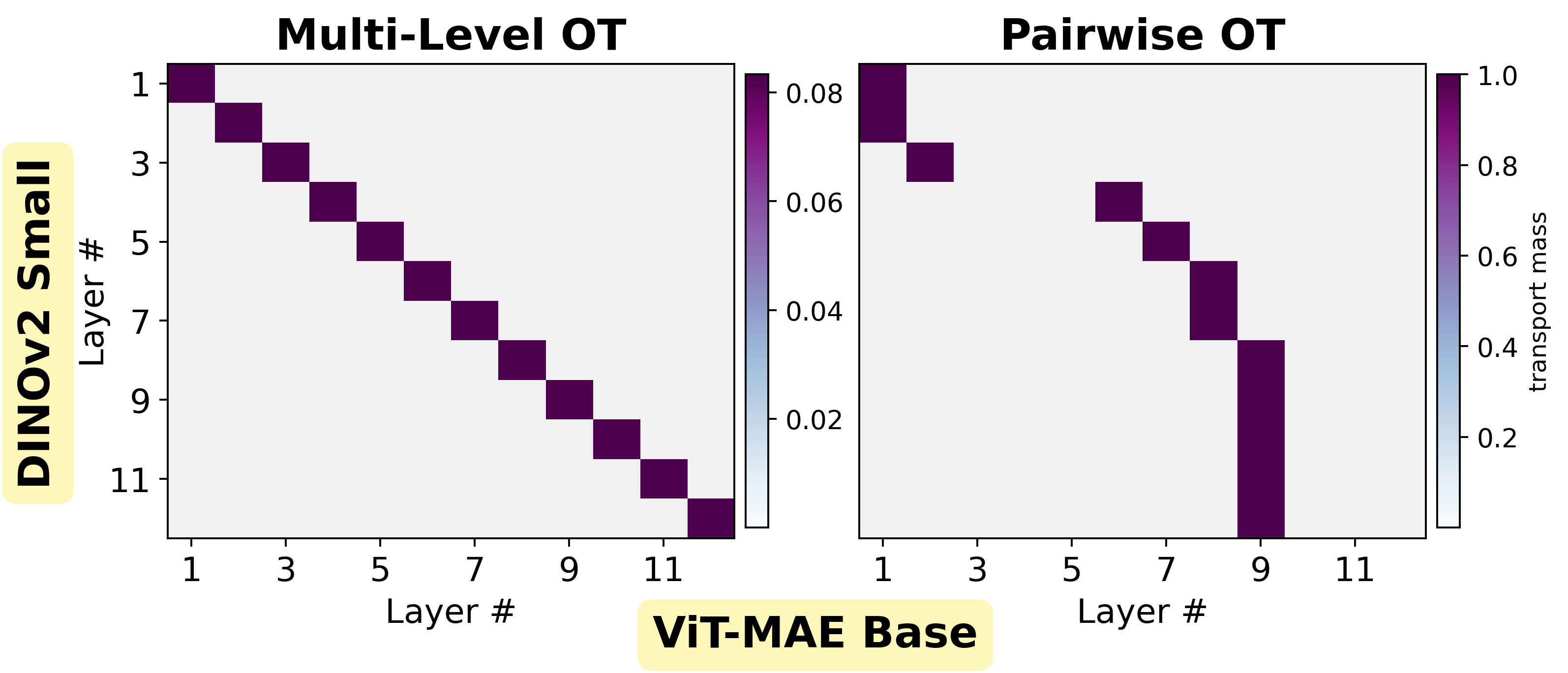}\hfill
\includegraphics[width=0.48\textwidth]{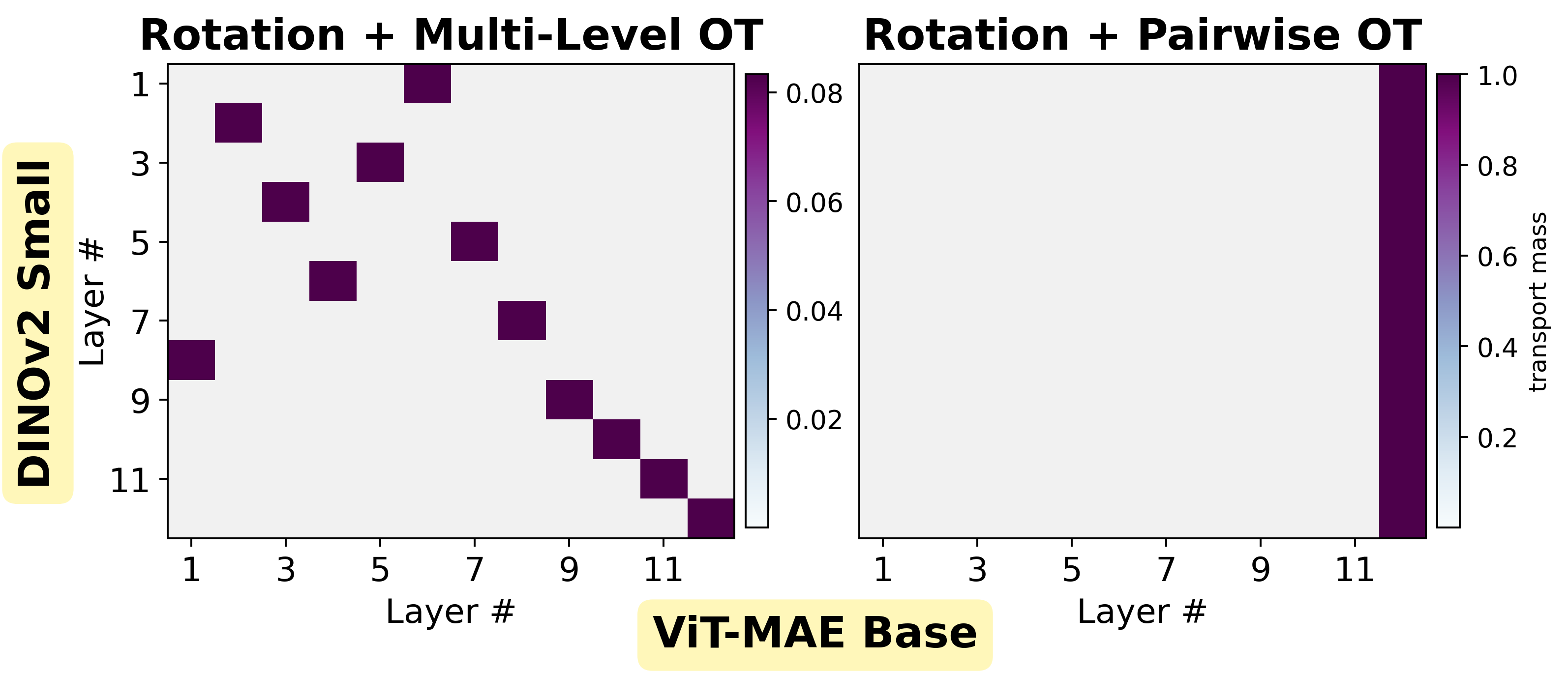}
\caption{\textbf{Transport plans for vision model alignment.} DINOv2 Small $\leftrightarrow$ ViT-MAE Base (a) without rotation (MOT) and (b) with rotation augmentation (MOT+R).}
\label{fig:dinov2-small-vit-mae-base}
\end{figure}

\begin{figure}[H]
\centering
\includegraphics[width=0.48\textwidth]{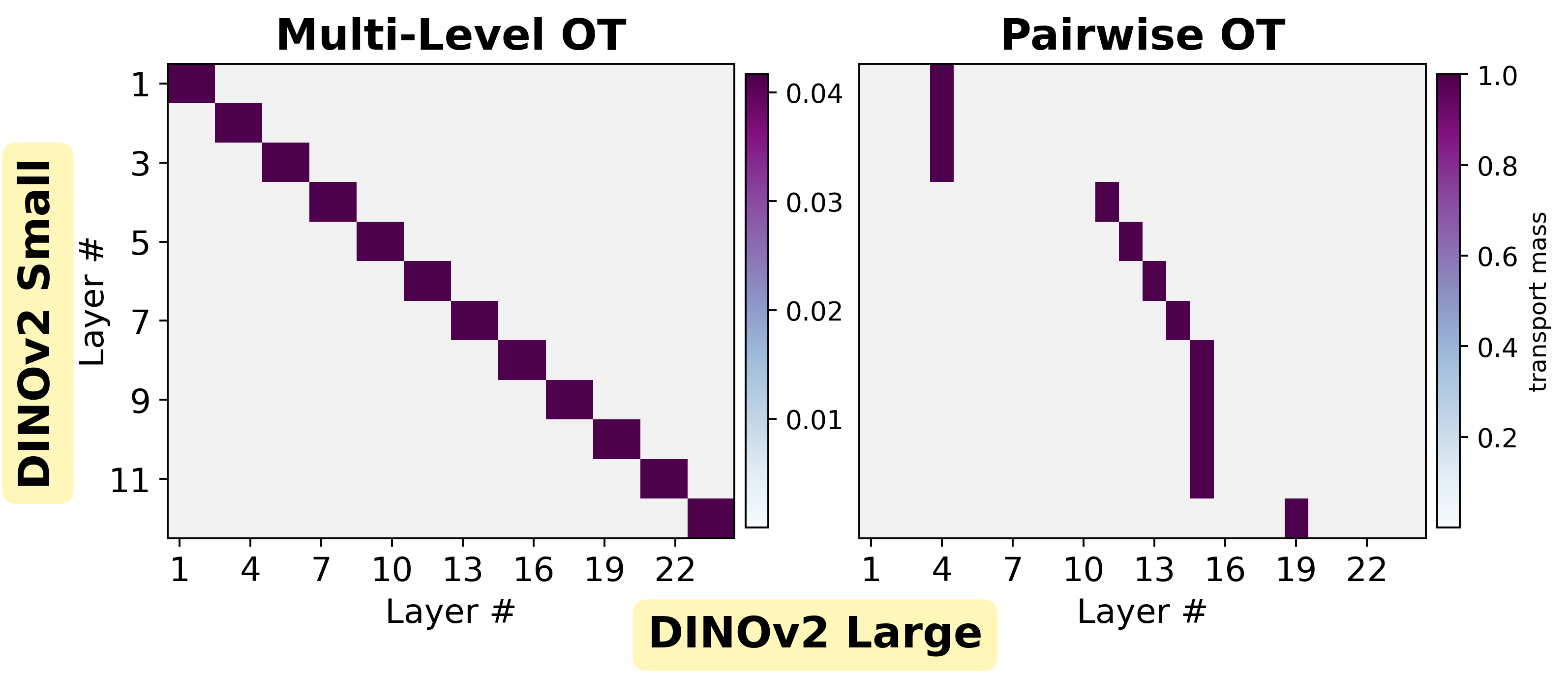}\hfill
\includegraphics[width=0.48\textwidth]{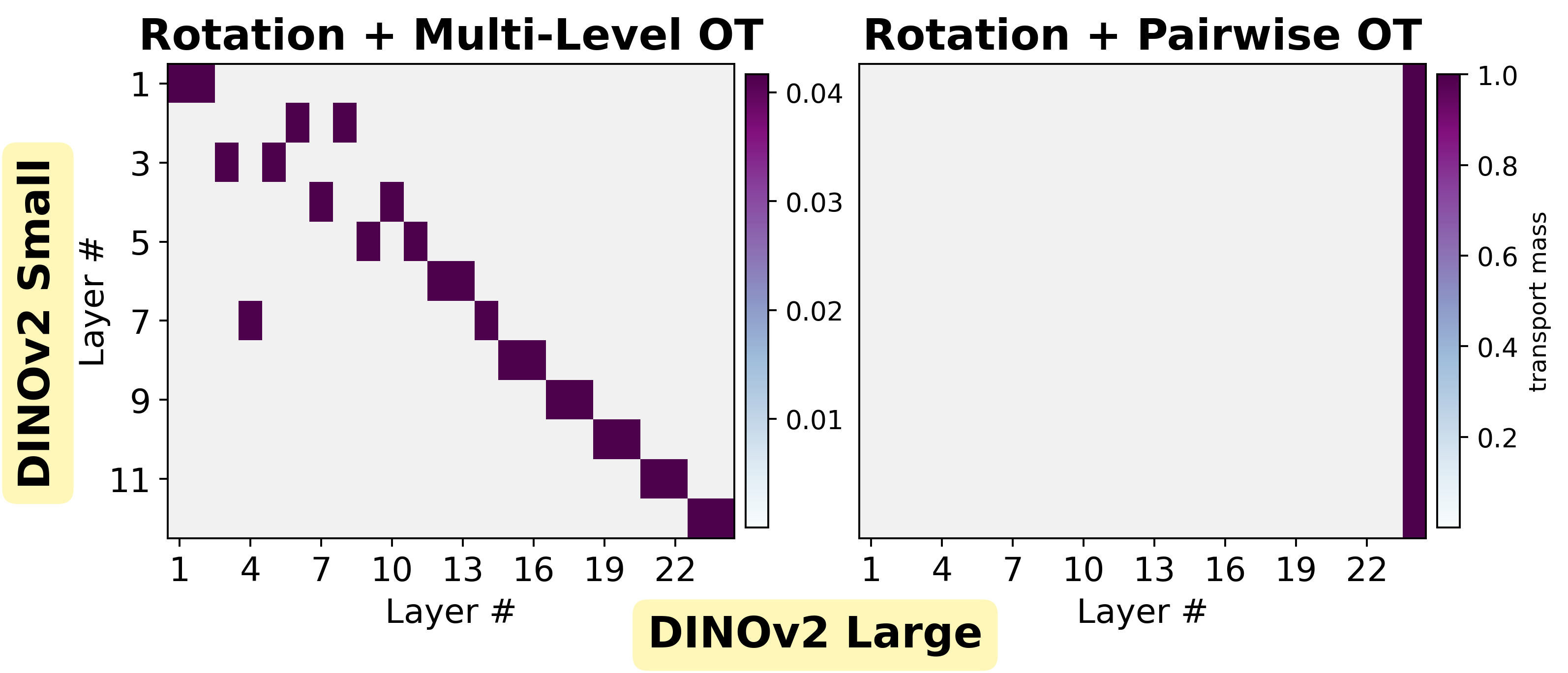}
\caption{\textbf{Transport plans for vision model alignment.} DINOv2 Small $\leftrightarrow$ DINOv2 Large (a) without rotation (MOT) and (b) with rotation augmentation (MOT+R).}
\label{fig:dinov2-small-dinov2-large}
\end{figure}

\begin{figure}[H]
\centering
\includegraphics[width=0.48\textwidth]{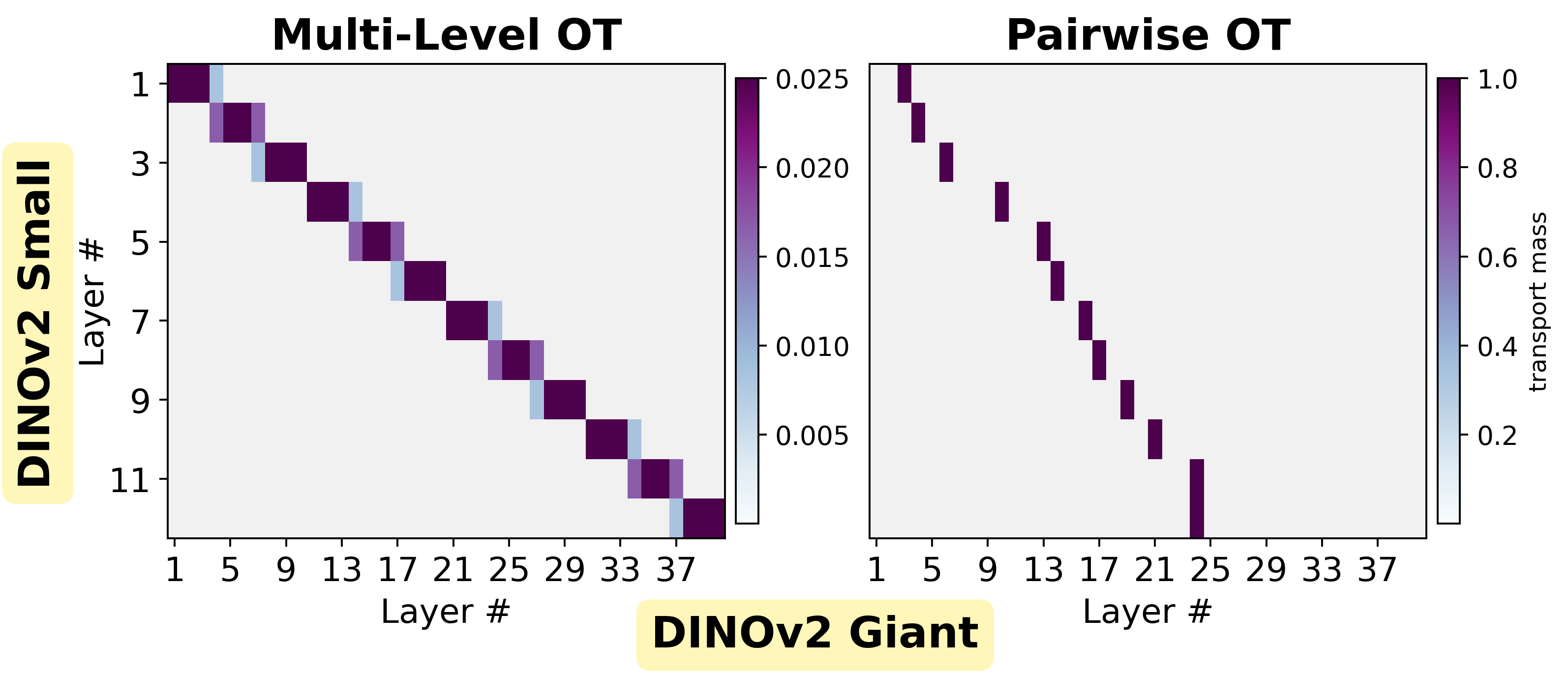}\hfill
\includegraphics[width=0.48\textwidth]{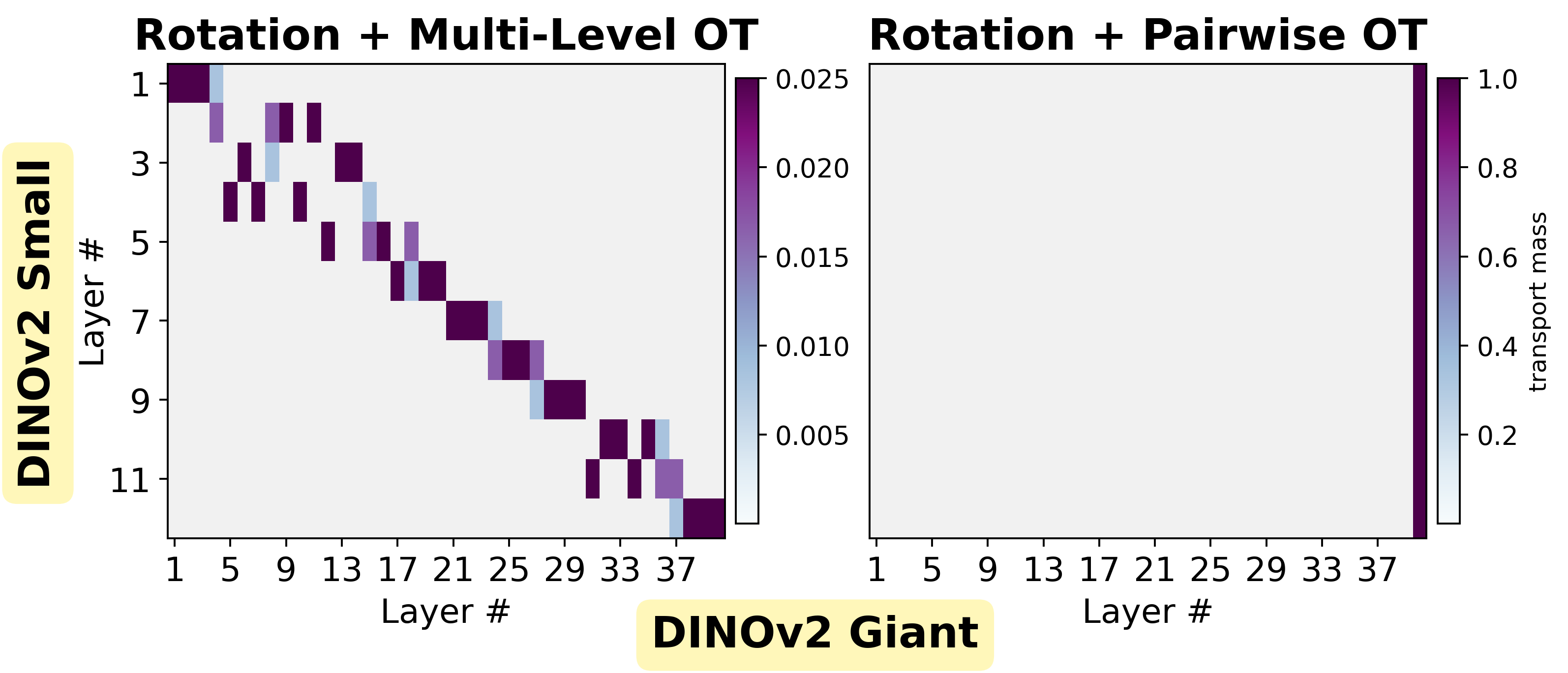}
\caption{\textbf{Transport plans for vision model alignment.} DINOv2 Small $\leftrightarrow$ DINOv2 Giant (a) without rotation (MOT) and (b) with rotation augmentation (MOT+R).}
\label{fig:dinov2-small-dinov2-giant}
\end{figure}

\begin{figure}[H]
\centering
\includegraphics[width=0.48\textwidth]{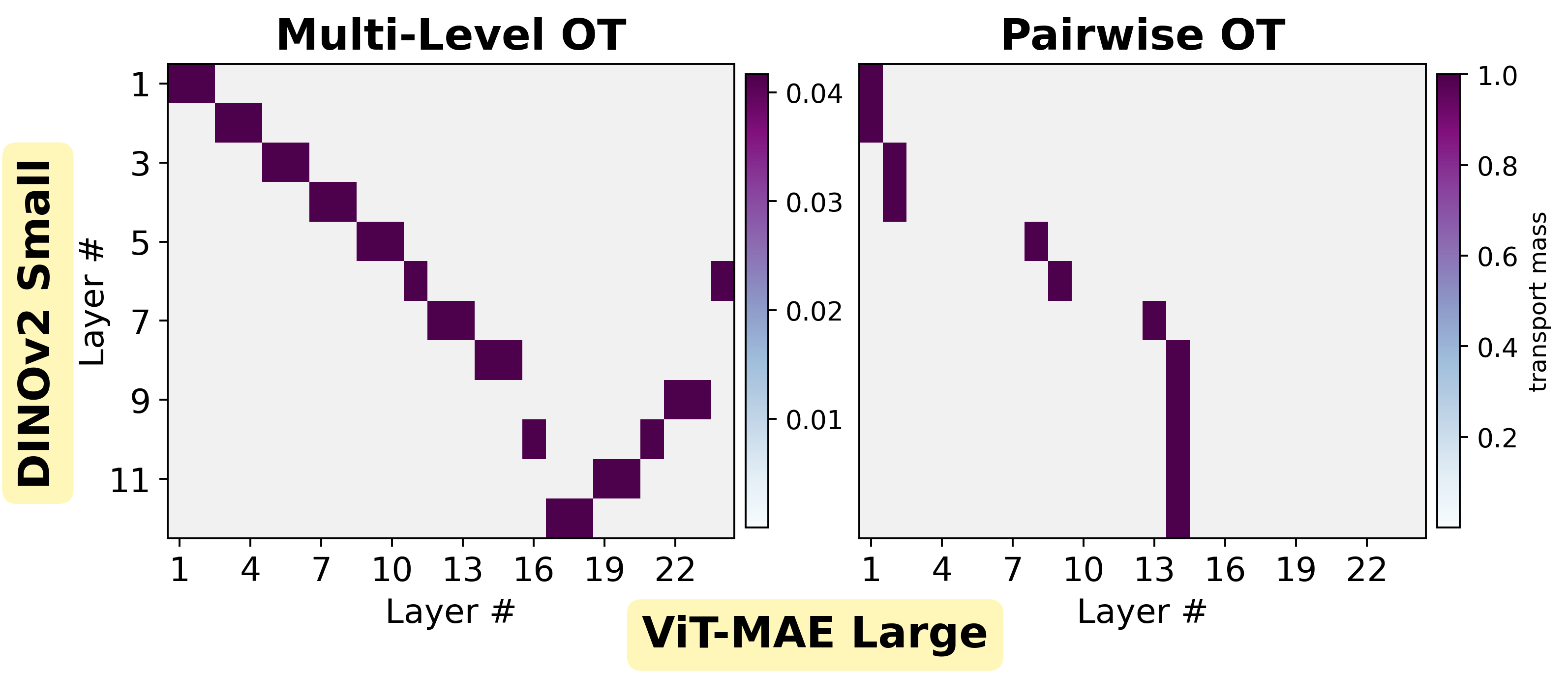}\hfill
\includegraphics[width=0.48\textwidth]{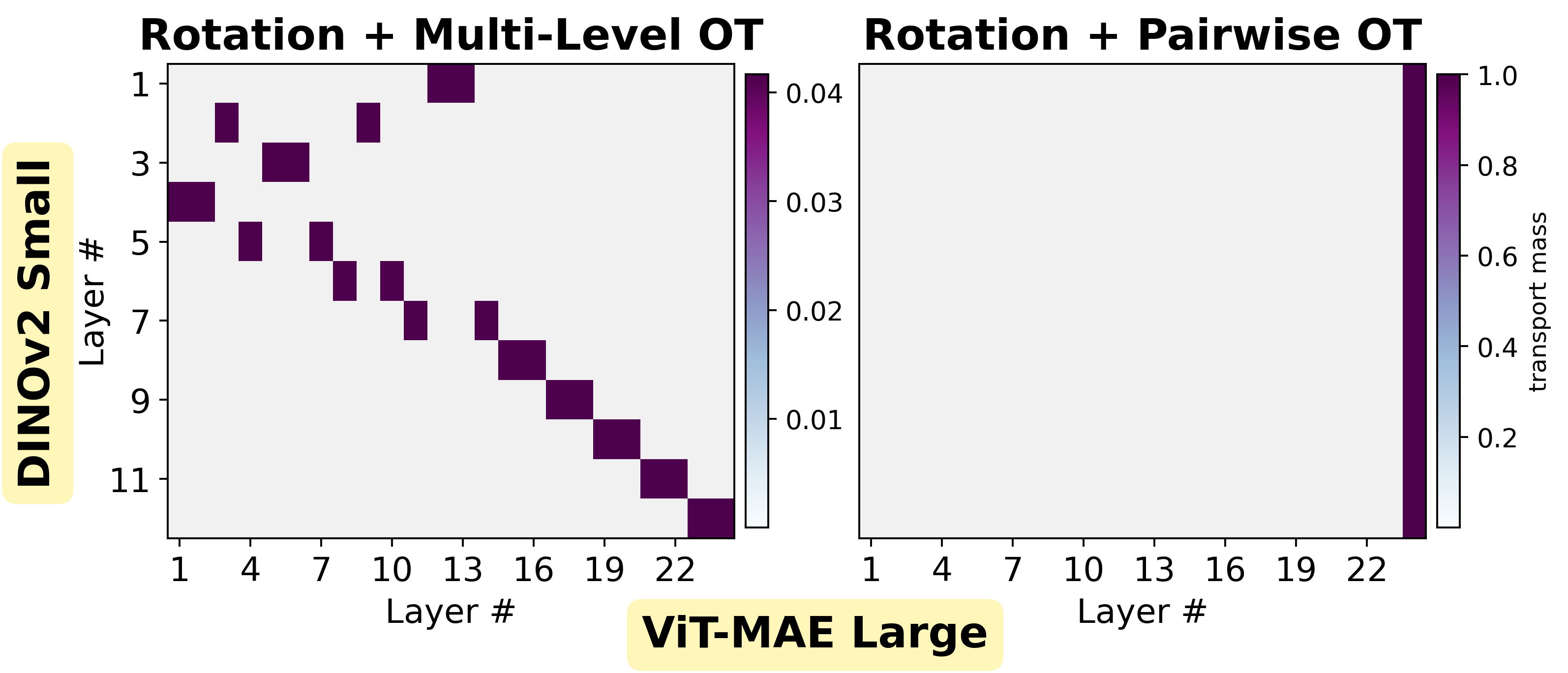}
\caption{\textbf{Transport plans for vision model alignment.} DINOv2 Small $\leftrightarrow$ ViT-MAE Large (a) without rotation (MOT) and (b) with rotation augmentation (MOT+R).}
\label{fig:dinov2-small-vit-mae-large}
\end{figure}

\begin{figure}[H]
\centering
\includegraphics[width=0.48\textwidth]{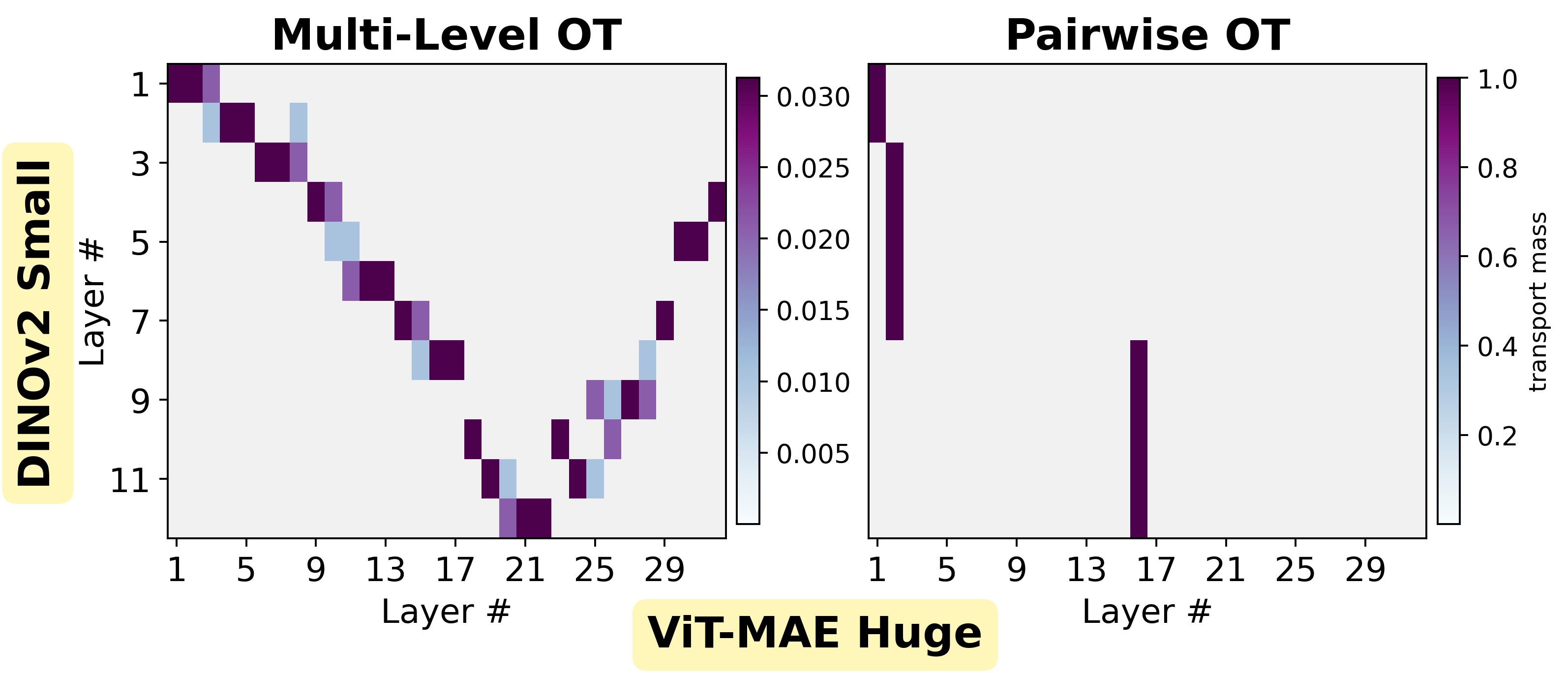}\hfill
\includegraphics[width=0.48\textwidth]{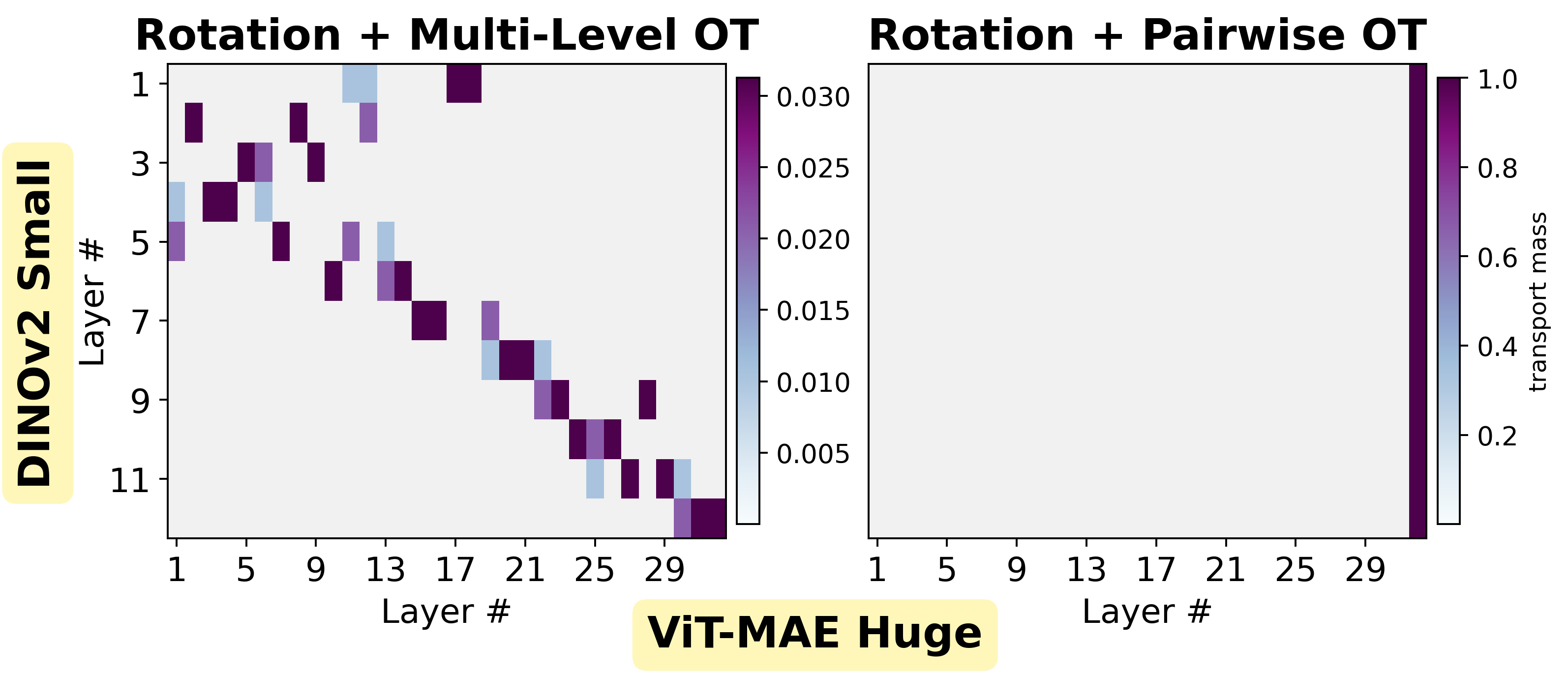}
\caption{\textbf{Transport plans for vision model alignment.} DINOv2 Small $\leftrightarrow$ ViT-MAE Huge (a) without rotation (MOT) and (b) with rotation augmentation (MOT+R).}
\label{fig:dinov2-small-vit-mae-huge}
\end{figure}

\begin{figure}[H]
\centering
\includegraphics[width=0.48\textwidth]{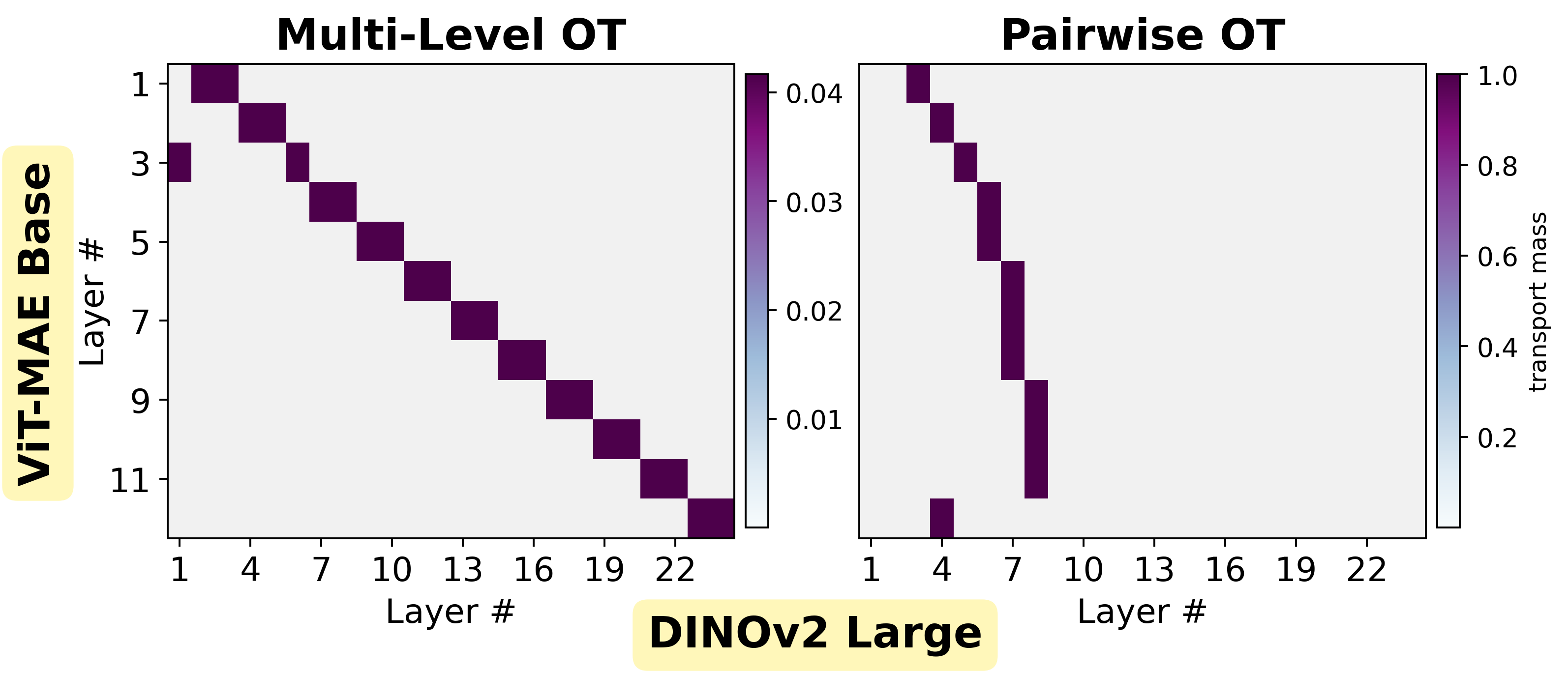}\hfill
\includegraphics[width=0.48\textwidth]{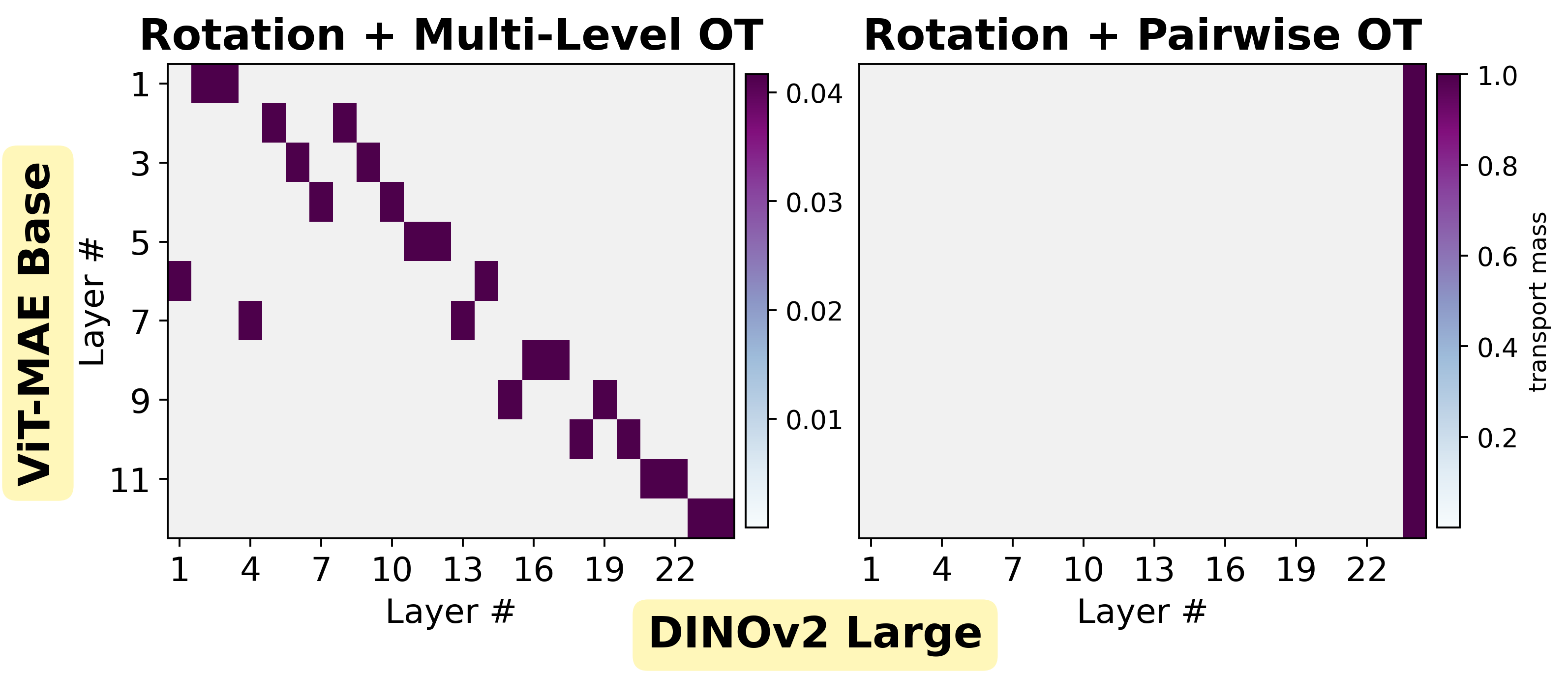}
\caption{\textbf{Transport plans for vision model alignment.} ViT-MAE Base $\leftrightarrow$ DINOv2 Large (a) without rotation (MOT) and (b) with rotation augmentation (MOT+R).}
\label{fig:vit-mae-base-dinov2-large}
\end{figure}

\begin{figure}[H]
\centering
\includegraphics[width=0.48\textwidth]{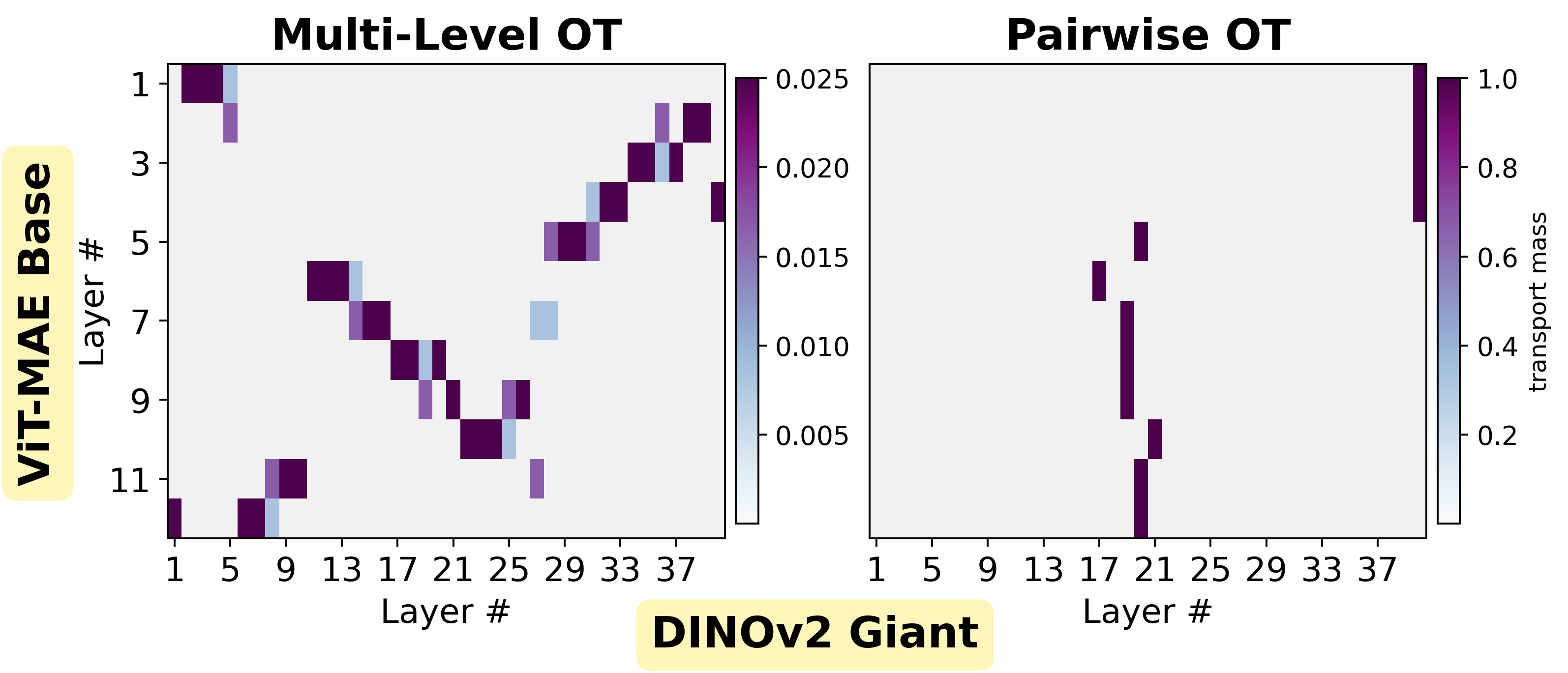}\hfill
\includegraphics[width=0.48\textwidth]{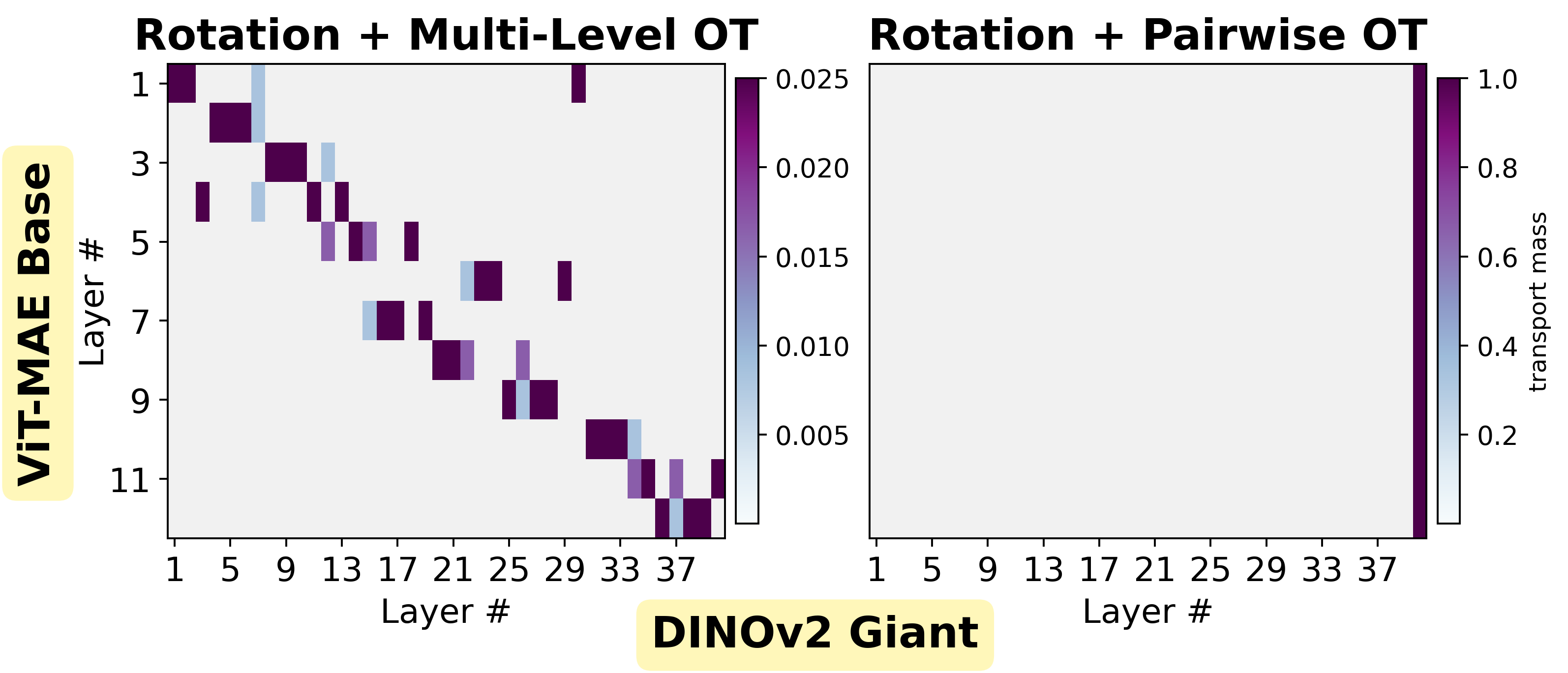}
\caption{\textbf{Transport plans for vision model alignment.} ViT-MAE Base $\leftrightarrow$ DINOv2 Giant (a) without rotation (MOT) and (b) with rotation augmentation (MOT+R).}
\label{fig:vit-mae-base-dinov2-giant}
\end{figure}

\begin{figure}[H]
\centering
\includegraphics[width=0.48\textwidth]{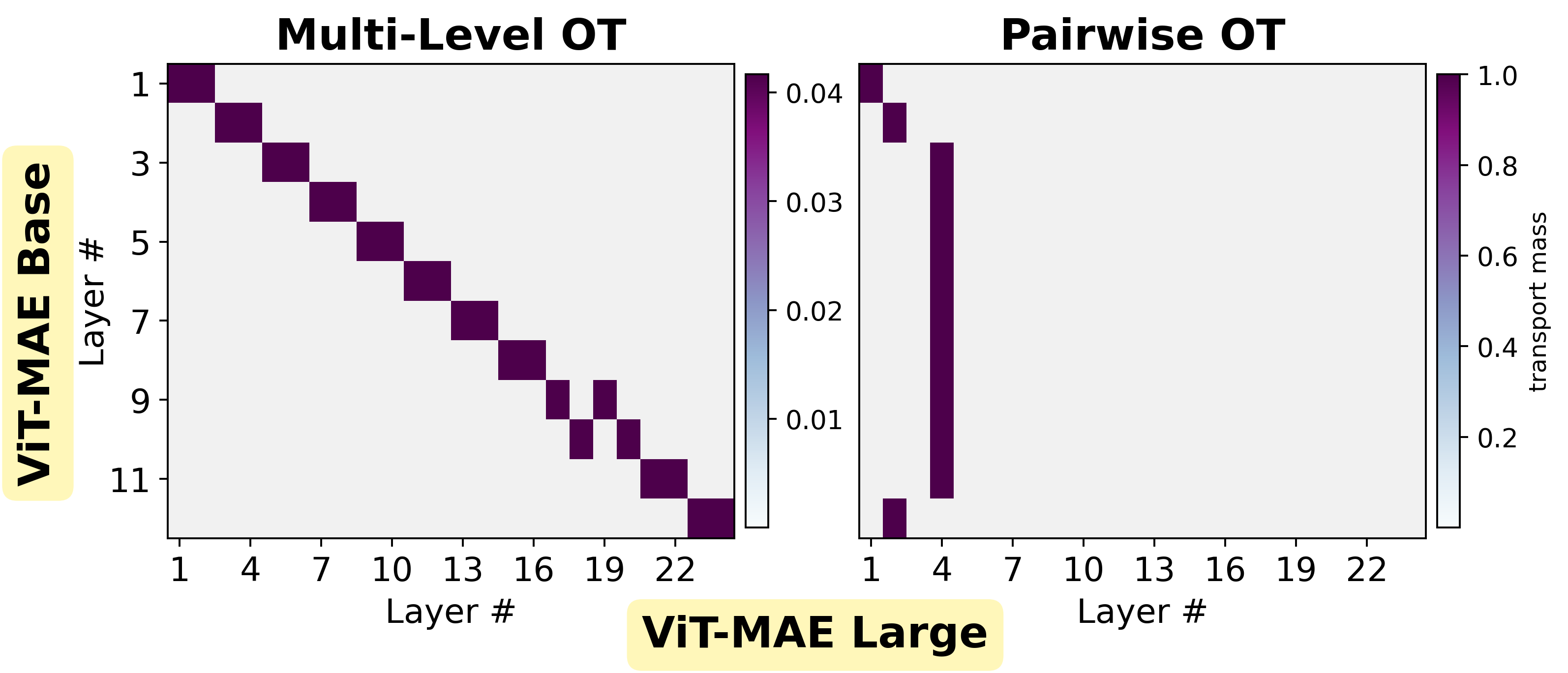}\hfill
\includegraphics[width=0.48\textwidth]{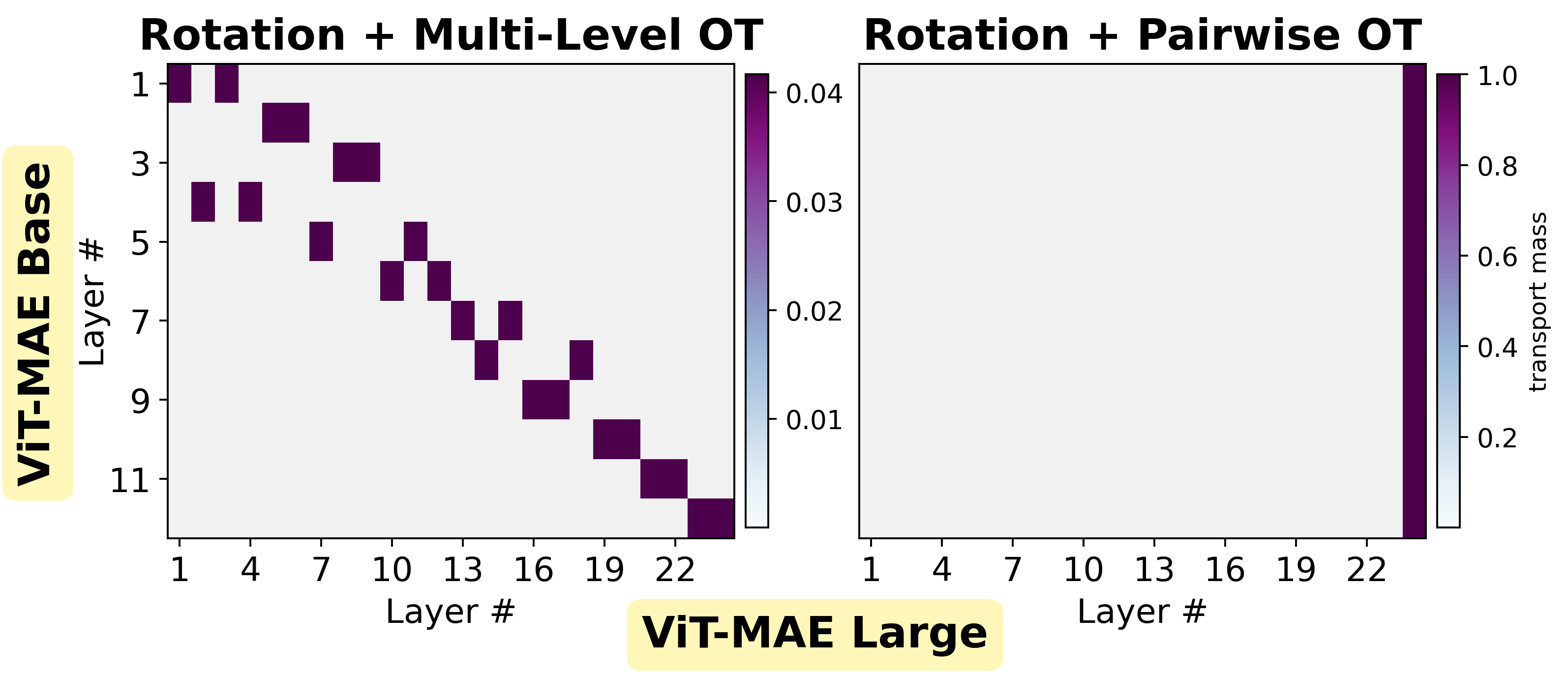}
\caption{\textbf{Transport plans for vision model alignment.} ViT-MAE Base $\leftrightarrow$ ViT-MAE Large (a) without rotation (MOT) and (b) with rotation augmentation (MOT+R).}
\label{fig:vit-mae-base-vit-mae-large}
\end{figure}

\begin{figure}[H]
\centering
\includegraphics[width=0.48\textwidth]{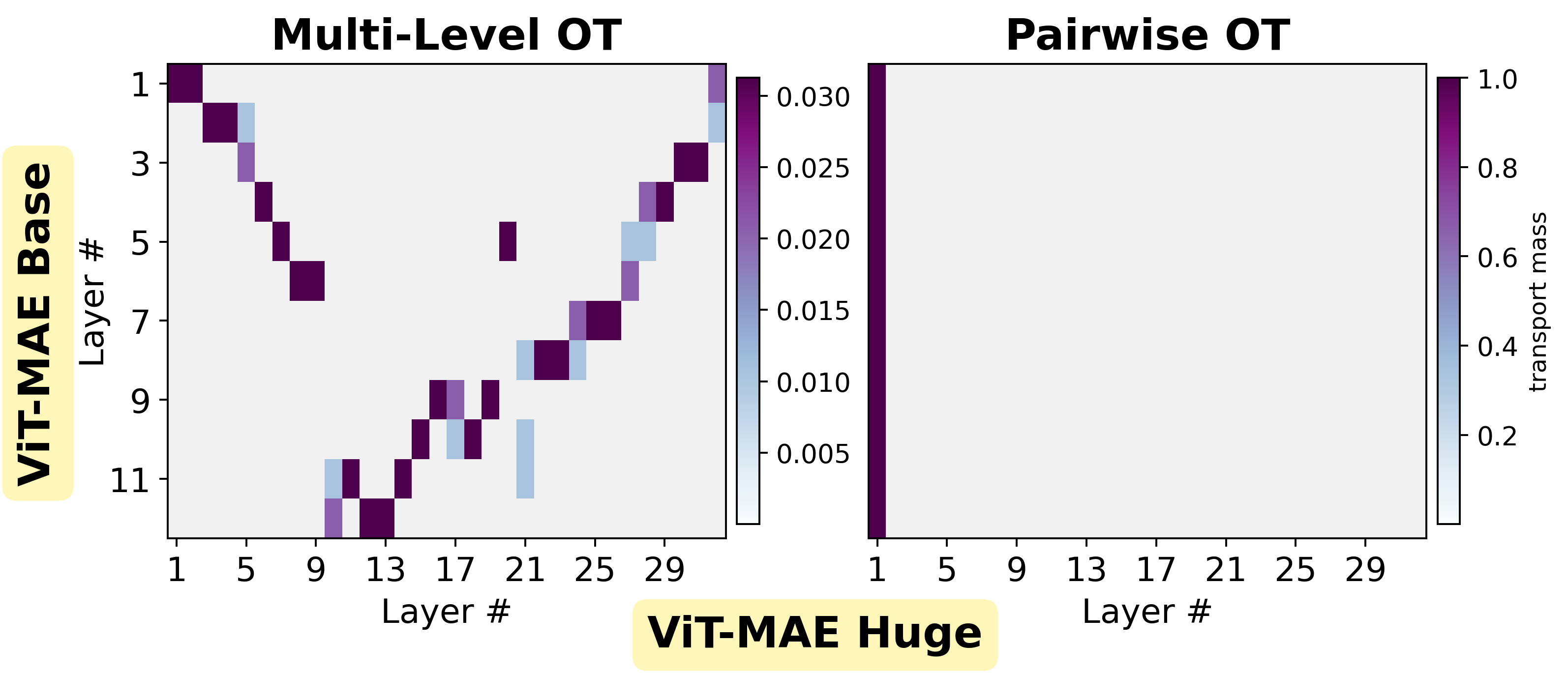}\hfill
\includegraphics[width=0.48\textwidth]{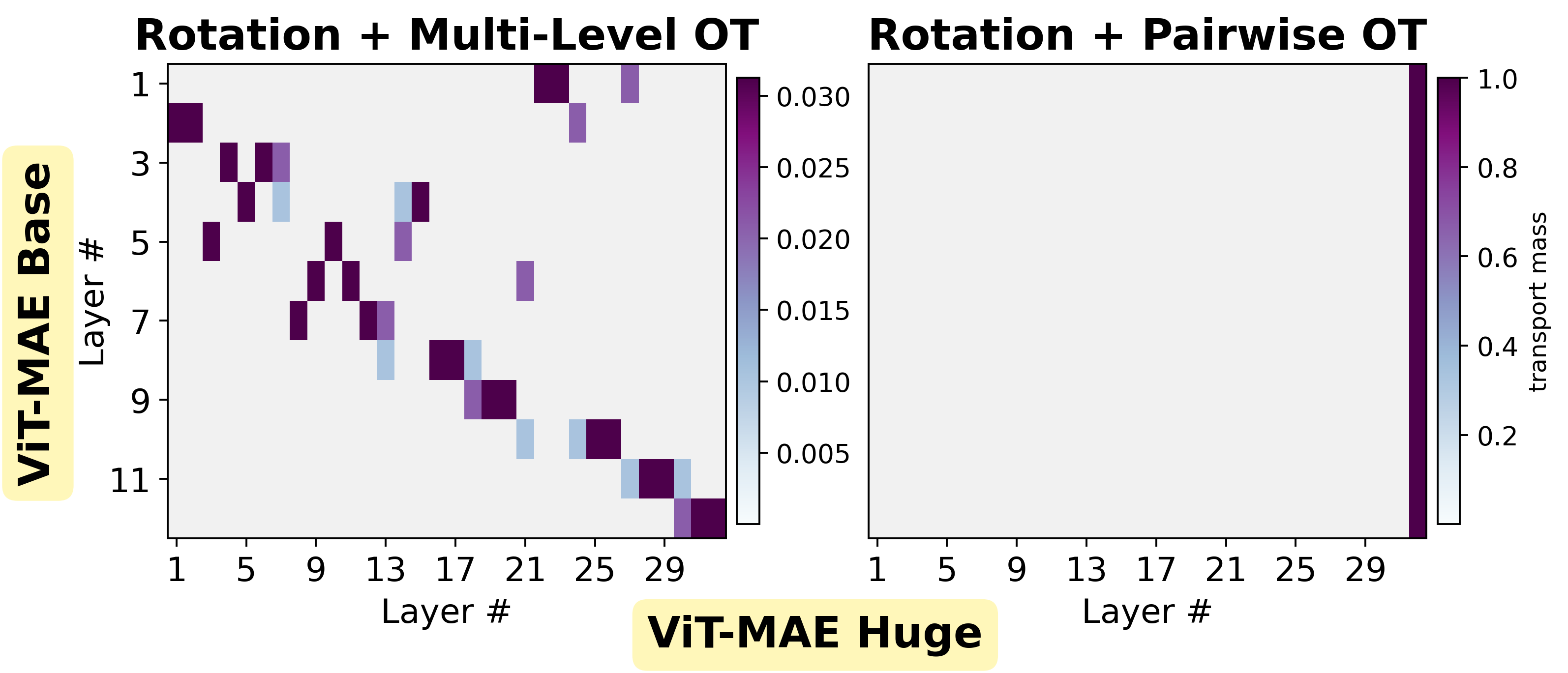}
\caption{\textbf{Transport plans for vision model alignment.} ViT-MAE Base $\leftrightarrow$ ViT-MAE Huge (a) without rotation (MOT) and (b) with rotation augmentation (MOT+R).}
\label{fig:vit-mae-base-vit-mae-huge}
\end{figure}

\begin{figure}[H]
\centering
\includegraphics[width=0.48\textwidth]{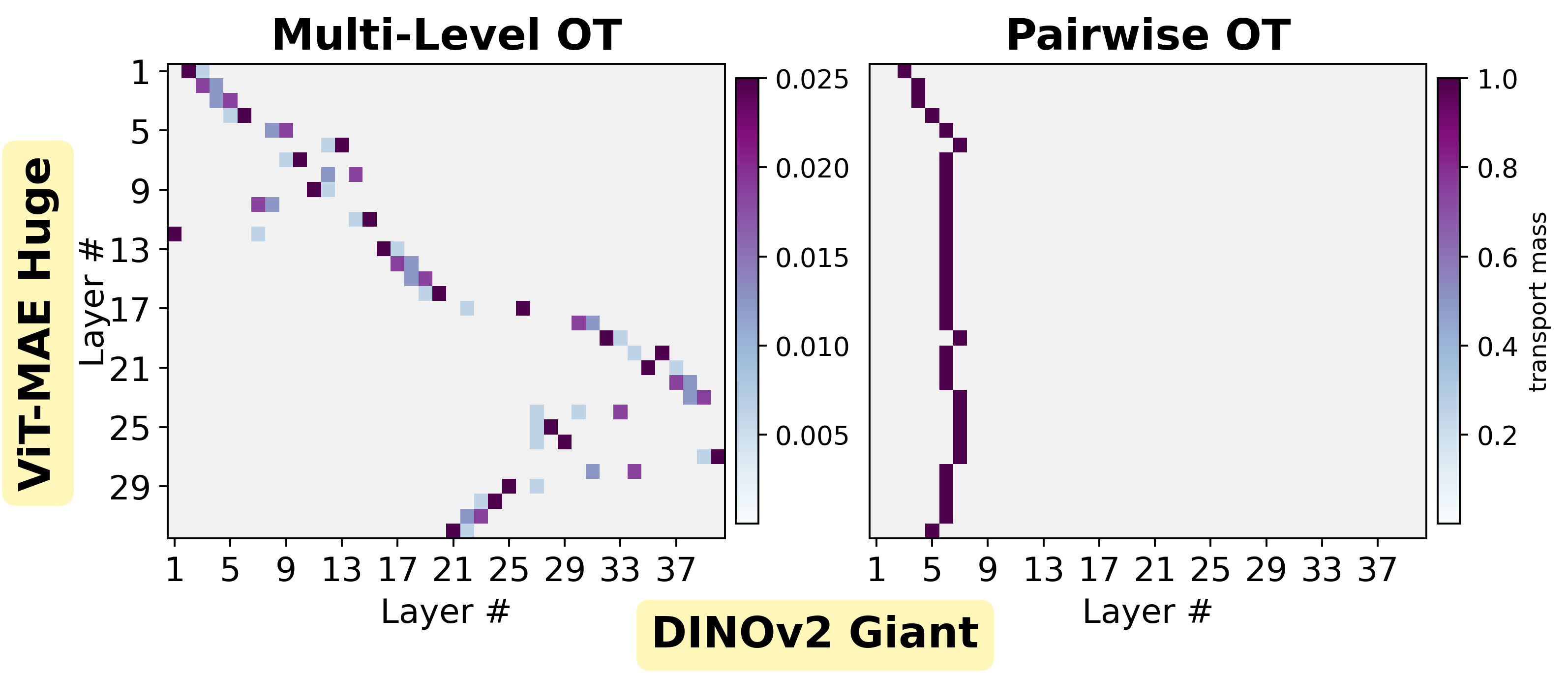}\hfill
\includegraphics[width=0.48\textwidth]{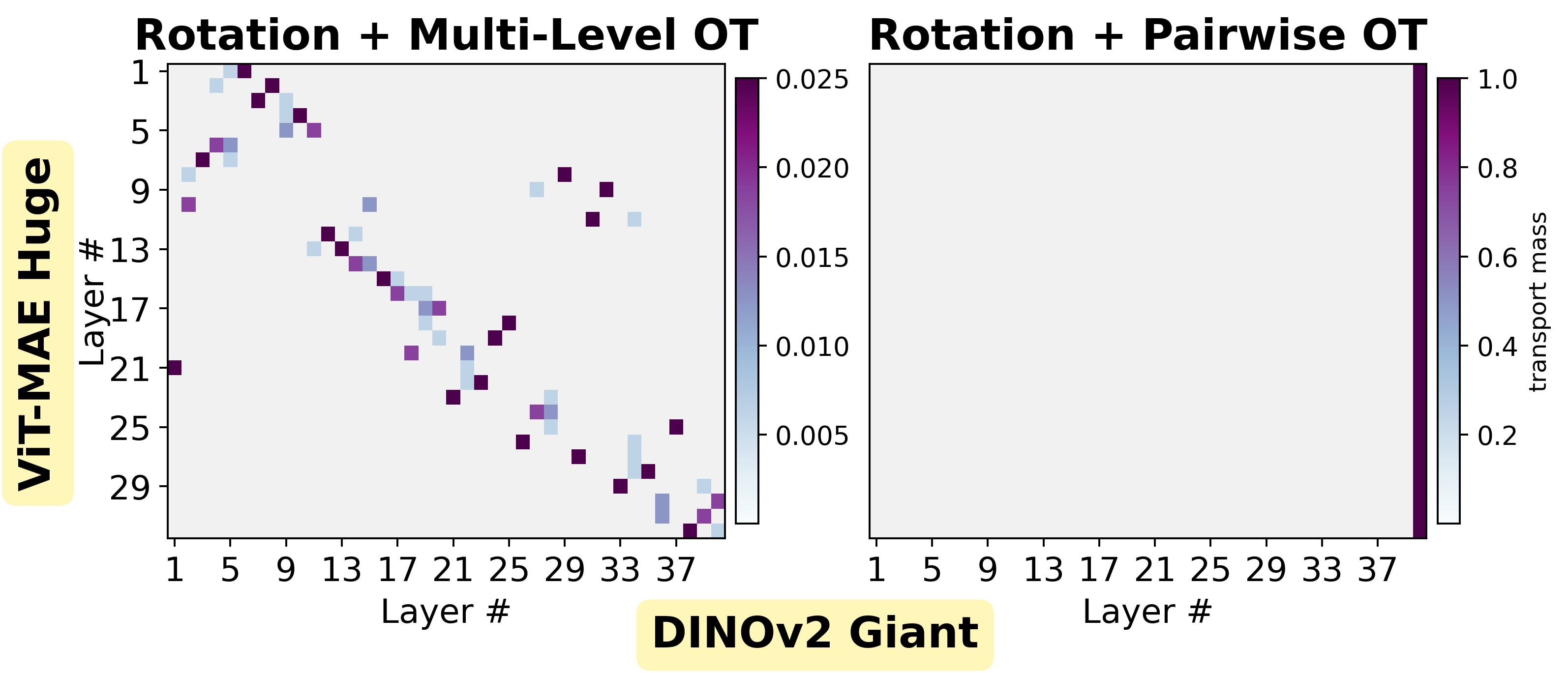}
\caption{\textbf{Transport plans for vision model alignment.} ViT-MAE Huge $\leftrightarrow$ DINOv2 Giant (a) without rotation (MOT) and (b) with rotation augmentation (MOT+R).}
\label{fig:vit-mae-huge-dinov2-giant}
\end{figure}

\newpage
\section{Representation Similarity between Vision Models and Vision Cortex}
\label{sec:FMRIModelsExtraResults}
\textbf{Experimental Setup}. We assess representational similarity between human visual cortex and vision transformers by comparing fMRI responses with model activations elicited by the same set of stimuli. Specifically, we use responses from the Natural Scenes Dataset (\citep{allen2022massive}), focusing on 1,000 shared images viewed by participants in the fMRI experiment. The same images are presented to pretrained vision transformers, and layer-wise representations are extracted by averaging patch embeddings across each input. Following the approach used in the cortex–cortex analysis, we treat distinct visual areas (V1–V4) as “layers” and individual voxels as “neurons.” We then compute both MOT and its rotation-augmented variant (MOT+R) to align cortical responses with model representations. This design enables a direct comparison of hierarchical organization across biological and artificial systems under matched visual input.

\begin{table}[H]
\centering
\small
\begin{tabular}{@{}ll S S S S@{}}
\toprule
\multicolumn{1}{c}{Model 1} & \multicolumn{1}{c}{Model 2} & \multicolumn{1}{c}{MOT Metric} & \multicolumn{1}{c}{Pairwise Best OT} & \multicolumn{1}{c}{MOT + R} & \multicolumn{1}{c}{Pairwise Best + R} \\
\midrule
Subject & DINOv2 Base & 0.092 & 0.084 & \textbf{0.145} & 0.094 \\
Subject & DINOv2 Giant & 0.090 & 0.071 & \textbf{0.163} & 0.110 \\
Subject & ViT-MAE Base & 0.127 & 0.114 & \textbf{0.151} & 0.121 \\
Subject & ViT-MAE Huge & 0.072 & 0.099 & \textbf{0.154} & 0.122 \\
\bottomrule
\end{tabular}
\caption{Results on MOT metric vs. baselines (test split).}
\label{tab:fmrimodels}
\end{table}

\begin{figure}[H]
\centering
\includegraphics[width=0.48\textwidth]{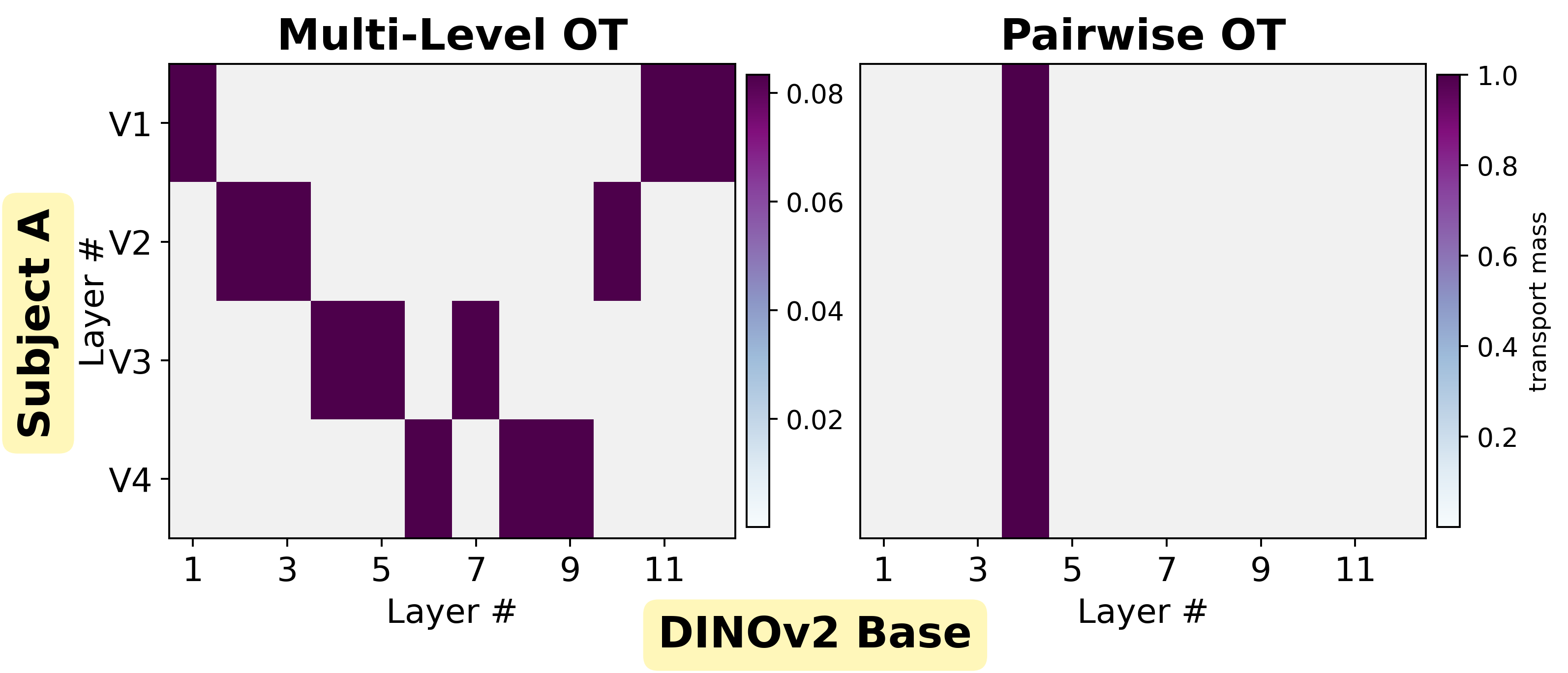}\hfill
\includegraphics[width=0.48\textwidth]{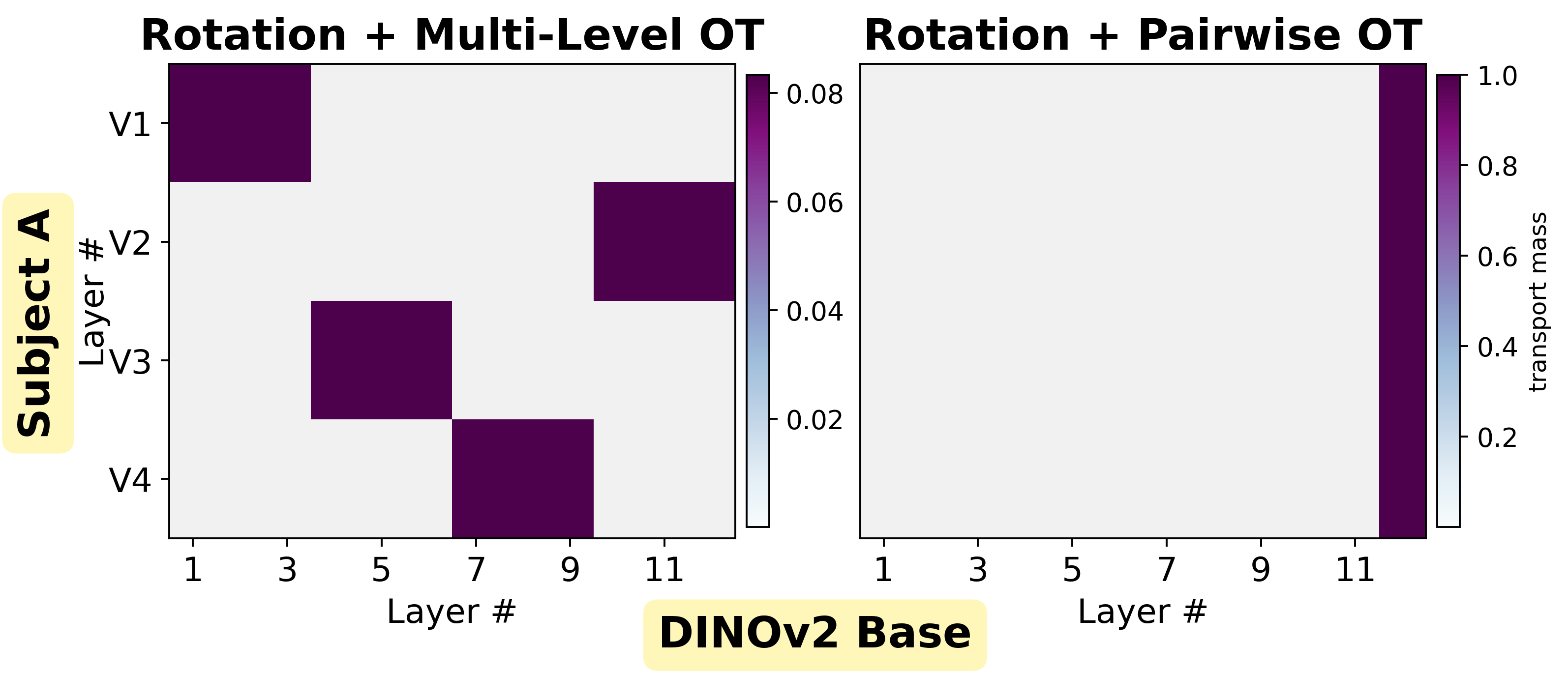}
\caption{Subject A $\leftrightarrow$ DINOv2 Base (a) without rotation (MOT) and (b) with rotation augmentation (MOT+R).}
\label{fig:subject_1-dinov2-base}
\end{figure}

\begin{figure}[H]
\centering
\includegraphics[width=0.48\textwidth]{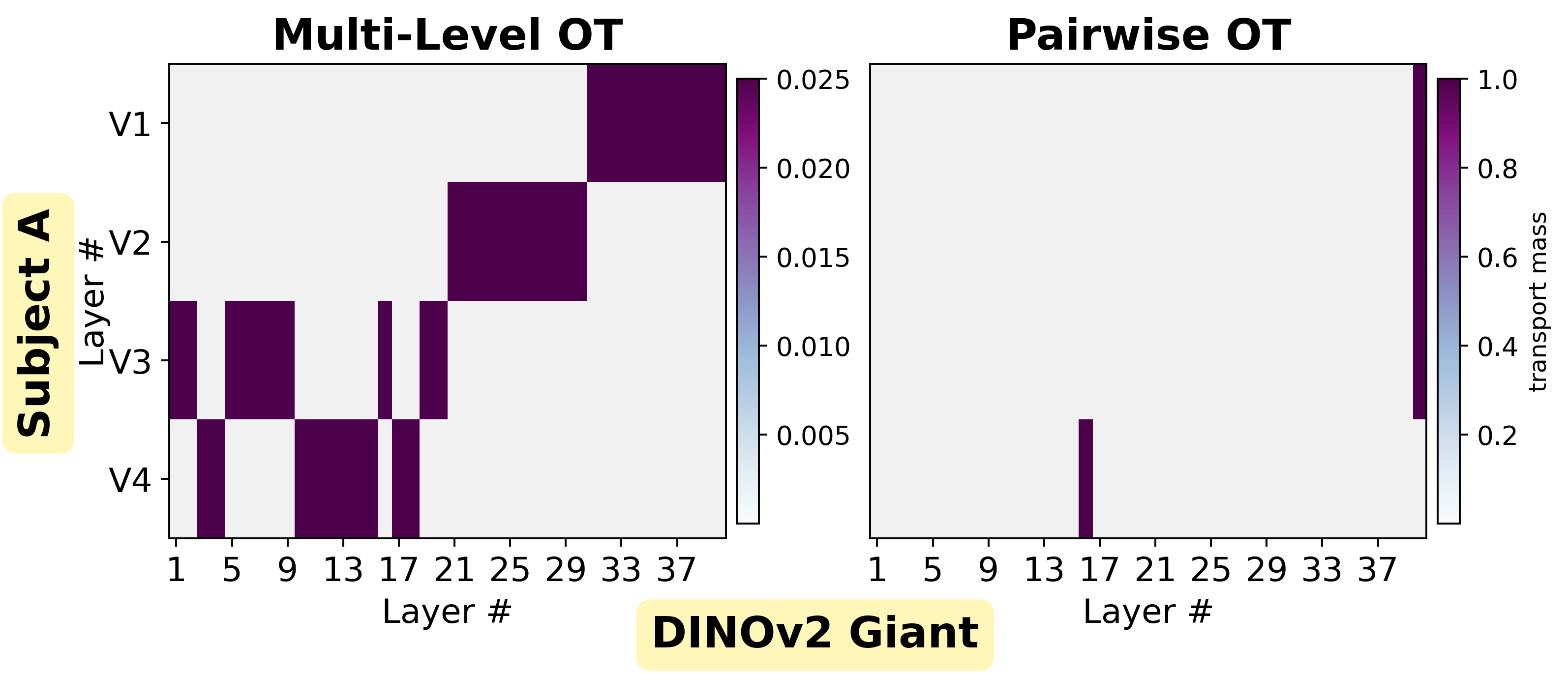}\hfill
\includegraphics[width=0.48\textwidth]{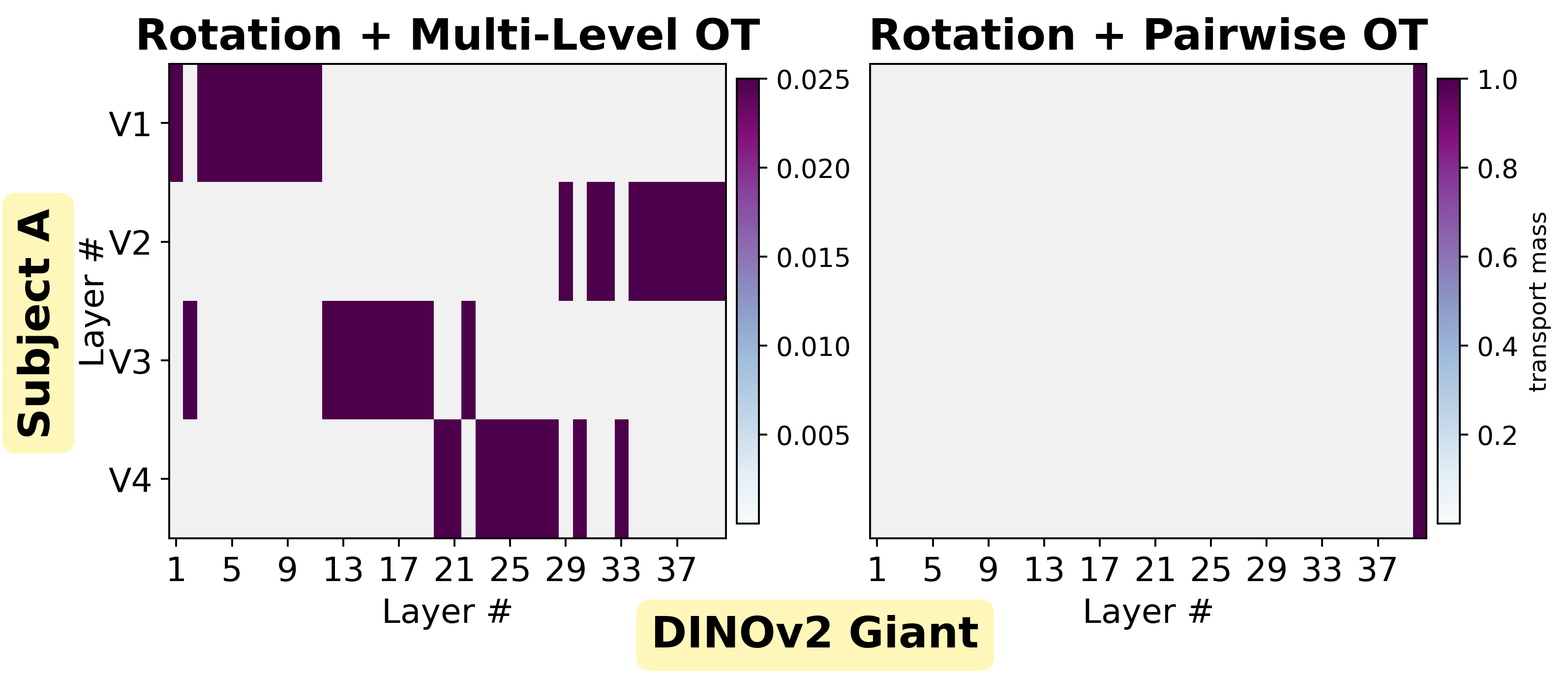}
\caption{Subject A $\leftrightarrow$ DINOv2 Giant (a) without rotation (MOT) and (b) with rotation augmentation (MOT+R).}
\label{fig:subject_1-dinov2-giant}
\end{figure}

\begin{figure}[H]
\centering
\includegraphics[width=0.48\textwidth]{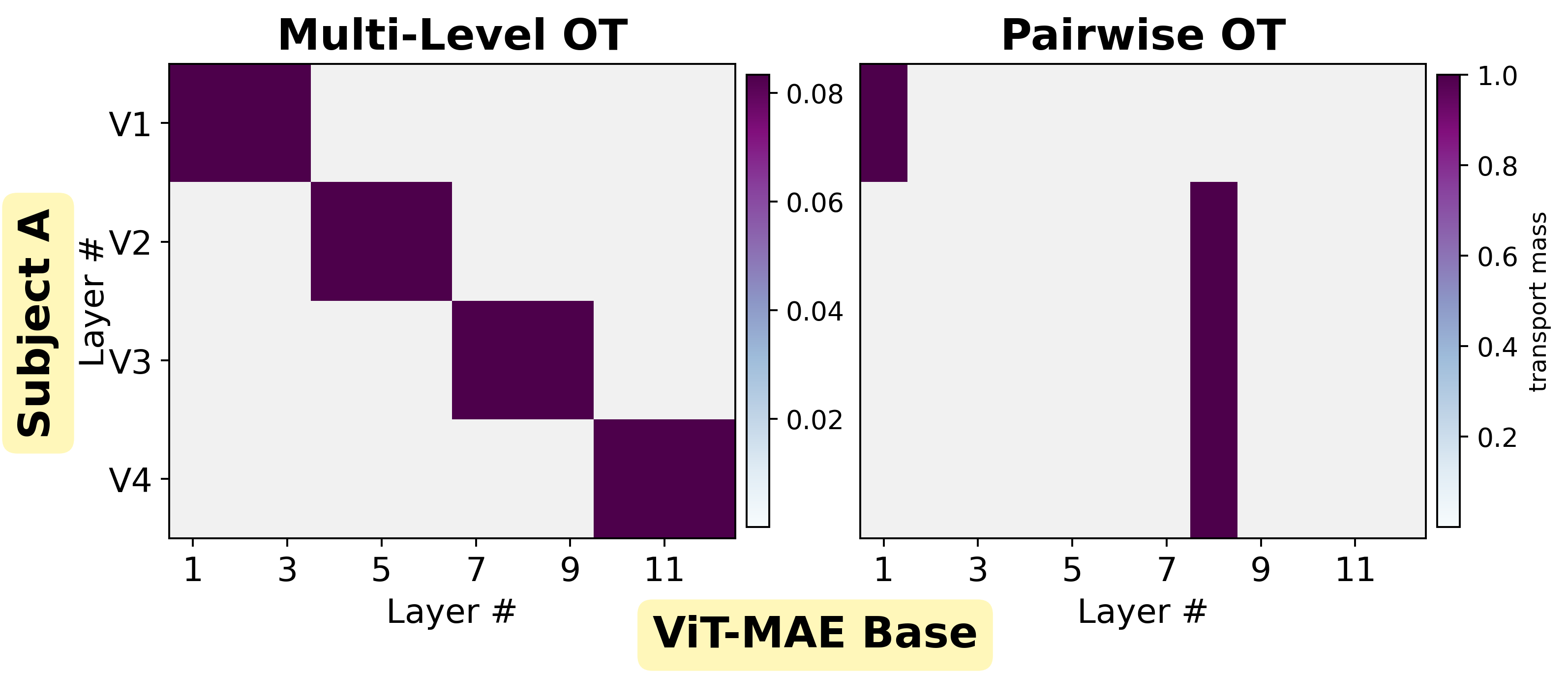}\hfill
\includegraphics[width=0.48\textwidth]{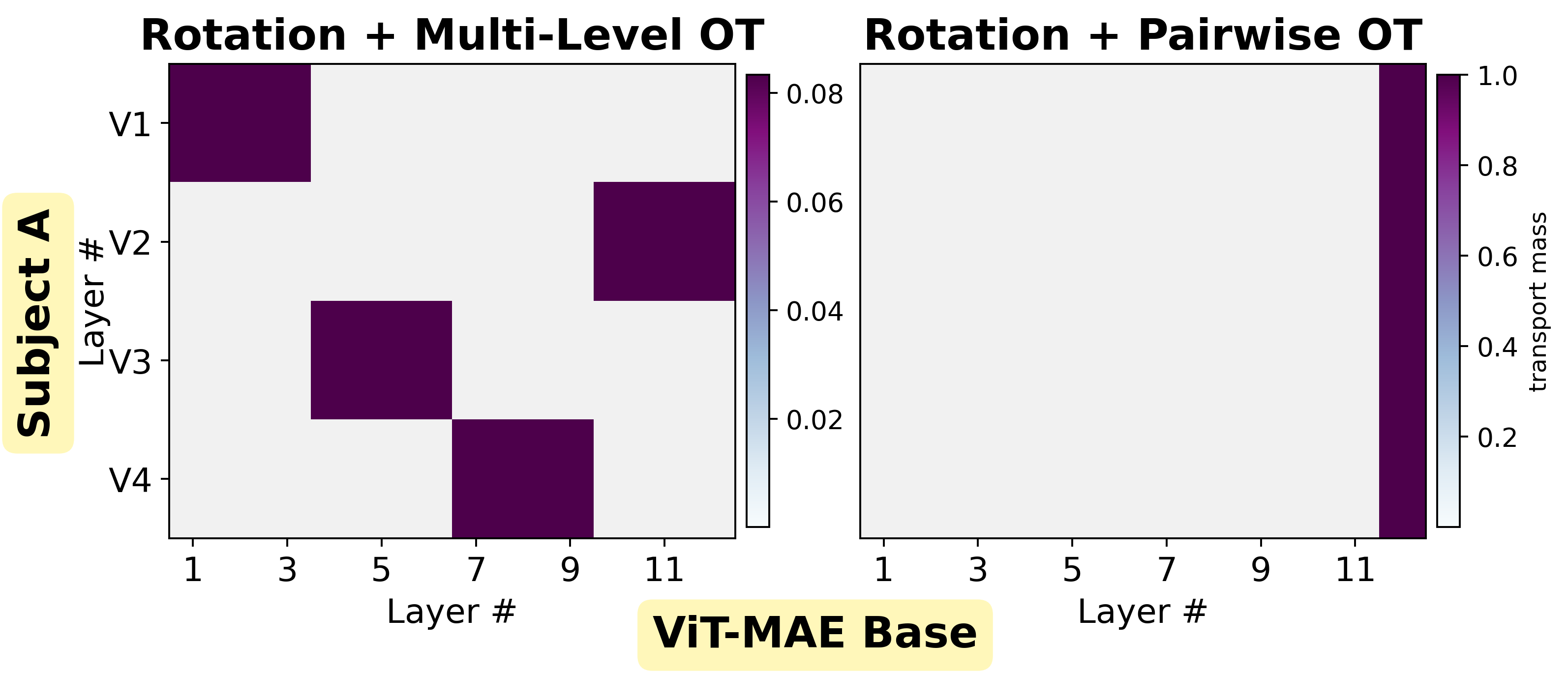}
\caption{Subject A $\leftrightarrow$ ViT-MAE Base (a) without rotation (MOT) and (b) with rotation augmentation (MOT+R).}
\label{fig:subject_1-vit-mae-base}
\end{figure}

\begin{figure}[H]
\centering
\includegraphics[width=0.48\textwidth]{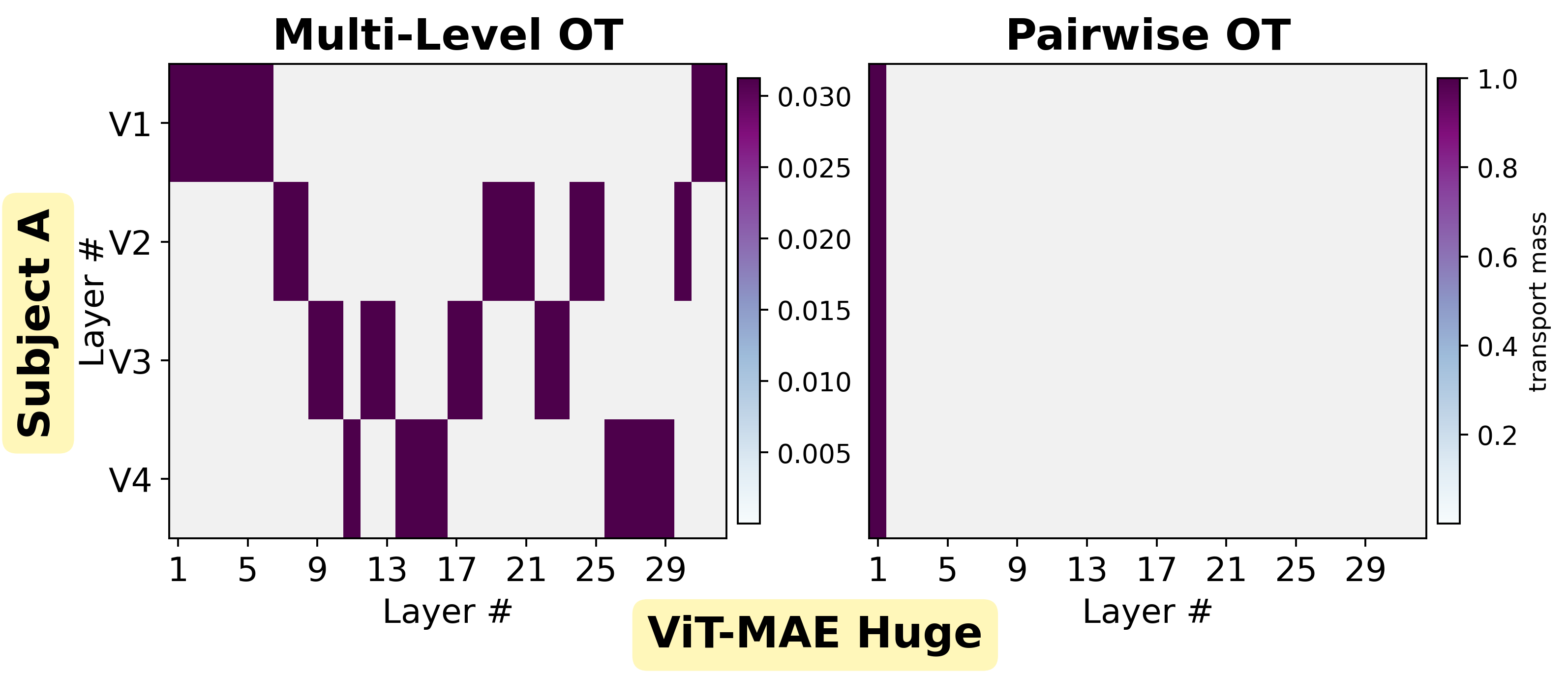}\hfill
\includegraphics[width=0.48\textwidth]{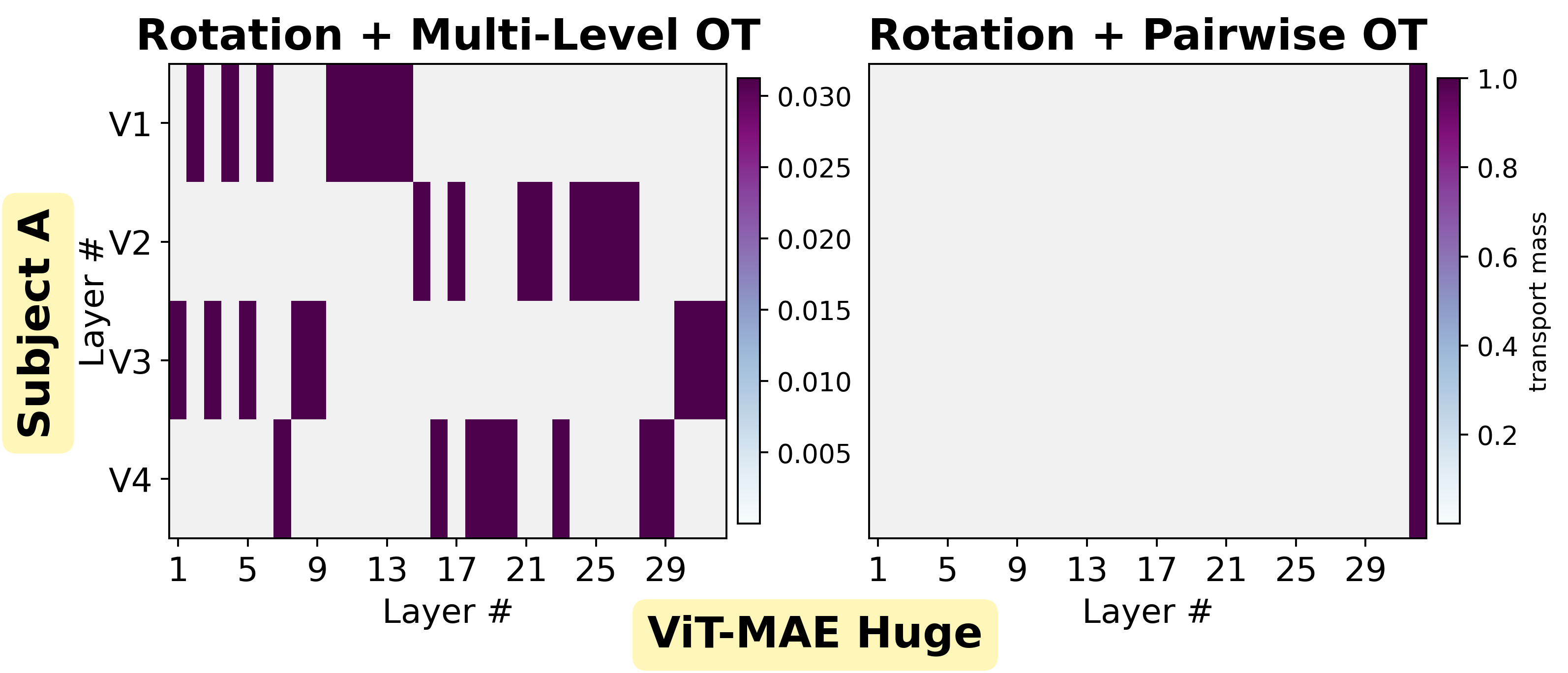}
\caption{Subject A $\leftrightarrow$ ViT-MAE Huge (a) without rotation (MOT) and (b) with rotation augmentation (MOT+R).}
\label{fig:subject_1-vit-mae-huge}
\end{figure}

\begin{figure}[H]
\centering
\includegraphics[width=0.48\textwidth]{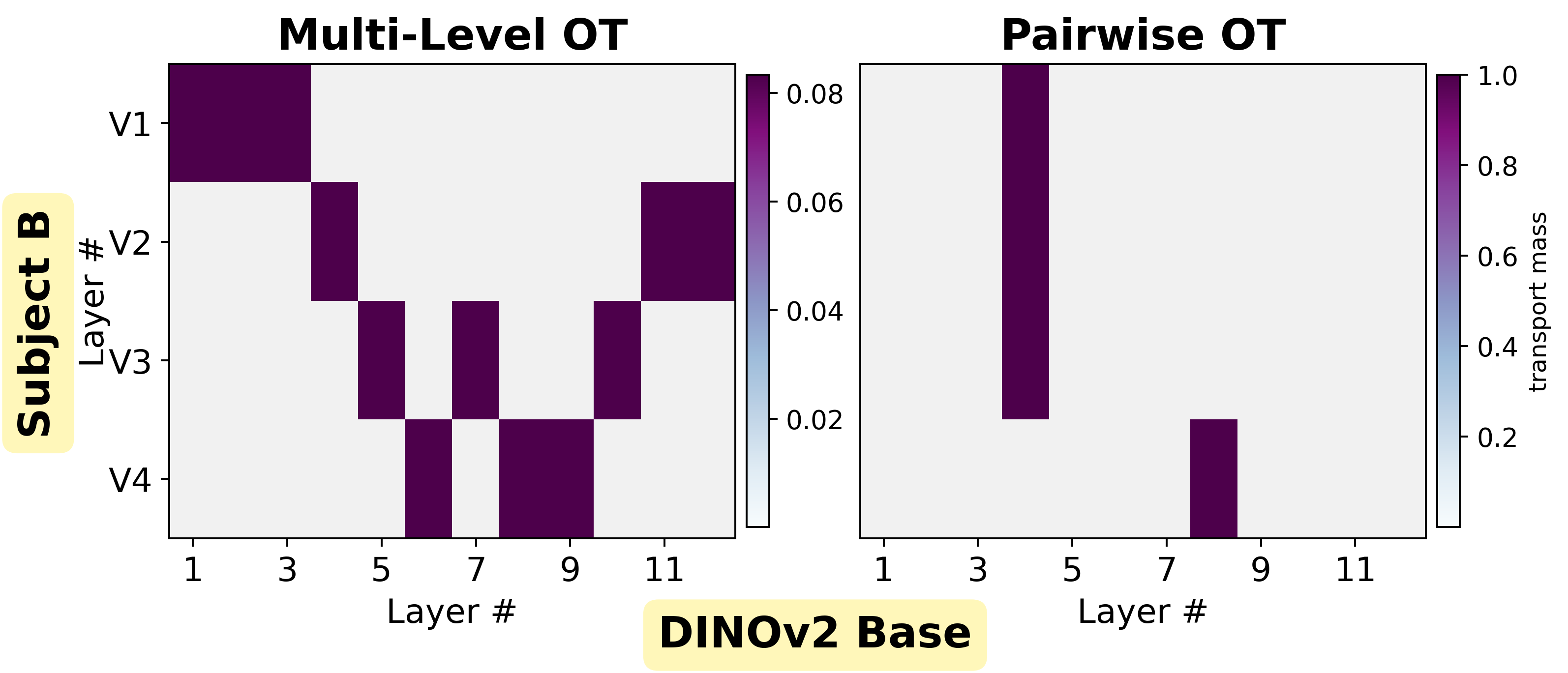}\hfill
\includegraphics[width=0.48\textwidth]{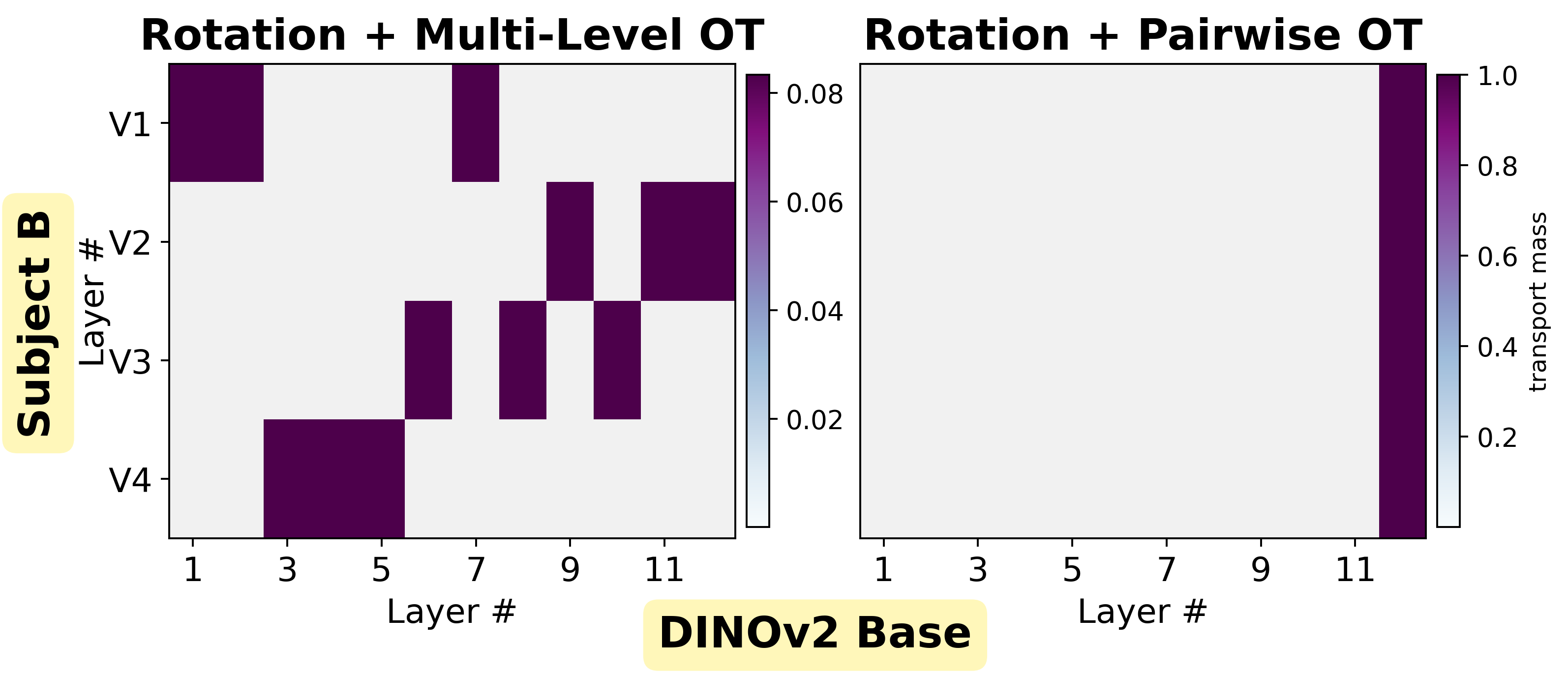}
\caption{Subject B $\leftrightarrow$ DINOv2 Base (a) without rotation (MOT) and (b) with rotation augmentation (MOT+R).}
\label{fig:subject_2-dinov2-base}
\end{figure}

\begin{figure}[H]
\centering
\includegraphics[width=0.48\textwidth]{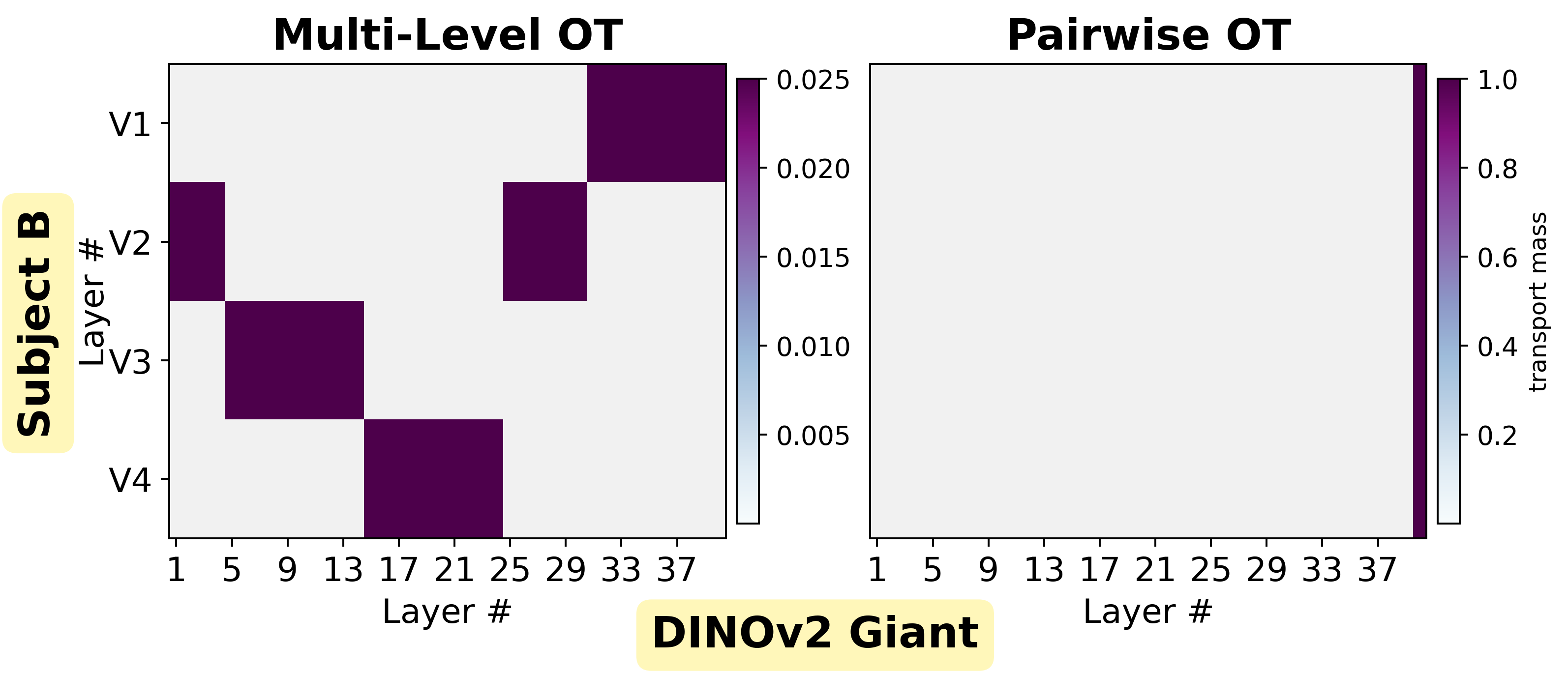}\hfill
\includegraphics[width=0.48\textwidth]{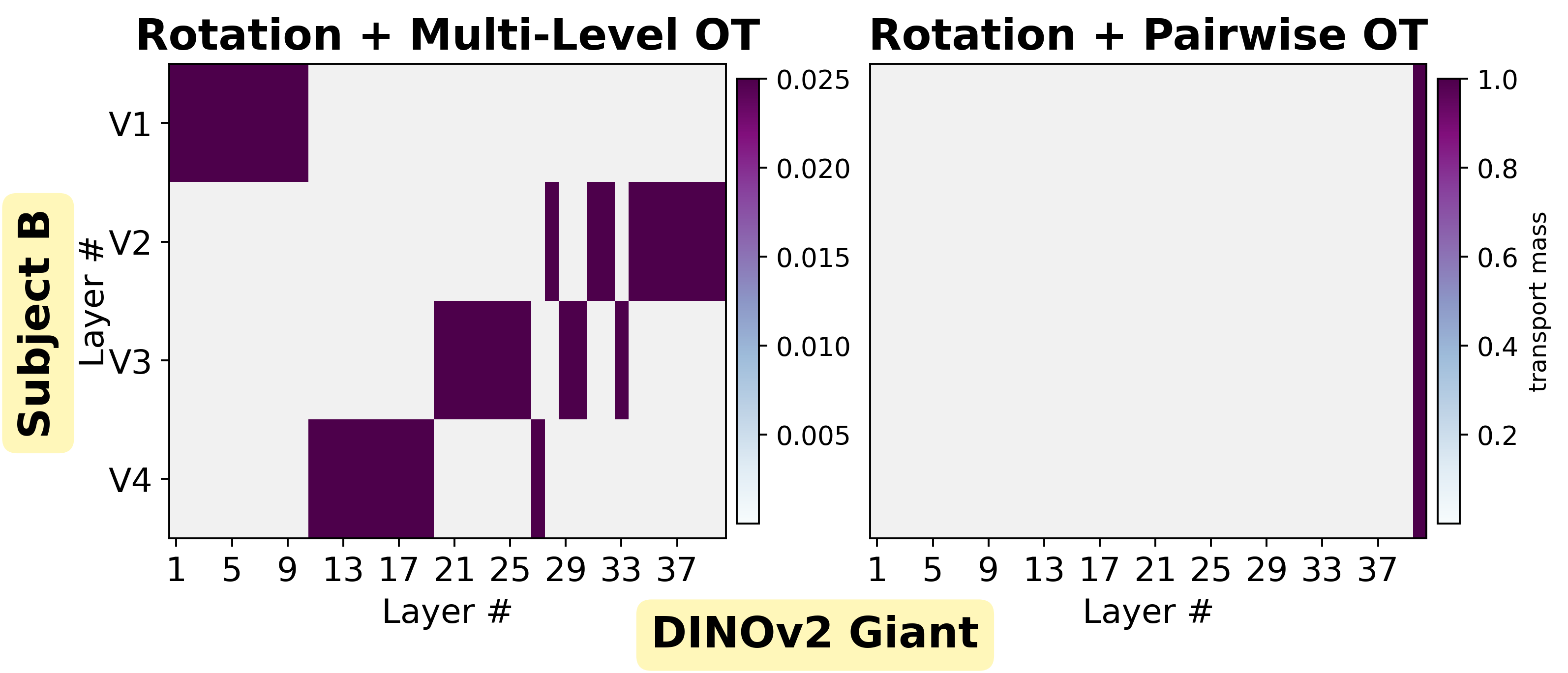}
\caption{Subject B $\leftrightarrow$ DINOv2 Giant (a) without rotation (MOT) and (b) with rotation augmentation (MOT+R).}
\label{fig:subject_2-dinov2-giant}
\end{figure}

\begin{figure}[H]
\centering
\includegraphics[width=0.48\textwidth]{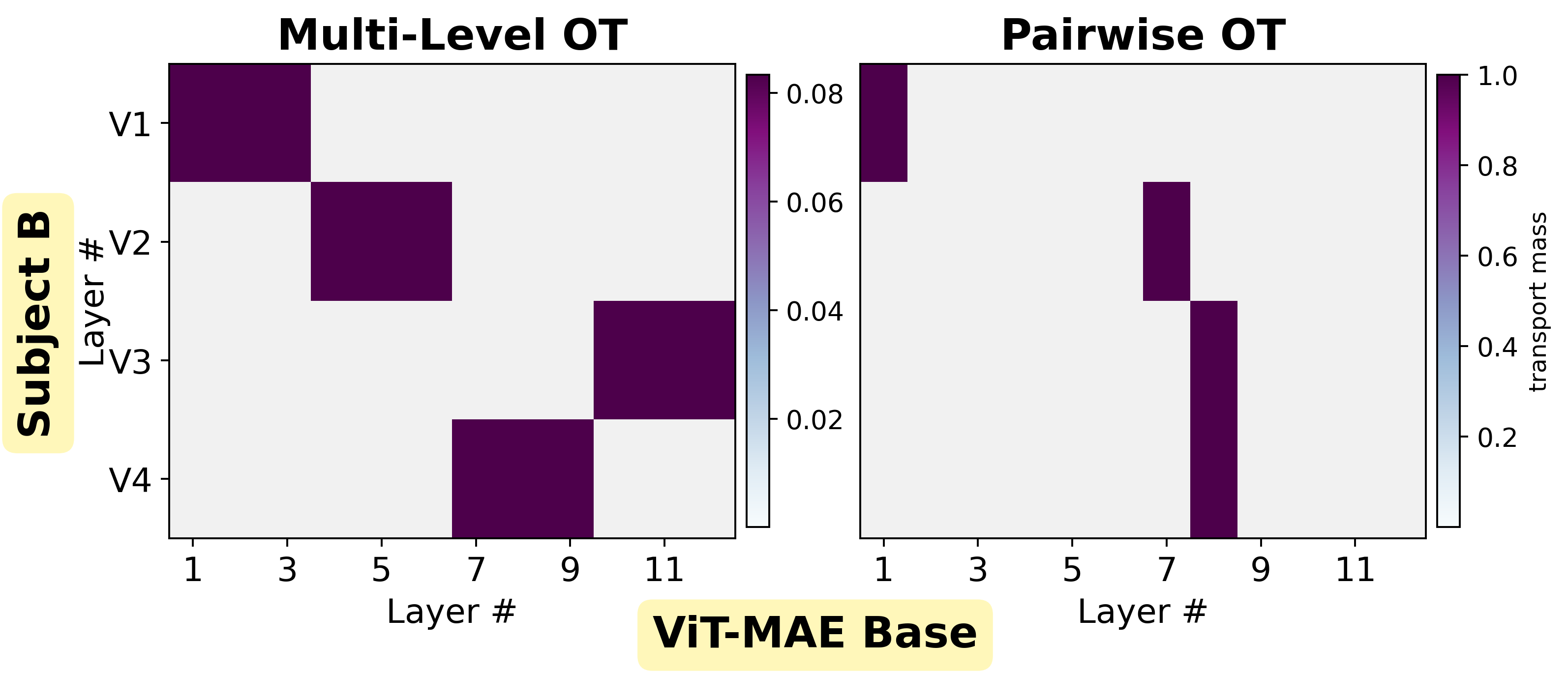}\hfill
\includegraphics[width=0.48\textwidth]{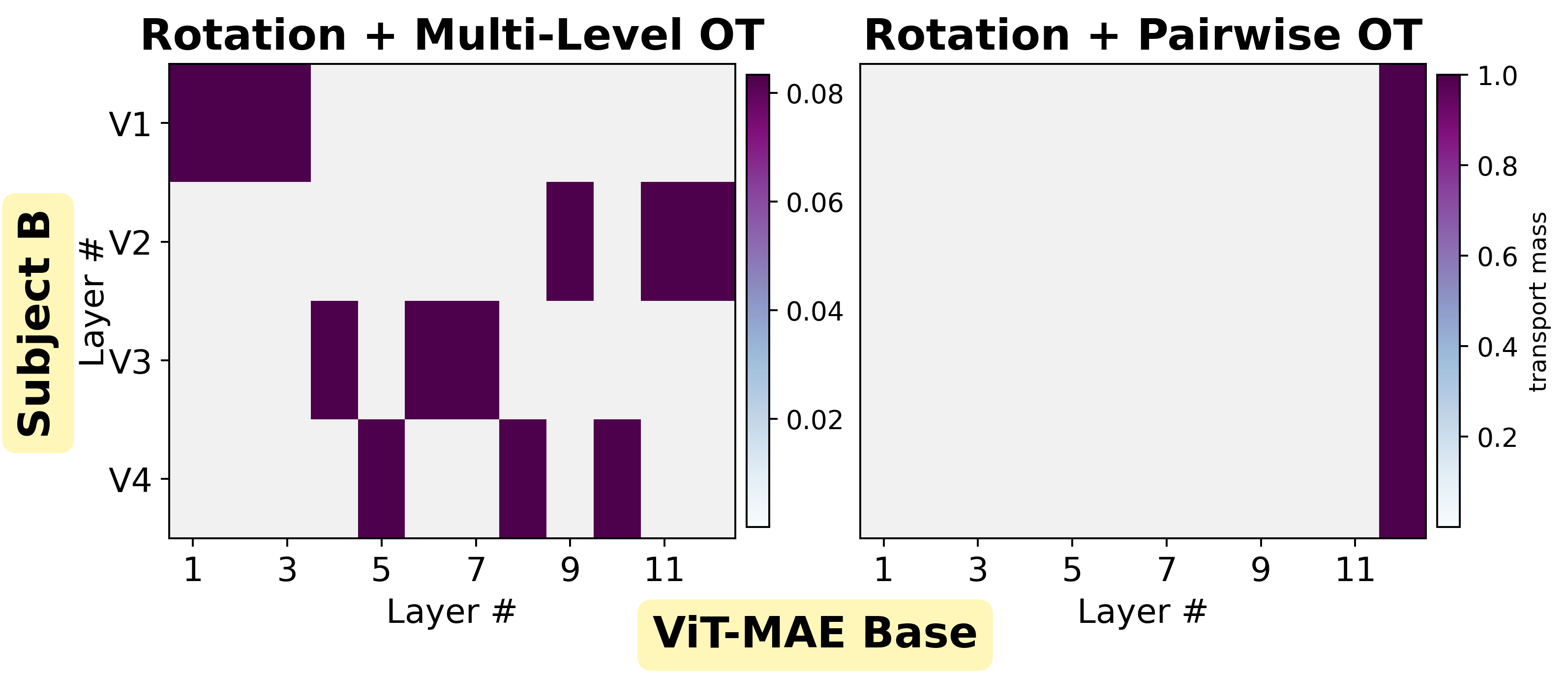}
\caption{Subject B $\leftrightarrow$ ViT-MAE Base (a) without rotation (MOT) and (b) with rotation augmentation (MOT+R).}
\label{fig:subject_2-vit-mae-base}
\end{figure}

\begin{figure}[H]
\centering
\includegraphics[width=0.48\textwidth]{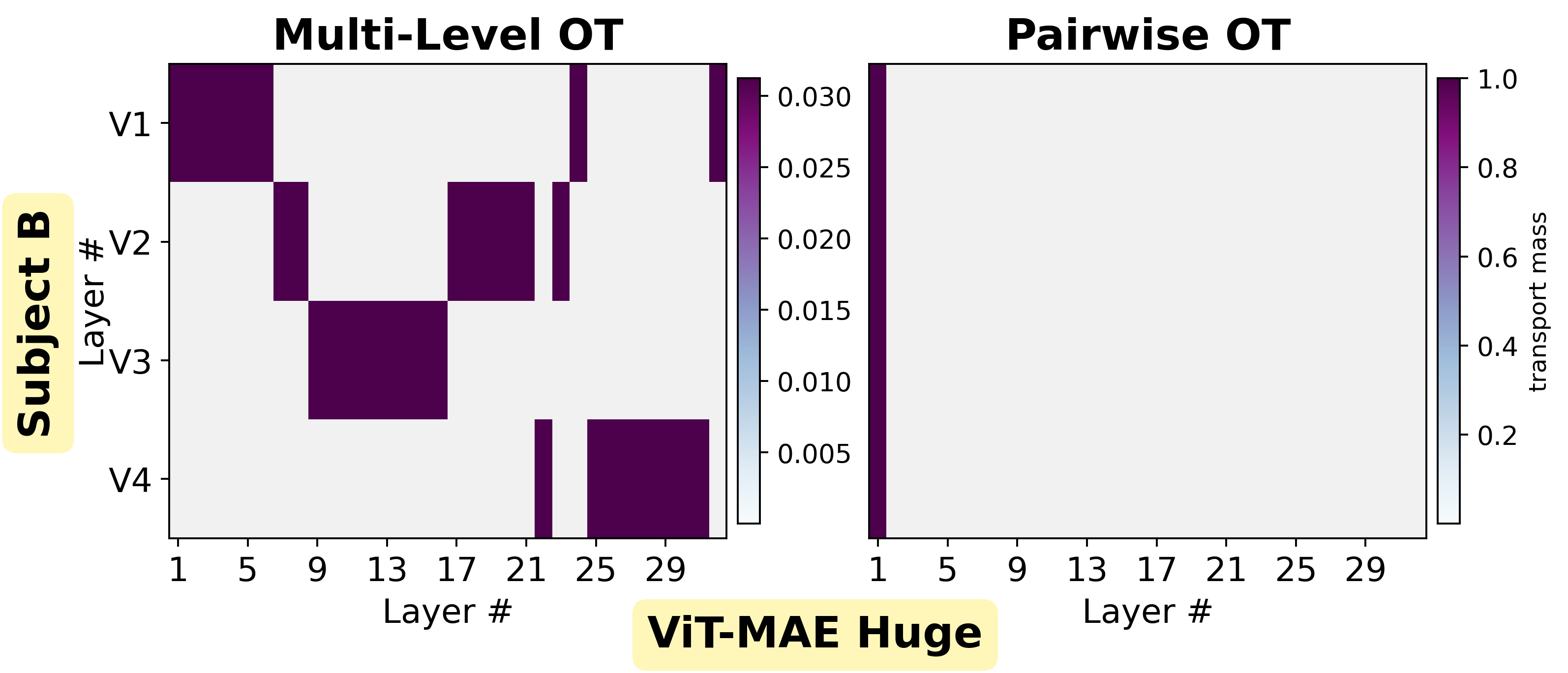}\hfill
\includegraphics[width=0.48\textwidth]{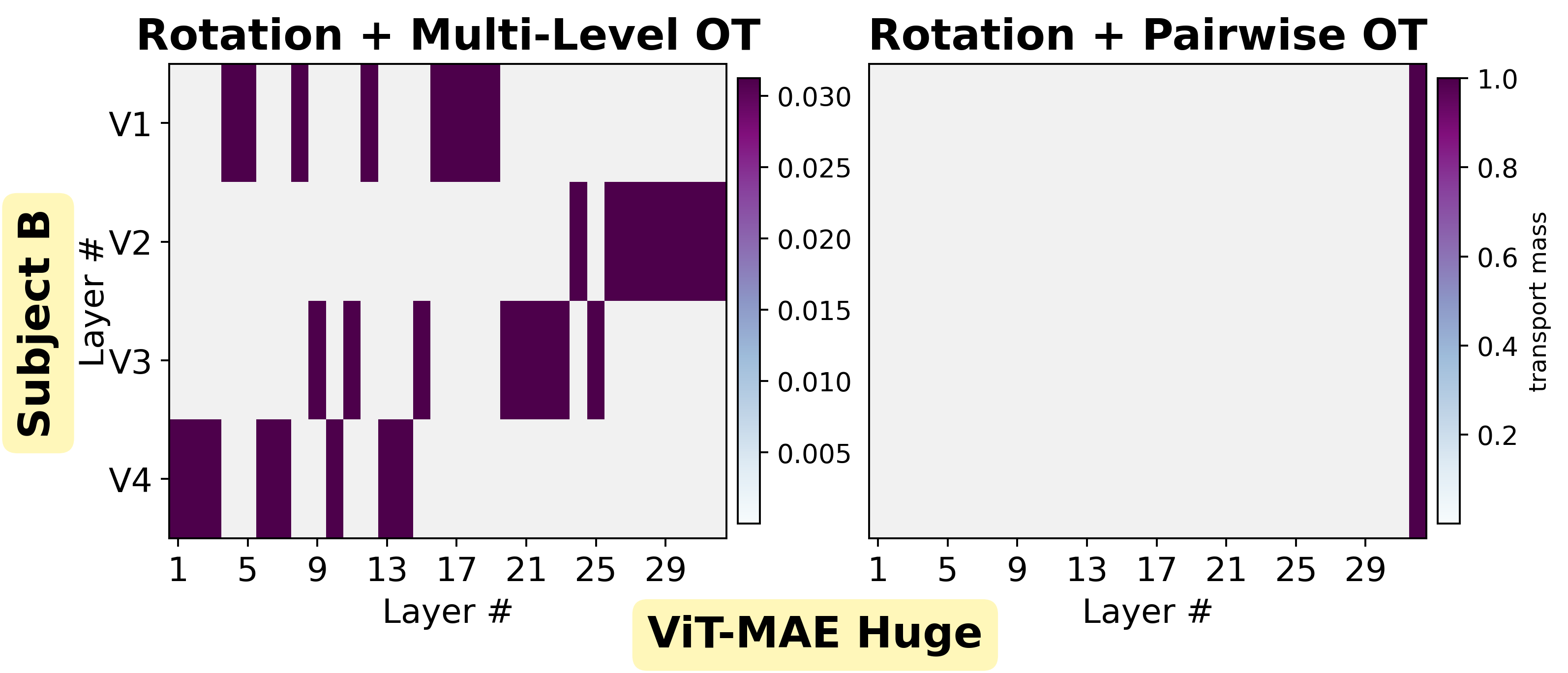}
\caption{Subject B $\leftrightarrow$ ViT-MAE Huge (a) without rotation (MOT) and (b) with rotation augmentation (MOT+R).}
\label{fig:subject_2-vit-mae-huge}
\end{figure}

\begin{figure}[H]
\centering
\includegraphics[width=0.48\textwidth]{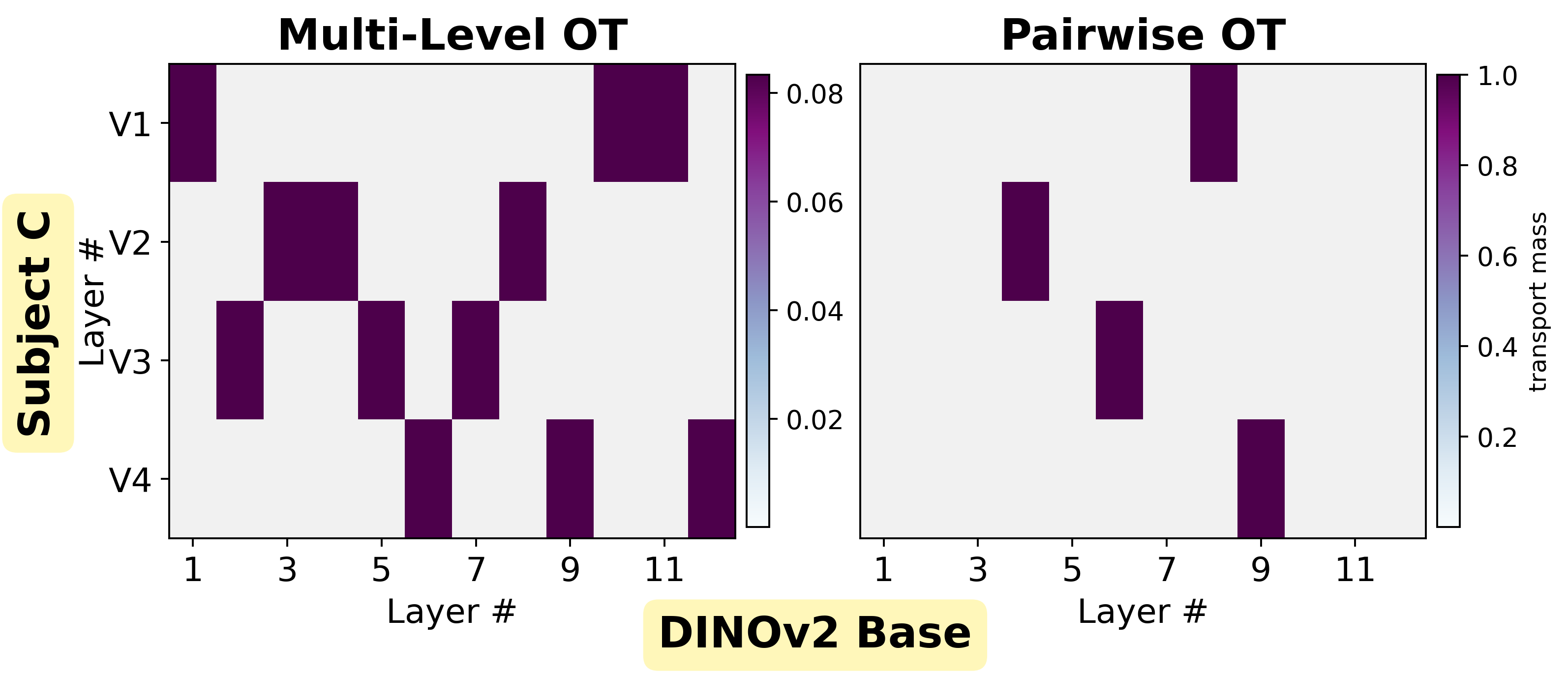}\hfill
\includegraphics[width=0.48\textwidth]{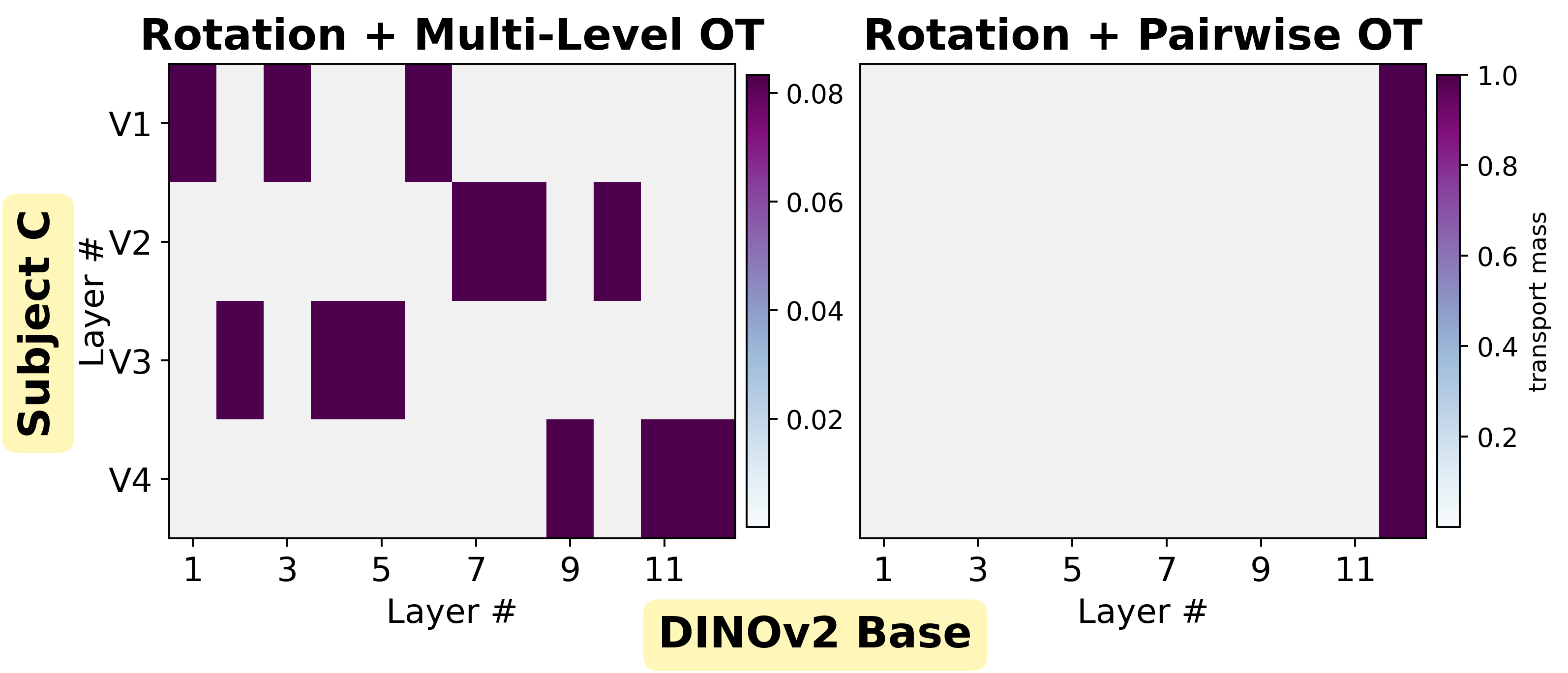}
\caption{Subject C $\leftrightarrow$ DINOv2 Base (a) without rotation (MOT) and (b) with rotation augmentation (MOT+R).}
\label{fig:subject_5-dinov2-base}
\end{figure}

\begin{figure}[H]
\centering
\includegraphics[width=0.48\textwidth]{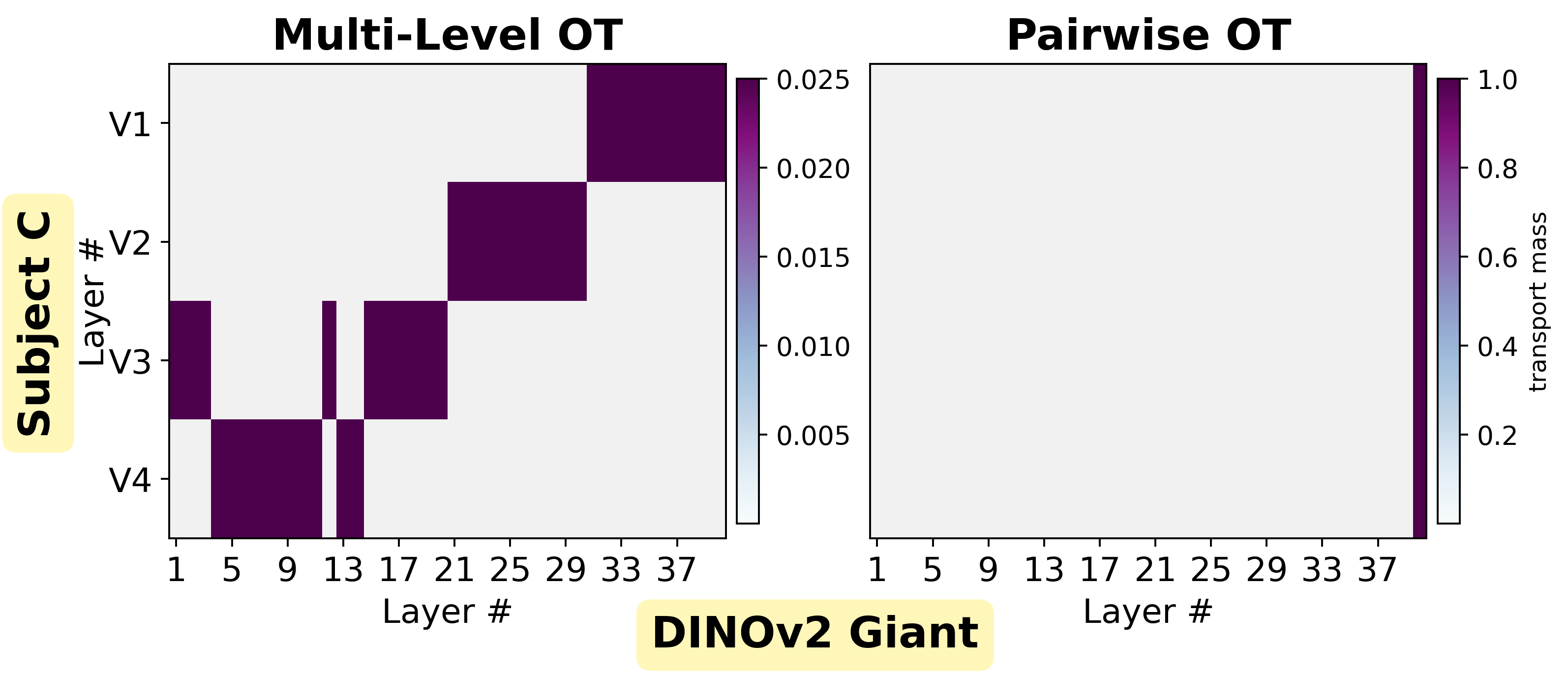}\hfill
\includegraphics[width=0.48\textwidth]{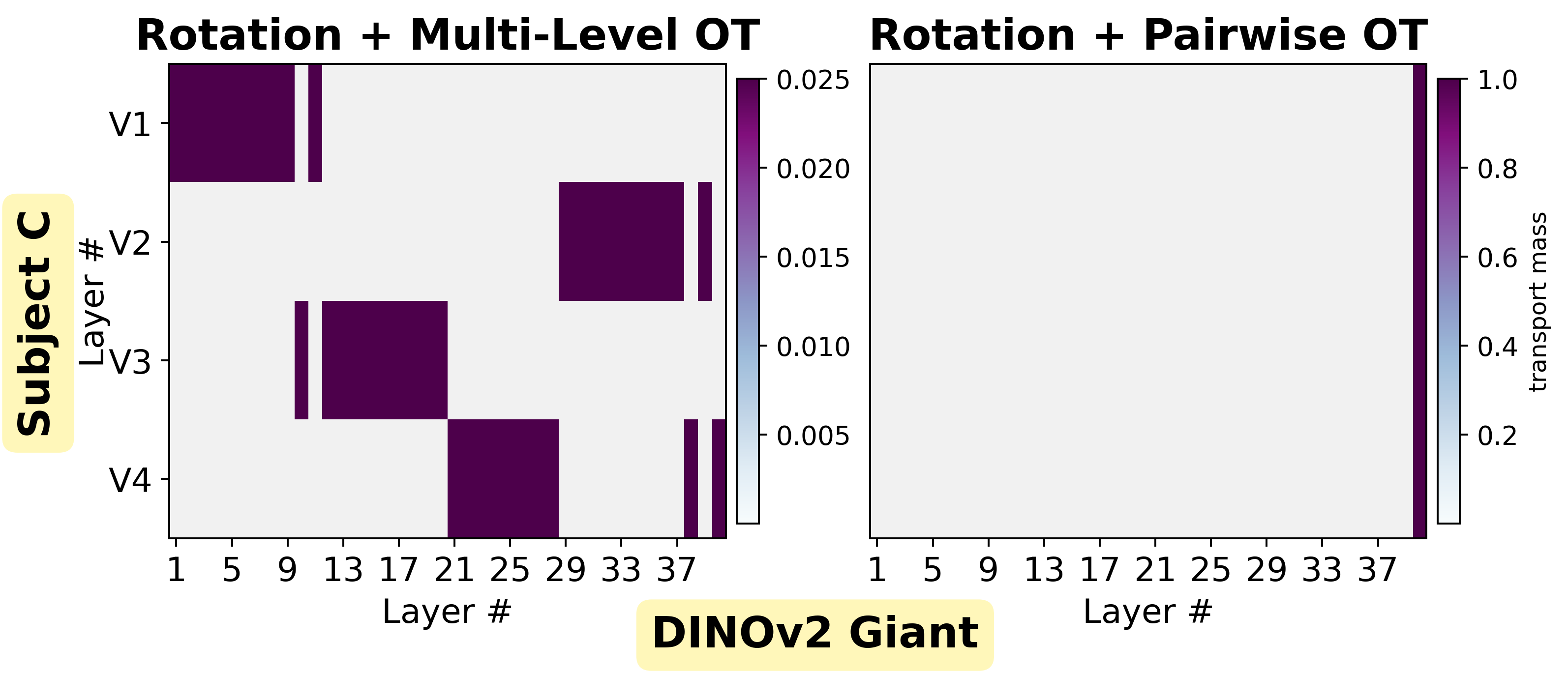}
\caption{Subject C $\leftrightarrow$ DINOv2 Giant (a) without rotation (MOT) and (b) with rotation augmentation (MOT+R).}
\label{fig:subject_5-dinov2-giant}
\end{figure}

\begin{figure}[H]
\centering
\includegraphics[width=0.48\textwidth]{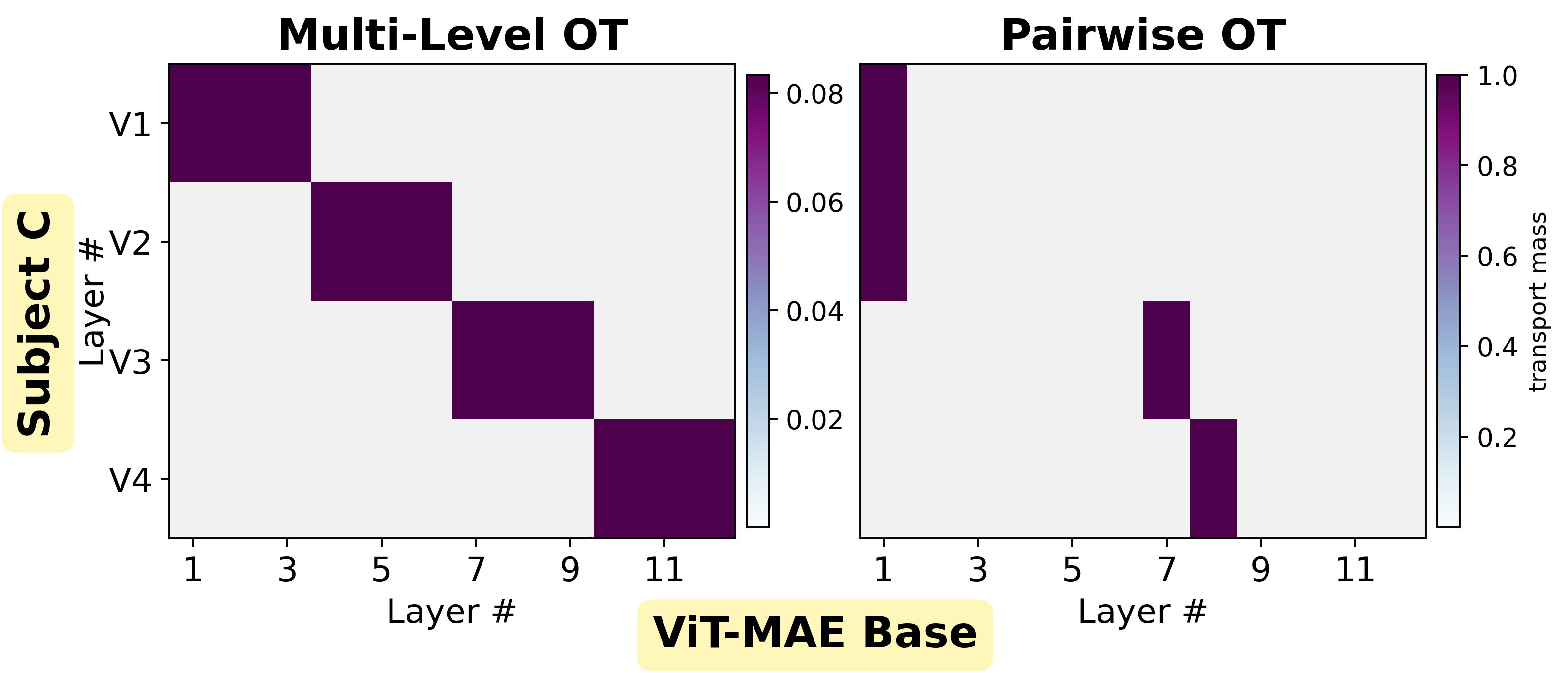}\hfill
\includegraphics[width=0.48\textwidth]{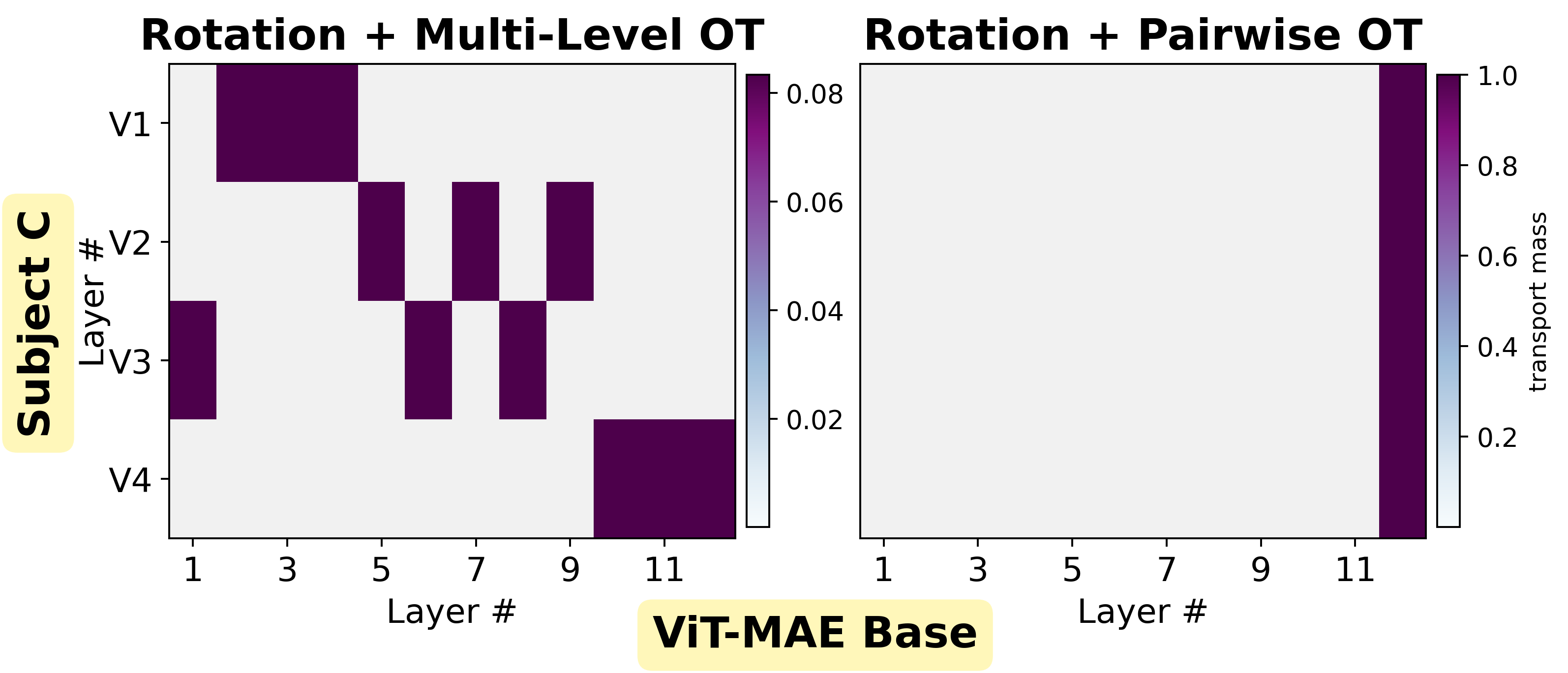}
\caption{Subject C $\leftrightarrow$ ViT-MAE Base (a) without rotation (MOT) and (b) with rotation augmentation (MOT+R).}
\label{fig:subject_5-vit-mae-base}
\end{figure}

\begin{figure}[H]
\centering
\includegraphics[width=0.48\textwidth]{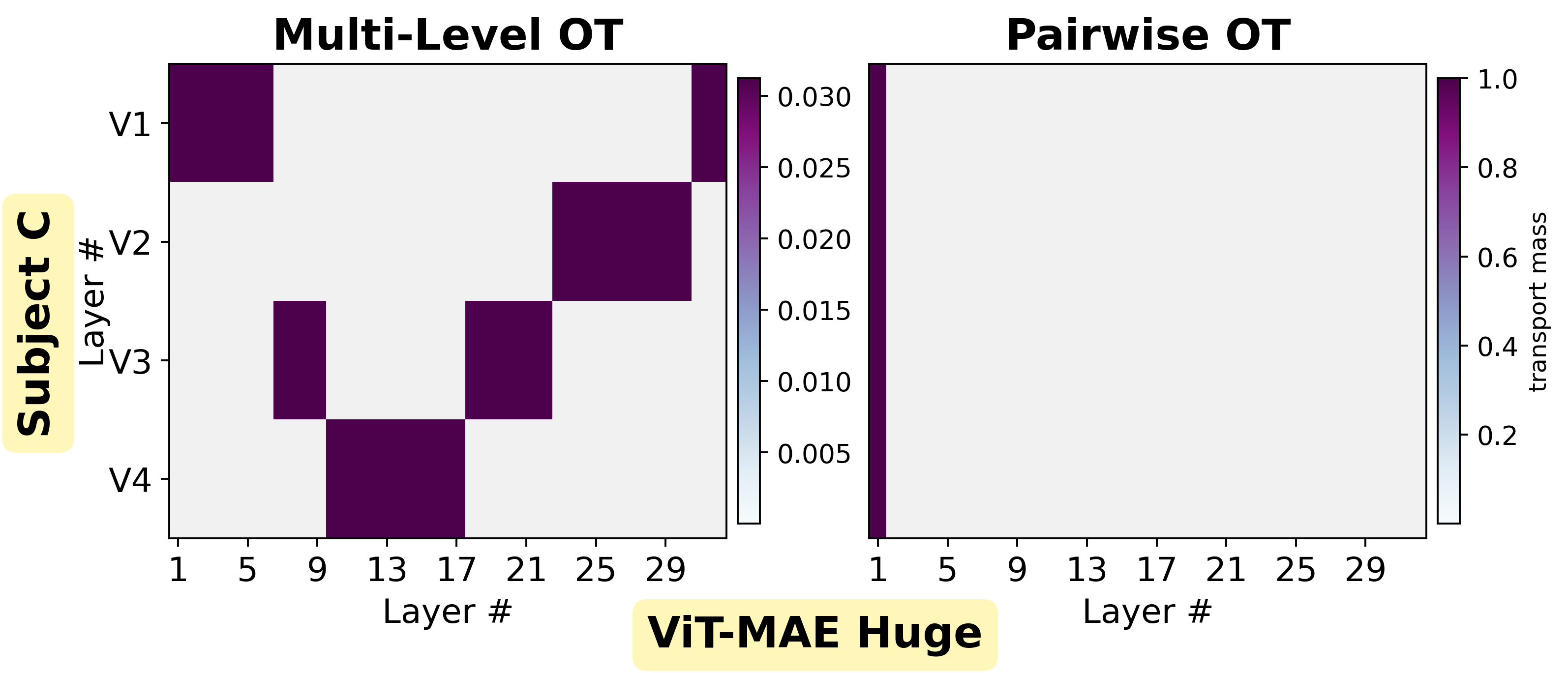}\hfill
\includegraphics[width=0.48\textwidth]{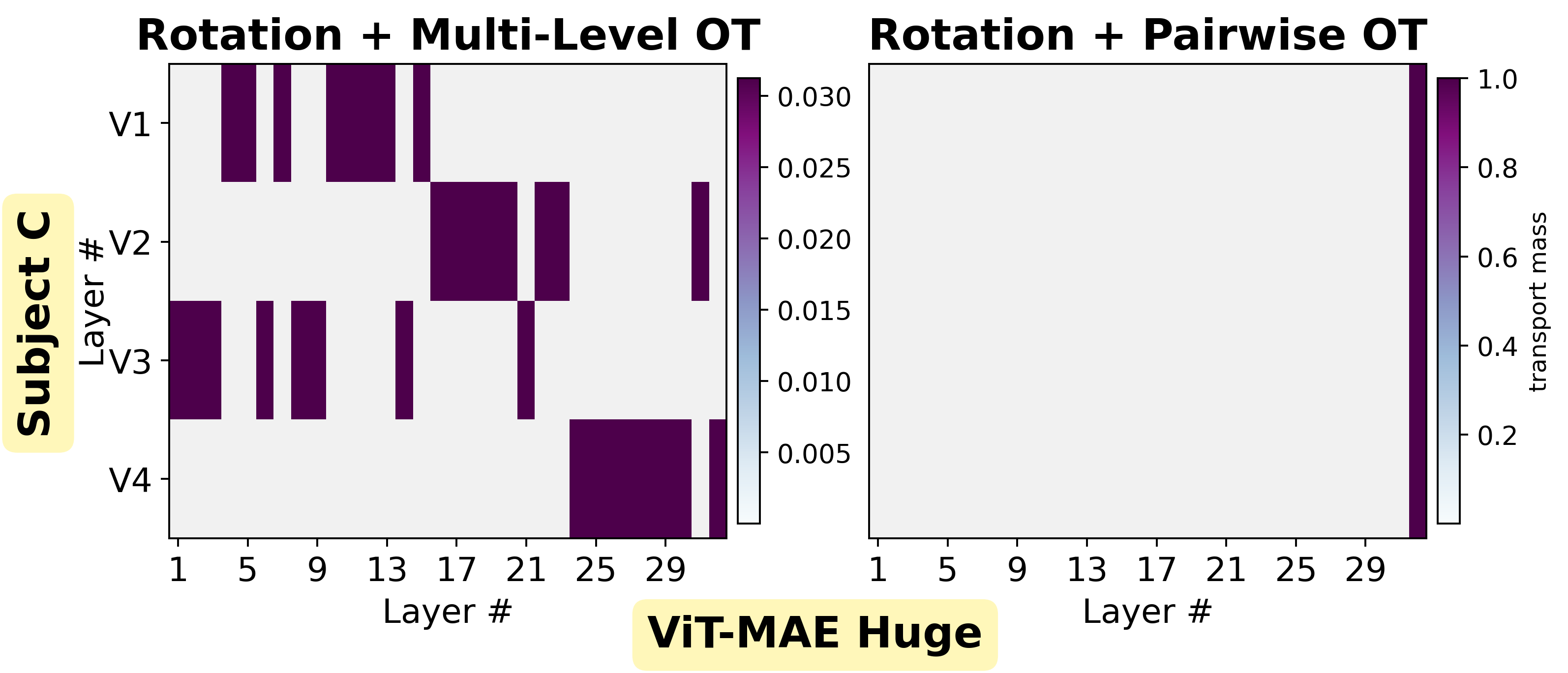}
\caption{Subject C $\leftrightarrow$ ViT-MAE Huge (a) without rotation (MOT) and (b) with rotation augmentation (MOT+R).}
\label{fig:subject_5-vit-mae-huge}
\end{figure}

\begin{figure}[H]
\centering
\includegraphics[width=0.48\textwidth]{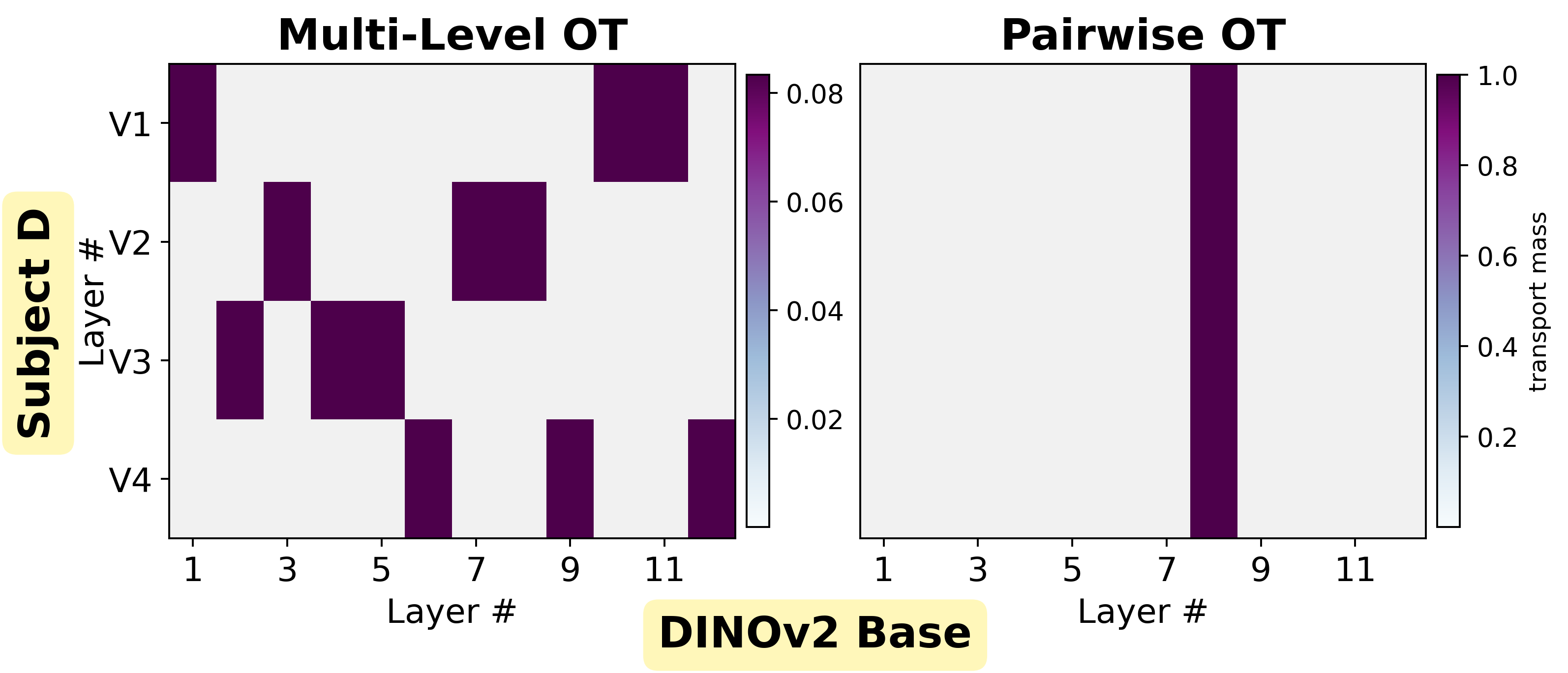}\hfill
\includegraphics[width=0.48\textwidth]{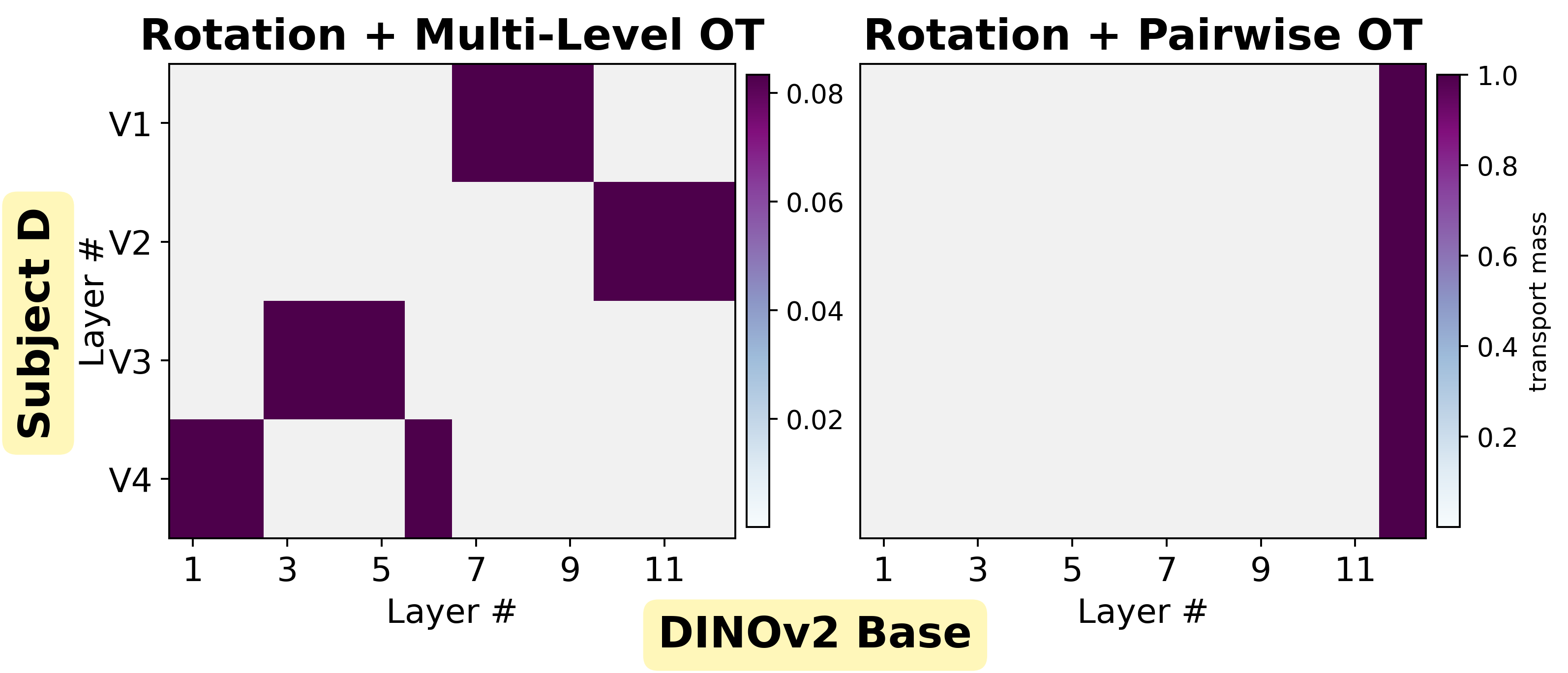}
\caption{Subject D $\leftrightarrow$ DINOv2 Base (a) without rotation (MOT) and (b) with rotation augmentation (MOT+R).}
\label{fig:subject_7-dinov2-base}
\end{figure}

\begin{figure}[H]
\centering
\includegraphics[width=0.48\textwidth]{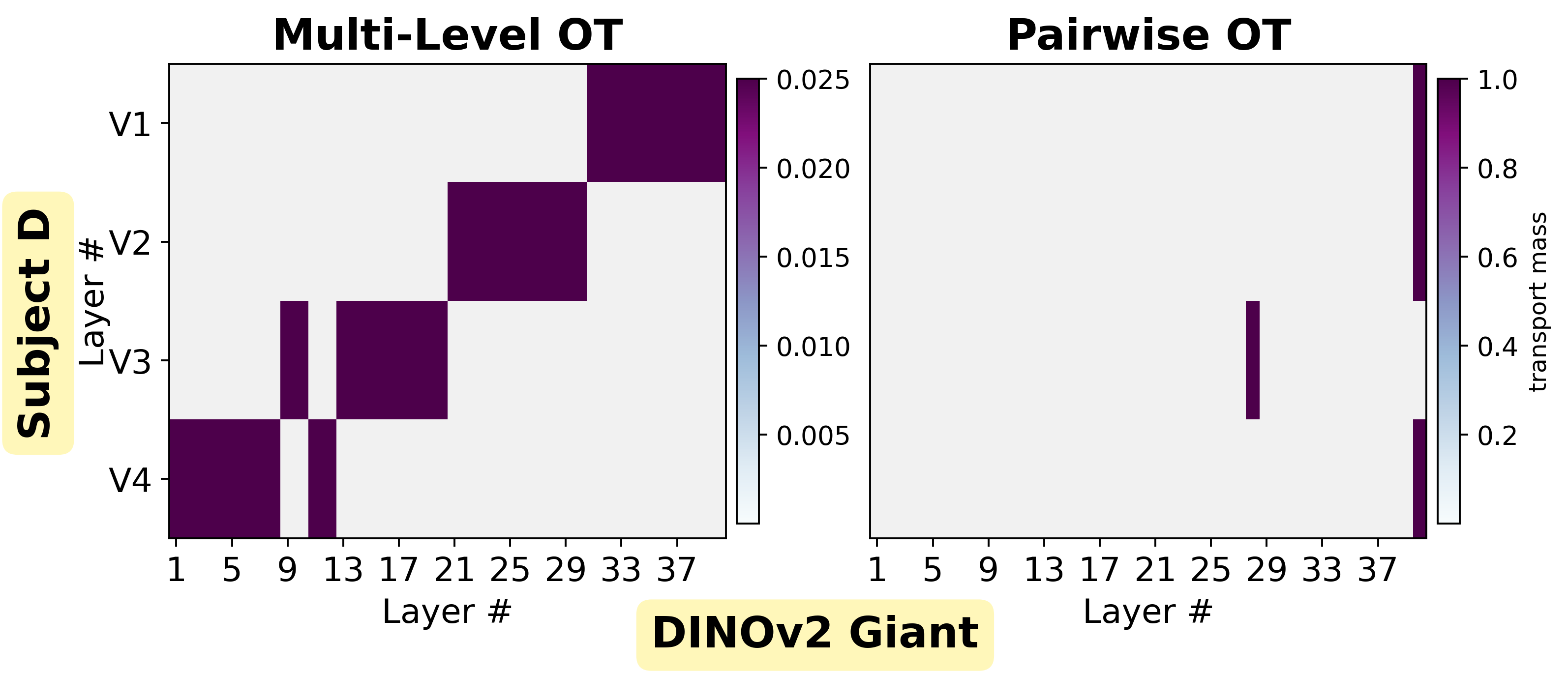}\hfill
\includegraphics[width=0.48\textwidth]{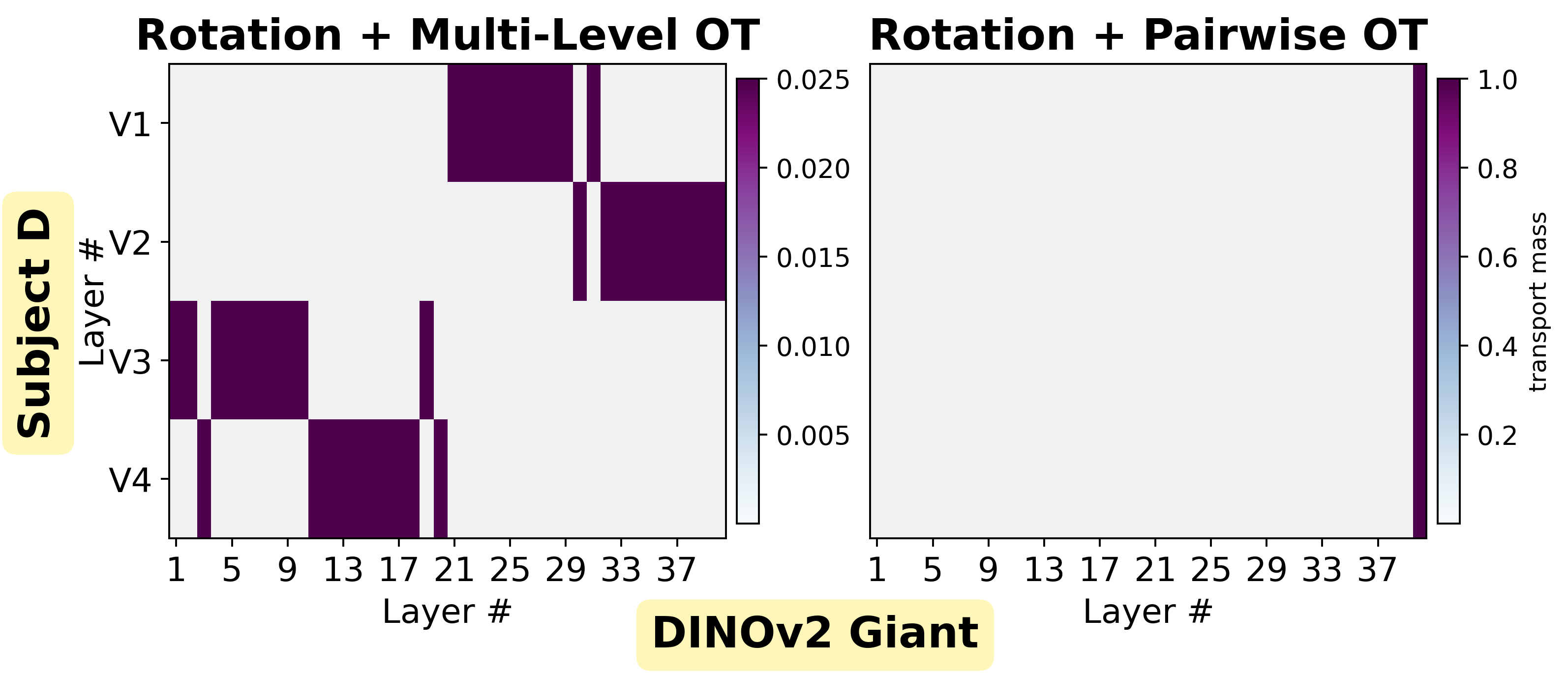}
\caption{Subject D $\leftrightarrow$ DINOv2 Giant (a) without rotation (MOT) and (b) with rotation augmentation (MOT+R).}
\label{fig:subject_7-dinov2-giant}
\end{figure}

\begin{figure}[H]
\centering
\includegraphics[width=0.48\textwidth]{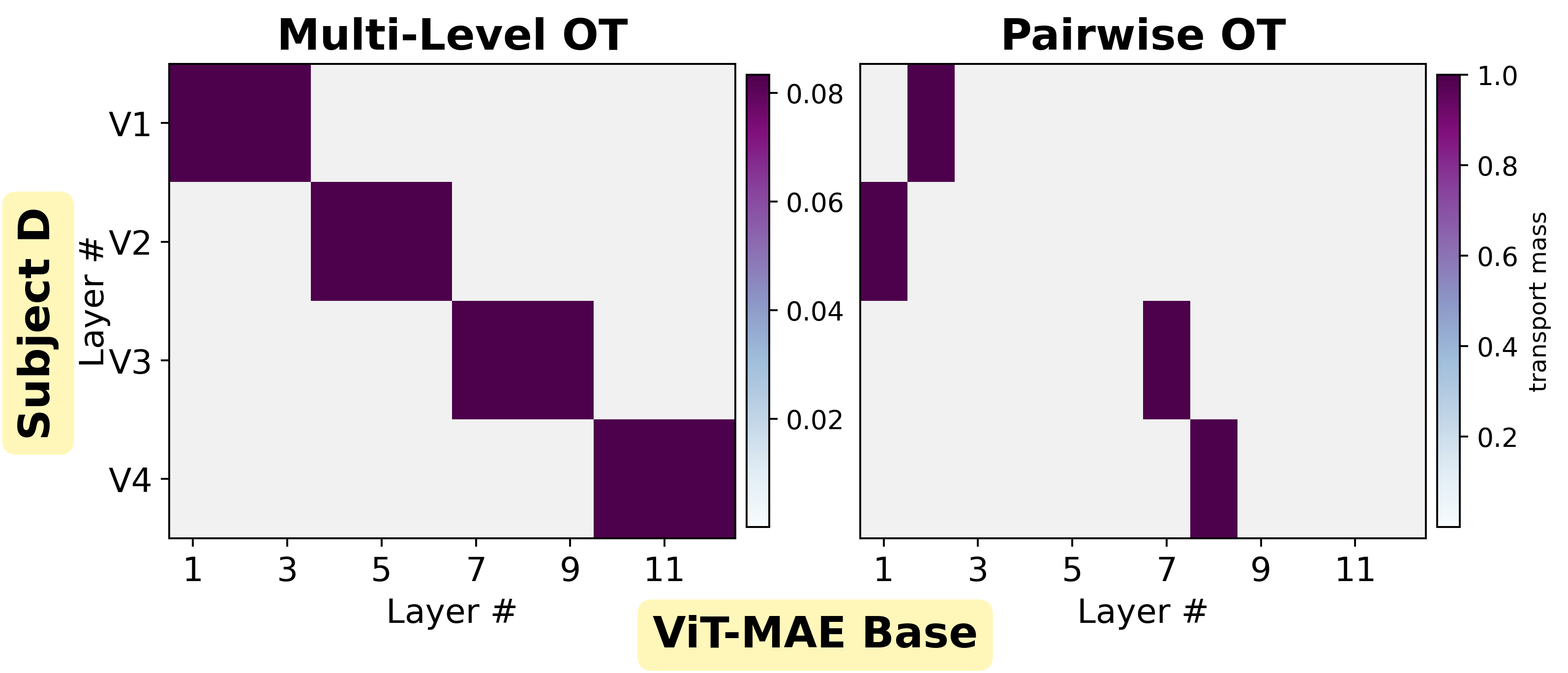}\hfill
\includegraphics[width=0.48\textwidth]{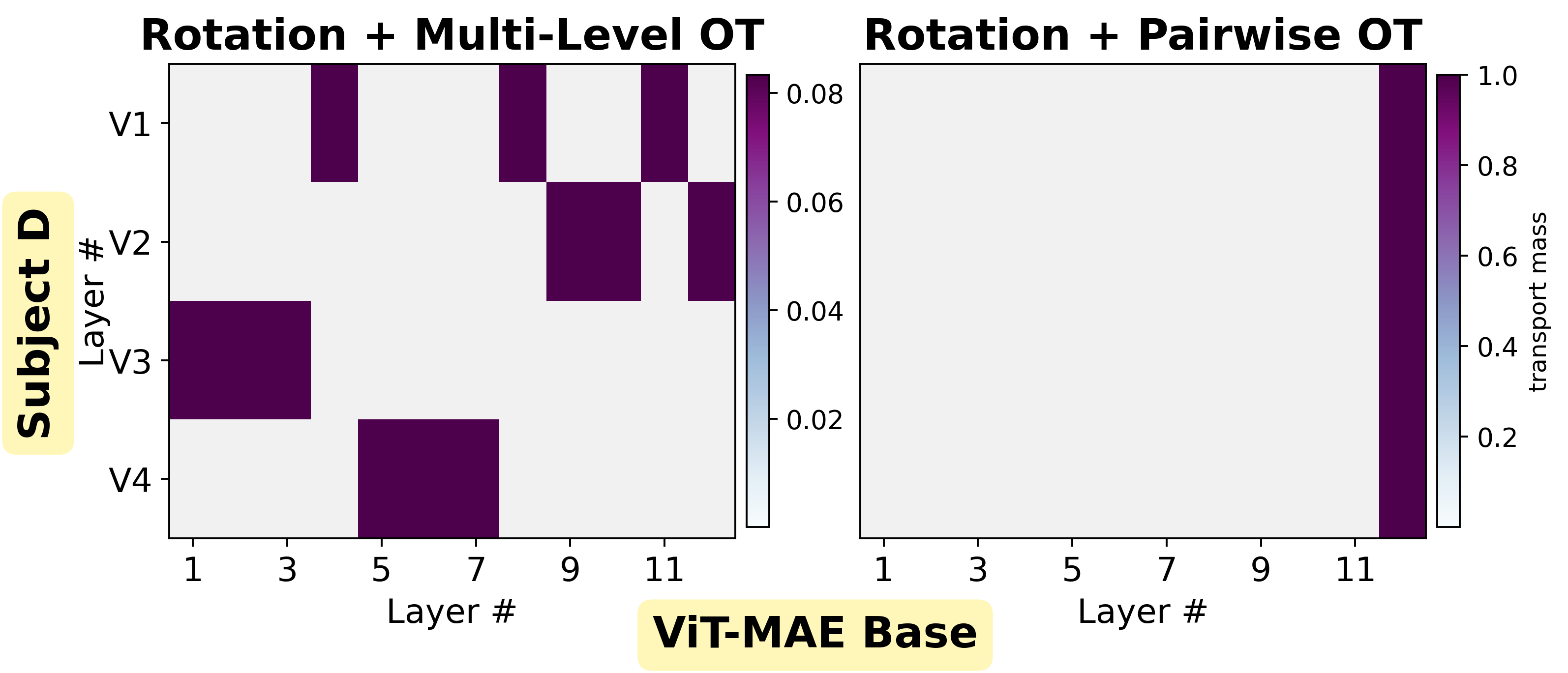}
\caption{Subject D $\leftrightarrow$ ViT-MAE Base (a) without rotation (MOT) and (b) with rotation augmentation (MOT+R).}
\label{fig:subject_7-vit-mae-base}
\end{figure}

\begin{figure}[H]
\centering
\includegraphics[width=0.48\textwidth]{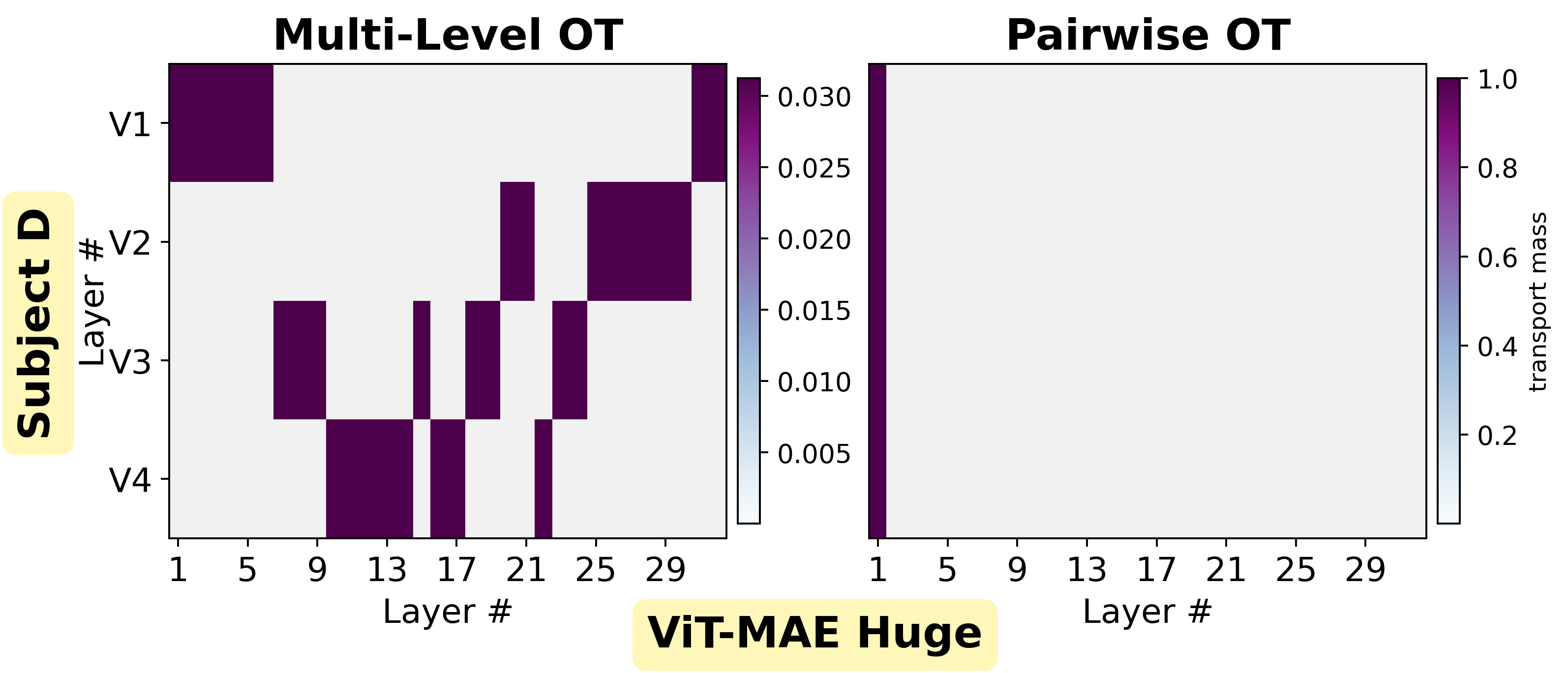}\hfill
\includegraphics[width=0.48\textwidth]{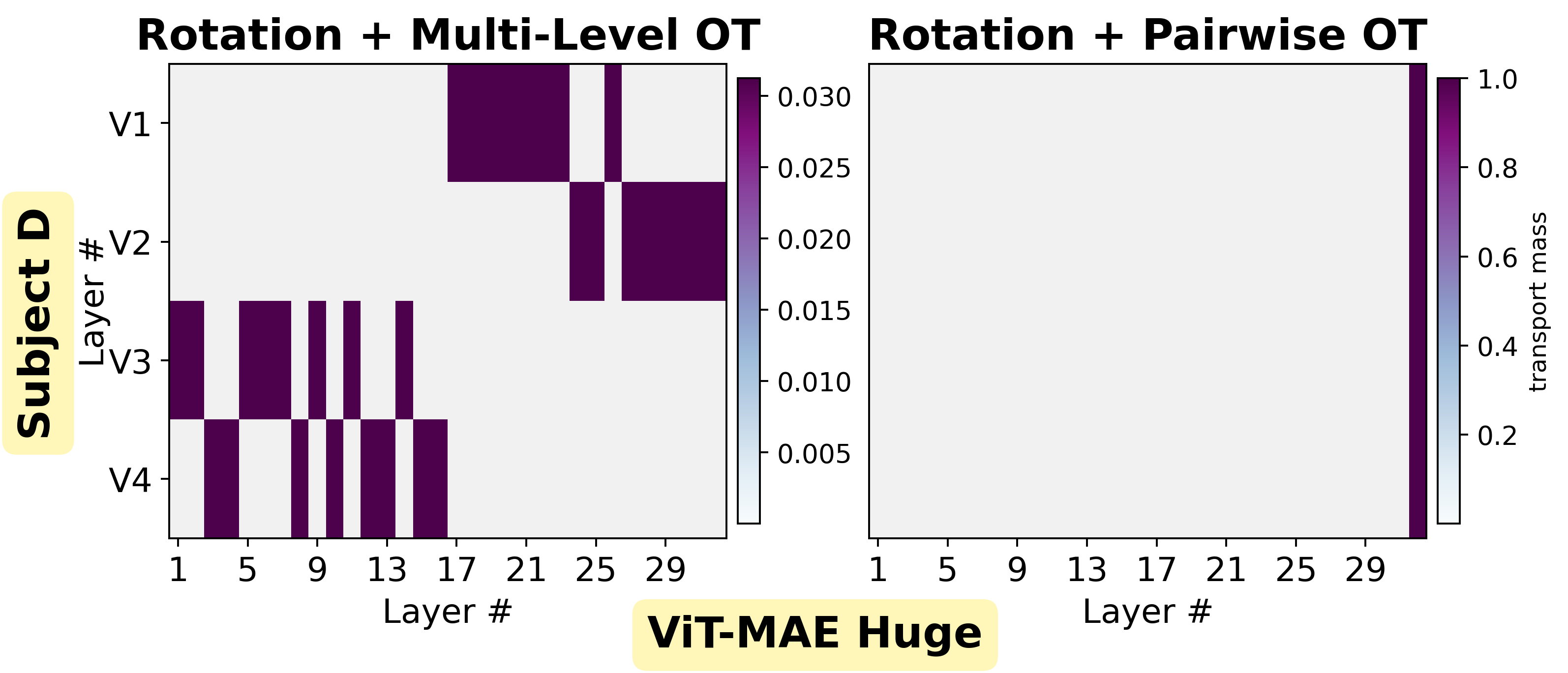}
\caption{Subject D $\leftrightarrow$ ViT-MAE Huge (a) without rotation (MOT) and (b) with rotation augmentation (MOT+R).}
\label{fig:subject_7-vit-mae-huge123456789}
\end{figure}

\textbf{Results}. Table ~\ref{tab:fmrimodels} reports reconstruction scores for MOT, MOT+R, and corresponding baselines averaged across the four subjects. Among all methods, MOT+R achieves the highest reconstruction score, indicating that incorporating rotation is critical for capturing shared structure between cortical and model representations. Subject-specific results (Tables ~\ref{tab:vision_fmri} and ~\ref{tab:vision_fmri_rotation}) further confirm this trend, with MOT+R consistently outperforming both vanilla MOT and pairwise baselines. Transport plans visualized in Figures ~\ref{fig:subject_1-dinov2-base}– ~\ref{fig:subject_7-vit-mae-huge123456789} reveal partial but inconsistent layer-wise correspondences: in some cases, early cortical regions align with early model layers and higher regions map to deeper layers, while in others the mappings appear noisier. 

\begin{table}[H]
\centering
\small
\begin{tabular}{@{}ll S S S S S S S@{}}
\toprule
\multicolumn{1}{c}{Model 1} & \multicolumn{1}{c}{Model 2} & \multicolumn{1}{c}{MOT Metric} & \multicolumn{1}{c}{Random (Perm-P)} & \multicolumn{1}{c}{Single-Best OT} & \multicolumn{1}{c}{Pairwise Best OT} \\
\midrule
Subject A & DINOv2 Giant & 0.085 & 0.087 & 0.046 & 0.063  \\
Subject A & DINOv2 Base & 0.081 & 0.072 & 0.075 & 0.080 \\
Subject A & ViT-MAE Huge & 0.065 & 0.057 & 0.063 & 0.105  \\
Subject A & ViT-MAE Base & 0.141 & 0.131 & 0.123 & 0.124 \\
\midrule
Subject B & DINOv2 Giant & 0.087 & 0.089 & 0.054 & 0.065 \\
Subject B & DINOv2 Base & 0.088 & 0.080 & 0.075 & 0.082 \\
Subject B & ViT-MAE Huge & 0.077 & 0.072 & 0.073 & 0.093 \\
Subject B & ViT-MAE Base & 0.115 & 0.109 & 0.101 & 0.104  \\
\midrule
Subject C & DINOv2 Giant & 0.115 & 0.124 & 0.072 & 0.092  \\
Subject C & DINOv2 Base & 0.115 & 0.108 & 0.095 & 0.100 \\
Subject C & ViT-MAE Huge & 0.092 & 0.088 & 0.082 & 0.114 \\
Subject C & ViT-MAE Base & 0.137 & 0.134 & 0.121 & 0.124  \\
\midrule
Subject D & DINOv2 Giant & 0.073 & 0.082 & 0.049 & 0.063  \\
Subject D & DINOv2 Base & 0.083 & 0.078 & 0.072 & 0.072  \\
Subject D & ViT-MAE Huge & 0.054 & 0.049 & 0.046 & 0.086  \\
Subject D & ViT-MAE Base & 0.114 & 0.111 & 0.101 & 0.106  \\
\bottomrule
\end{tabular}
\caption{Results on MOT metric vs. baselines (test split).}
\label{tab:vision_fmri}
\end{table}

\begin{table}[H]
\centering
\small
\begin{tabular}{@{}ll S S S S S S S@{}}
\toprule
\multicolumn{1}{c}{Model 1} & \multicolumn{1}{c}{Model 2} & \multicolumn{1}{c}{MOT + R} & \multicolumn{1}{c}{Single-Best + R} & \multicolumn{1}{c}{Pairwise Best + R} \\
\midrule
Subject A & DINOv2 Giant & 0.161 & 0.140 & 0.099 \\
Subject A & DINOv2 Base  & 0.146 & 0.140 & 0.083 \\
Subject A & ViT-MAE Huge & 0.165 & 0.151 & 0.127 \\
Subject A & ViT-MAE Base & 0.157 & 0.151 & 0.124 \\
\midrule
Subject B & DINOv2 Giant & 0.147 & 0.132 & 0.096 \\
Subject B & DINOv2 Base  & 0.127 & 0.120 & 0.084 \\
Subject B & ViT-MAE Huge & 0.125 & 0.109 & 0.104 \\
Subject B & ViT-MAE Base & 0.128 & 0.118 & 0.101 \\
\midrule
Subject C & DINOv2 Giant & 0.205 & 0.183 & 0.147 \\
Subject C & DINOv2 Base  & 0.185 & 0.181 & 0.128 \\
Subject C & ViT-MAE Huge & 0.189 & 0.179 & 0.151 \\
Subject C & ViT-MAE Base & 0.186 & 0.180 & 0.152 \\
\midrule
Subject D & DINOv2 Giant & 0.139 & 0.123 & 0.099 \\
Subject D & DINOv2 Base  & 0.121 & 0.114 & 0.081 \\
Subject D & ViT-MAE Huge & 0.135 & 0.122 & 0.107 \\
Subject D & ViT-MAE Base & 0.134 & 0.129 & 0.109 \\
\bottomrule
\end{tabular}
\caption{Results on rotational MOT metric vs. baselines (test split).}
\label{tab:vision_fmri_rotation}
\end{table}

\section{Global Alignment}
\label{Global_Alignment}
In this section, we report reconstruction scores on a held-out validation dataset for the vision-cortex alignment study under a global alignment formulation. In this setting, all layers from both networks are flattened so that all neurons are treated as belonging to a single layer. We compare the resulting reconstruction scores to those obtained with our MOT reconstruction scores.

Global alignment scores are computed using two alignment metrics: optimal transport and linear predictivity. Reconstruction performance is then evaluated on held-out validation data. The results for this experiment are presented in Table \ref{tab:vision_global_ot}.

\begin{table}[H]
\centering
\small
\begin{tabular}{@{}llccc@{}}
\toprule
\multicolumn{1}{c}{Model 1} & \multicolumn{1}{c}{Model 2} & \multicolumn{1}{c}{Global OT} & \multicolumn{1}{c}{Global Linear Predictivity} & \multicolumn{1}{c}{MOT} \\
\midrule
Subject A & Subject B & 0.184 & 0.305 & 0.244 \\
Subject A & Subject C & 0.177 & 0.295 & 0.199 \\
Subject A & Subject D & 0.168 & 0.289 & 0.198 \\
Subject B & Subject C & 0.183 & 0.306 & 0.212 \\
Subject C & Subject D & 0.158 & 0.306 & 0.197 \\
Subject B & Subject D & 0.163 & 0.297 & 0.201 \\
\bottomrule
\end{tabular}
\caption{Comparison of global OT, Linear Predictivity, and MOT, evaluated by reconstruction correlation on held-out fMRI responses.}

\label{tab:vision_global_ot}
\end{table}

\section{Quantifying Hierarchy}
\label{sec:hierarchy_score}

Given a layer-to-layer transport plan $P \in \mathbb{R}^{L \times M}$,
we define a scalar ``hierarchy'' score that quantifies how strongly the transport mass is concentrated along a diagonal correspondence between source and target layers.

We first embed the layer indices into the unit interval via
\[
r_\ell = \frac{\ell - 1}{L - 1}, 
\qquad
c_m   = \frac{m - 1}{M - 1},
\qquad
\ell = 1,\dots,L,\; m = 1,\dots,M,
\]
so that the ideal diagonal correspondence between layers lies along the line $r = c$.
For each entry $(\ell,m)$ we define its normalized distance to this diagonal as
\[
d_{\ell m} = \bigl| r_\ell - c_m \bigr| \in [0,1].
\]

We interpret the magnitudes of the transport plan as a discrete mass distribution,
\[
s_{\ell m} = |P_{\ell m}|,
\qquad
w_{\ell m} = \frac{s_{\ell m}}{\sum_{\ell',m'} s_{\ell' m'}},
\]
and compute the average distance of this mass from the diagonal,
\[
D(P) = \sum_{\ell=1}^L \sum_{m=1}^M w_{\ell m} \, d_{\ell m}.
\]
Finally, we define the hierarchy score as
\[
H(P) = 1 - D(P).
\]

By construction $H(P) \in [0,1]$ and it depends only on the relative placement of transport mass with respect to the diagonal. Table \ref{tab:quant_hierarchy1} reports $H(P)$ for each transport plan as a summary measure of its hierarchy.

\begin{table}[H]
\centering
\small
\begin{tabular}{@{}ll S S S S@{}}
\toprule
\multicolumn{1}{c}{Model 1} & \multicolumn{1}{c}{Model 2} & \multicolumn{1}{c}{MOT} & \multicolumn{1}{c}{Pairwise OT} & \multicolumn{1}{c}{MOT+R} & \multicolumn{1}{c}{Pairwise Best + R} \\
\midrule
DINOv2 Small & ViT-MAE Large & \textbf{0.90} & 0.84 & 0.89 & 0.50 \\
ViT-MAE Base & ViT-MAE Large & \textbf{0.97} & 0.61 & 0.93 & 0.50 \\
DINOv2 Small & ViT-MAE Base & \textbf{1.00} & 0.88 & 0.85 & 0.50 \\
ViT-MAE Base & DINOv2 Large & \textbf{0.96} & 0.70 & 0.90 & 0.50 \\
DINOv2 Small & DINOv2 Large & \textbf{0.98} & 0.88 & 0.94 & 0.50 \\
ViT-MAE Huge & DINOv2 Giant & 0.86 & 0.62 & \textbf{0.86} & 0.50 \\
ViT-MAE Base & DINOv2 Giant & 0.60 & 0.55 & \textbf{0.91} & 0.50 \\
DINOv2 Small & DINOv2 Giant & \textbf{0.97} & 0.83 & 0.95 & 0.50 \\
ViT-MAE Base & ViT-MAE Huge & 0.68 & 0.50 & \textbf{0.83} & 0.50 \\
DINOv2 Small & ViT-MAE Huge & 0.82 & 0.72 & \textbf{0.88} & 0.50 \\
\midrule
\addlinespace[2pt]
Llama-3.2 1B & Llama-3.2 3B & \textbf{0.96} & 0.93 &  &  \\
Qwen-2.5 0.5B & Llama-3.2 3B & \textbf{0.86} & 0.63 &  &  \\
Llama-3.2 3B & Qwen-2.5 3B & \textbf{0.83} & 0.80 &  &  \\
Llama-3.2 1B & Qwen-2.5 3B & \textbf{0.85} & 0.59 &  &  \\
Qwen-2.5 0.5B & Llama-3.2 1B & \textbf{0.86} & 0.67 &  &  \\
Qwen-2.5 0.5B & Qwen-2.5 3B & \textbf{0.93} & 0.60 &  &  \\
\midrule
\addlinespace[2pt]
Subject B & Subject D & \textbf{1.00} & 0.93 &  &  \\
Subject C & Subject D & \textbf{1.00} & 0.98 &  &  \\
Subject B & Subject C & \textbf{1.00} & 0.93 &  &  \\
Subject A & Subject B & \textbf{1.00} & 0.98 &  &  \\
Subject A & Subject C & \textbf{1.00} & 0.98 &  &  \\
Subject A & Subject D & \textbf{1.00} & 0.98 &  &  \\
\bottomrule
\end{tabular}
\caption{MOT and pairwise (rotation where applicable) hierarchy scores for vision, language, and fMRI alignment.}
\label{tab:quant_hierarchy1}
\end{table}

We observe that, across all transport plans, the hierarchy metric is highest for plans generated by MOT (or MOT+R), compared with those produced by pairwise best (or pairwise rotational) methods. This indicates that MOT and MOT+R based approaches uncover hierarchical correspondences that pairwise methods don't reveal.

\section{Transport plans for Linear Predictivity based mapping pairs}
\begin{table}[H]
\centering
\small
\begin{tabular}{@{}ll SS@{}}
\toprule
\multicolumn{1}{c}{Model 1} & \multicolumn{1}{c}{Model 2} &
\multicolumn{1}{c}{Pairwise LP Reconstruction score} &
\multicolumn{1}{c}{MOT Metric} \\
\midrule
Llama-3.2 1B   & Llama-3.2 3B   & 0.737 & 0.558 \\
Qwen-2.5 0.5B  & Llama-3.2 3B   & 0.694 & 0.531 \\
Llama-3.2 3B   & Qwen-2.5 3B    & 0.784 & 0.383 \\
Llama-3.2 1B   & Qwen-2.5 3B    & 0.841 & 0.432 \\
Qwen-2.5 0.5B  & Llama-3.2 1B   & 0.848 & 0.522 \\
Qwen-2.5 0.5B  & Qwen-2.5 3B    & 0.848 & 0.510 \\
\midrule
\addlinespace[2pt]
Subject B & Subject D & 0.210 & 0.201 \\
Subject C & Subject D & 0.230 & 0.197 \\
Subject B & Subject C & 0.199 & 0.212 \\
Subject A & Subject B & 0.238 & 0.244 \\
Subject A & Subject C & 0.179 & 0.199 \\
Subject A & Subject D & 0.206 & 0.198 \\
\bottomrule
\end{tabular}
\caption{\textbf{Pairwise Linear Predictivity and MOT reconstruction scores on test splits}. Top: LLM model-pair alignment. Bottom: fMRI subject-pair alignment.}
\label{tab:lp_llm_fmri}
\end{table}

\begin{figure}[H]
\centering
\includegraphics[width=0.32\textwidth]{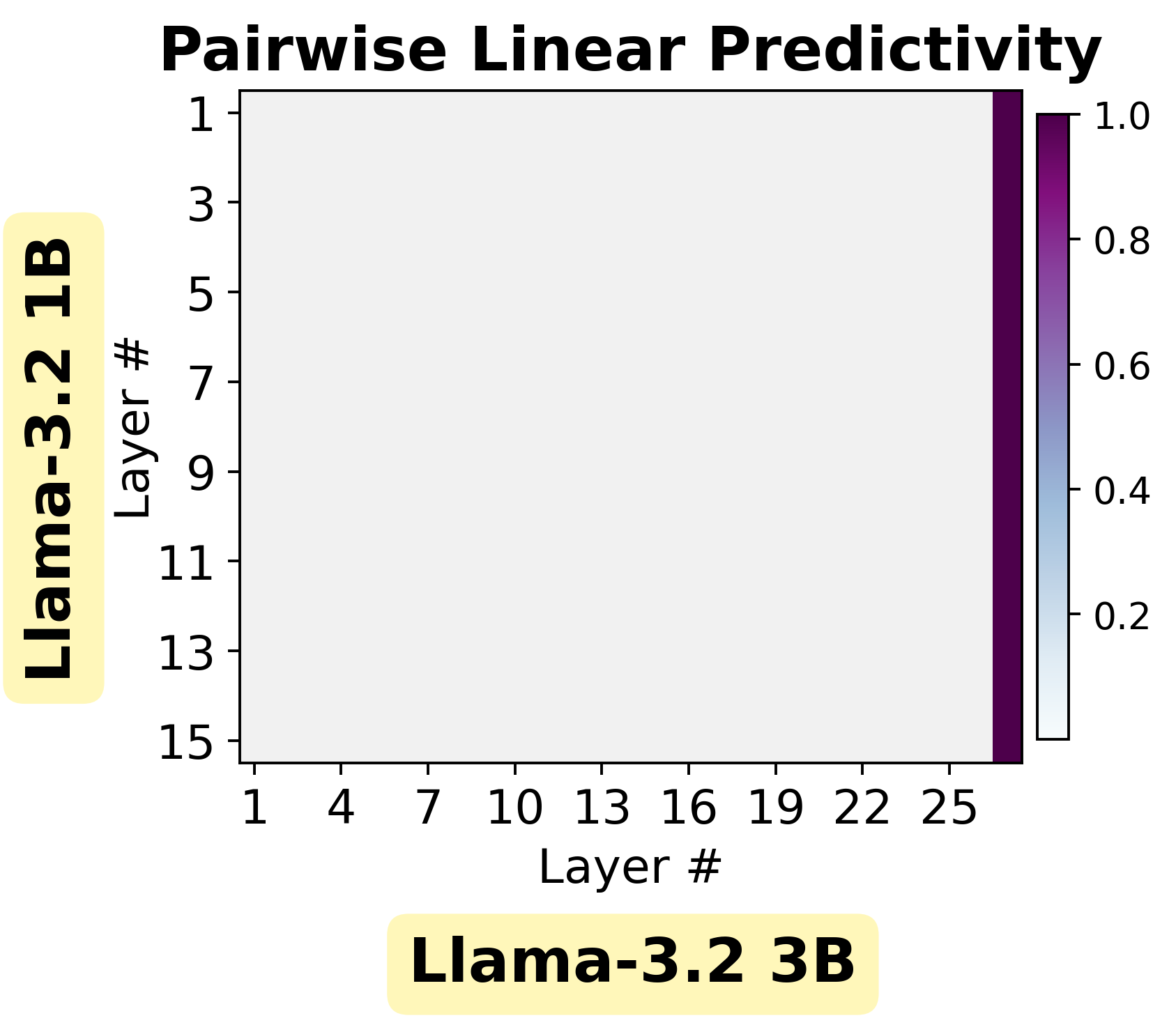}\hfill
\includegraphics[width=0.32\textwidth]{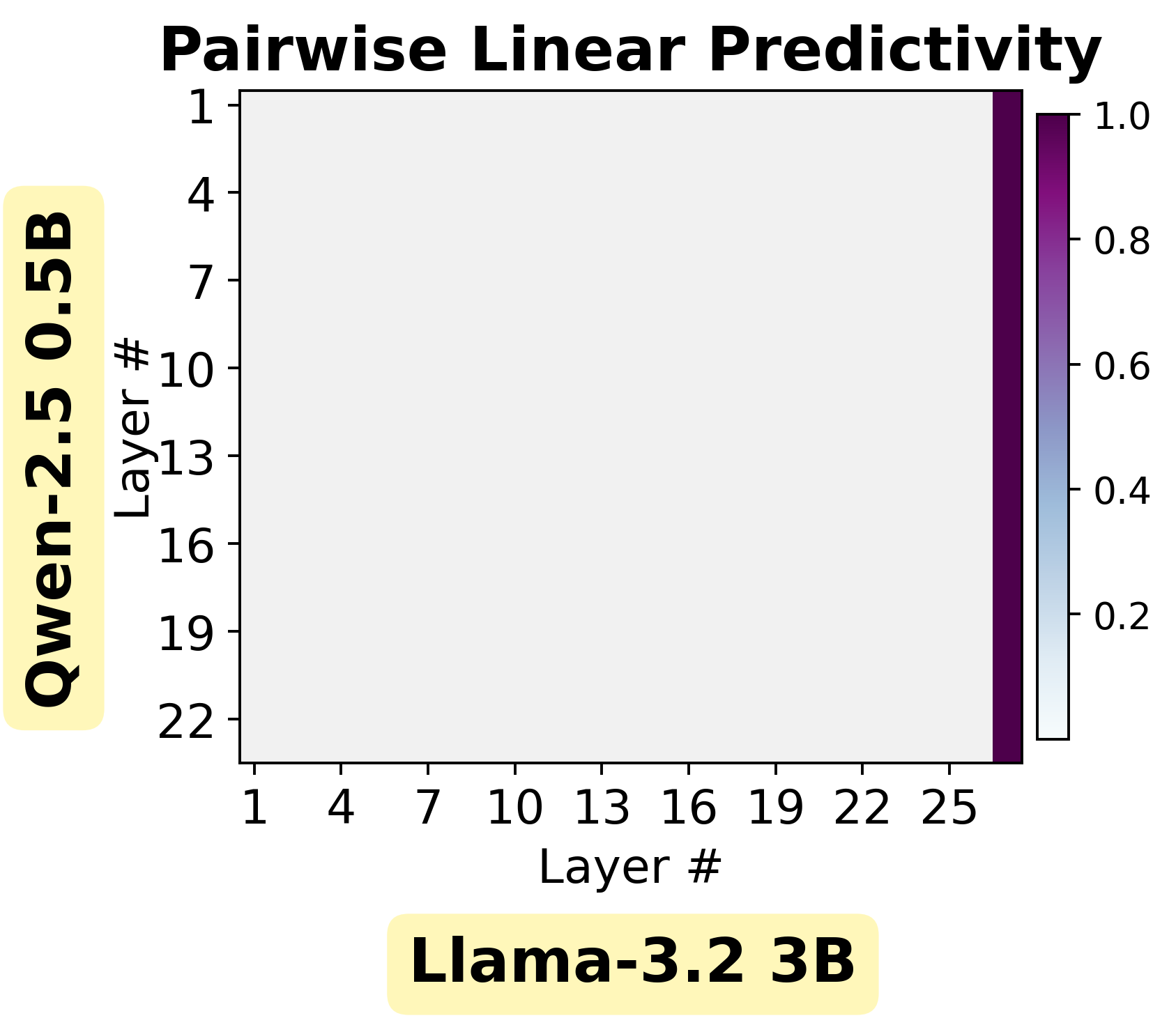}\hfill
\includegraphics[width=0.32\textwidth]{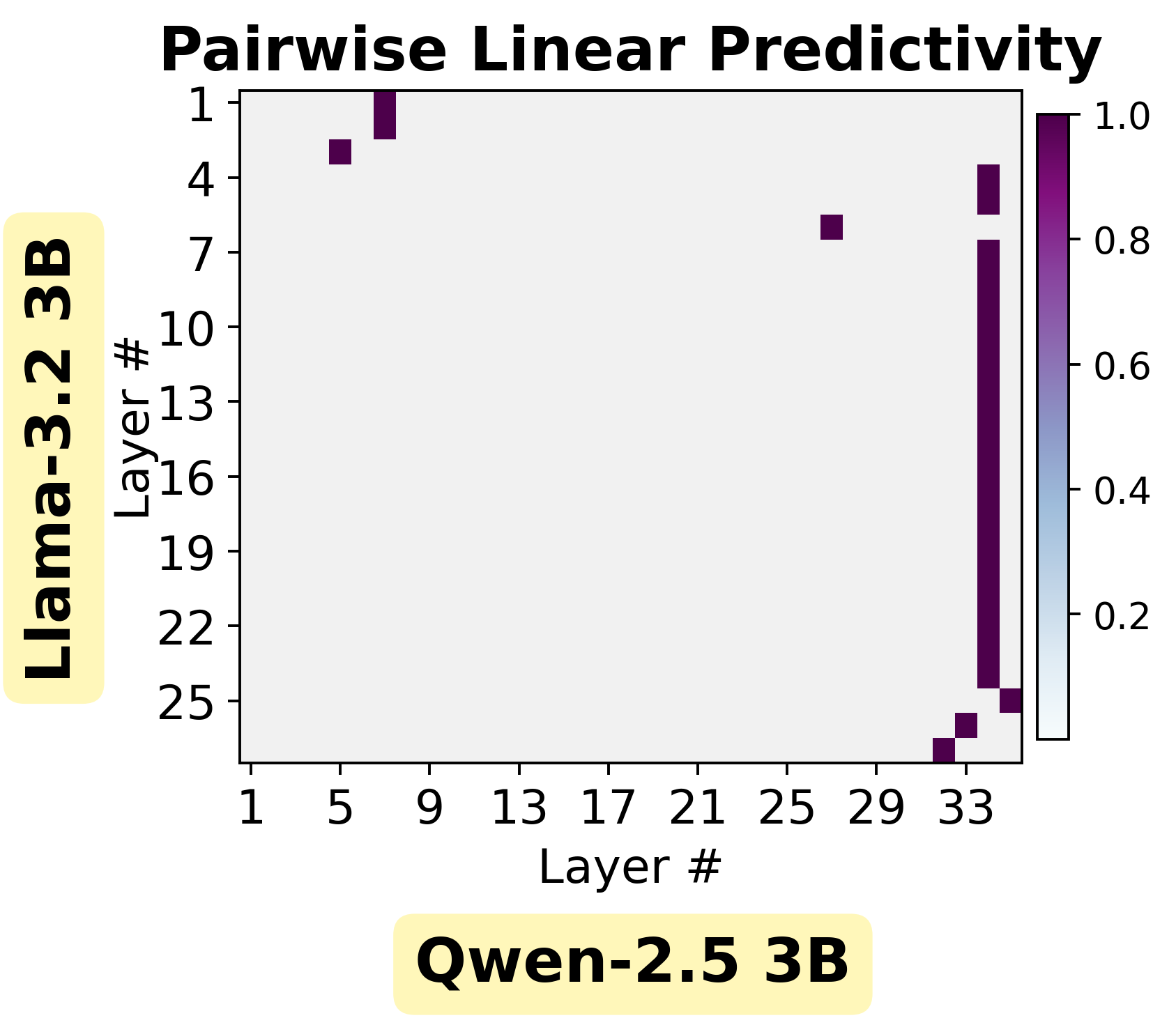}
\medskip\medskip

\includegraphics[width=0.32\textwidth]{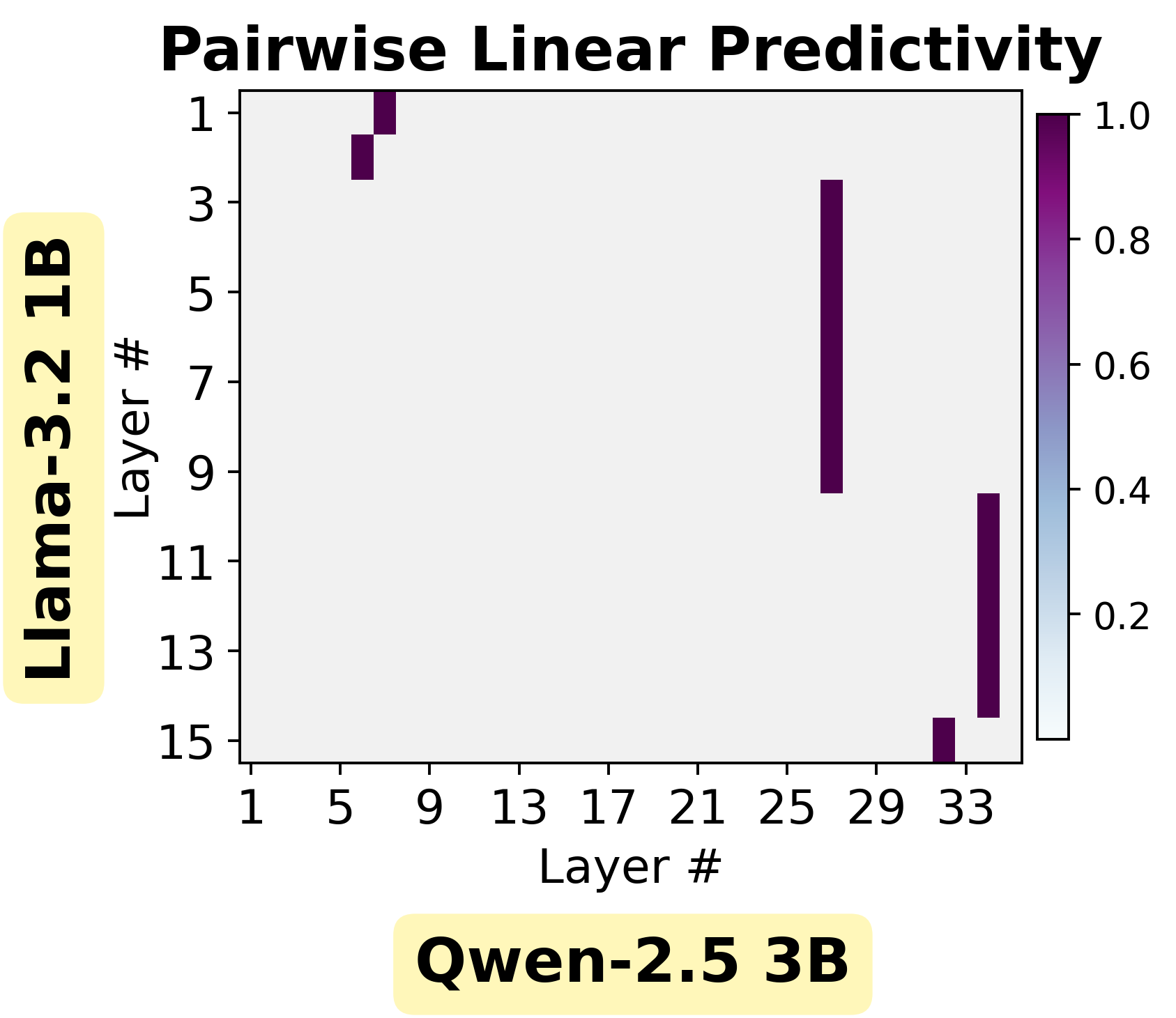}\hfill
\includegraphics[width=0.32\textwidth]{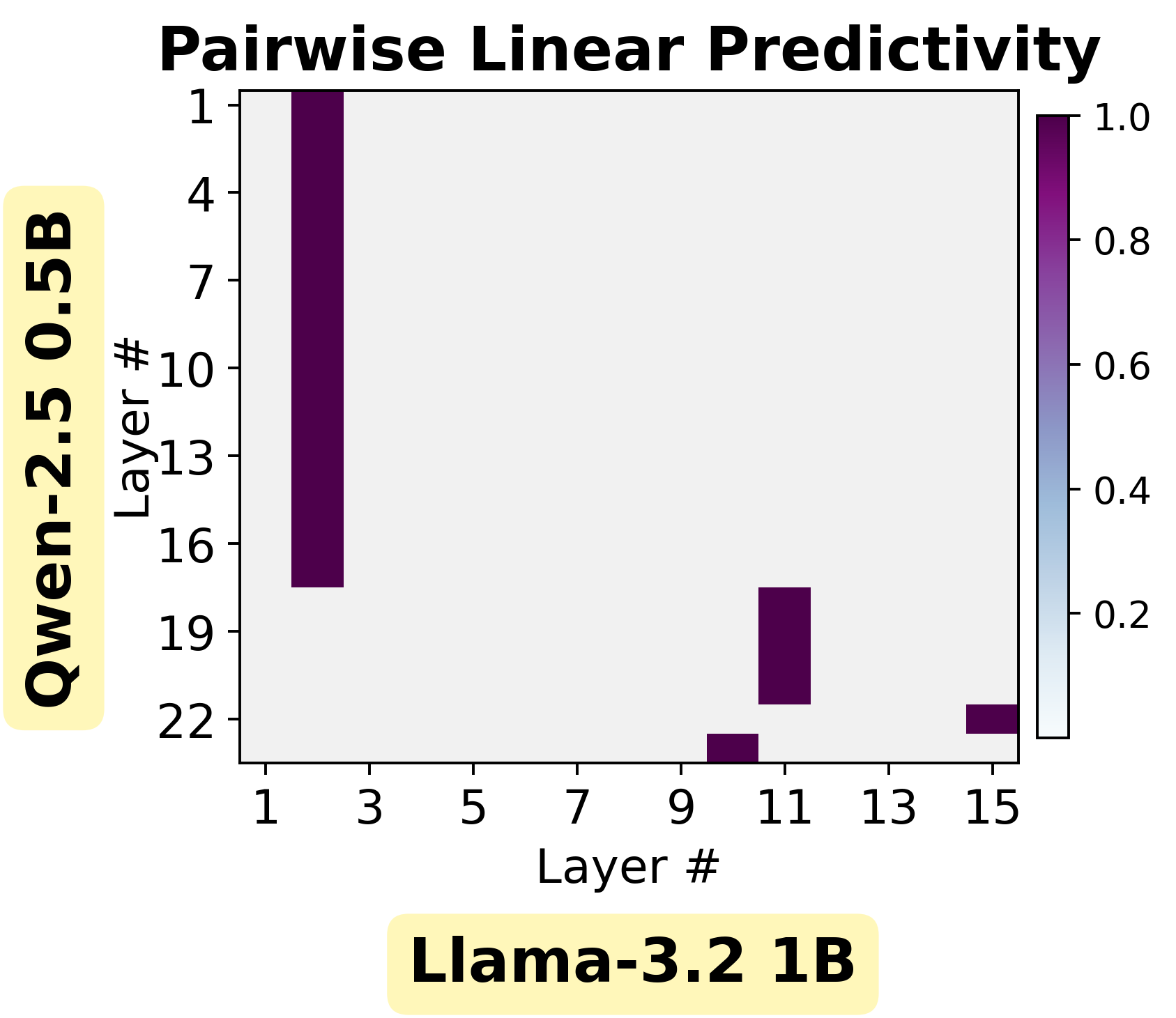}\hfill
\includegraphics[width=0.32\textwidth]{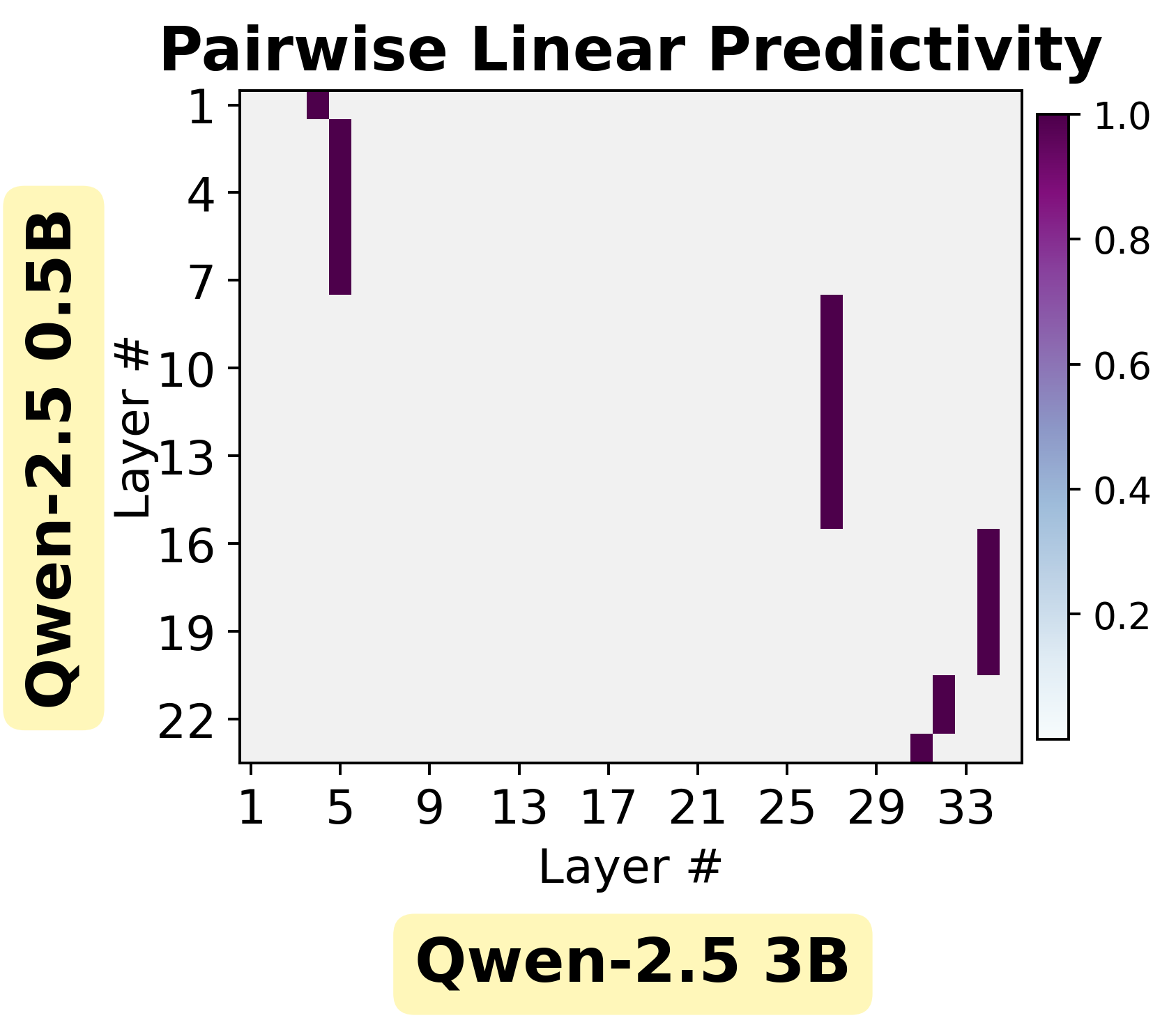}

\bigskip\bigskip

\includegraphics[width=0.32\textwidth]{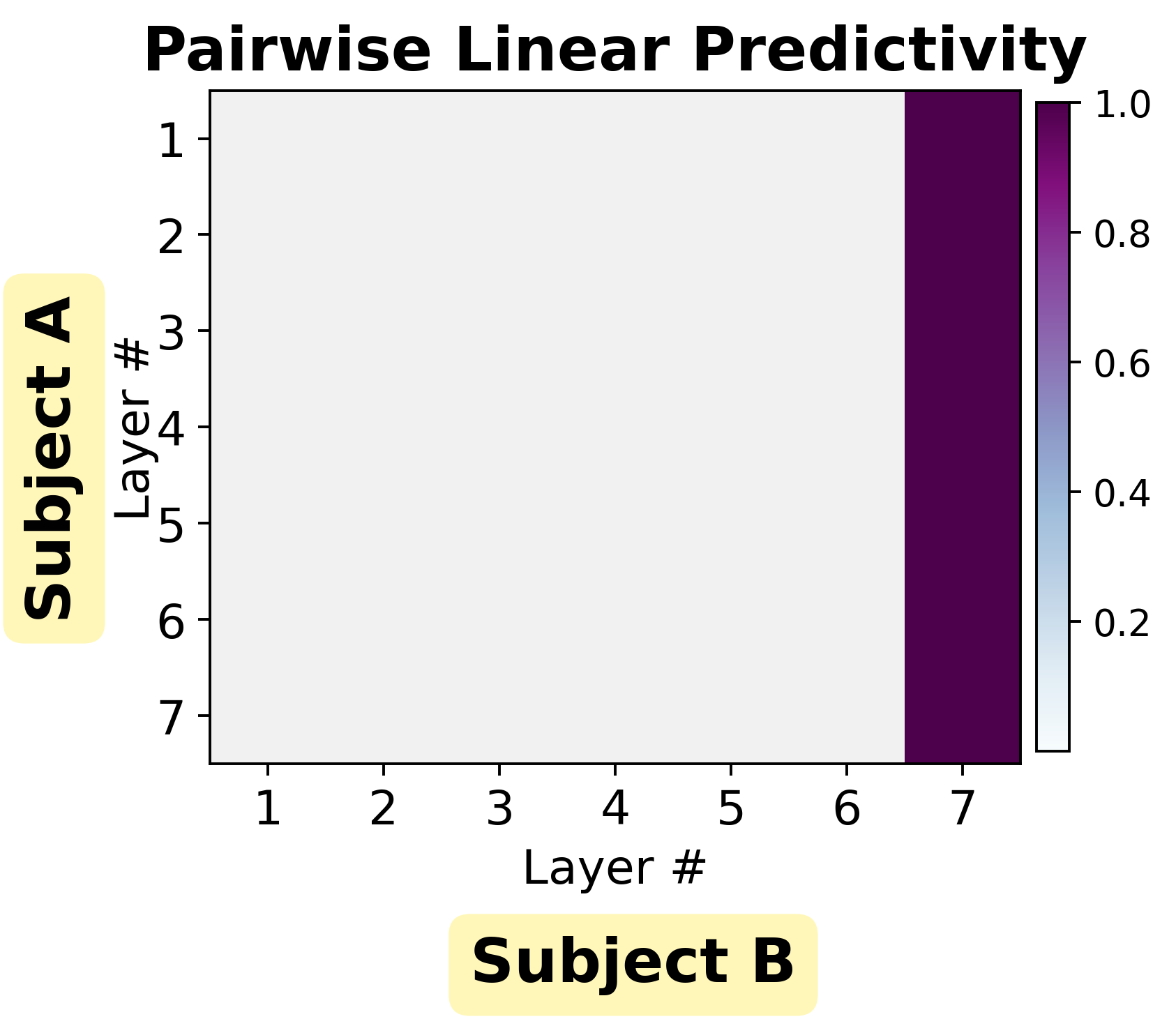}\hfill
\includegraphics[width=0.32\textwidth]{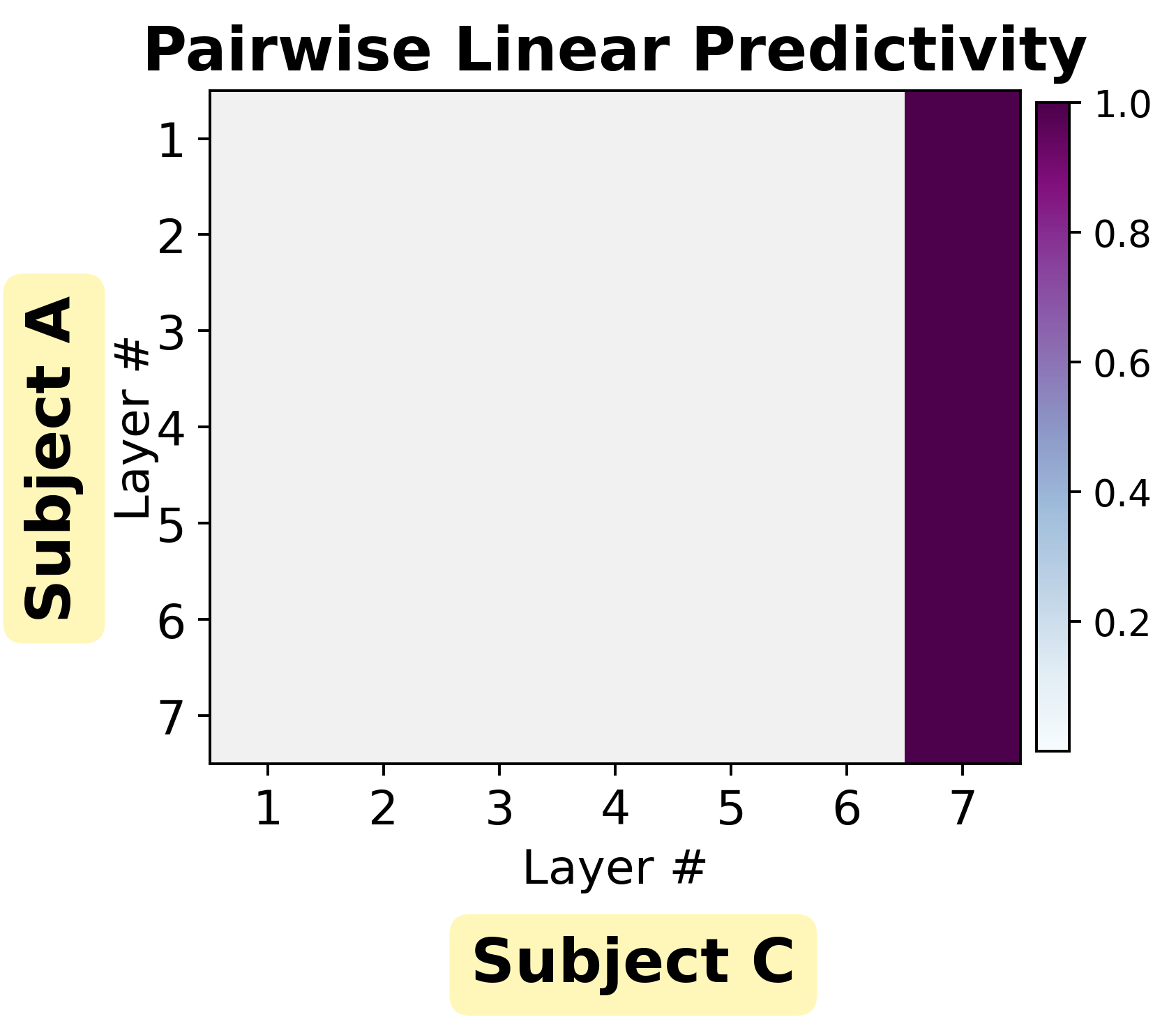}\hfill
\includegraphics[width=0.32\textwidth]{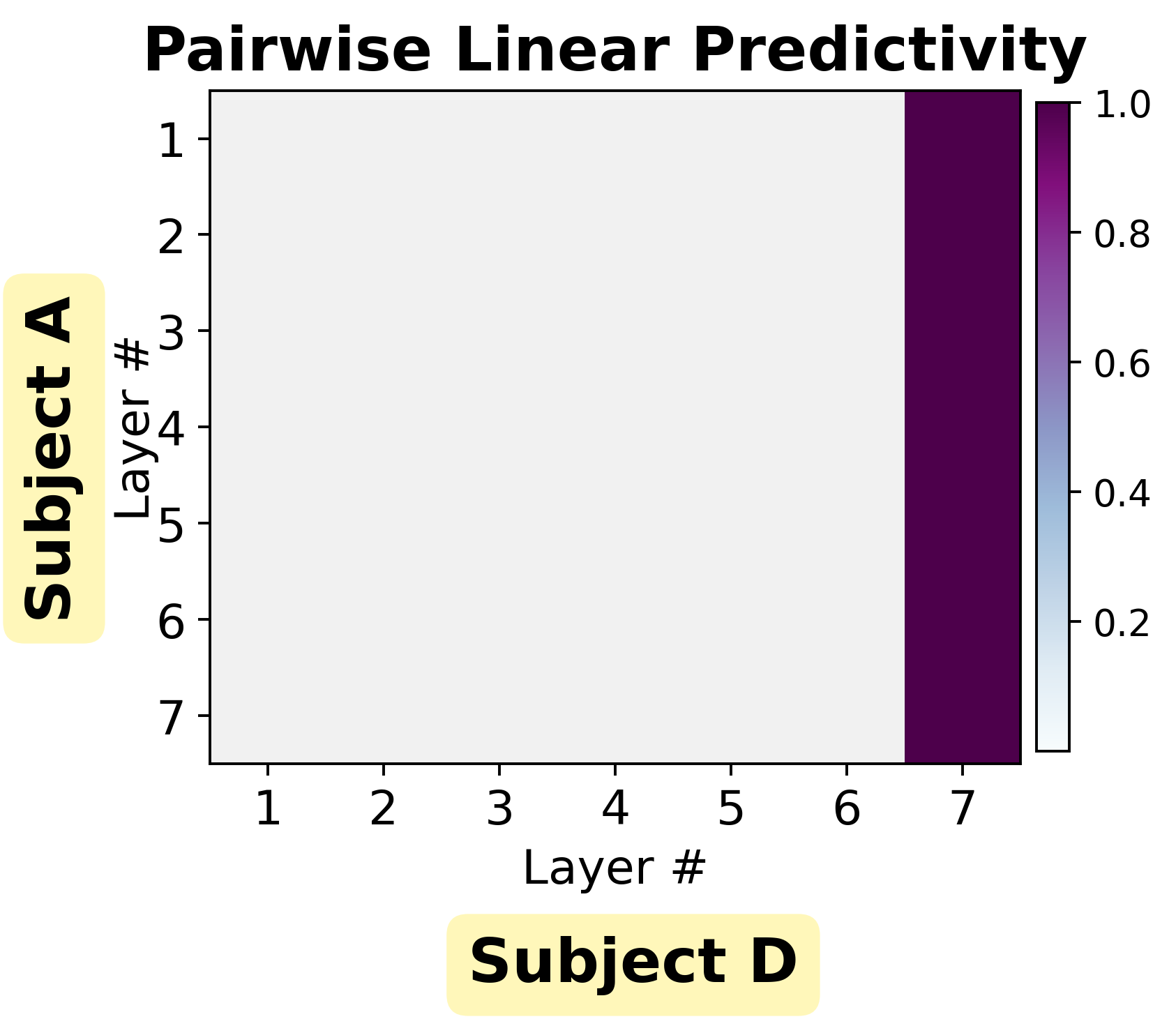}
\medskip\medskip

\includegraphics[width=0.32\textwidth]{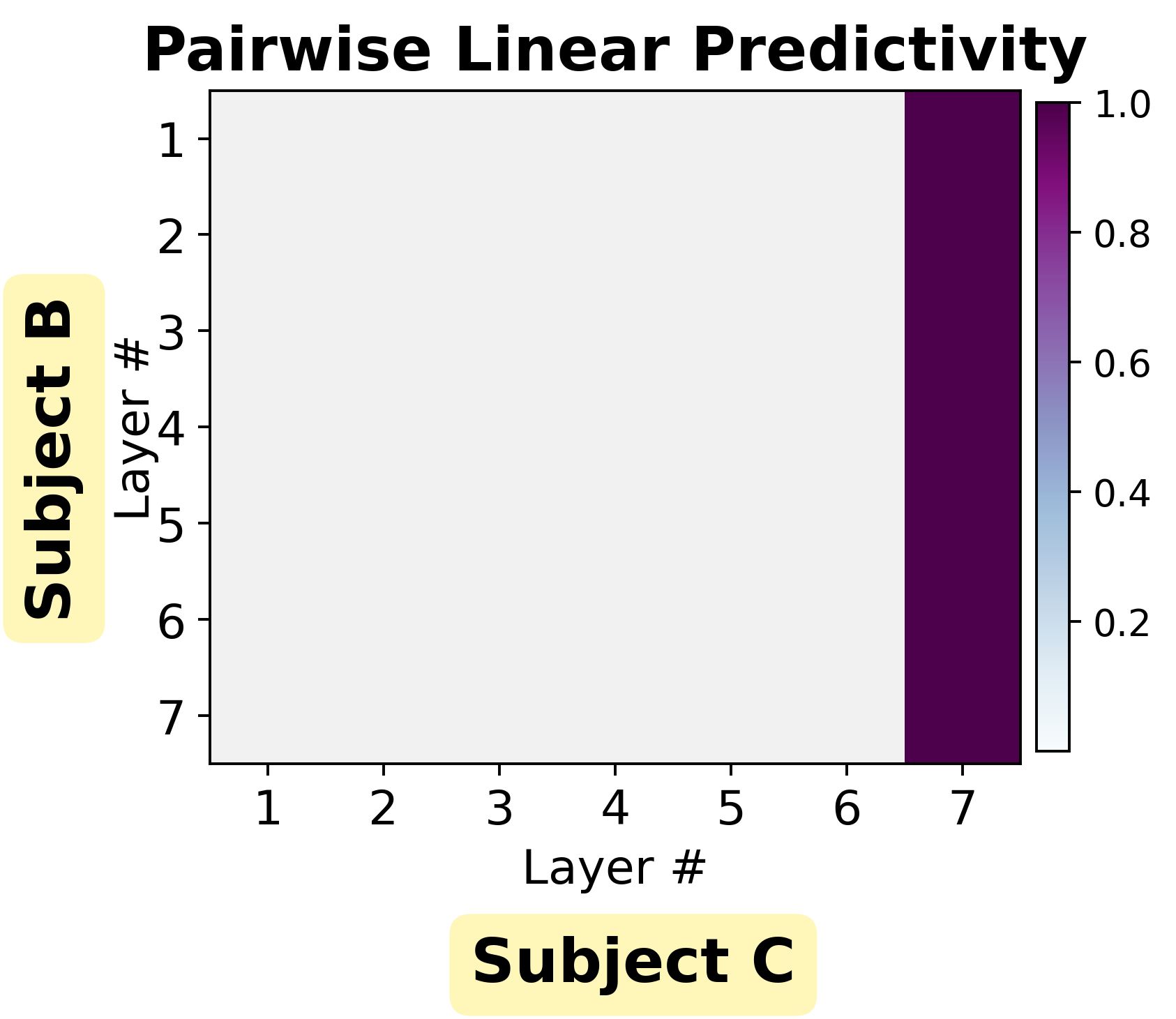}\hfill
\includegraphics[width=0.32\textwidth]{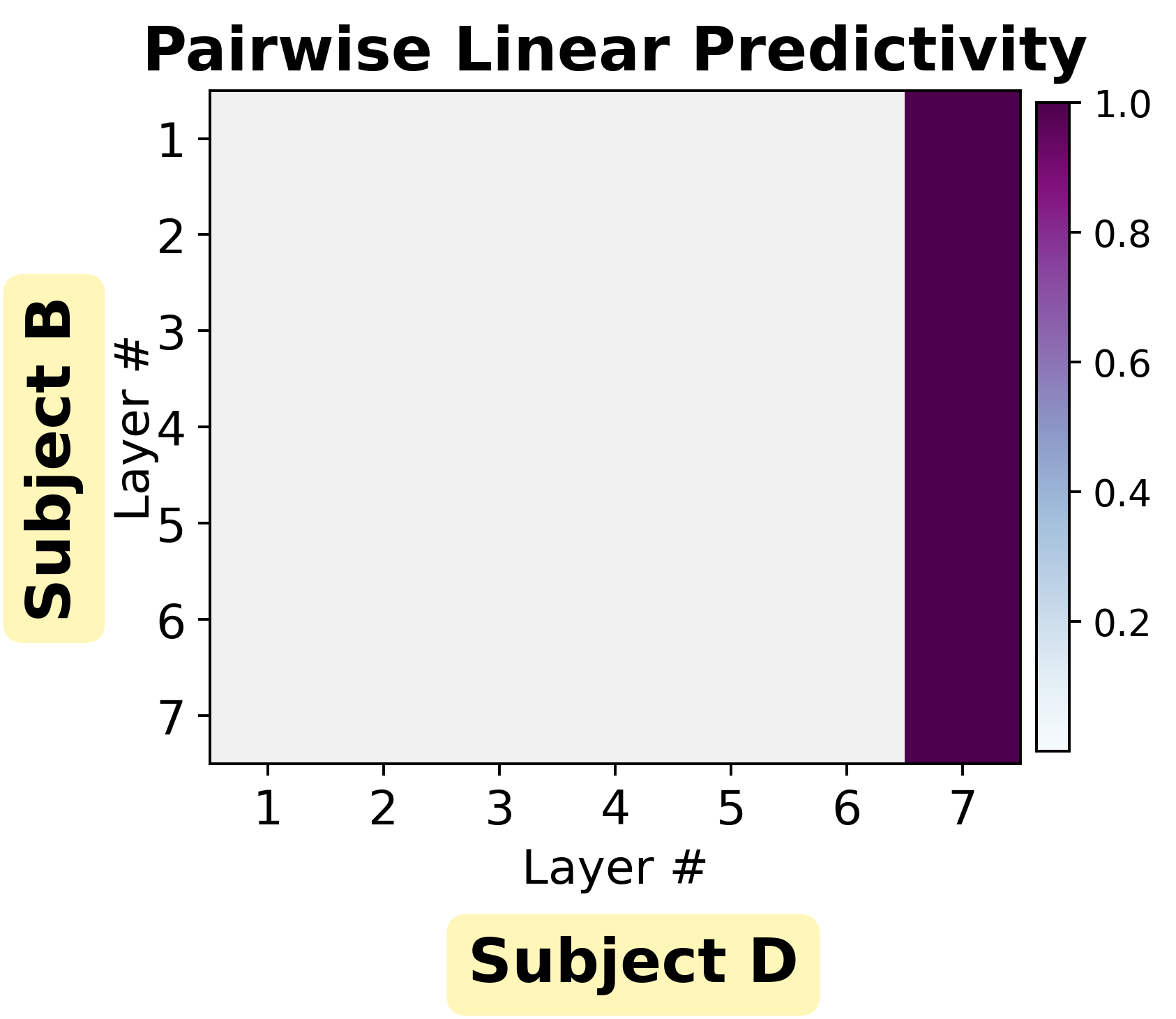}\hfill
\includegraphics[width=0.32\textwidth]{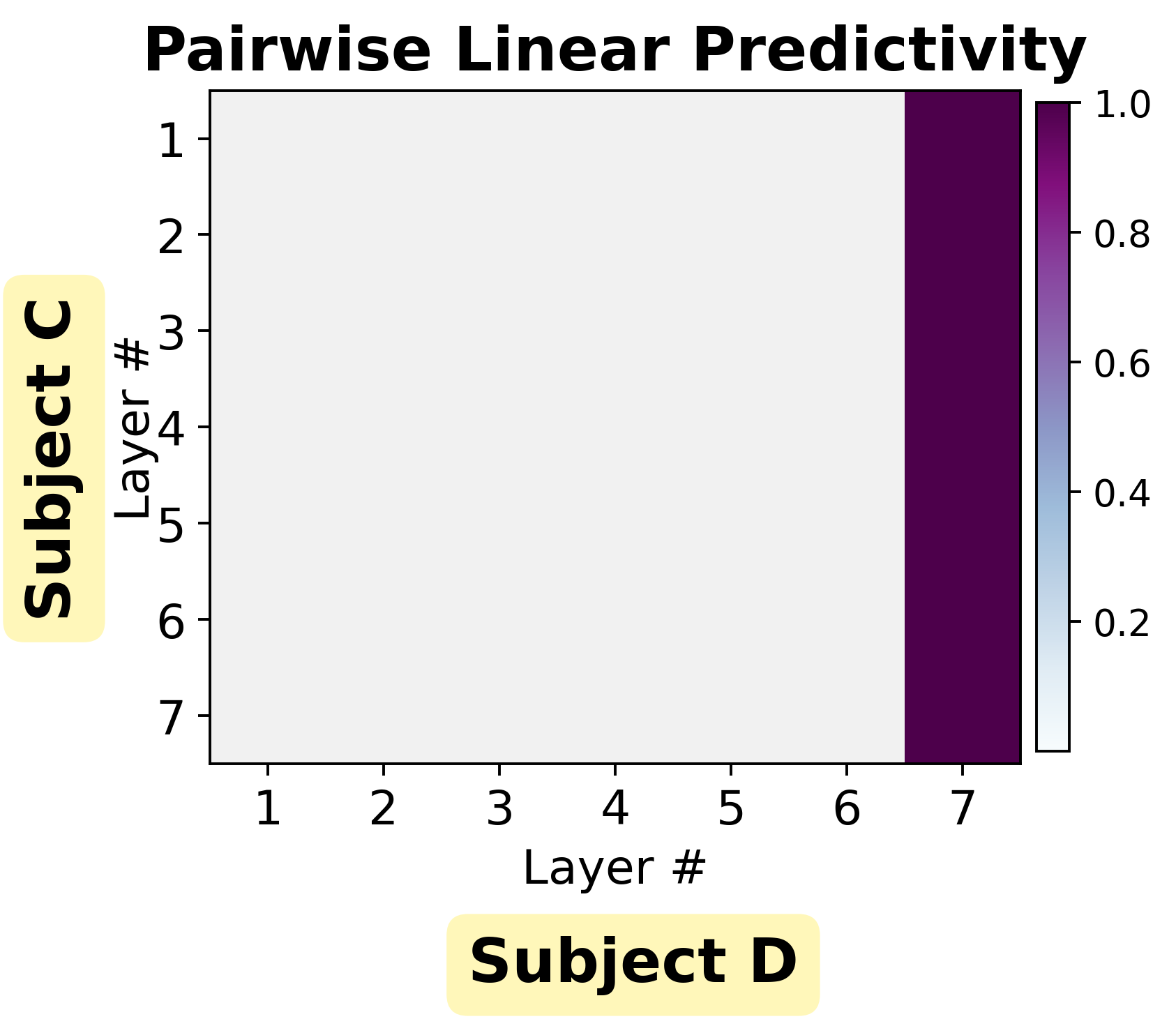}

\caption{\textbf{Transport plans for linear predictivity-based mappings in LLM and fMRI spaces.}
Top two rows: pairwise mappings learned under linear predictivity constraints for all LLM pairs:
(a) Llama-3.2 1B $\leftrightarrow$ Llama-3.2 3B,
(b) Qwen-2.5 0.5B $\leftrightarrow$ Llama-3.2 3B,
(c) Llama-3.2 3B $\leftrightarrow$ Qwen-2.5 3B,
(d) Llama-3.2 1B $\leftrightarrow$ Qwen-2.5 3B,
(e) Qwen-2.5 0.5B $\leftrightarrow$ Llama-3.2 1B, and
(f) Qwen-2.5 0.5B $\leftrightarrow$ Qwen-2.5 3B.
Bottom two rows: pairwise mappings learned under linear predictivity constraints for all fMRI subject pairs:
(g) Subject A $\leftrightarrow$ Subject B,
(h) Subject A $\leftrightarrow$ Subject C,
(i) Subject A $\leftrightarrow$ Subject D,
(j) Subject B $\leftrightarrow$ Subject C,
(k) Subject B $\leftrightarrow$ Subject D, and
(l) Subject C $\leftrightarrow$ Subject D.
In both the LLM and fMRI settings, linear predictivity-based pairwise mappings do not recover structured layer-wise correspondences, highlighting the importance of MOT for learning robust and generalizable mappings across architectures, model scales, and subjects.}
\label{fig:linearpredictivity_transport}
\end{figure}

\newpage
\section{Transport plans for RSA-based mapping pairs}
\begin{figure}[H]
\centering
\includegraphics[width=0.32\textwidth]{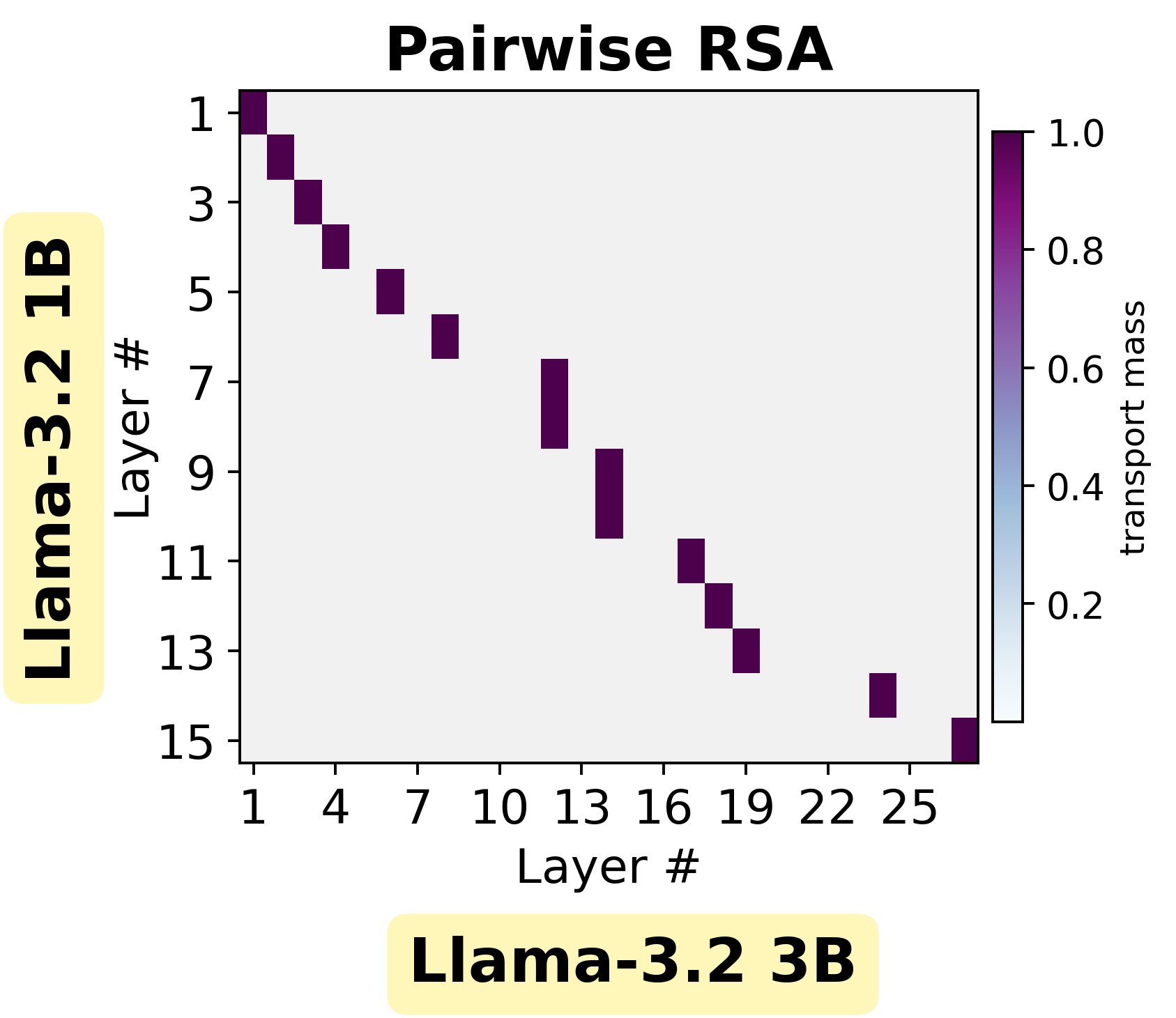}\hfill
\includegraphics[width=0.32\textwidth]{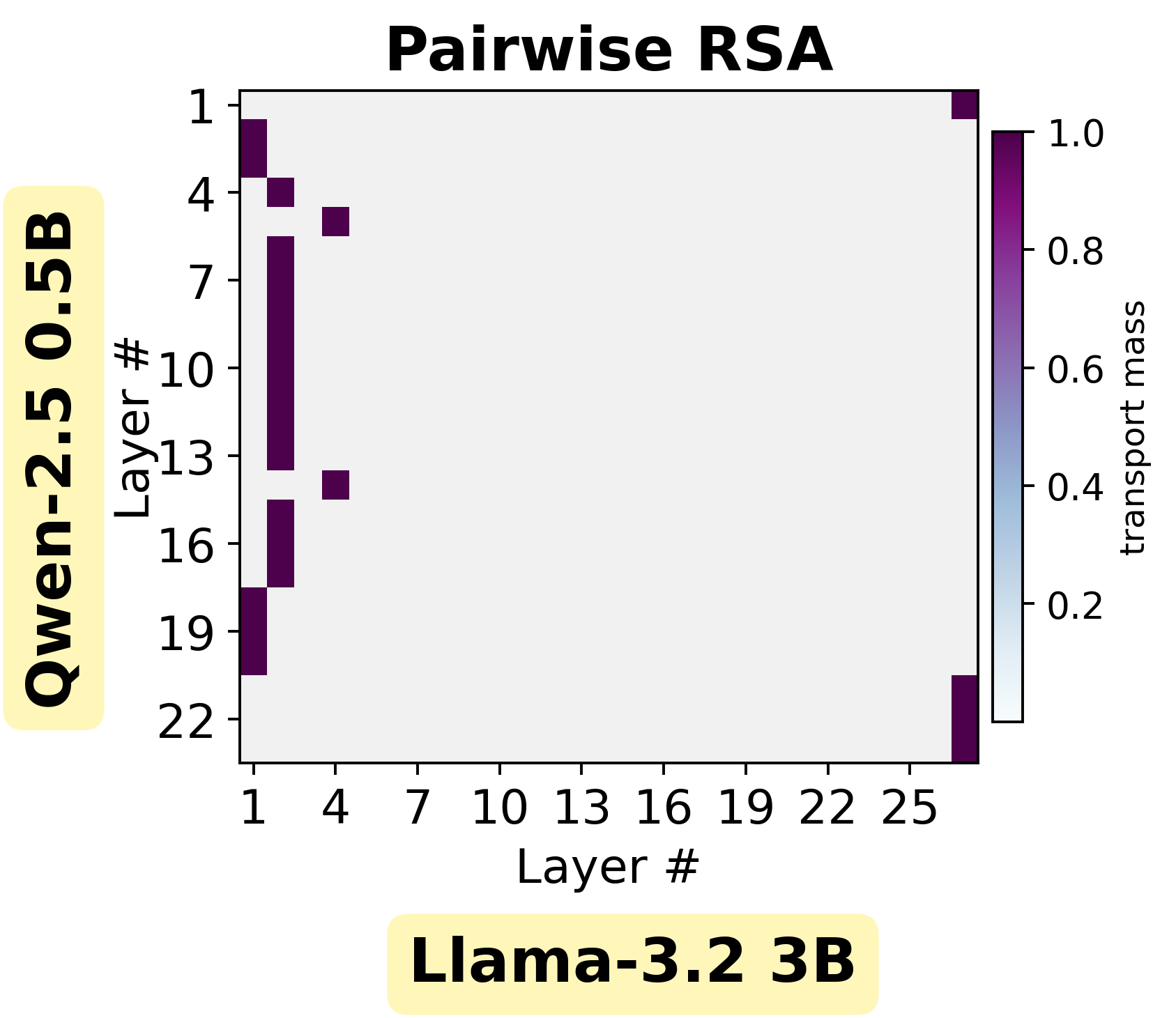}\hfill
\includegraphics[width=0.32\textwidth]{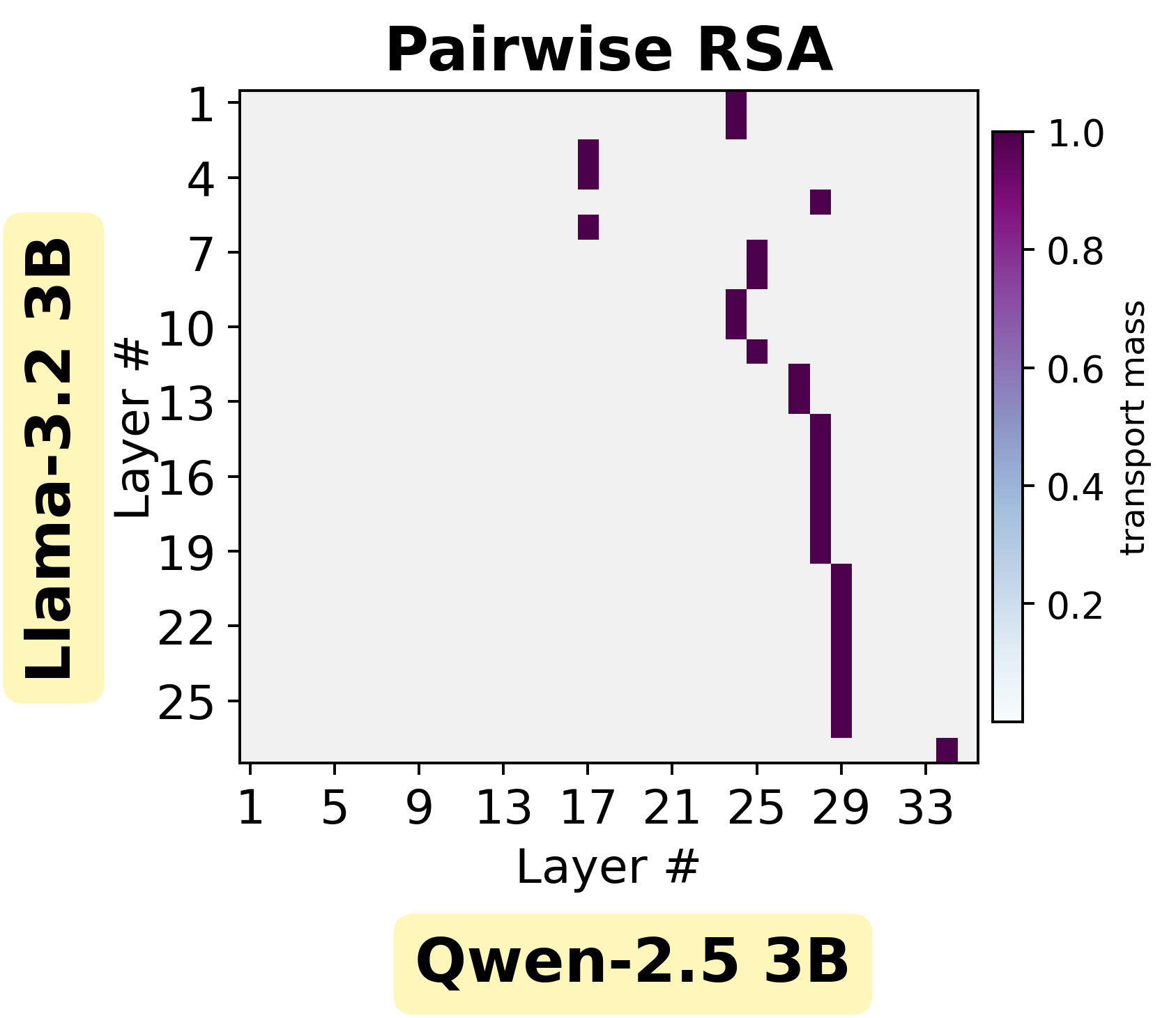}
\medskip\medskip

\includegraphics[width=0.32\textwidth]{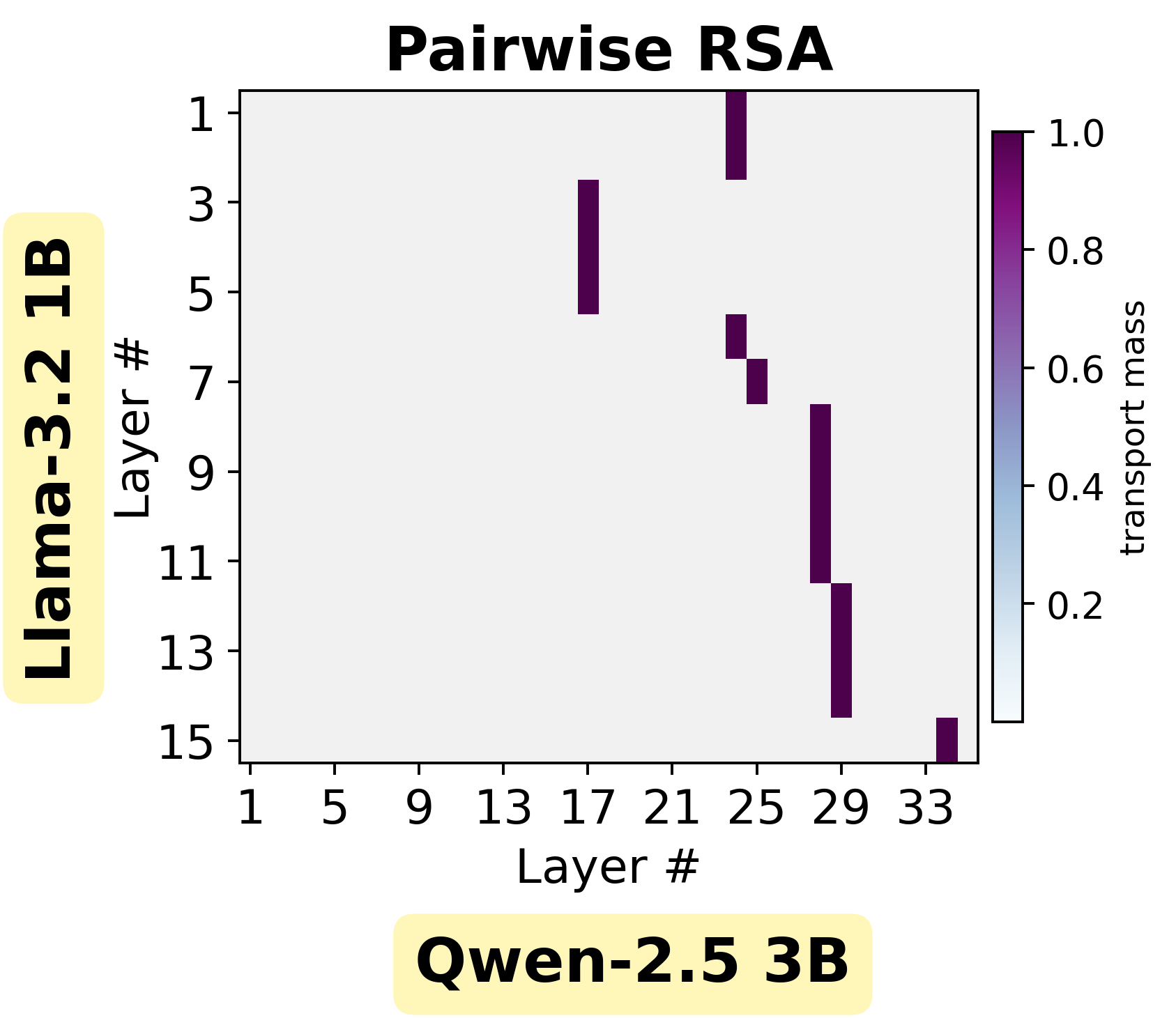}\hfill
\includegraphics[width=0.32\textwidth]{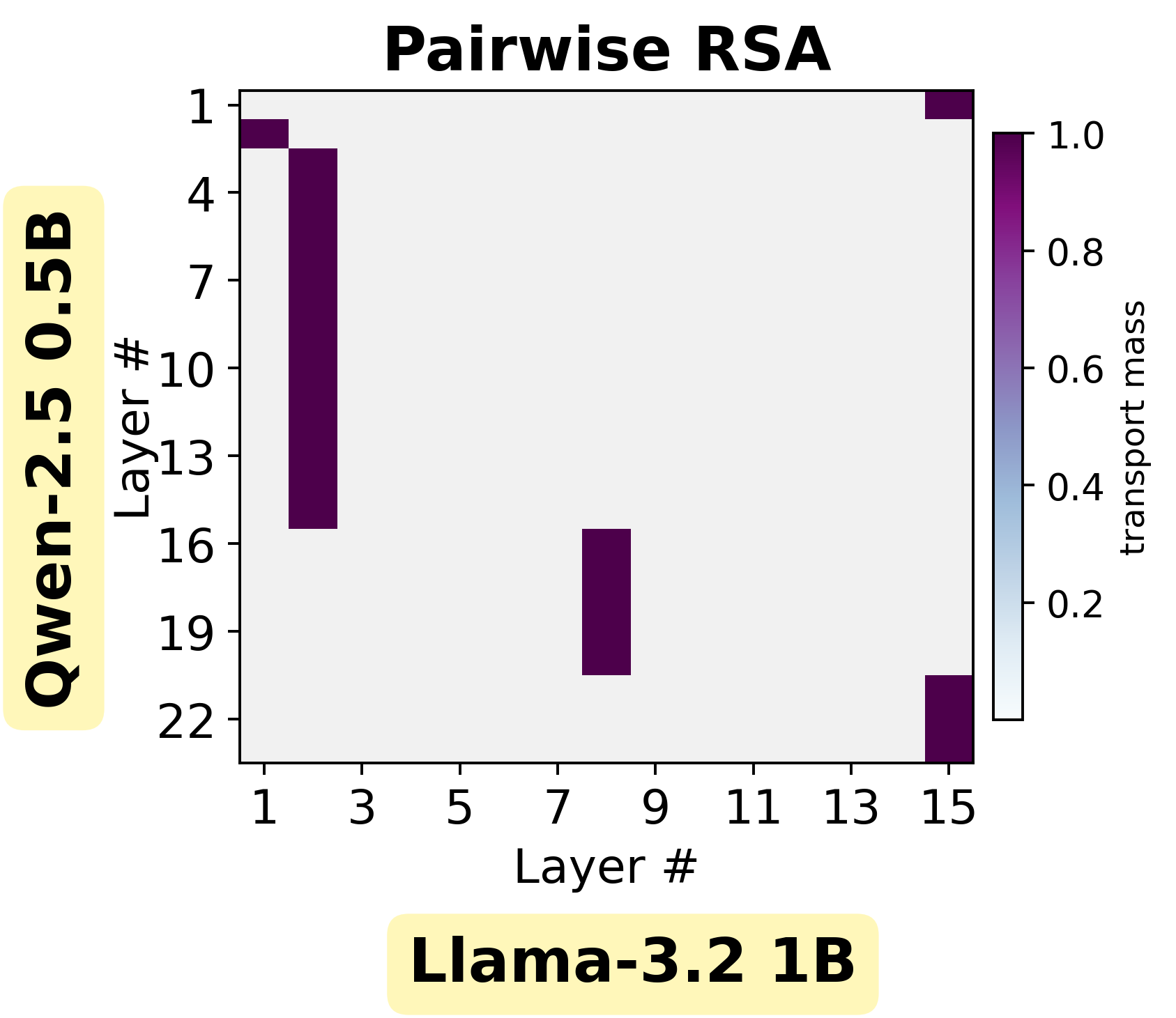}\hfill
\includegraphics[width=0.32\textwidth]{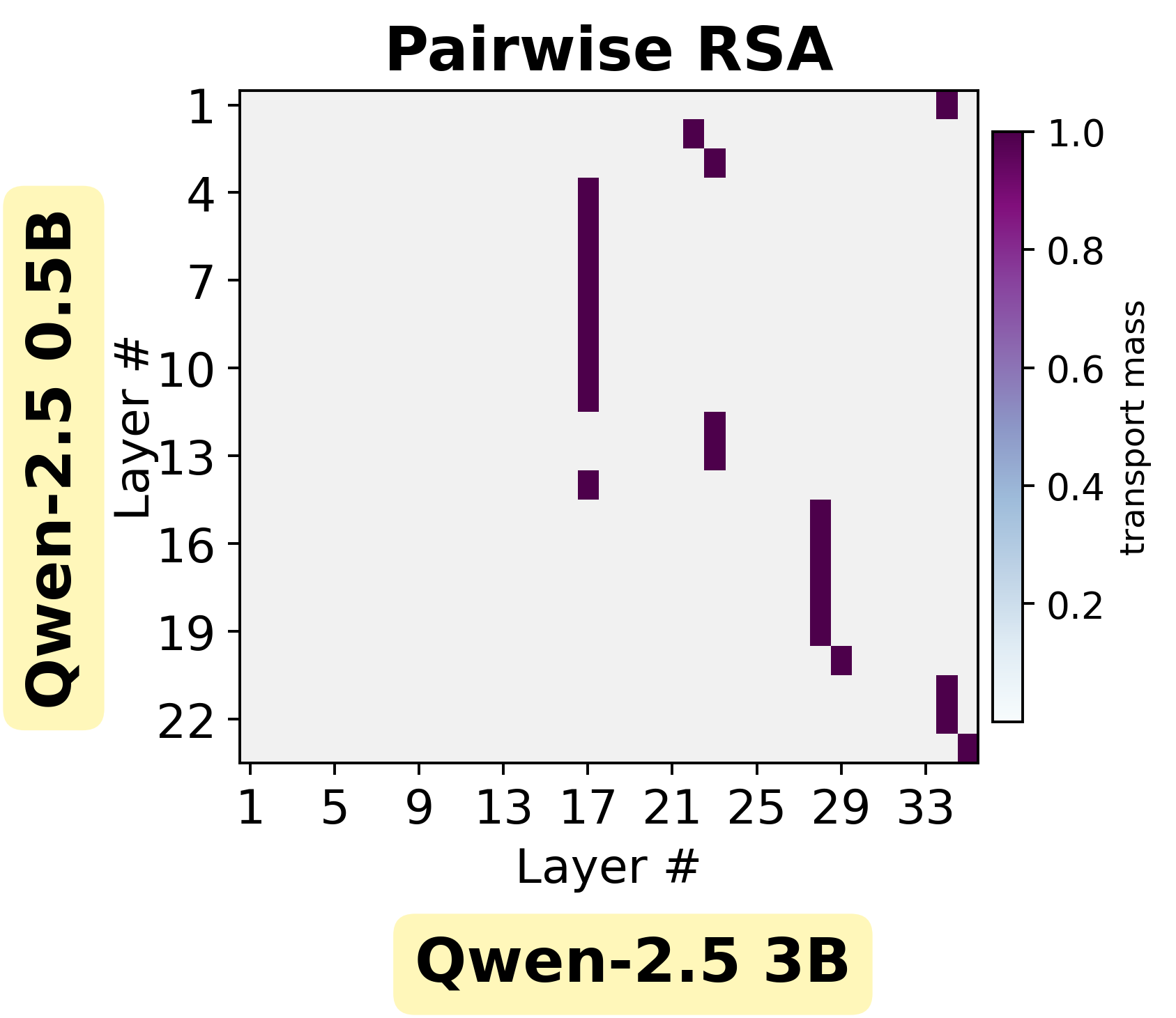}

\bigskip\bigskip

\includegraphics[width=0.32\textwidth]{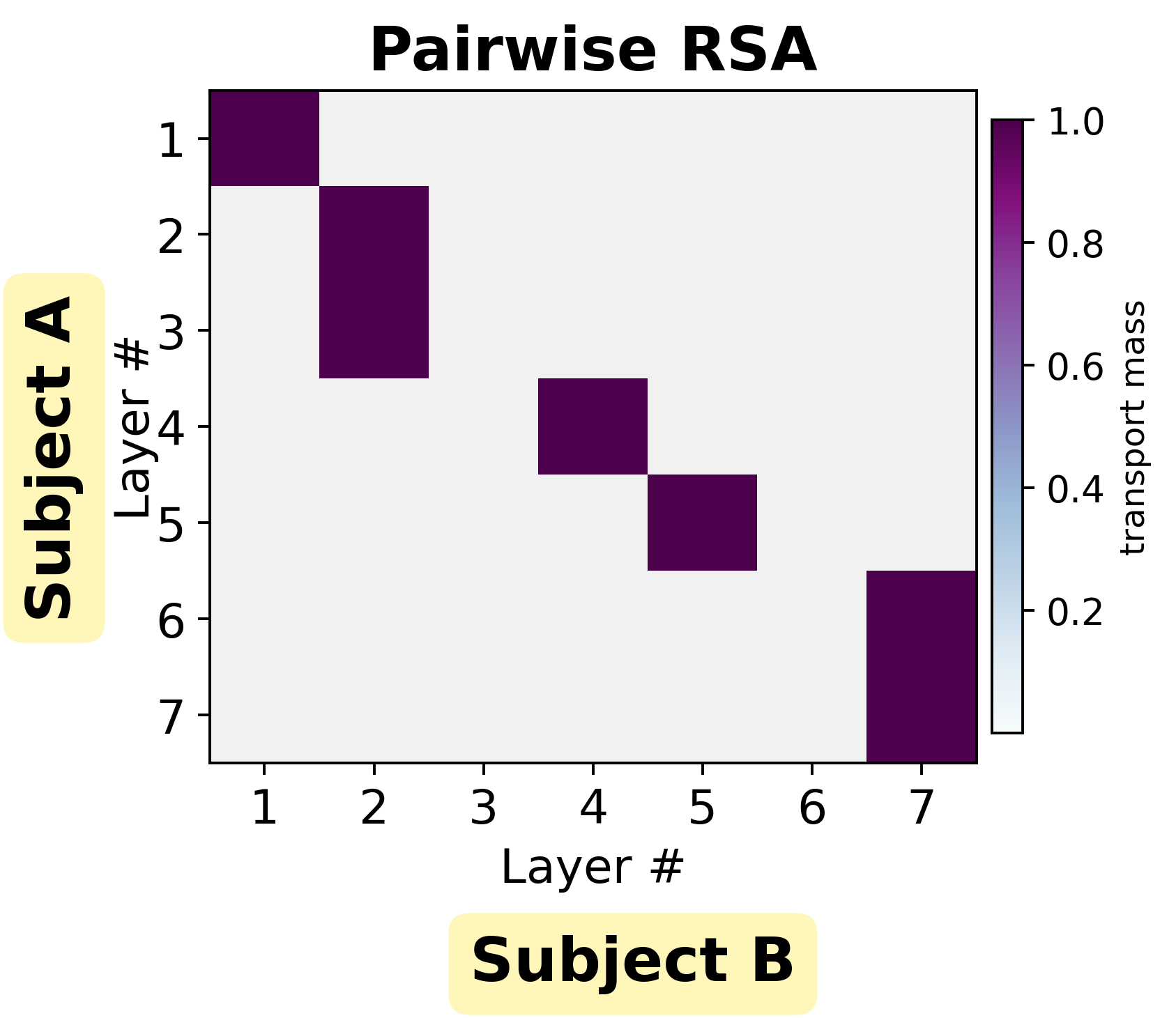}\hfill
\includegraphics[width=0.32\textwidth]{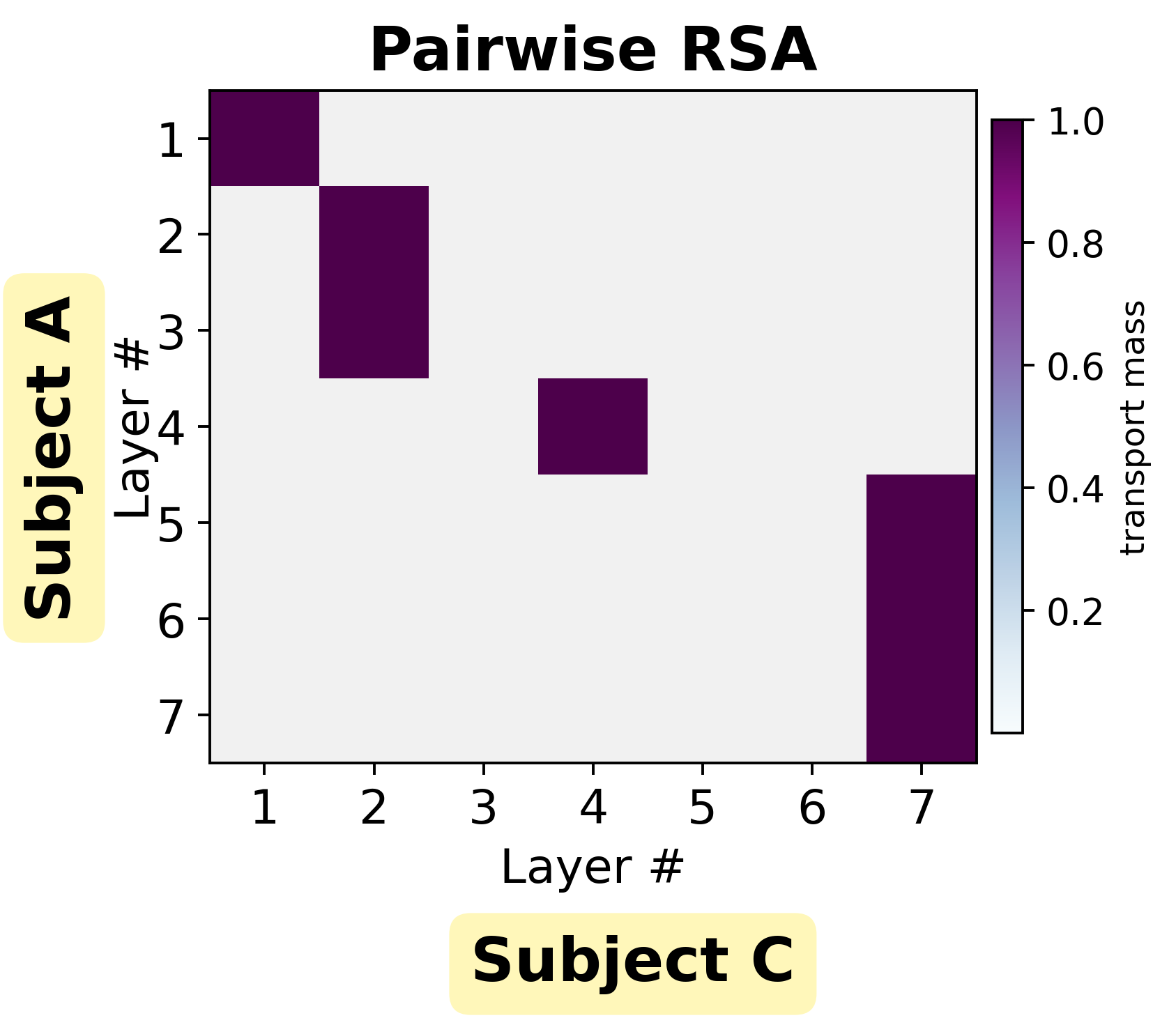}\hfill
\includegraphics[width=0.32\textwidth]{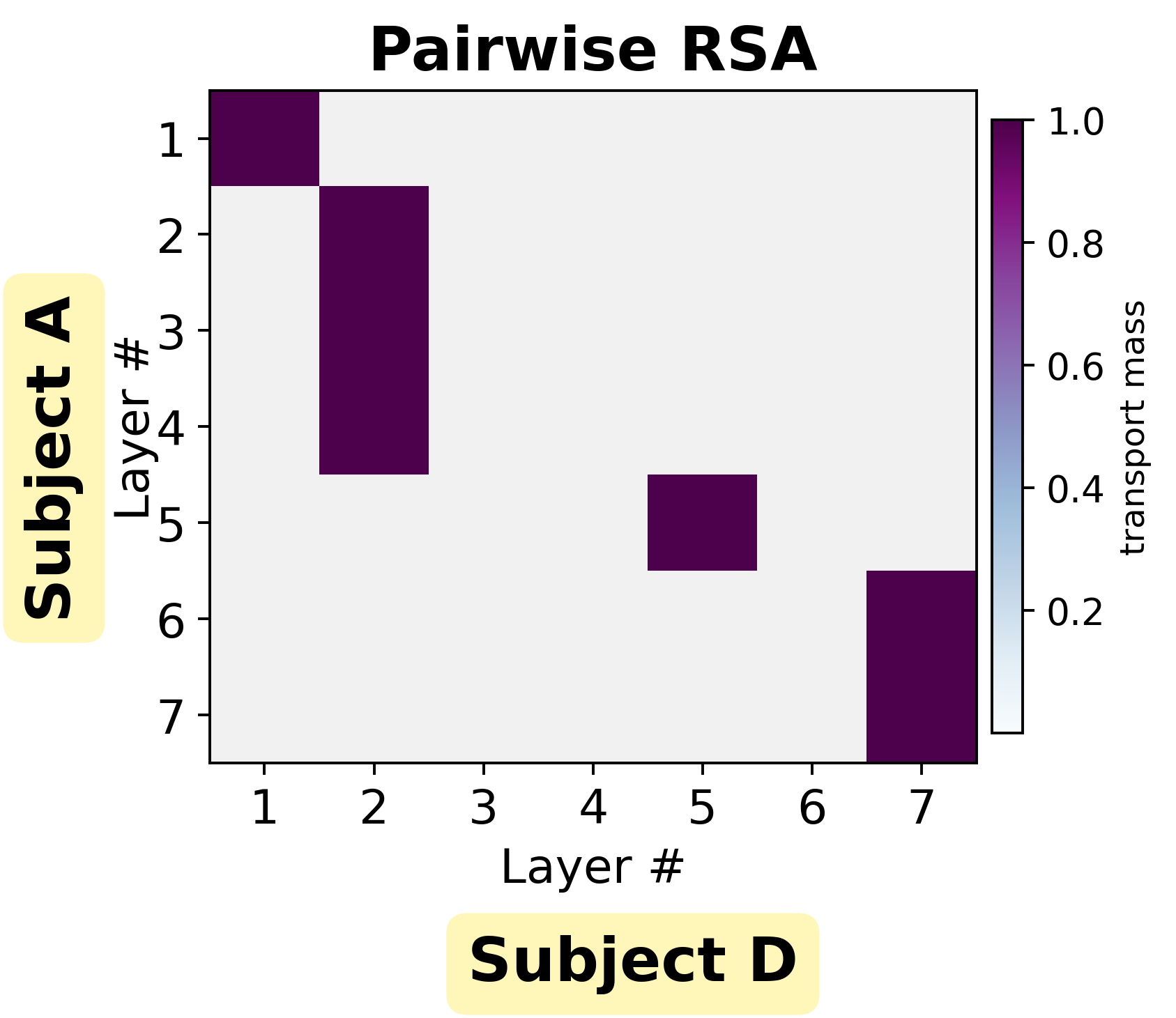}
\medskip\medskip

\includegraphics[width=0.32\textwidth]{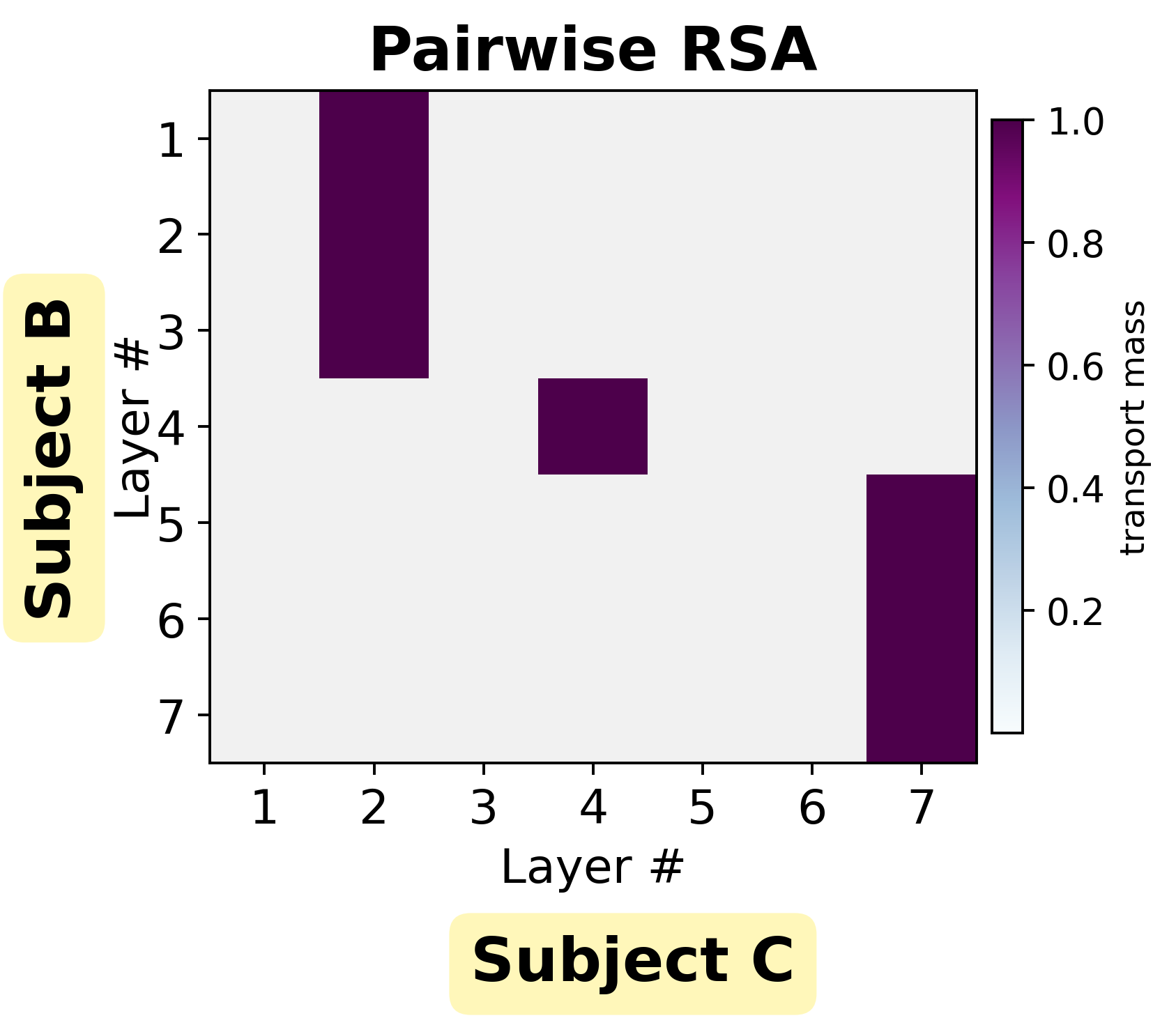}\hfill
\includegraphics[width=0.32\textwidth]{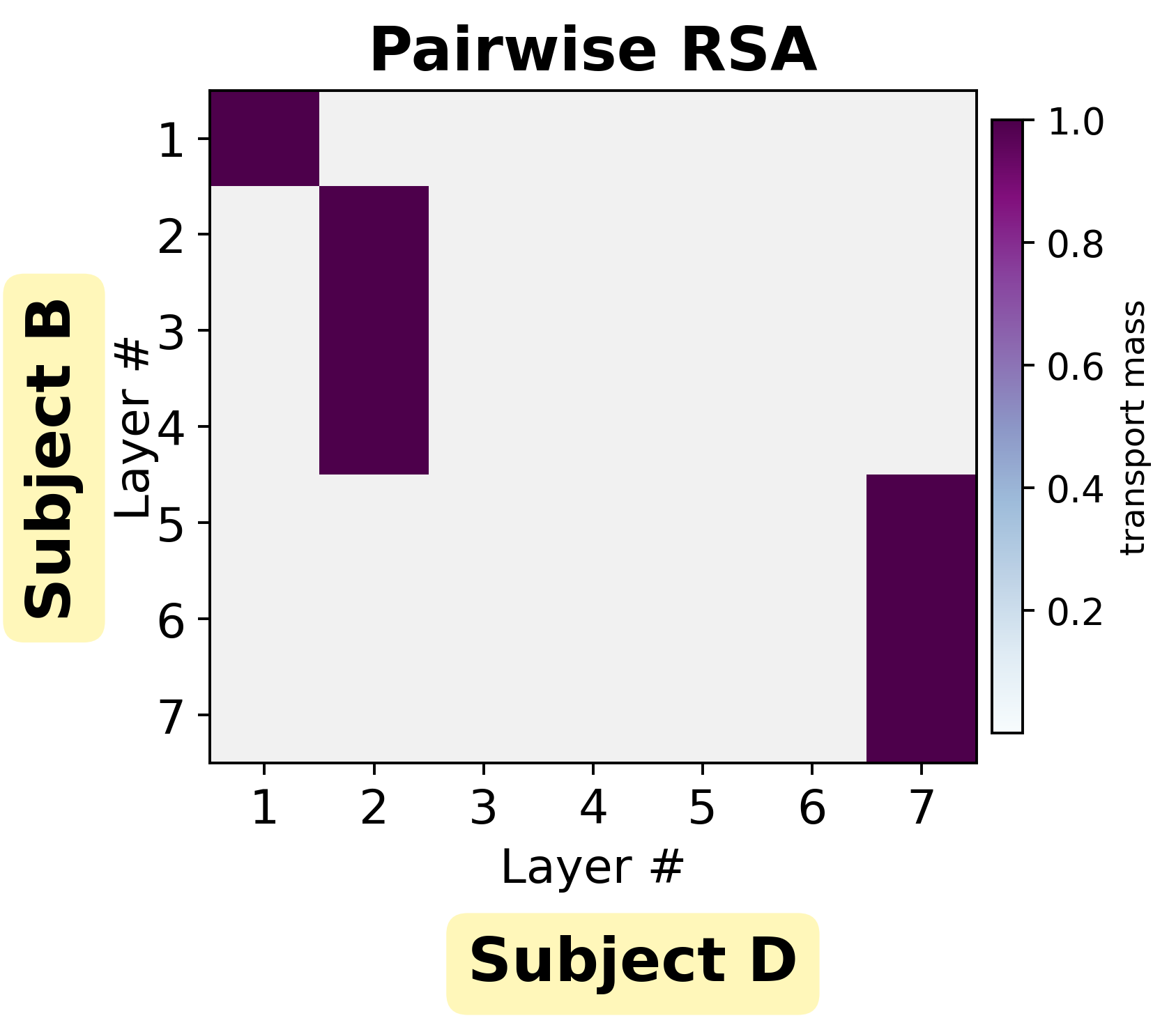}\hfill
\includegraphics[width=0.32\textwidth]{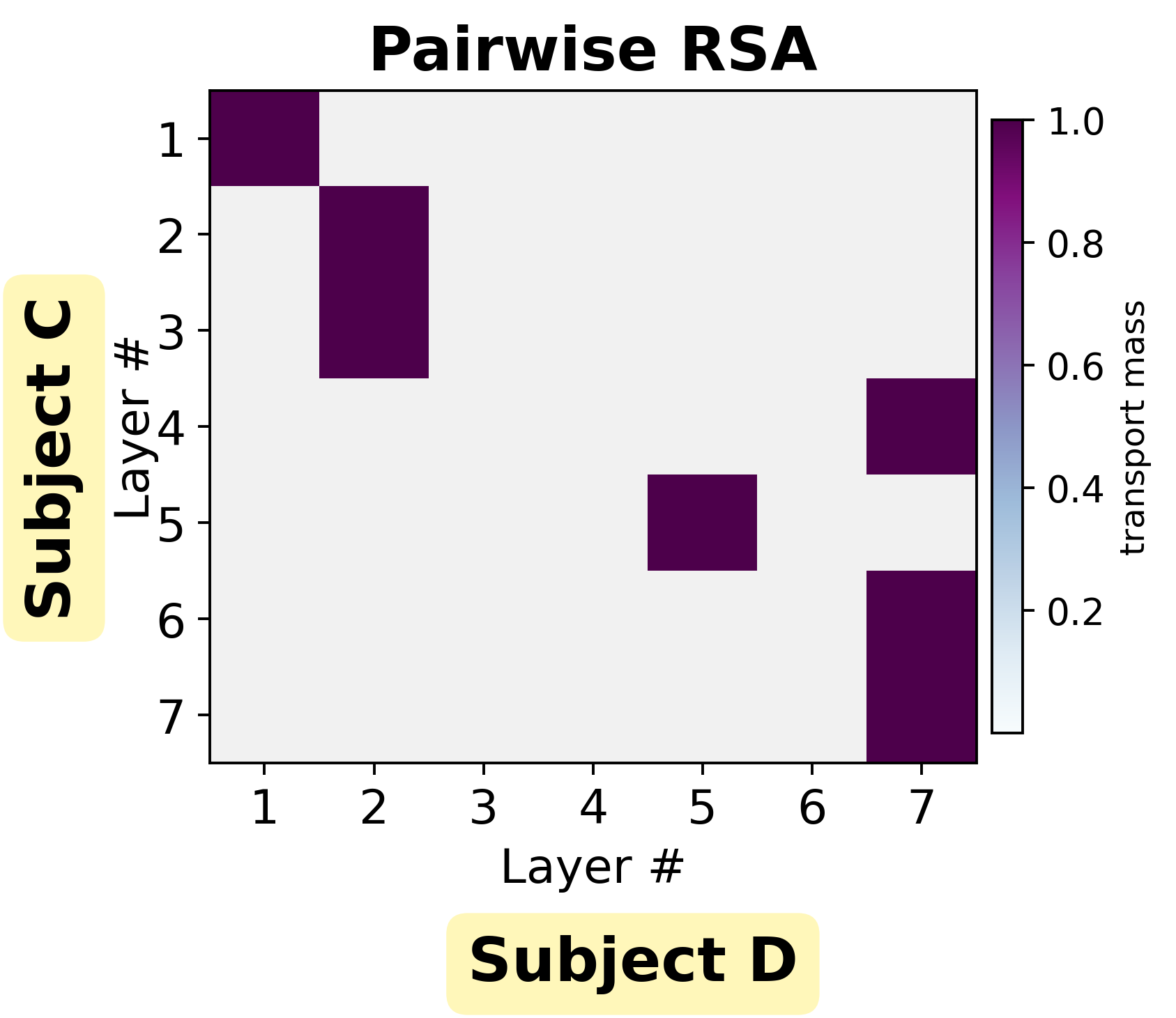}

\caption{\textbf{Transport plans for RSA-based mappings in LLM and fMRI spaces.}
Top two rows: pairwise mappings learned under RSA-based constraints for all LLM pairs:
(a) Llama-3.2 1B $\leftrightarrow$ Llama-3.2 3B,
(b) Qwen-2.5 0.5B $\leftrightarrow$ Llama-3.2 3B,
(c) Llama-3.2 3B $\leftrightarrow$ Qwen-2.5 3B,
(d) Llama-3.2 1B $\leftrightarrow$ Qwen-2.5 3B,
(e) Qwen-2.5 0.5B $\leftrightarrow$ Llama-3.2 1B, and
(f) Qwen-2.5 0.5B $\leftrightarrow$ Qwen-2.5 3B.
Bottom two rows: pairwise mappings learned under RSA-based constraints for all fMRI subject pairs:
(g) Subject A $\leftrightarrow$ Subject B,
(h) Subject A $\leftrightarrow$ Subject C,
(i) Subject A $\leftrightarrow$ Subject D,
(j) Subject B $\leftrightarrow$ Subject C,
(k) Subject B $\leftrightarrow$ Subject D, and
(l) Subject C $\leftrightarrow$ Subject D.
In both the LLM and fMRI settings, RSA-based pairwise mappings do not recover structured layer-wise correspondences, highlighting the importance of MOT for learning robust and generalizable mappings across architectures, model scales, and subjects.}
\label{fig:rsa_transport}
\end{figure}

\newpage
\section{Robustness of metrics to sub-sampling}
\begin{figure}[H]
\centering
\includegraphics[width=0.65\textwidth]{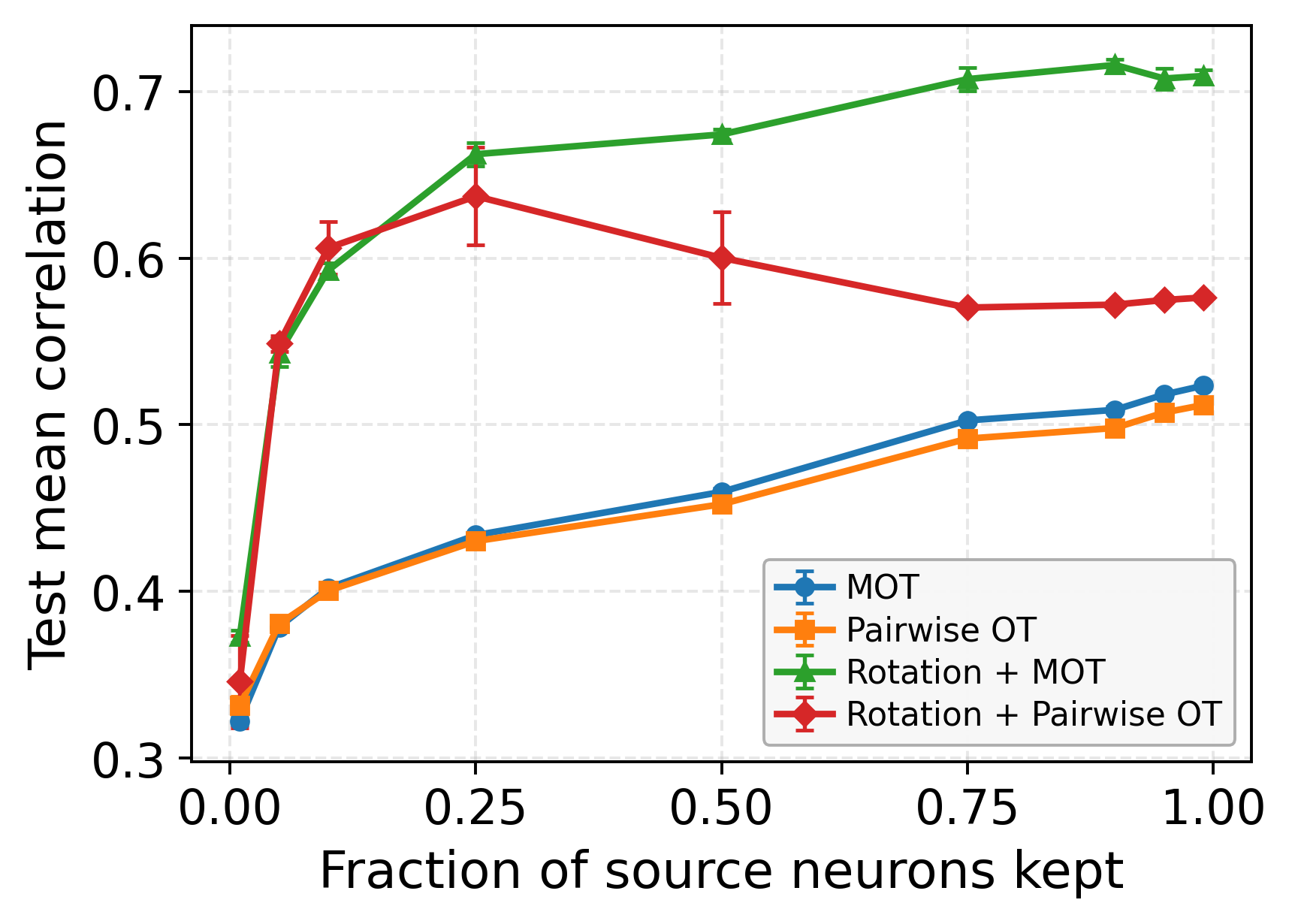}
\caption{\textbf{Neuron sub-sampling in LLM alignment.}
Reconstruction performance when aligning Qwen-2.5 0.5B to Llama-3.2 1B as a function of the fraction of Llama neurons kept. The curves show reconstruction scores for MOT and MOT+R (with their corresponding pairwise baselines), averaged over five different seeds.}
\label{fig:llm-sub-sampling}
\end{figure}

We consider a pair of LLMs (Qwen-2.5 0.5B and Llama-3.2 1B) and systematically sub-sample neurons from the larger model (Llama3.2 1B) to study how this affects reconstruction of the smaller model’s activations. For each fraction of neurons kept, we randomly select that fraction of neurons from Llama-3.2 1B, re-fit the transport plans between the two models, and evaluate the reconstruction score on held-out test data (a 20\% split, as before), repeating this procedure over five  different seeds. Figure \ref{fig:llm-sub-sampling} shows the resulting reconstruction scores for MOT and MOT+R (with their corresponding pairwise baselines).

We observe that reconstruction scores for both MOT/MOT+R and the corresponding baselines increase with the fraction of neurons kept. Importantly, sub-sampling does not cause MOT to collapse; its performance degrades smoothly. For MOT+R, keeping only 10\% of neurons yields a reconstruction score of nearly 0.5, indicating substantial robustness to aggressive sub-sampling. In the upper range of fraction of neurons kept, MOT+R achieves substantially higher reconstruction scores than both its rotation-aware baseline and the corresponding pairwise OT and MOT scores. These results suggest that neuron sub-sampling is a practical way to trade off computation and accuracy, and that the metrics remain informative even under significant sub-sampling.

\section{Use of Large Language Models}
LLMs were used in this work to assist with writing tasks, specifically for ensuring grammatical correctness and enhancing the clarity of the text.
\end{document}